\documentclass[10pt,twocolumn,letterpaper]{article}

\usepackage[pagenumbers]{cvpr} 

\usepackage{graphicx}
\usepackage{amsmath}
\usepackage{amssymb}
\usepackage{booktabs}
\usepackage{enumitem}
\usepackage{bbm}
\usepackage{multirow}
\usepackage{float}
\usepackage{pifont}
\newcommand{\cmark}{\ding{51}}
\newcommand{\xmark}{\ding{55}}

\usepackage[pagebackref,breaklinks,colorlinks]{hyperref}

\usepackage[capitalize]{cleveref}
\crefname{section}{Sec.}{Secs.}
\Crefname{section}{Section}{Sections}
\Crefname{table}{Table}{Tables}
\crefname{table}{Tab.}{Tabs.}

\newcommand{\eq}[1]{(\ref{eq:#1})}

\newcommand{\real}{\mathbb{R}}
\newcommand{\ind}{\mathbbm{1}}
\newcommand{\diag}{\operatorname{diag}}
\newcommand{\norm}[1]{\left\|{#1}\right\|}

\newcommand{\bc}{\mathbf{c}}
\newcommand{\bd}{\mathbf{d}}

\newcommand{\bq}{\mathbf{q}}
\newcommand{\bR}{\mathbf{R}}
\newcommand{\bs}{\mathbf{s}}
\newcommand{\bt}{\mathbf{t}}

\newcommand{\bx}{\mathbf{x}}
\newcommand{\by}{\mathbf{y}}
\newcommand{\bz}{\mathbf{z}}

\newcommand{\bmu}{\boldsymbol{\mu}}

\newcommand{\bSigma}{\boldsymbol{\Sigma}}

\newcommand{\cL}{\mathcal{L}}

\newcommand{\cN}{\mathcal{N}}

\newcommand{\cR}{\mathcal{R}}

\newcommand{\cT}{\mathcal{T}}

\newcommand{\cV}{\mathcal{V}}

\newcommand{\cX}{\mathcal{X}}
\newcommand{\cY}{\mathcal{Y}}
\newcommand{\cZ}{\mathcal{Z}}

\usepackage{amsfonts}

\DeclareMathOperator*{\argmin}{argmin~}

\makeatletter
\DeclareRobustCommand\onedot{\futurelet\@let@token\@onedot}
\def\@onedot{\ifx\@let@token.\else.\null\fi\xspace}
\def\eg{e.g\onedot} 
\def\ie{i.e\onedot}

\def\wrt{wrt\onedot}

\def\etal{et~al\onedot}

\makeatother

\newcommand{\boldparagraph}[1]{\vspace{0.2cm}\noindent{\bf #1:} }

\newcommand{\figref}[1]{Fig.~\ref{#1}}
\newcommand{\tabref}[1]{Tab.~\ref{#1}}
\newcommand{\secref}[1]{Sec.~\ref{#1}}

\usepackage{xcolor}
\definecolor{darkgreen}{rgb}{0,0.7,0}
\definecolor{orange}{RGB}{255,127,0}
\definecolor{ourpurple}{RGB}{127,127,204}
\definecolor{palgreen}{RGB}{51,179,179}
\definecolor{magenta}{RGB}{199,21,133}

\definecolor{darkred}{RGB}{220,50,32}
\definecolor{darkblue}{RGB}{0,90,181}
\newcommand{\darkred}[1]{\noindent{\color{darkred}{#1}}}
\newcommand{\darkblue}[1]{\noindent{\color{darkblue}{#1}}}

\begin{document}

\title{PartNeRF: Generating Part-Aware Editable 3D Shapes without 3D Supervision}

\author{%
  Konstantinos Tertikas\thanks{Work done during internship at Stanford.}\,\,$^{1,3}$ \quad Despoina Paschalidou$^{2}$ \quad
  Boxiao Pan$^{2}$ \quad Jeong Joon Park$^{2}$ \\ Mikaela Angelina Uy$^{2}$ \quad Ioannis Emiris$^{3,1}$\quad
  Yannis Avrithis$^{4}$ \quad Leonidas Guibas$^{2}$\\
  $^1$National and Kapodistrian University of Athens\quad
  $^2$Stanford University\\
  $^3$Athena RC, Greece \quad
  $^4$Institute of Advanced Research in Artificial Intelligence (IARAI)\\
  {\tt\small \{ktertikas,emiris\}@di.uoa.gr \quad \{paschald,bxpan,jjpark3d,mikacuy,guibas\}@stanford.edu} 
  }
\twocolumn[{%
\maketitle

\begin{center}
    \centering
    \captionsetup{type=figure}
    \includegraphics[width=0.9\linewidth]{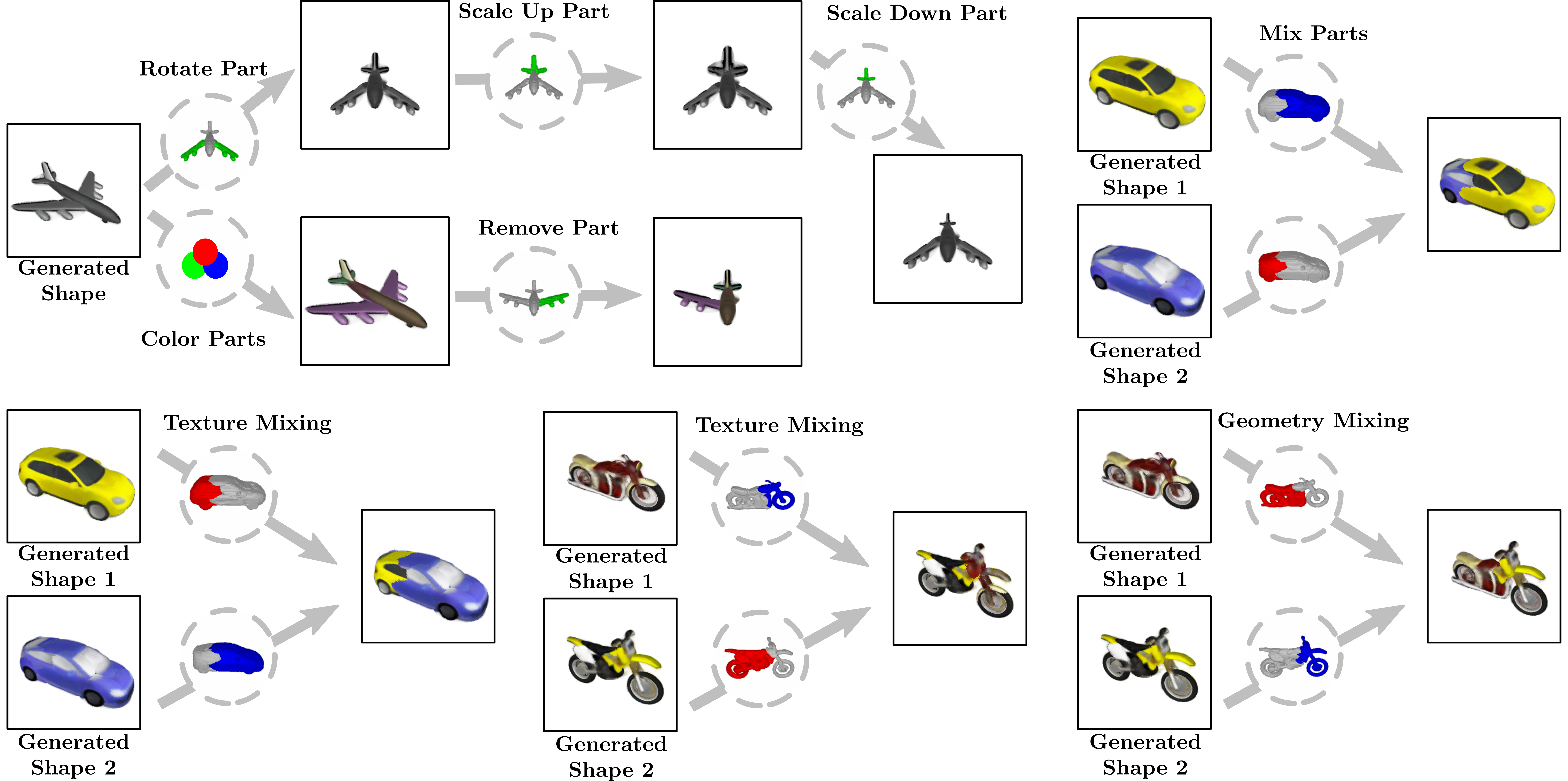}
    \captionof{figure}{{\bf Part-Aware Controllable 3D Shape Generation and Editing}. We
    address the task of part-aware 3D shape generation and editing without
    explicit 3D supervision. Prior part-aware generative models
    \cite{Hertz2022SIGGRAPH, Hao2020CVPR} assume 3D supervision, at training,
    and only allow changing the shape of the object. In this work, we
    introduce PartNeRF, a generative model capable of editing the shape and appearance of
    generated shapes that are parametrized as a collection of locally defined NeRFs.}
    \label{fig:teaser}
\end{center}%
}]
\def\thefootnote{*}\footnotetext{Work done during internship at Stanford.}\def\thefootnote{\arabic{footnote}}

\begin{abstract}

Impressive progress in generative models and implicit representations gave rise
to methods that can generate 3D shapes of high quality. However, being able to
locally control and edit shapes is another essential property that can unlock
several content creation applications. Local control can be achieved with part-aware
models, but existing methods require 3D supervision and cannot produce
textures. In this work, we devise PartNeRF, a novel part-aware generative model for
editable 3D shape synthesis that does not require any explicit 3D
supervision. Our model generates objects as a set of locally defined NeRFs,
augmented with an affine transformation. This enables several editing
operations such as applying transformations on parts, mixing parts from
different objects etc. To ensure distinct, manipulable parts we enforce a hard
assignment of rays to parts that makes sure that the color of each ray is
only determined by a single NeRF. As a result, altering one part does not
affect the appearance of the others. Evaluations on various ShapeNet
categories demonstrate the ability of our model to generate editable 3D objects of
improved fidelity, compared to previous part-based generative approaches that
require 3D supervision or models relying on NeRFs.

\end{abstract}

\section{Introduction}
\label{sec:intro}

Generating realistic and editable 3D content is a long-standing
problem in computer vision and graphics that has recently gained more attention
due to the increased demand for 3D objects in AR/VR, robotics and gaming
applications. However, manual creation of 3D models is a laborious
endeavor that requires technical skills from highly experienced
artists and product designers. On the other hand, editing 3D shapes, typically
involves re-purposing existing 3D models, by manually changing faces and 
vertices of a mesh and modifying its respective UV-map
\cite{Schmidt2016ComputerGraphicsForum}.
To accommodate this process, several recent works introduced
generative models that go beyond generation and allow editing
the generated instances \cite{
Yang2020ARXIV, Liu2021ICCV, Yang2021ICCV, Li2021SIGGRAPH,
Deng2021CVPR, Niemeyer2021CVPR, Sun2022CVPR, Cheng2022ARXIV, Yuan2022CVPR,
Lazova2022ARXIV}. Shape editing involves
making \emph{local changes} on the shape and the appearance of different parts
of an object. Therefore, having a basic understanding of the
decomposition of the object into parts facilitates controlling \emph{what to
edit}.

While Generative Adversarial Networks (GANs) \cite{Goodfellow2014NIPS} have
emerged as a powerful tool for synthesizing photorealistic images
\cite{Brock2019ICLR, Choi2018CVPR, Choi2020CVPR, Karras2018ICLR,
Karras2020CVPR, Karras2021NeurIPS}, scaling them to 3D data is non-trivial as
they ignore the physics of image formation process along a viewing direction.
To address this, 3D-aware GANs
incorporate 3D representations such as voxel grids
\cite{Henzler2019ICCV, Nguyen2019ICCV, Nguyen2020NeurIPS} in
generative settings or combine them with differentiable renderers
\cite{Liao2020CVPRa, Zhang2021ICLR}.
While they faithfully recover the geometry and appearance, they do not allow
changing specific parts of the object.

Inspired by the rapid evolution of neural implicit rendering techniques
\cite{Mildenhall2020ECCV},
recent works
\cite{Schwarz2020NEURIPS, Niemeyer2021CVPR, Xue2022CVPR, Chan2022CVPR,
Gu2022ICLR} proposed to combine them with GANs in order to allow for
multi-view-consistent generations of high quality.
Despite their impressive performance on novel view synthesis,
their editing capabilities are limited. To this end, editing operations in the
latent space have been explored \cite{Elsner2021ARXIV, Liu2021ICCV,
Jang2021ICCV, Yang2022ECCV, Yuan2022CVPR} but these approaches lack intuitive control
over the shape. By decomposing shapes into parts, other works facilitate
structure-aware shape manipulations \cite{Mo2020CVPR, Wu2020CVPR,
Hertz2022SIGGRAPH, Petrov2022ARXIV}. However, they require 3D supervision during training
and can only operate on textureless shapes.

To address these limitations, we devise PartNeRF, a novel part-aware generative model,
implemented as an auto-decoder~\cite{Bojanowski2018ICML}.  Our model enables
part-level control, which facilitates various editing operations on the
shape and appearance of the generated instance. These operations include
rigid and non-rigid transformations on
the object parts, part mixing from different objects, removing/adding parts and editing the
appearance of specific parts of the object.

Our key idea is to represent the objects using
a set of locally defined Neural Radiance Fields (NeRFs) that are arranged
such that the object can
be plausibly rendered from a novel view. To enable part-level control, we
enforce a \emph{hard assignment} between parts and rays that ensures that altering
one part does not affect the shape and appearance of other parts. Our model does not require any explicit 3D
supervision; we only assume supervision from images and
object masks captured from known cameras. We evaluate
PartNeRF on various ShapeNet categories and demonstrate that it
can generate textured shapes of higher fidelity than methods that assume 3D
supervision as well as approaches that rely on NeRFs. Furthermore, we showcase
various editing operations that were not previously possible.

In summary, we make the following \textbf{contributions}: We propose the first
part-aware generative model that parametrizes parts as NeRFs. Among the
existing generative pipelines that consider the decomposition of
objects into parts, our work is the first that does not require explicit 3D
supervision and enables editing operations both on the shape and the  texture
of the generated object.
Code and data is available at \url{https://ktertikas.github.io/part_nerf}.

\section{Related Work}
\label{sec:related}

We now discuss the most relevant literature on 3D generative
models in the context of generating editable shapes.

\boldparagraph{Neural Implicit Representations}%
Neural Implicit Representations
\cite{Mescheder2019CVPR, Park2019CVPR, Chen2019CVPR}
have demonstrated impressive capabilities on various tasks ranging from 3D
reconstruction with \cite{Oechsle2021ICCV, Saito2019ICCV, Saito2020CVPR,
Oechsle2020THREEDV} and without texture \cite{Mescheder2019CVPR, Park2019CVPR,
Chen2019CVPR, Michalkiewicz2019ICCV, Atzmon2019NIPS, Xu2019NIPS, Peng2020ECCV,
Atzmon2020CVPR, Chibane2020CVPR, Jiang2020CVPR, Gropp2020ICML, Remelli2020ARXIV}
to video encoding
\cite{Chen2021NeurIPS}, 3D-aware generative
modelling \cite{Schwarz2020NEURIPS,
Chan2021CVPR, Niemeyer2021CVPR, Meng2021ICCV, DeVries2021ICCV, Gu2022ICLR, Chan2022CVPR},
inverse graphics \cite{Niemeyer2020CVPR, Yariv2020ARXIV} and novel view
synthesis \cite{Mildenhall2020ECCV, Yu2021CVPR, Henzler2021CVPR, Wang2021CVPR,
Reizenstein2021ICCV}. In contrast to explicit representations \ie point clouds,
meshes and voxels, implicit representations encode the shape's geometry
and appearance in the weights of a neural network.
Among the most
extensively used implicit-based models are NeRFs
\cite{Mildenhall2020ECCV}, which combine an implicit neural network with
volumetric rendering \cite{Kajiya1984SIGGRAPH} to perform novel view synthesis.
Due to their compelling results,
numerous works have been introduced to improve
the training and rendering time
\cite{Liu2020NeuRIPS, Zhang2020ARXIVb, Rebain2021CVPR, Lindell2021CVPR,
Tancik2021CVPR, Garbin2021ICCV, Reiser2021ICCV, Yu2021ICCV,
Muller2022SIGGRAPH}, the underlying geometry \cite{Oechsle2021ICCV,
Wang2021NeurIPS}, to better handle lighting variations \cite{Brualla2021CVPR,
Bi2020ARXIV, Srinivasan2021CVPR, Boss2021ICCV} and to encode shape priors for
better generalization \cite{Yu2021CVPR, Henzler2021CVPR, Trevithick2021ICCV}
and generation \cite{Schwarz2020NEURIPS, Chan2021CVPR}. For a thorough overview
on NeRF-based approaches we refer readers to \cite{Tewari2022CGF, Xie2022CGF}.
In our work, we introduce a generative model for editable 3D shapes with
texture. Specifically, we parametrize objects as a structured set of local
NeRFs that are trained from posed images and object masks.

\begin{figure*}
    \centering
    \includegraphics[width=1.0\textwidth]{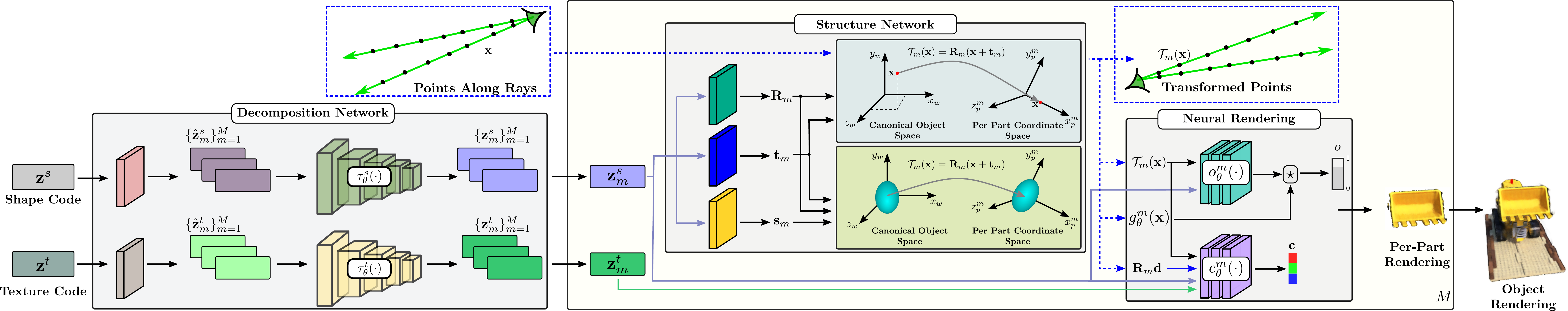}
    \caption{
        {\bf{Method Overview.}} Our generative model is implemented as an
        auto-decoder and it comprises three main components: The
        \emph{Decomposition Network} takes two object specific learnable embeddings
        $\{\bz_s, \bz_t\}$ that represent its shape and texture and maps
        them to a set of $M$ latent codes that control the shape and texture of
        each part. First, we map $\bz^s$ and $\bz^t$  to $M$ per-part
        embeddings $\{\hat{\bz}_m^s\}_{m=1}^M$ and $\{\hat{\bz}_m^t\}_{m=1}^M$ using $M$
        linear projections, which are then fed to two
        transformer encoders: $\tau^s_\theta$ and $\tau^t_\theta$, that predict the final
        per-part shape and texture embeddings, $\{\bz_m^s\}_{m=1}^M$ and
        $\{\bz_m^t\}_{m=1}^M$. Next, the \emph{Structure Network}
        maps the per-part shape feature representation $\bz_m^s$ to a rotation
        matrix $\bR_m$, a translation vector $\bt_m$ and a scale vector $\bs_m$
        that define the coordinate system of the m-th part and its spatial extent.
        The last component of our model is the \emph{Neural Rendering} module that
        takes the 3D points along each ray, transformed to the coordinate frame
        of its associated part, and maps them to an occupancy and a color value.
        We use plate notation to denote repetition over the $M$ parts.
    }
    \label{fig:method_overview}
    \vspace{-1.2em}
\end{figure*}

\boldparagraph{3D-Aware Image Generation}%
Incorporating 3D representations in generative settings
\cite{Gadelha2017THREEDV, Henderson2019IJCV, Henzler2019ICCV,
Henderson2020NIPS, Lunz2020ARXIV, Nguyen-Phuoc2019ICCV, Nguyen-Phuoc2020ARXIV,
Gu2022ICLR, DeVries2021ICCV, Hao2021ICCV, Meng2021ICCV, Niemeyer2021THREEDV,
Zhou2021ARXIV} has significantly improved the quality of the generated images
and increased control over various aspects of the image formation
process. Likewise, radiance fields have been combined
with GANs to allow for photorealistic image synthesis of
objects \cite{Schwarz2020NEURIPS, Chan2021CVPR} and scenes
\cite{Niemeyer2021CVPR}. More recently, \cite{Chan2022CVPR}
introduced a triplane-based architecture that leverages both
implicit and explicit representations and can generate high resolution
images. Concurrently, \cite{Gao2022NEURIPS} combined
differentiable surface modelling \cite{Gao2020NeurIPS} with a differentiable
renderer to generate high-quality textured meshes.
Unlike
\cite{Schwarz2020NEURIPS, Chan2021CVPR, Chan2022CVPR, Gao2022NEURIPS} that
are part agnostic, our model generates objects
with part-level control, hence unlocking editing operations not
previously possible.
Our formulation is closely related to the compositional
representation of \cite{Niemeyer2021CVPR} but has two important differences.
First, we enforce a hard assignment between rays and parts
ensuring that the color of each part is only determined by one NeRF, thus enabling local editing.
Moreover, we model the shape and texture of each part separately, hence
enabling even more control.

\boldparagraph{Shape Editing using NeRFs}%
Our work falls into the category of implicit-based shape editing approaches
that directly operate on the radiance field \cite{Liu2021ICCV, Yuan2022CVPR,
Yang2022ECCV}. Recent methods \cite{Yuan2022CVPR, Yang2022ECCV} extract meshes
from a pre-trained NeRF and rely on deformation techniques
\cite{Sorkine2007SGP, Joshi2007TG, Floater2003ComputerAidedGeometricDesign,
Lipman2008TG} to guide the rendering
process. Unlike our approach, these models are scene specific and cannot capture shape and
texture variations across an object class.
An alternative line of research explored using separate embeddings for
capturing the shape and texture variations of 3D shapes \cite{Liu2021ICCV,
Jang2021ICCV}. They demonstrated various editing operations such as color
modifications or removal of certain parts of the shape. However, as they do not
consider parts, they rely on heuristics for controlling what needs to be changed.
Instead, by incorporating parts, our model
provides more intuitive control, when editing a 3D shape. Also similar to our work,
\cite{Noguchi2022CVPR} uses multi-view videos and decomposes the object using a
set of ellipsoids. However, this model is scene-specific and
cannot generate novel shapes. In the context of face editing \cite{Kacper2022CVPR,
Jingxiang2022CVPR, Jingxiang2022ARXIV}, different models require video
sequences and partial semantic masks
\cite{Kacper2022CVPR}, or posed images along with full semantic masks
\cite{Jingxiang2022CVPR, Jingxiang2022ARXIV}. In contrast, our work only requires
posed images and 2D object masks at training.

\boldparagraph{Primitive-based Representations}%
Shape abstraction techniques represent 3D shapes using semantically consistent
primitive arrangements across different instances in a class. This most often requires 3D supervision
\cite{Gao2019SIGGRAPHASIA, Li2017SIGGRAPH, Li2019CVPR, Li2020ECCV, Niu2018CVPR,
Mo2019SIGGRAPH, Sharma2020ECCV, Zou2017ICCV, Deng2020CVPR, Deprelle2019NIPS, Genova2019ICCV,
Kawana2020NIPS, Paschalidou2019CVPR, Paschalidou2020CVPR, Tulsiani2017CVPRa},
although \cite{Yao2021ICCV, Yao2022ARXIV} demonstrate that is possible to learn only from images.
Unlike our parts, geometric primitives are typically simple shapes such as cuboids
\cite{Tulsiani2017CVPRa}, spheres \cite{Hao2020CVPR} or superquadrics
\cite{Paschalidou2019CVPR, Paschalidou2020CVPR}. Due to their simple
shape parametrization these primitives cannot
capture complex geometries. To address this, recent works propose to
increase the number of primitives \cite{Deng2020CVPR, Kawana2020NIPS} or
represent shapes using a family of homeomorphic mappings
\cite{Paschalidou2021CVPR}, or a structured set of implicit functions
\cite{Genova2020CVPR}. Similar to
\cite{Paschalidou2021CVPR} our parts can capture complex
geometries, but as we parametrize them with locally defined
NeRFs, we do not require explicit 3D supervision.

\boldparagraph{Part-based Shape Editing}%
Part-based generative models \cite{ Hao2020CVPR, Wu2020CVPR,
Li2021SIGGRAPH, Hertz2022SIGGRAPH, Petrov2022ARXIV} can generate plausible
3D shapes \cite{Mo2019SIGGRAPH, Mo2020CVPR, Yang2020ARXIV} and perform various editing
operations such as part mixing \cite{Yin2020THREEDV, Lin2022NEURIPS,
Petrov2022ARXIV} and shape manipulation \cite{Hao2020CVPR, Hertz2022SIGGRAPH}.
Closely related to our work are \cite{Hao2020CVPR, Hertz2022SIGGRAPH} which employ
primitives, such
as spheres \cite{Hao2020CVPR} and 3D Gaussians \cite{hertz2022spaghetti},
to enable shape editing by explicitly
\cite{Hao2020CVPR} or implicitly \cite{Hertz2022SIGGRAPH} transforming
them. However, both require 3D supervision and can only alter
attributes related to the shape of the object.

\section{Method}
\label{sec:method}

Our goal is to design a 3D part-aware generative model that can be trained
without explicit 3D supervision. Moreover, we
want our generated shapes to be editable, namely to be able to make local
changes on the shape and texture of specific parts of the object. To this
end, we represent objects using $M$ \emph{locally defined} parts that are
parametrized with a NeRF~\cite{Mildenhall2020ECCV}. Defining NeRFs locally, \ie
in their own coordinate system, enables direct part-level control simply by
applying transformations on the per-part coordinate system.

However, to achieve distinct, manipulable parts, it is essential that each
object part is represented by a single NeRF.  To enforce this, we introduce a
\emph{hard assignment} between rays and parts by associating a ray with the
first part it intersects (\figref{fig:ray_part_association}). Namely, the color
of each ray is predicted from \emph{a single NeRF}, thus preventing
combinations of parts reasoning about the color of a ray.  Note that our
formulation differs from \cite{Niemeyer2021CVPR}, as we explicitly associate
rays with parts, instead of simply combining them. This ensures that when
editing one part, the shape and appearance of the others does not change
(\figref{fig:teaser}).

\subsection{Neural Radiance Fields}
\label{subsec:nerf}

Given a set of posed images of objects in a semantic class, each
accompanied by an object mask, which is simply a binary image indicating
whether each pixel is inside the object or not, we define $\cR$, the complete
set of rays from all views. For each
ray $r = \{\bx_0^r + t \bd^r: t \ge 0\}$ with origin $\bx_0^r$ and viewing
direction $\bd^r$, we denote $C(r) \in \mathbb{R}^3$ and $I(r) \in \{0,1\}$ the
color value of the RGB image and the binary value of the mask, respectively, at
the corresponding pixel. Finally, we sample a set of $N$ points $\cX_r =
\{\bx_1^r, \dots, \bx_N^r\}$ along $r$, which are ordered by increasing
distance from the origin $\bx_0^r$, used for estimating the color along this
ray using numerical quadrature \cite{Wallis1656}.

\boldparagraph{Neural Radiance Fields}%
NeRFs~\cite{Mildenhall2020ECCV} represent a scene as a continuous function,
parametrized with an MLP, that maps a 3D point $\bx \in \real^3$ and a viewing
direction $\bd \in \mathbb{S}^2$ into a \emph{color} $\bc \in \real^3$ and a
volume \emph{density} $\sigma \in \real^{+}$. Before passing the inputs $\bx$
and $\bd$ to the MLP, they are projected to a higher dimensional space by
applying a fixed \emph{positional encoding}~\cite{Vaswani2017NIPS} to each
one of their elements.  Given the predicted color and densities
$\{\bc_i^r, \sigma_i^r\}_{i=1}^N$ for the $N$ sampled points $\cX_r$ along ray
$r$, its rendered color can be derived from
\begin{equation}
    \hat{C}(r) = \sum_{i=1}^N \exp\Big( -\sum_{j<i} \sigma_j^r \delta_j^r\Big)
             (1 -\exp(-\sigma_i^r \delta_i^r))\bc_i^r,
    \label{eq:rendering_nerf}
\end{equation}
where $\delta_i^r$ is the distance between two adjacent samples along $r$.
At training, the MLP is optimized by minimizing the error between observed and
rendered images.

Alternatively, \cite{Oechsle2021ICCV} propose predicting
occupancies instead of densities, hence their rendering equation becomes
\begin{equation}
	\hat{C}(r) = \sum_{i=1}^N
		o_i^r \prod_{j<i} \left(1 - o_j^r \right)\bc_i^r,
    \label{eq:rendering_unisurf}
\end{equation}
where $o_i^r = 1- \exp(-\sigma_i^r \delta_i^r)$ is the occupancy value at point
$\bx_i^r$ and $\bc_i^r$ its color. Similar to \cite{Oechsle2021ICCV} we also
predict occupancy values, as this facilitates associating rays with parts,
hence enabling part-level control, as discussed in \secref{subsec:part_nerf}.

\subsection{Parts as Neural Radiance Fields}
\label{subsec:part_nerf}

We represent a 3D object using $M$ parts, where each part is parametrized as a
NeRF. Note that we assume a fixed number of parts across all objects, namely
the generated objects cannot have a variable number of parts. To learn the
latent space of each NeRF, we follow \cite{Liu2021ICCV} and condition it
on two part-specific learnable latent codes: one for the shape and one for the texture,
$\bz_m^s, \bz_m^t$. These codes are obtained from a per-object
specific learnable embedding (see \secref{subsec:network_architecture}).
Disentangling the shape from the texture allows modifying one property without
affecting the other.

Moreover, as we are interested in editing specific parts of the object, we want
to be able to modify the pose, size, and appearance of each part independently.
This can be enforced by making sure that each NeRF receives geometric inputs in
its own local coordinate system. To this end, we augment each part with:
(i) an \emph{affine transformation} $\cT_m(\bx) = \bR_m (\bx + \bt_m)$ that maps
a 3D point $\bx$ to the local coordinate system of the $m$-th part, where
$\bt_m \in \real^3$ is the translation vector and $\bR_m \in SO(3)$ is the
rotation matrix and (ii) a \emph{scale} vector $\bs_m \in \real^3$,
representing its spatial extent. All are obtained from the per-part shape
code $\bz_m^s$.

\boldparagraph{Part Representation}%
Each part is represented as a continuous function that maps a 3D
point $\bx \in \real^3$, a viewing direction $\bd \in \mathbb{S}^2$, a shape
code $\bz^s \in \real^{L_s}$ and a texture code $\bz^t \in \real^{L_t}$ into a
color $\bc \in \mathbb{R}^3$ and an occupancy value $o \in [0, 1]$. Similar to
\cite{Oechsle2021ICCV}, we employ two separate networks: a \emph{color network}
$c_{\theta}$ and an \emph{occupancy network} $o_{\theta}$ to predict the
color and the occupancy value. Note that the occupancy value
$o$ is constrained to $[0,1]$ with a sigmoid. More formally, each NeRF
maps a 3D point $\bx$ along a viewing direction $\bd$ to a color $\bc$ and an
occupancy $o$ as:
\begin{align}
	c_\theta^m(\bx, \bd) &= c_\theta(\cT_m(\bx), \bR_m \bd, \bz_m^s, \bz_m^t) \label{eq:nerf_color_local} \\
	o_{\theta}^m(\bx)   &= o_\theta(\cT_m(\bx), \bz_m^s).                    \label{eq:nerf_occ_local}
\end{align}
Note that while we apply positional encoding on the inputs, we omit it from
\eqref{eq:nerf_color_local}$+$\eqref{eq:nerf_occ_local} to avoid notation
clutter.

To enforce that each part only captures continuous regions of the object,
we multiply its occupancy function with the occupancy function of an
axis-aligned 3D ellipsoid centered at the origin of the coordinate
system, with axis lengths given by the scale vector $\bs_m$. This results in the following
joint occupancy function for the $m$-th part
\begin{equation}
	h_\theta^m(\bx) = o_\theta^m(\bx) g_\theta^m(\bx),
    \label{eq:occupancy_joint}
\end{equation}
where $g_\theta^m(\bx) = g(T_m(\bx), \bs_m)$ denotes the occupancy function of
the $m$-th ellipsoid that is simply
\begin{equation}
    g(\bx, \bs) = \sigma \left(\beta \left(1 - \norm{\diag(\bs)^{-1} \bx}^2 \right) \right),
    \label{eq:ellipsoid}
\end{equation}
where $\sigma(\cdot)$ is the sigmoid function and $\beta$ controls the sharpness of
the transition. To estimate $g_\theta^m(\bx)$, we first transform
$\bx$ to the coordinate frame of the $m$-th part. This ensures that any
transformation of the part, also transforms its ellipsoid.

\begin{figure}
    \centering
    \includegraphics[width=1.0\columnwidth]{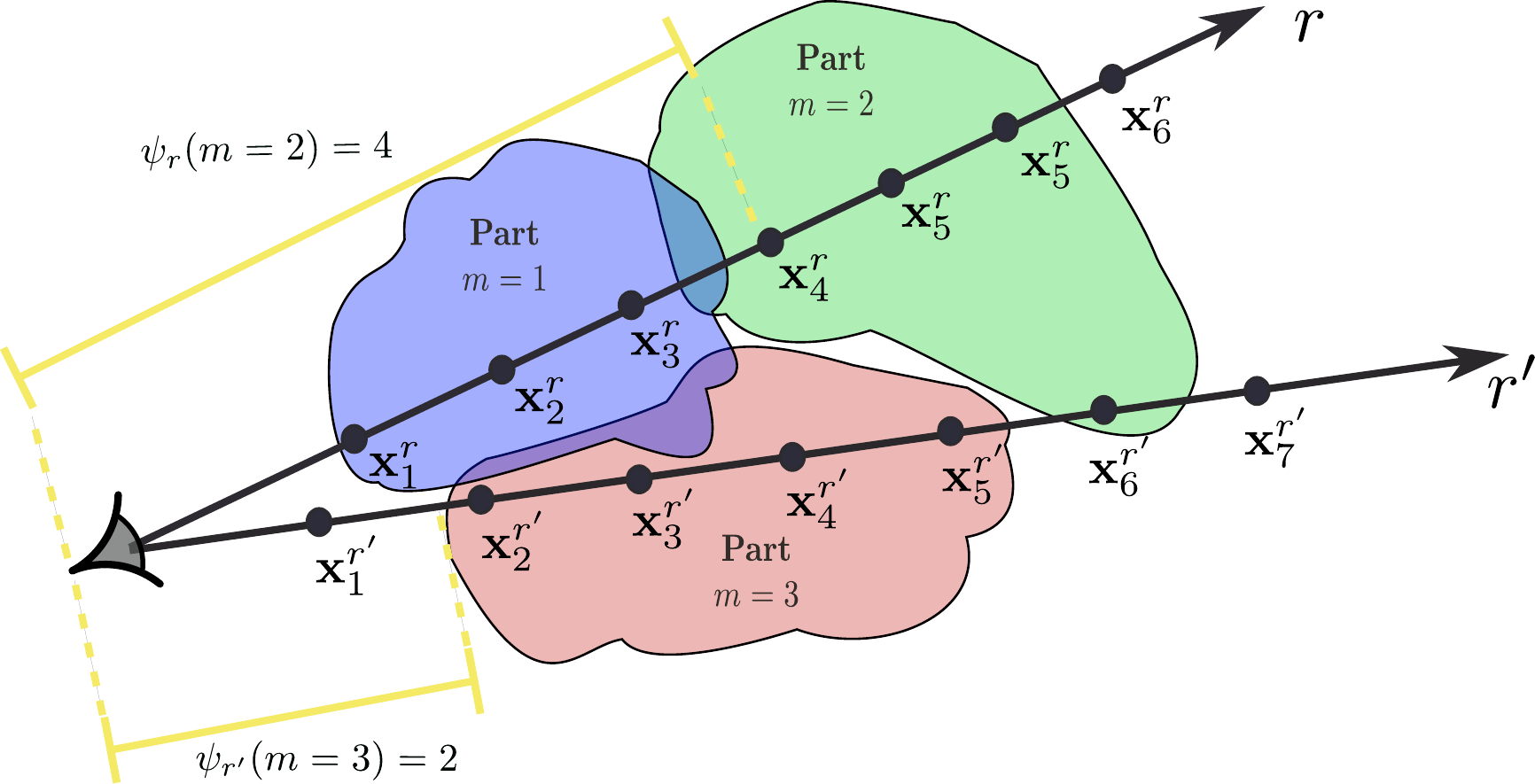}
    \caption{\small 
        {\bf{Ray-Part Association.}} We illustrate the hard assignment between
        rays and parts in a 2D example with 3 parts and two rays $r$ and $r'$.
        Since the association between rays and parts is determined based on the
        first part that a ray interests, the associations that emerge from
        \eqref{eq:per_part_rays} are $\cR_1=\{r\},
        \cR_2=\{\emptyset\}$ and $\cR_3=\{r'\}$.
        }
    \label{fig:ray_part_association}
    \vspace{-1.2em}
\end{figure}

\boldparagraph{Part Rendering}%
The rendering equation of the $m$-th part, given a set of $N$ sampled
points along ray $r$ now becomes
\begin{equation}
	\hat{C}_m(r) = \sum_{i=1}^N
		h_\theta^m(\bx_i^r) \prod_{j<i} \left(1 - h_\theta^m (\bx_i^r) \right)
		c_\theta^m(\bx_i^r, \bd^r).
    \label{eq:rendering_ours}
\end{equation}

\boldparagraph{Object Rendering}%
To ensure distinct, manipulable parts, we introduce a \emph{hard assignment}
between rays and parts, by associating a ray with the first part it intersects.
Given the ordered set of points $\cX_r$ sampled along ray
$r$, we define the index of the first point inside the part
that $r$ intersects as
\begin{equation}
	\psi_r(m) = \min \left\{ i \in \{1, \dots, N\} \, : \, h_\theta^m(\bx_i^r) \ge \tau \right\},
    \label{eq:ray_part_intersection}
\end{equation}
where $\tau$ is a threshold used to determine whether a point is \emph{inside}
the $m$-th part or not. This is illustrated in \figref{fig:ray_part_association}. We
can now define the set of rays $\cR_m$ associated with the $m$-th part, as the
set of rays that first intersect with it, namely:
\begin{equation}
    \cR_m = \Bigl\{r \in \cR \, : \,
    m = \argmin_{k \in \{0\dots M\}} \psi_r(k)\Bigr\}.
    \label{eq:per_part_rays}
    \vspace{-0.5em}
\end{equation}
Using the assignment of rays to parts and the per-part rendering equation
\eqref{eq:rendering_ours}, we can formulate the rendering equation for the
entire object, using $M$ NeRFs as follows
\begin{equation}
	\hat{C}(r) = \sum_{m=1}^M \ind_{r \in \cR_m} \hat{C}_m(r).
    \label{eq:rendering_equation_image}
\end{equation}
Namely, we use the $m$-th NeRF to render ray $r$ if it is assigned to the
$m$-th part. If a ray is not associated with any part its color is black. The hard ray-part assignment is an essential property
that ensures that editing one part does not alter the appearance
of other parts (see \figref{fig:ablation}).

\begin{figure}
    \centering
    \scriptsize
    \begin{minipage}[b]{0.19\linewidth}
		\centering
		No Editing
    \end{minipage}%
    \hfill%
    \begin{minipage}[b]{0.19\linewidth}
		\centering
		Rotation
    \end{minipage}%
    \hfill%
    \begin{minipage}[b]{0.19\linewidth}
		\centering
		Translation
    \end{minipage}%
    \hfill%
    \begin{minipage}[b]{0.19\linewidth}
		\centering
		Scaling
    \end{minipage}%
    \hfill%
    \begin{minipage}[b]{0.19\linewidth}
		\centering
		Color
    \end{minipage}%
    \hfill%
    \vspace{-1em}
    \vskip\baselineskip%
    \begin{subfigure}[b]{0.19\linewidth}
		\centering
		\includegraphics[width=\linewidth,trim=0 1cm 0 1cm,clip]{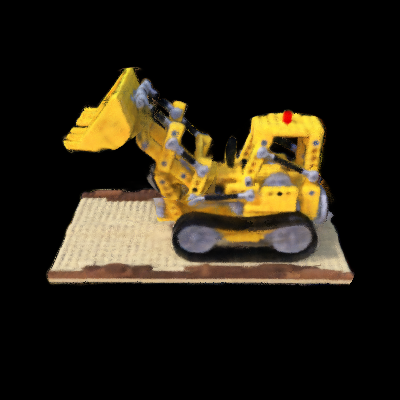}
    \end{subfigure}%
    \hfill%
    \begin{subfigure}[b]{0.19\linewidth}
		\centering
		\includegraphics[width=\linewidth,trim=0 1cm 0 1cm,clip]{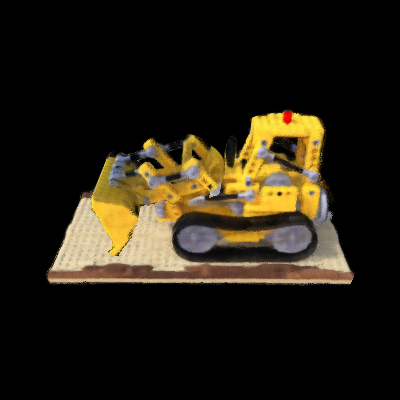}
    \end{subfigure}%
    \hfill%
    \begin{subfigure}[b]{0.19\linewidth}
		\centering
		\includegraphics[width=\linewidth,trim=0 1cm 0 1cm,clip]{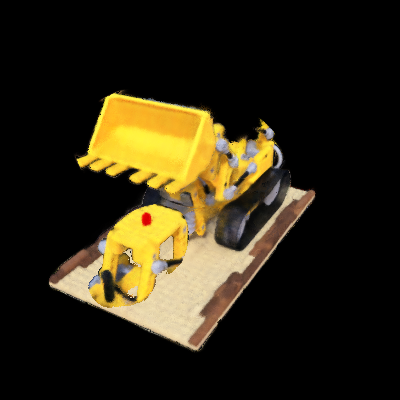}
    \end{subfigure}%
    \hfill%
    \begin{subfigure}[b]{0.19\linewidth}
		\centering
		\includegraphics[width=\linewidth,trim=0 1cm 0 1cm,clip]{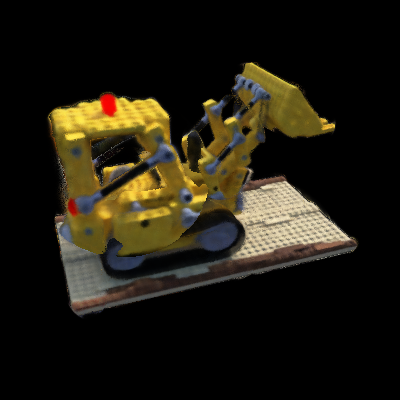}
    \end{subfigure}%
    \hfill%
    \begin{subfigure}[b]{0.19\linewidth}
		\centering
		\includegraphics[width=\linewidth,trim=0 1cm 0 1cm,clip]{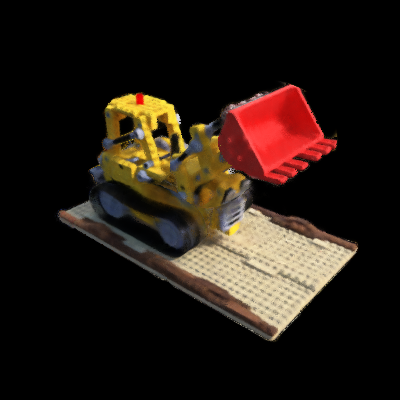}
    \end{subfigure}%
    \hfill%
    \vspace{-1.2em}
    \vskip\baselineskip%
    \begin{subfigure}[b]{0.19\linewidth}
		\centering
		\includegraphics[width=\linewidth,trim=0 1cm 0 2cm,clip]{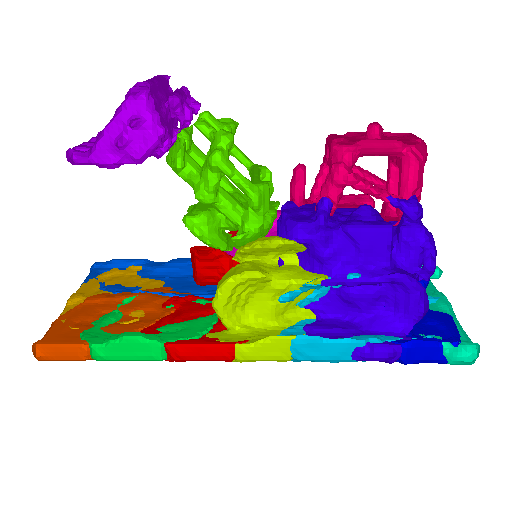}
    \end{subfigure}%
    \hfill%
    \begin{subfigure}[b]{0.19\linewidth}
		\centering
		\includegraphics[width=\linewidth,trim=0 1cm 0 2cm,clip]{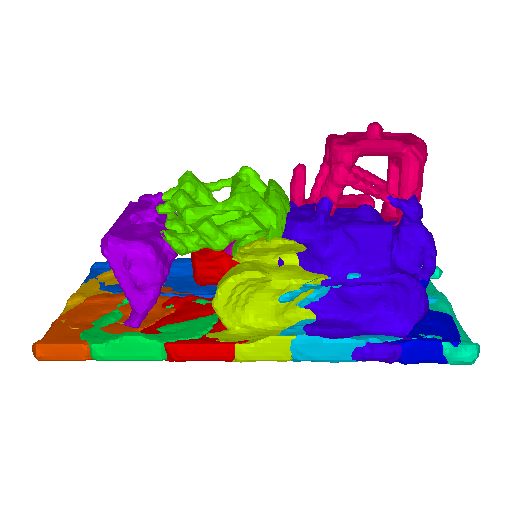}
    \end{subfigure}%
    \hfill%
    \begin{subfigure}[b]{0.19\linewidth}
		\centering
		\includegraphics[width=\linewidth,trim=0 1cm 0 2cm,clip]{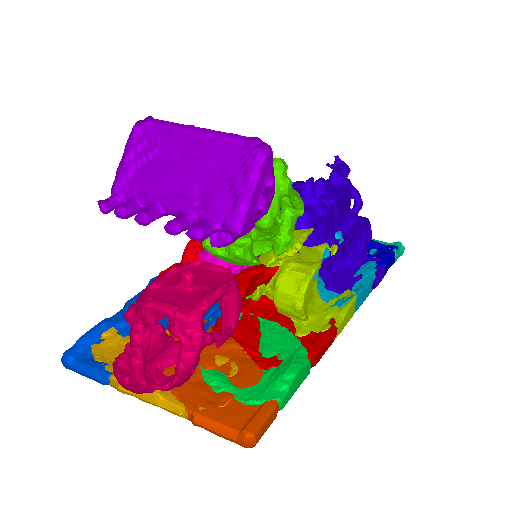}
    \end{subfigure}%
    \hfill%
    \begin{subfigure}[b]{0.19\linewidth}
		\centering
		\includegraphics[width=\linewidth,trim=0 1cm 0 2cm,clip]{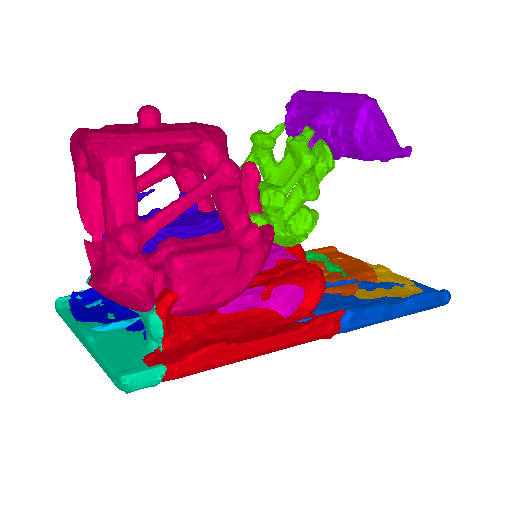}
    \end{subfigure}%
    \hfill%
    \begin{subfigure}[b]{0.19\linewidth}
		\centering
		\includegraphics[width=\linewidth,trim=0 1cm 0 2cm,clip]{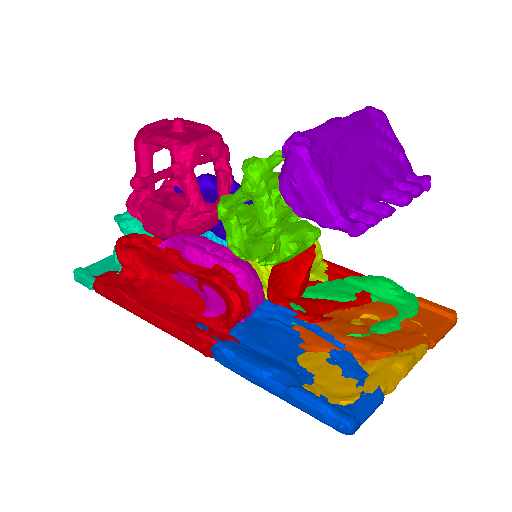}
    \end{subfigure}%
    \hfill%
    \vskip\baselineskip%
    \vspace{-2.6em}
    \caption{{\bf Scene-Specific Editing}.
    The top and bottom row show the rendered images and the part-based geometries respectively.
    The 1st column shows the tractor from a novel view, before editing.
    In the 2nd, we select the bucket and \emph{rotate} it downwards,
    whereas in the 3rd we \emph{translate} the cockpit to the floor.
    In the 4th, we perform \emph{isotropic scaling} of the cockpit and in
    the last we \emph{change the color} of the bucket to red.
    }
                    \label{fig:scene_specific_editing}
    \vspace{-1.4em}
\end{figure}

\subsection{Network Architecture}
\label{subsec:network_architecture}

We implement PartNeRF using an auto-decoder
\cite{Bojanowski2018ICML, Park2019CVPR}. The input to our model
is two learnable embeddings $\bz^s, \bz^t$, per training sample, that
represent its shape and texture. Our network consists of three components: (i)
the \emph{decomposition network} that maps $\bz^s$ and $\bz^t$ to $M$
latent codes that control the per-part shape and texture, (ii) the
\emph{structure network} that predicts the pose and the scale for each part $m$
and (iii) the \emph{neural rendering network} that renders a 2D image using the
$M$ locally defined NeRFs. The overall architecture is illustrated in \figref{fig:method_overview}.

\boldparagraph{Decomposition Network}%
The decomposition network takes the two object specific embeddings $\bz^s,
\bz^t \in \mathbb{R}^{L_d}$ and maps them to $M$ per-part embeddings of
the same dimensionality, $L_d$, using $M$ linear projections $f_\theta(\cdot)$.
Subsequently, these embeddings are mapped to part-specific latent codes using
two multi-head attention transformers, $\tau^s_\theta$ and $\tau^t_\theta$
without positional encoding \cite{Vaswani2017NIPS} as follows
\begin{align}
    \{\bz_m^s\}_{m=1}^M &= \tau^s_\theta(f_\theta(\bz^s)) \\
    \{\bz_m^t\}_{m=1}^M &= \tau^t_\theta(f_\theta(\bz^t)).
\end{align}
$\{\bz_m^s\}_{m=1}^M$ and $\{\bz_m^t\}_{m=1}^M$ are the latent
codes that control the shape and texture of each part respectively.

\boldparagraph{Structure Network}%
The structure network $s_\theta$ maps the shape latent code $\bz_m^s \in
\mathbb{R}^{L_d}$ to a translation vector $\bt_m \in \mathbb{R}^3$, a rotation
matrix $\bR_m \in SO(3)$ and a scale vector $\bs_m \in \mathbb{R}^3$, with
\begin{equation}
	\{\bt_m, \bR_m, \bs_m\} = s_\theta(\bz_m^s)
    \label{eq:structure_net}
\end{equation}
an MLP shared across parts.
Similar to \cite{Paschalidou2019CVPR}, we parametrize $\bR_m$
using quaternions~\cite{Hamilton1848}. As discussed in
\secref{subsec:part_nerf}, we determine the set of rays $\cR_m$ that are
assigned to each part $m$ from \eqref{eq:per_part_rays} and transform them
to its coordinate system.

\boldparagraph{Neural Rendering}%
Given the transformed points to the per-part coordinate system, we
predict colors and occupancies with
\eqref{eq:occupancy_joint}$+$\eqref{eq:nerf_color_local}. Next, we perform
volumetric rendering using \eq{rendering_ours} and render the
object using $M$ NeRFs with \eq{rendering_equation_image}.

\begin{figure}
    \centering
    \begin{subfigure}[b]{0.24\linewidth}
		\centering
        \scriptsize No Editing
    \end{subfigure}%
    \hfill%
    \begin{subfigure}[b]{0.24\linewidth}
		\centering
        \scriptsize Rotation
    \end{subfigure}%
    \hfill%
    \begin{subfigure}[b]{0.24\linewidth}
		\centering
        \scriptsize Translation
    \end{subfigure}%
    \hfill%
    \begin{subfigure}[b]{0.24\linewidth}
		\centering
        \scriptsize Scaling
    \end{subfigure}
    \vskip\baselineskip%
    \vspace{-1.2em}
    \begin{subfigure}[b]{0.24\linewidth}
		\centering
		\includegraphics[width=\linewidth, trim=0 1cm 0 1cm, clip]{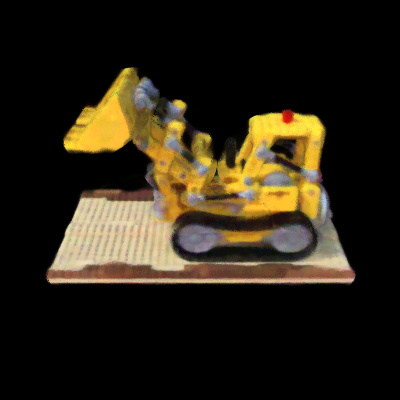}
    \end{subfigure}%
    \hfill%
    \begin{subfigure}[b]{0.24\linewidth}
		\centering
		\includegraphics[width=\linewidth, trim=0 1cm 0 1cm, clip]{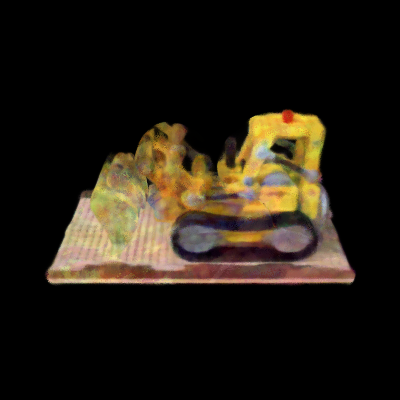}
    \end{subfigure}%
    \hfill%
    \begin{subfigure}[b]{0.24\linewidth}
		\centering
		\includegraphics[width=\linewidth, trim=0 1cm 0 1cm, clip]{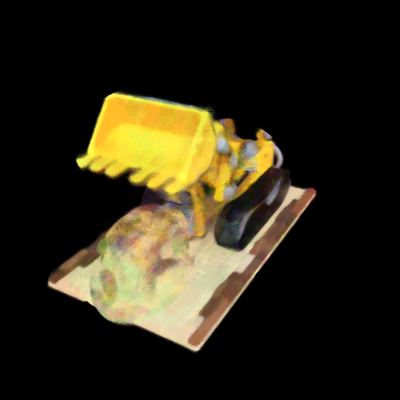}
    \end{subfigure}%
    \hfill%
    \begin{subfigure}[b]{0.24\linewidth}
		\centering
		\includegraphics[width=\linewidth, trim=0 1cm 0 1cm, clip]{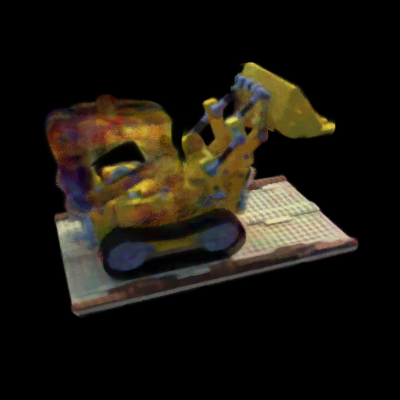}
    \end{subfigure}%
    \vskip\baselineskip%
    \vspace{-1.2em}
    \caption{{\bf Soft Ray-Part Assignment}. We demonstrate that enforcing a soft ray-part
    assignment results in parts that do not preserve their texture across transformations.}
    \label{fig:ablation}
    \vspace{-1.2em}
\end{figure}

\subsection{Training}
\label{subsec:training}

Our optimization objective $\cL$ is the sum over six terms combined with two regularizers on the shape and texture embeddings $\bz^s, \bz^t$, namely
\begin{align}
	\cL = \ &
		\cL_{rgb}(\cR) + \cL_{mask}(\cR) +
		\cL_{occ}(\cR) + \cL_{cov}(\cR) \ + \nonumber \\
		&
		\cL_{overlap}(\cR) + \cL_{control} +
		\norm{\bz^s}_2 + \norm{\bz^t}_2. \label{eq:optimization_objective}
\end{align}
As supervision, we use the observed RGB color
$C(r) \in \mathbb{R}^3$ and the object mask $I(r) \in \{0, 1\}$ for each ray $r
\in \cR$. We also associate $r$ with a binary label $\ell_r = I(r)$.
Namely, we label a ray $r$ as
\emph{inside}, if $\ell_r=1$ and \emph{outside} if $\ell_r=0$.

\boldparagraph{Reconstruction Loss}%
We measure the error between the observed $C(r)$
and the rendered $\hat{C}(r)$ color for the ray $r$ as
\begin{equation}
    \cL_{rgb}(\cR) = \sum_{r \in \cR}\Vert \hat{C}(r) - C(r)\Vert_{2}^{2}.
    \label{eq:rgb_loss}
\end{equation}

\boldparagraph{Mask Loss}%
Likewise, we measure the squared error between the observed $I(r)$ and the
rendered $\hat{I}(r)$ pixel value of the object mask for ray $r$ as
\begin{equation}
    \cL_{mask}(\cR) = \sum_{r \in \cR}\Vert \hat{I}(r) - I(r)\Vert_{2}^{2}.
    \label{eq:mask_loss}
\end{equation}
Note that $\hat{I}(r)$ can be derived from
\eqref{eq:rendering_equation_image}$+$\eqref{eq:rendering_ours} simply by
omitting the multiplication with the predicted color, namely
\begin{equation}
	\hat{I}(r) =
		\sum_{m=1}^M \ind_{r \in \cR_m}
		\sum_{i=1}^N h_\theta^m(\bx_i^r) \prod_{j<i} \left(1 - h_\theta^m (\bx_j^r) \right).
\label{eq:rendering_mask}
\end{equation}

\boldparagraph{Occupancy Loss}%
This loss makes sure that the generated parts do not occupy empty space. To this
end, we employ a binary cross-entropy classification loss $\cL_{ce}(\cdot ;
\cdot)$ between the predicted and the target labels for all rays
\begin{equation}
    \begin{aligned}
    \cL_{occ}(\cR) = \frac{1}{|\cR|}\sum_{r \in \cR} \Big(
        \cL_{ce}(\hat{\ell}_r, \ell_r)+ \cL_{ce}(\tilde{\ell}_r, \ell_r)
        \Big),
    \end{aligned}
    \label{eq:occupancy_loss}
\end{equation}
where $\hat{\ell}_r$ and $\tilde{\ell_r}$ are the predicted labels along ray $r$
based on the predicted occupancies and ellipsoid occupancies respectively.
Intuitively, we consider a ray $r$ to be inside the object if it is
inside at least one part. In turn, in order for $r$ to be inside a part it
suffices if at least one point along the ray $\cX_r$ is inside this part. This
can be expressed as
\begin{align}
    \hat{\ell}_r &= \max_{m \in \{1 \dots M\}}\max_{\bx_i^r \in \cX_r} h_\theta^m(\bx_i^r)
    \label{eq:predicted_ray_label_occnet} \\
    \tilde{\ell}_r &= \max_{m \in \{1 \dots M\}}\max_{\bx_i^r \in \cX_r} g_\theta^m(\bx_i^r).
    \label{eq:predicted_ray_label_ellipsoid}
\end{align}

\begin{figure}
    \centering
    \scriptsize
    \begin{minipage}[b]{0.03\linewidth}
		\phantom{.}
    \end{minipage}
    \begin{subfigure}[b]{0.31\linewidth}
		\centering Motorbike
    \end{subfigure}%
    \hfill%
    \begin{subfigure}[b]{0.31\linewidth}
		\centering Car
    \end{subfigure}%
    \hfill%
    \begin{subfigure}[b]{0.31\linewidth}
		\centering Chair
    \end{subfigure}
    \vskip\baselineskip%
    \vspace{-1.0em}
    \begin{minipage}[b]{0.03\linewidth}
        \rotatebox{90}{~ {Pi-GAN}}
    \end{minipage}
    \begin{subfigure}[b]{0.155\linewidth}
		\centering
		\includegraphics[width=\linewidth]{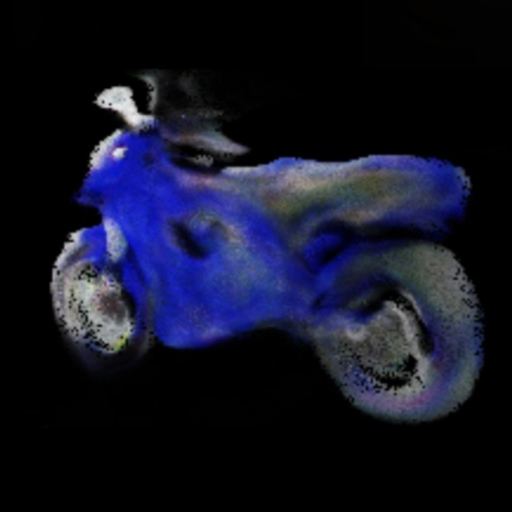}
    \end{subfigure}%
    \hfill%
    \begin{subfigure}[b]{0.155\linewidth}
		\centering
		\includegraphics[width=\linewidth]{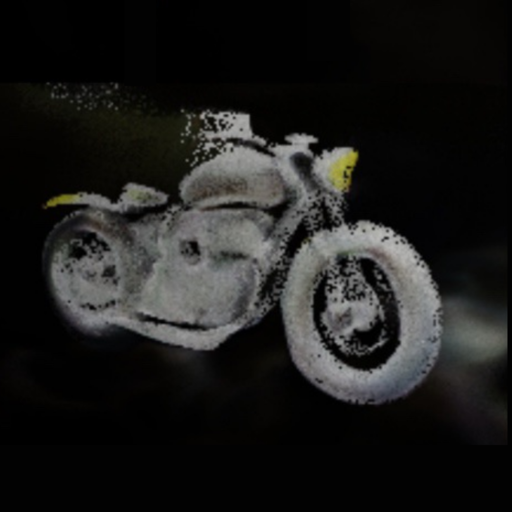}
    \end{subfigure}%
    \hfill%
    \begin{subfigure}[b]{0.155\linewidth}
		\centering
		\includegraphics[width=\linewidth]{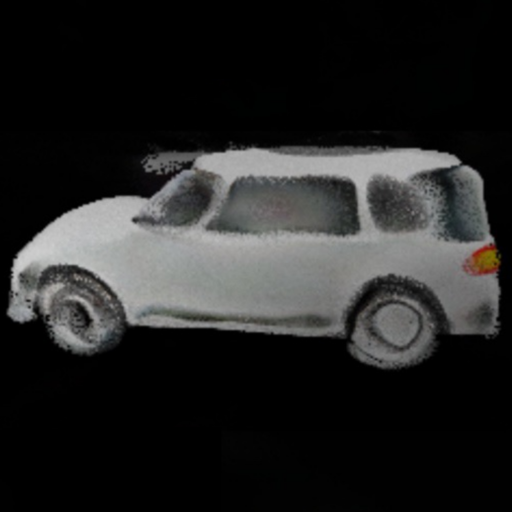}
    \end{subfigure}%
    \hfill%
    \begin{subfigure}[b]{0.155\linewidth}
		\centering
		\includegraphics[width=\linewidth]{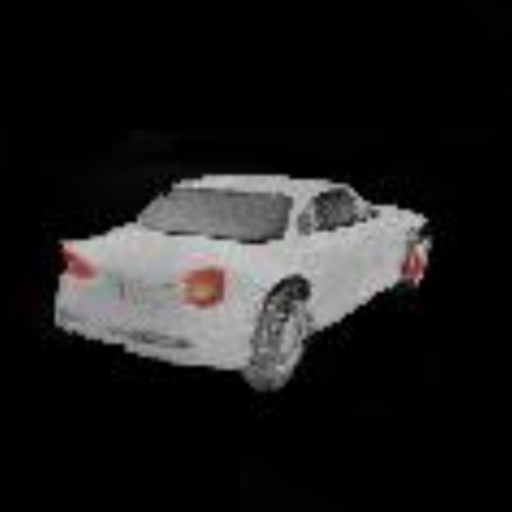}
    \end{subfigure}%
    \hfill%
    \begin{subfigure}[b]{0.155\linewidth}
		\centering
		\includegraphics[width=\linewidth]{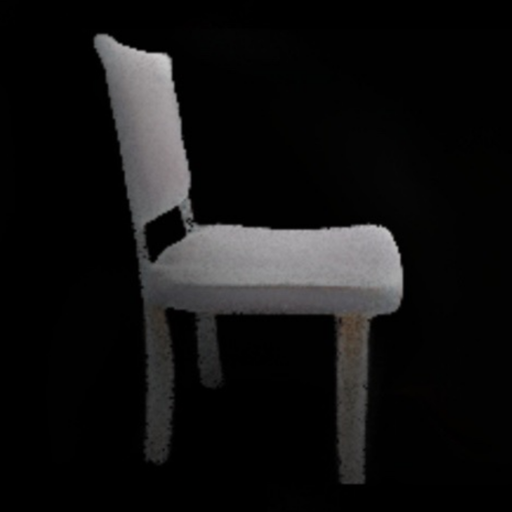}
    \end{subfigure}%
    \hfill%
    \begin{subfigure}[b]{0.155\linewidth}
		\centering
		\includegraphics[width=\linewidth]{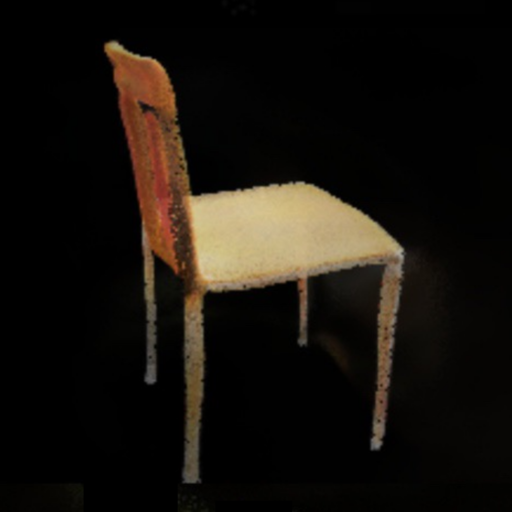}
    \end{subfigure}%
    \vskip\baselineskip%
    \vspace{-1.1em}
    \begin{minipage}[b]{0.03\linewidth}
        \rotatebox{90}{~ {GRAF}}
    \end{minipage}
    \begin{subfigure}[b]{0.155\linewidth}
		\centering
		\includegraphics[width=\linewidth]{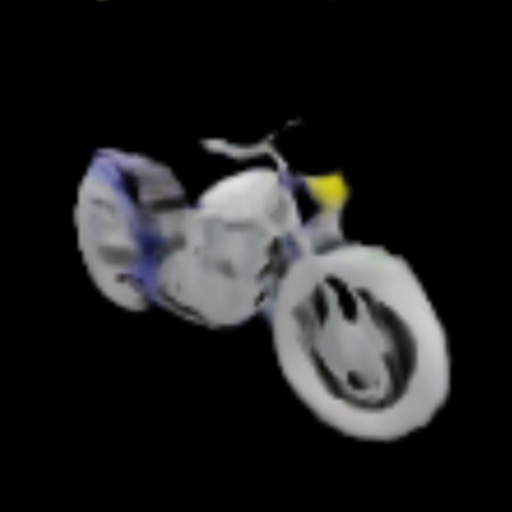}
    \end{subfigure}%
    \hfill%
    \begin{subfigure}[b]{0.155\linewidth}
		\centering
		\includegraphics[width=\linewidth]{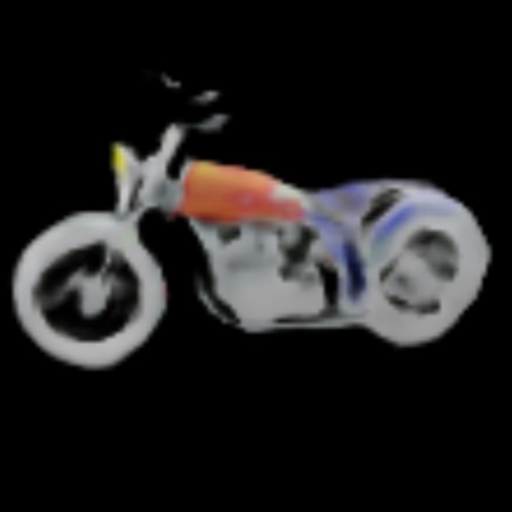}
    \end{subfigure}%
    \hfill%
    \begin{subfigure}[b]{0.155\linewidth}
		\centering
		\includegraphics[width=\linewidth]{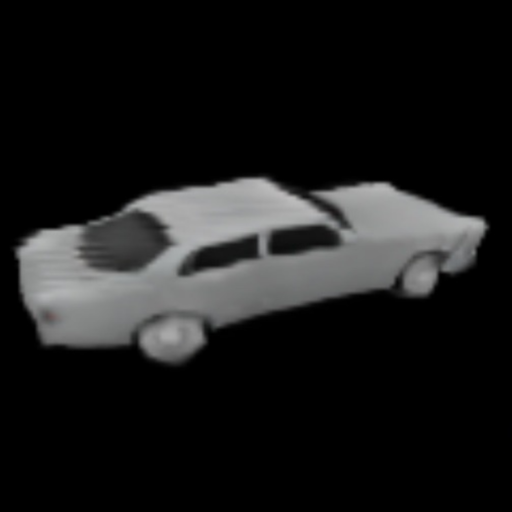}
    \end{subfigure}%
    \hfill%
    \begin{subfigure}[b]{0.155\linewidth}
		\centering
		\includegraphics[width=\linewidth]{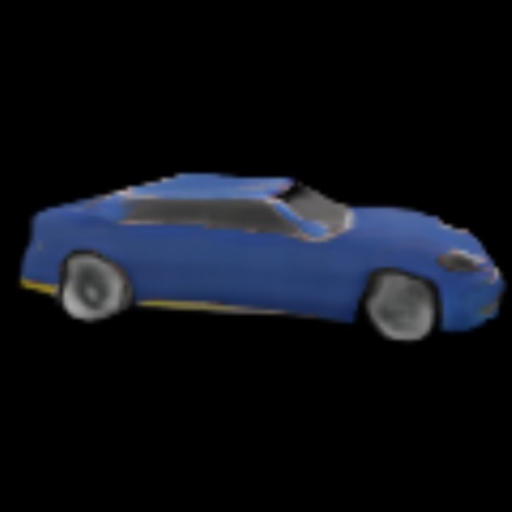}
    \end{subfigure}%
    \hfill%
    \begin{subfigure}[b]{0.155\linewidth}
		\centering
		\includegraphics[width=\linewidth]{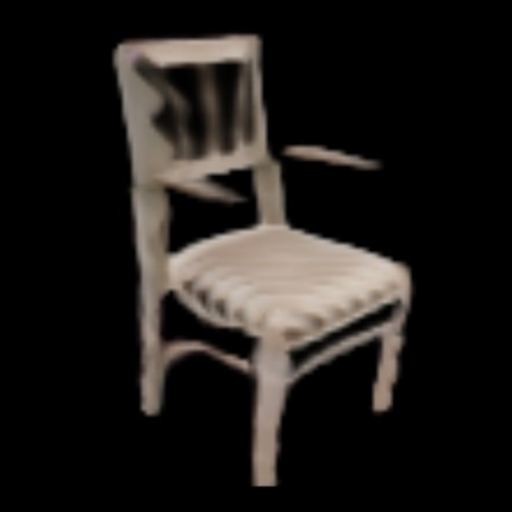}
    \end{subfigure}%
    \hfill%
    \begin{subfigure}[b]{0.155\linewidth}
		\centering
		\includegraphics[width=\linewidth]{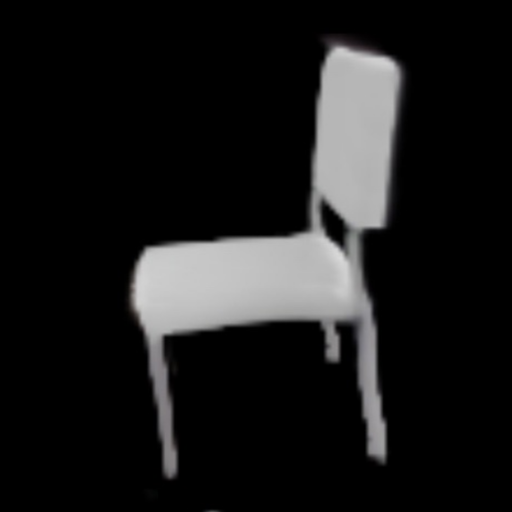}
    \end{subfigure}%
    \vskip\baselineskip%
    \vspace{-1.1em}
    \begin{minipage}[b]{0.03\linewidth}
        \rotatebox{90}{~ {EG3D}}
    \end{minipage}
    \begin{subfigure}[b]{0.155\linewidth}
		\centering
		\includegraphics[width=\linewidth]{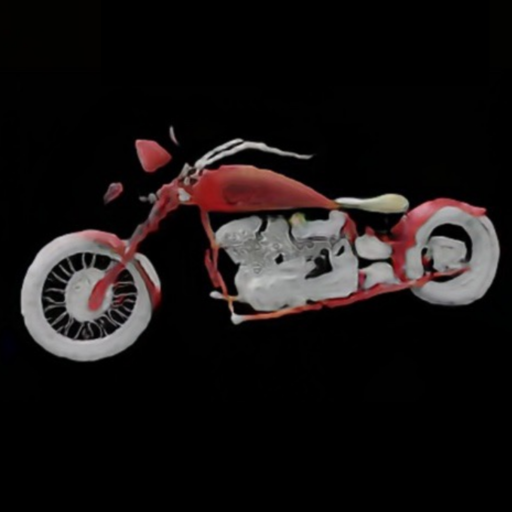}
    \end{subfigure}%
    \hfill%
    \begin{subfigure}[b]{0.155\linewidth}
		\centering
		\includegraphics[width=\linewidth]{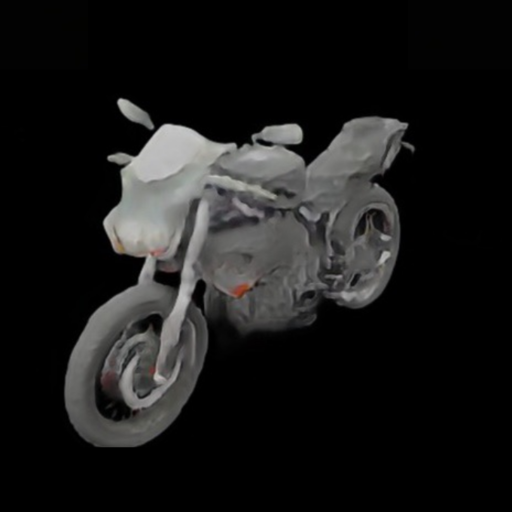}
    \end{subfigure}%
    \hfill%
    \begin{subfigure}[b]{0.155\linewidth}
		\centering
		\includegraphics[width=\linewidth]{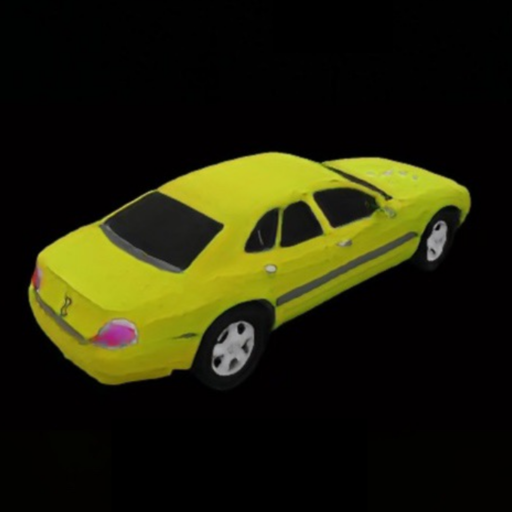}
    \end{subfigure}%
    \hfill%
    \begin{subfigure}[b]{0.155\linewidth}
		\centering
		\includegraphics[width=\linewidth]{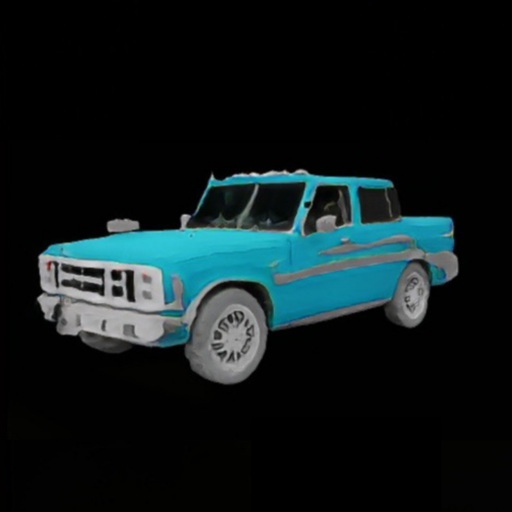}
    \end{subfigure}%
    \hfill%
    \begin{subfigure}[b]{0.155\linewidth}
		\centering
		\includegraphics[width=\linewidth]{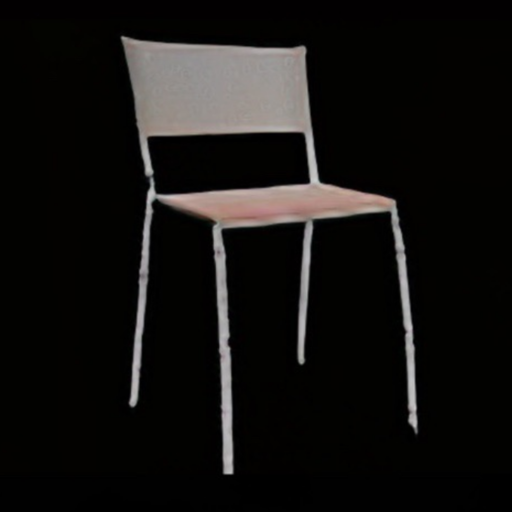}
    \end{subfigure}%
    \hfill%
    \begin{subfigure}[b]{0.155\linewidth}
		\centering
		\includegraphics[width=\linewidth]{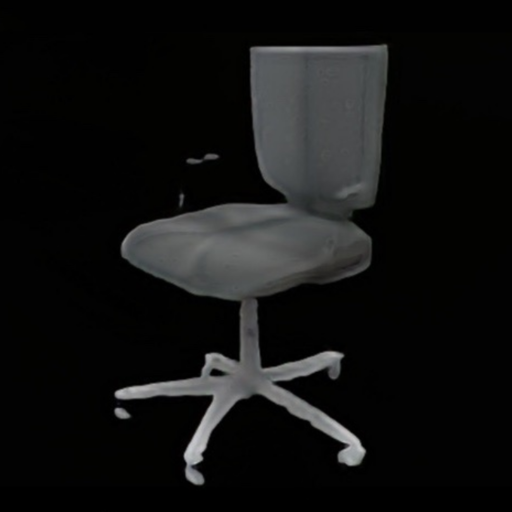}
    \end{subfigure}%
    \vskip\baselineskip%
    \vspace{-1.1em}
    \begin{minipage}[b]{0.03\linewidth}
        \rotatebox{90}{~ {GET3D}}
    \end{minipage}
    \begin{subfigure}[b]{0.155\linewidth}
		\centering
		\includegraphics[width=\linewidth]{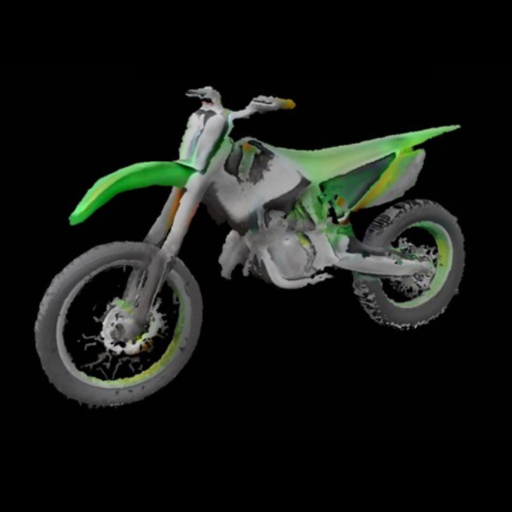}
    \end{subfigure}%
    \hfill%
    \begin{subfigure}[b]{0.155\linewidth}
		\centering
		\includegraphics[width=\linewidth]{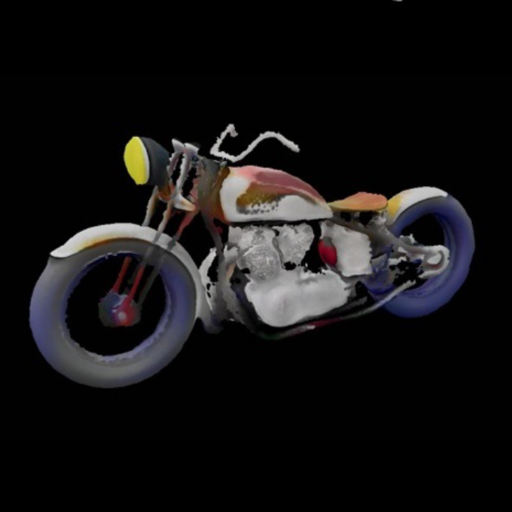}
    \end{subfigure}%
    \hfill%
    \begin{subfigure}[b]{0.155\linewidth}
		\centering
		\includegraphics[width=\linewidth]{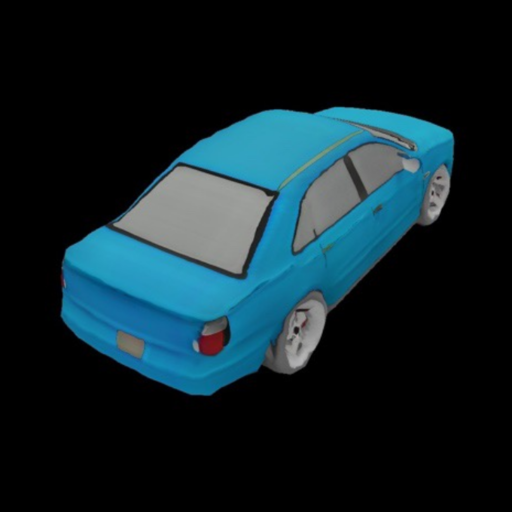}
    \end{subfigure}%
    \hfill%
    \begin{subfigure}[b]{0.155\linewidth}
		\centering
		\includegraphics[width=\linewidth]{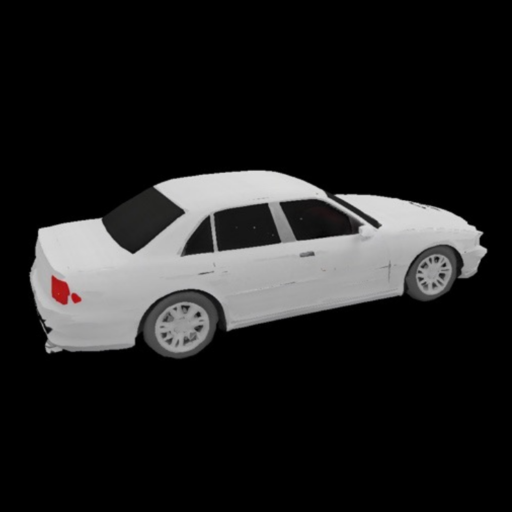}
    \end{subfigure}%
    \hfill%
    \begin{subfigure}[b]{0.155\linewidth}
		\centering
		\includegraphics[width=\linewidth]{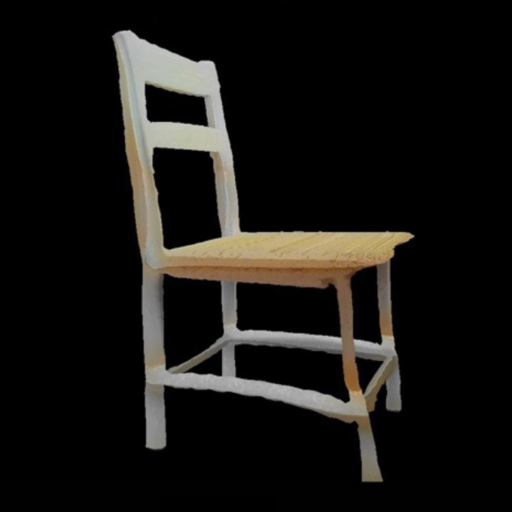}
    \end{subfigure}%
    \hfill%
    \begin{subfigure}[b]{0.155\linewidth}
		\centering
		\includegraphics[width=\linewidth]{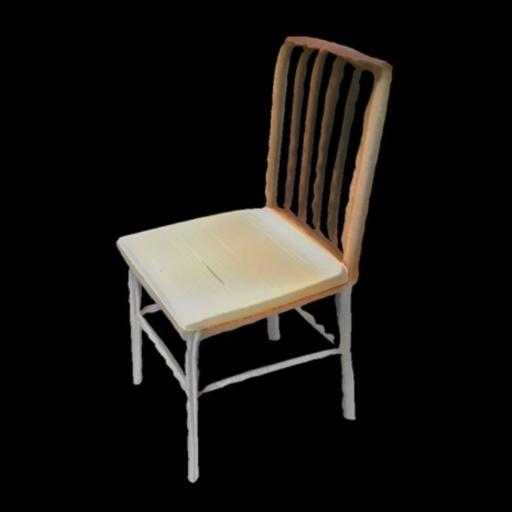}
    \end{subfigure}%
    \vskip\baselineskip%
    \vspace{-1.1em}
    \begin{minipage}[b]{0.03\linewidth}
        \rotatebox{90}{~ {Ours}}
    \end{minipage}
    \begin{subfigure}[b]{0.155\linewidth}
		\centering
		\includegraphics[trim=20 30 40 30,clip,width=\linewidth]{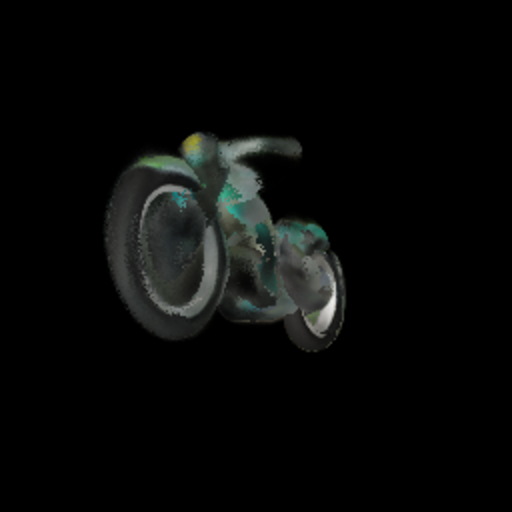}
    \end{subfigure}%
    \hfill%
    \begin{subfigure}[b]{0.155\linewidth}
		\centering
		\includegraphics[width=\linewidth]{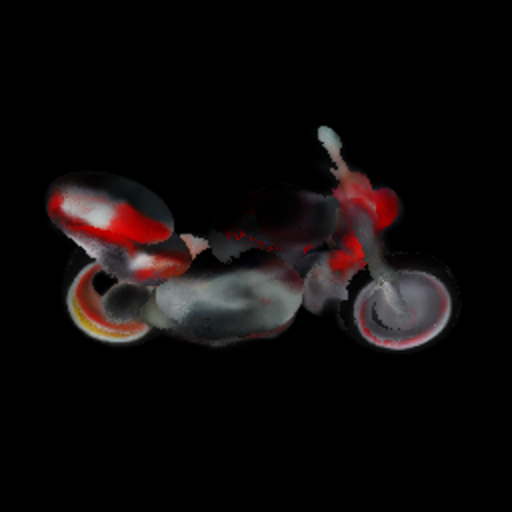}
    \end{subfigure}%
    \hfill%
    \begin{subfigure}[b]{0.155\linewidth}
		\centering
		\includegraphics[width=\linewidth]{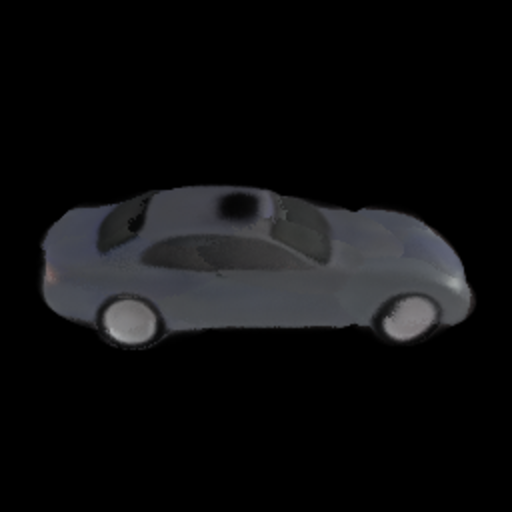}
    \end{subfigure}%
    \hfill%
    \begin{subfigure}[b]{0.155\linewidth}
		\centering
		\includegraphics[width=\linewidth]{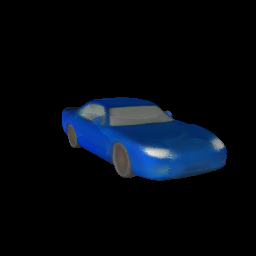}
    \end{subfigure}%
    \hfill%
    \begin{subfigure}[b]{0.155\linewidth}
		\centering
		\includegraphics[width=\linewidth]{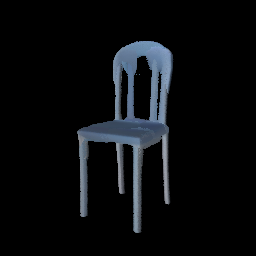}
    \end{subfigure}%
    \hfill%
    \begin{subfigure}[b]{0.155\linewidth}
		\centering
		\includegraphics[width=\linewidth]{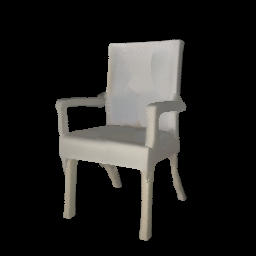}
    \end{subfigure}%
    \vskip\baselineskip%
    \vspace{-1.2em}
    \caption{{\bf Shape Generation}. We compare our model with 3D
    generative models that are part agnostic but can generate textured meshes. Here we show two
    randomly generated samples per category.}
    \label{fig:shapenet_qualitative_comparison}
    \vspace{-1.2em}
\end{figure}

\begin{table}
\resizebox{\columnwidth}{!}{%
    \centering
    \begin{tabular}{lc|ccc|ccc}
        \toprule
        \multirow{2}{*}{Method} & \multirow{2}{*}{Rendering} &
        \multicolumn{3}{c}{MMD-CD ($\downarrow$)} & \multicolumn{3}{c}{COV-CD ($ \%, \uparrow$)} \\ \cmidrule{3-8}
        & & Motorbike & Car & Chair & Motorbike & Car & Chair\\
        \midrule
        GET3D & Differentiable & 1.72 & 0.71 & 3.72 & 67.12 & 58.39 & 69.91 \\
        \midrule
        Pi-GAN & Volumetric & 21.80 & 25.54 & 6.65 & 6.85 & 0.55 & 39.65 \\
        GRAF & Volumetric & 2.40 & 10.63 & 6.80 & 50.68 & 1.57 & 39.28 \\
        EG3D & Volumetric & 2.21 & \bf{0.72} & 4.72 & 34.25 & \bf{49.52} & 50.14 \\
        \midrule
        Ours & Volumetric & \bf{1.68} & 1.74 & \bf{4.42} & \bf{56.06} & 21.10 & \bf{67.20} \\
        \bottomrule
    \end{tabular}
    }
    \vspace{-0.2em}
    \caption{{\bf Comparison with 3D Generative Models.}
    We measure MMD-CD ($\downarrow$) and COV-CD ($\uparrow$). Note that
    none of these baselines considers parts nor
    allows any part-level shape editing.}
    \label{tab:shape_generation_quantitative}
    \vspace{-1.2em}
\end{table}

\boldparagraph{Coverage Loss}%
This loss ensures that the parts cover the object, thus
preventing degenerate arrangements with small parts or parts outside
the object. To implement this, we encourage parts to contain
at least $k$ \emph{inside} rays. This can be expressed as a binary
cross-entropy loss between the predicted per-part and target labels for all
rays in $\cR_m^k$,
\begin{equation}
	\cL_{cov}(\cR) =
		\frac{1}{M}\sum_{m=1}^{M} \sum_{r \in \cR_{m}^{k}}
		\cL_{ce}(\hat{\ell}_r^m, \ell_r),
\label{eq:coverage_loss}
\end{equation}
where $\displaystyle \hat{\ell}_r^m = \max_{\bx_i^r \in \cX_r}
h_\theta^m(\bx_i^r)$ the predicted label for ray $r$ \wrt part $m$ and
$\cR_m^k$ the set of the $k$ inside rays with the greatest predicted occupancy
values for this part.

\boldparagraph{Overlapping Loss}%
To encourage the generated parts to capture different regions of the object, we
penalize rays that are inside of more than $\lambda$ parts as follows
\begin{equation}
    \cL_{overlap}(\cR) = \frac{1}{|\cR|}\sum_{r \in \cR}
    \max\Big(0, \sum_{m=1}^{M}\hat{\ell}_r^m - \lambda \Big).
    \label{eq:overlap_loss}
\end{equation}

\boldparagraph{Control Loss}%
To ensure uniform control across the shape, we want parts with comparable
volumes. We implement this loss on the volumes $\cV(\cdot)$ 
of the ellipsoids, as follows:
\begin{equation}
    \cL_{control} = \frac{1}{M(M-1)} \sum_{i=1}^{M} \sum_{j=1}^{i} | \cV(\bs_i) - \cV(\bs_j) |.
    \label{eq:control_loss}
\end{equation}

\begin{figure}
    \centering
    \begin{subfigure}[b]{0.5\linewidth}
		\centering
        \scriptsize Tables
    \end{subfigure}%
    \hfill%
    \begin{subfigure}[b]{0.5\linewidth}
		\centering
        \scriptsize Airplanes
    \end{subfigure}
    \vskip\baselineskip%
    \vspace{-1.2em}
                \begin{subfigure}[b]{0.15\linewidth}
		\centering
		\includegraphics[width=\linewidth]{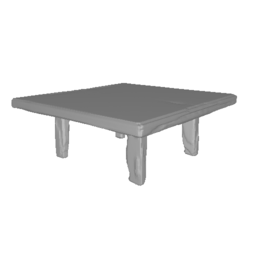}
    \end{subfigure}%
    \hfill%
    \begin{subfigure}[b]{0.15\linewidth}
		\centering
		\includegraphics[width=\linewidth]{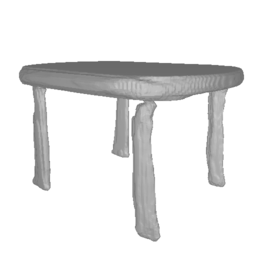}
    \end{subfigure}%
    \hfill%
    \begin{subfigure}[b]{0.15\linewidth}
		\centering
		\includegraphics[width=\linewidth]{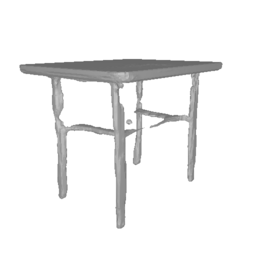}
    \end{subfigure}%
    \hfill%
    \begin{subfigure}[b]{0.15\linewidth}
		\centering
		\includegraphics[width=\linewidth]{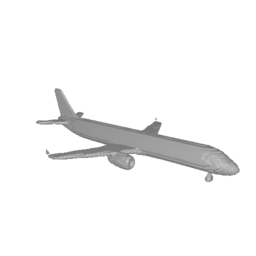}
    \end{subfigure}%
    \hfill%
    \begin{subfigure}[b]{0.15\linewidth}
		\centering
		\includegraphics[width=\linewidth]{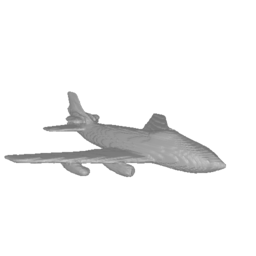}
    \end{subfigure}%
    \hfill%
    \begin{subfigure}[b]{0.15\linewidth}
		\centering
		\includegraphics[width=\linewidth]{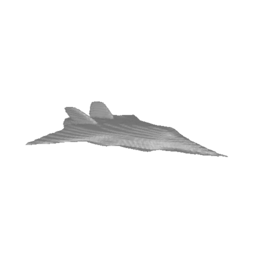}
    \end{subfigure}%
    \vskip\baselineskip%
    \vspace{-2.2em}
    \begin{minipage}[b]{\linewidth}
		\centering
		\scriptsize{SPAGHETTI \cite{Hertz2022SIGGRAPH}}
    \end{minipage}%
                \vskip\baselineskip%
    \vspace{-1.2em}
    \begin{subfigure}[b]{0.15\linewidth}
		\centering
		\includegraphics[width=\linewidth]{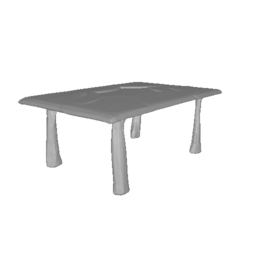}
    \end{subfigure}%
    \hfill%
    \begin{subfigure}[b]{0.15\linewidth}
		\centering
		\includegraphics[width=\linewidth]{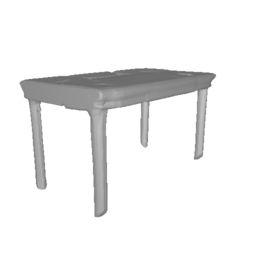}
    \end{subfigure}%
    \hfill%
    \begin{subfigure}[b]{0.15\linewidth}
		\centering
		\includegraphics[width=\linewidth]{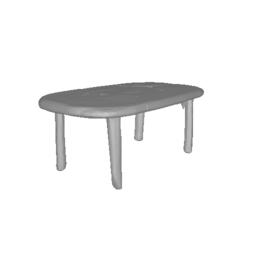}
    \end{subfigure}%
    \hfill%
    \begin{subfigure}[b]{0.15\linewidth}
		\centering
		\includegraphics[width=\linewidth]{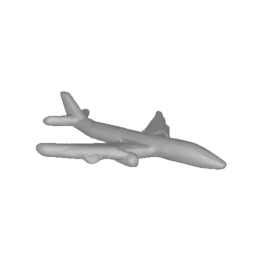}
    \end{subfigure}%
    \hfill%
    \begin{subfigure}[b]{0.15\linewidth}
		\centering
		\includegraphics[width=\linewidth]{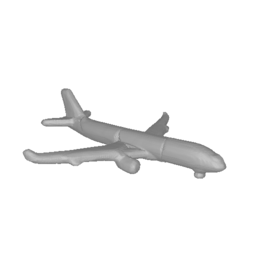}
    \end{subfigure}%
    \hfill%
    \begin{subfigure}[b]{0.15\linewidth}
		\centering
		\includegraphics[width=\linewidth]{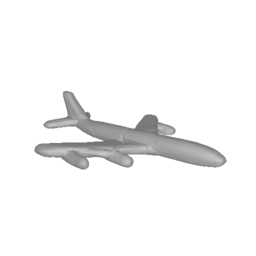}
    \end{subfigure}%
    \vskip\baselineskip%
    \vspace{-2.6em}
    \begin{minipage}[b]{\linewidth}
		\centering
		\scriptsize{Ours}
    \end{minipage}%
    \vskip\baselineskip%
    \vspace{-1.3em}
                \begin{subfigure}[b]{0.15\linewidth}
		\centering
		\includegraphics[width=\linewidth]{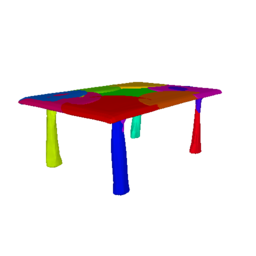}
    \end{subfigure}%
    \hfill%
    \begin{subfigure}[b]{0.15\linewidth}
		\centering
		\includegraphics[width=\linewidth]{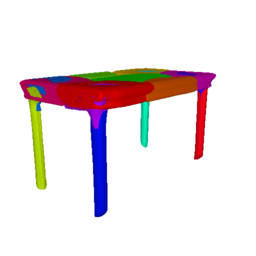}
    \end{subfigure}%
    \hfill%
    \begin{subfigure}[b]{0.15\linewidth}
		\centering
		\includegraphics[width=\linewidth]{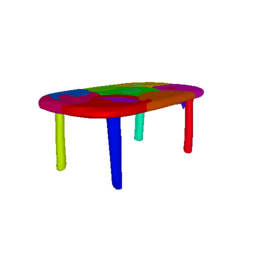}
    \end{subfigure}%
    \hfill%
    \begin{subfigure}[b]{0.15\linewidth}
		\centering
		\includegraphics[width=\linewidth]{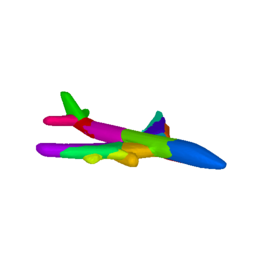}
    \end{subfigure}%
    \hfill%
    \begin{subfigure}[b]{0.15\linewidth}
		\centering
		\includegraphics[width=\linewidth]{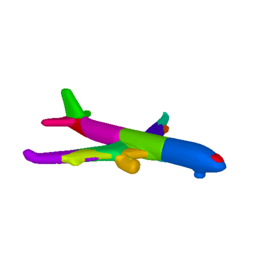}
    \end{subfigure}%
    \hfill%
    \begin{subfigure}[b]{0.15\linewidth}
		\centering
		\includegraphics[width=\linewidth]{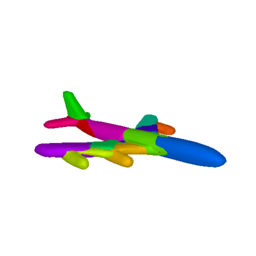}
    \end{subfigure}%
    \vskip\baselineskip%
    \vspace{-2.4em}
    \begin{minipage}[b]{\linewidth}
		\centering
		\scriptsize{Ours with Parts}
    \end{minipage}%
    \vskip\baselineskip%
    \vspace{-1.2em}
    \caption{{\bf Shape Generation}. We compare our model with
    \cite{Hertz2022SIGGRAPH} and show three randomly generated samples per
    category.}
    \label{fig:shapenet_qualitative_comparison_parts}
    \vspace{-0.4em}
\end{figure}

\begin{table}
\resizebox{\columnwidth}{!}{%
    \centering
    \begin{tabular}{lc|cc|cc}
        \toprule
        \multirow{2}{*}{Method} & \multirow{2}{*}{Supervision} &
        \multicolumn{2}{c}{MMD-CD ($\downarrow$)} & \multicolumn{2}{c}{COV-CD ($ \%, \uparrow$)} \\ \cmidrule{3-6}
        & & Airplane & Table & Airplane & Table \\
        \midrule
        DualSDF & 3D Shapes& 4.20 & 12.30 & 25.00 & 36.30 \\
        SPAGHETTI & 3D Shapes & 2.40 & 5.90 & 35.00 & \textbf{47.80} \\
        \midrule
        Ours & Multi-view &\textbf{1.37} & \textbf{4.48} & \textbf{37.90} & 40.60 \\
        \bottomrule
    \end{tabular}
    }
    \vspace{-0.2em}
    \caption{{\bf Comparison with Part-based Generative Models.} We measure
    MMD-CD ($\downarrow$) and the COV-CD ($\uparrow$). Unlike our model, both
    \cite{Hao2020CVPR, Hertz2022SIGGRAPH} require 3D supervision during
    training.}
    \label{tab:shape_generation_parts_quantitative}
    \vspace{-1.2em}
\end{table}

\section{Experimental Evaluation}
\label{sec:results}

We provide an extensive evaluation of PartNeRF comparing it
to relevant baselines in terms of the realism and diversity of the generated shapes.
We also showcase several editing operations of our model on multiple object categories.
Additional results, ablations and implementation details are provided in the
supplementary.

\boldparagraph{Datasets}%
We use five ShapeNet~\cite{Chang2015ARXIV} categories:
\emph{Motorbike, Chair, Table, Airplane and Car}.
To render our training data, we randomly sample camera poses from the upper
hemisphere of each shape and render images at $256^2$ resolution as in
\cite{Gao2022NEURIPS}.  For the
\emph{Car, Table, Airplane and Chair}, we use $24$ random views,
while for Motorbike we use $100$. To ensure
fair comparison with our baselines, we use the train-test splits of
\cite{Gao2022NEURIPS} for the \emph{Motorbike, Car, Chair} object
categories and the train-test splits of \cite{Hertz2022SIGGRAPH} for the
\emph{Airplane, Table} category. In all our experiments, we perform category
specific training. Finally, we showcase the scene-specific editing capabilities
of our model on the Lego tractor~\cite{Mildenhall2020ECCV}.
In all our experiments, we set $M=16$.

\boldparagraph{Baselines}%
We compare our model with several NeRF-based models:
GRAF~\cite{Schwarz2020NEURIPS}, Pi-GAN~\cite{Chan2021CVPR},
EG3D~\cite{Chan2022CVPR} and the concurrent GET3D~\cite{Gao2022NEURIPS} that
relies on differentiable rendering. Unlike our model, none of the
above considers parts.
We also compare with the part-based DualSDF~\cite{Hao2020CVPR} and
SPAGHETTI~\cite{Hertz2022SIGGRAPH} that require 3D supervision.

\boldparagraph{Metrics}%
We report the Coverage (COV) and the Minimum
Matching Distance (MMD) \cite{Achlioptas2018ICML} using Chamfer-$L_2$ distance.
MMD measures how likely it is that a generated shape looks
like a test shape. COV measures how many shape variations
are covered by the generated shapes.

\subsection{Scene-Specific Shape Editing}
We train PartNeRF on the tractor scene \cite{Mildenhall2020ECCV},
using all
$200$ training views. The results are shown in \figref{fig:scene_specific_editing}.
Initially, we select the part that
corresponds to the bucket of the tractor and apply a rotation,
such that the bucket is facing downwards.
Similarly, we select the cockpit and move it to a new location on the
floor, by adding a displacement to the part's translation vector.
During both transformations the rest of the
shape (geometry, parts and textures) does not alter. We then
apply a non-rigid transformation on the cockpit, scaling it uniformly across
all axes, by multiplying the part's rotation
with an isometric scale matrix. Again, while we change
the shape of one part its texture as well as the texture and shape of the
others does not change. Finally, we alter the color of the bucket by
explicitly setting the predicted color of its associated NeRF to red.

\begin{figure}
    \centering
    \begin{subfigure}[b]{0.15\linewidth}
		\centering
		\includegraphics[width=\linewidth]{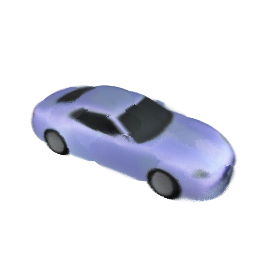}
    \end{subfigure}%
    \hfill%
    \begin{subfigure}[b]{0.15\linewidth}
		\centering
		\includegraphics[width=\linewidth]{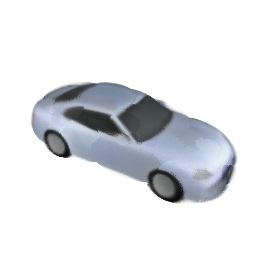}
    \end{subfigure}%
    \hfill%
    \begin{subfigure}[b]{0.15\linewidth}
		\centering
		\includegraphics[width=\linewidth]{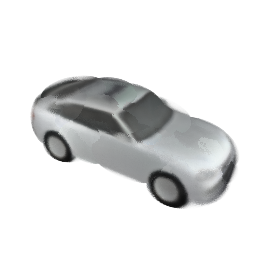}
    \end{subfigure}%
    \hfill%
    \begin{subfigure}[b]{0.15\linewidth}
		\centering
		\includegraphics[width=\linewidth]{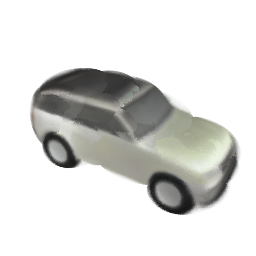}
    \end{subfigure}%
    \hfill%
    \begin{subfigure}[b]{0.15\linewidth}
		\centering
		\includegraphics[width=\linewidth]{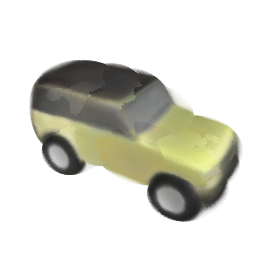}
    \end{subfigure}%
    \hfill%
    \begin{subfigure}[b]{0.15\linewidth}
		\centering
		\includegraphics[width=\linewidth]{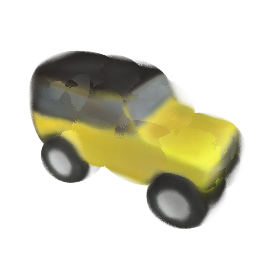}
    \end{subfigure}%
    \vskip\baselineskip%
    \vspace{-2.2em}
    \begin{subfigure}[b]{0.15\linewidth}
		\centering
		\includegraphics[trim=100 0 100 0,clip,width=\linewidth]{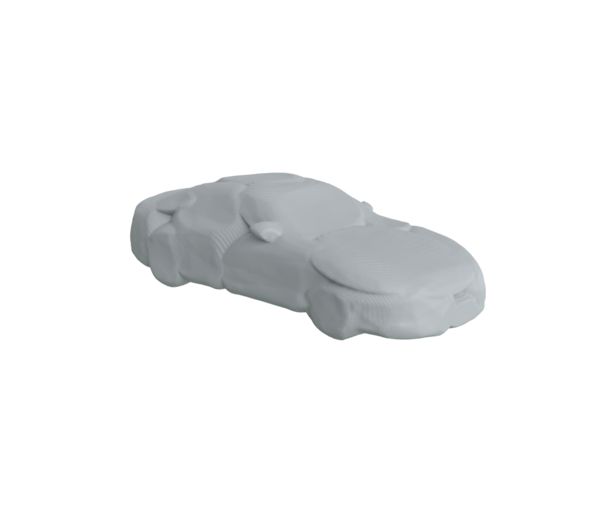}
    \end{subfigure}%
    \hfill%
    \begin{subfigure}[b]{0.15\linewidth}
		\centering
		\includegraphics[trim=100 0 100 0,clip,width=\linewidth]{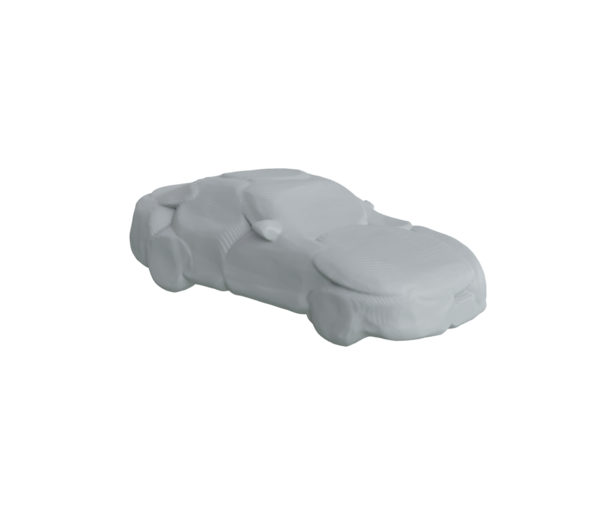}
    \end{subfigure}%
    \hfill%
    \begin{subfigure}[b]{0.15\linewidth}
		\centering
		\includegraphics[trim=100 0 100 0,clip,width=\linewidth]{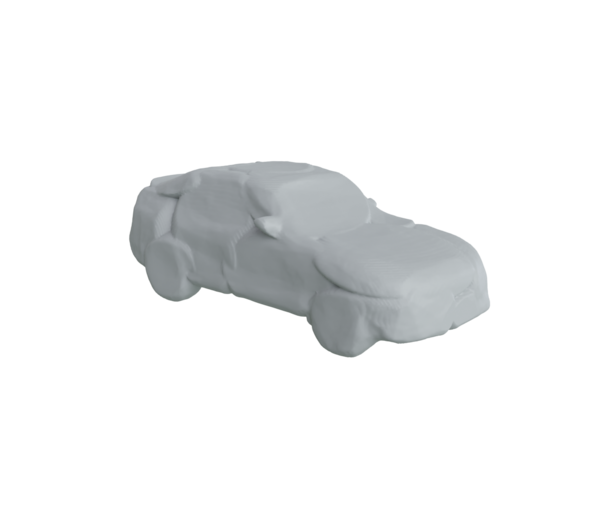}
    \end{subfigure}%
    \hfill%
    \begin{subfigure}[b]{0.15\linewidth}
		\centering
		\includegraphics[trim=100 0 100 0,clip,width=\linewidth]{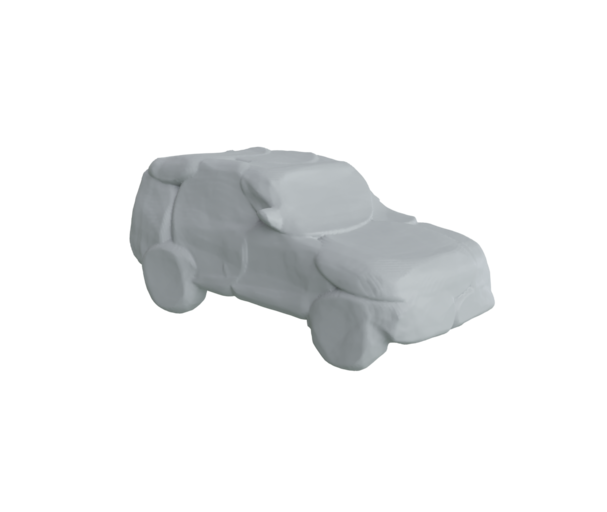}
    \end{subfigure}%
    \hfill%
    \begin{subfigure}[b]{0.15\linewidth}
		\centering
		\includegraphics[trim=100 0 100 0,clip,width=\linewidth]{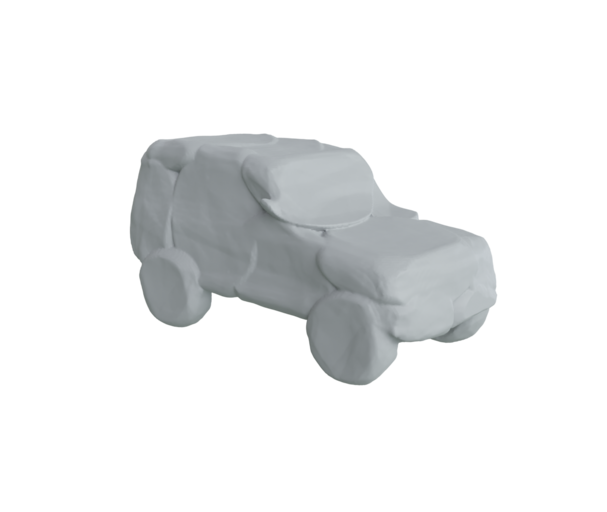}
    \end{subfigure}%
    \hfill%
    \begin{subfigure}[b]{0.15\linewidth}
		\centering
		\includegraphics[trim=100 0 100 0,clip,width=\linewidth]{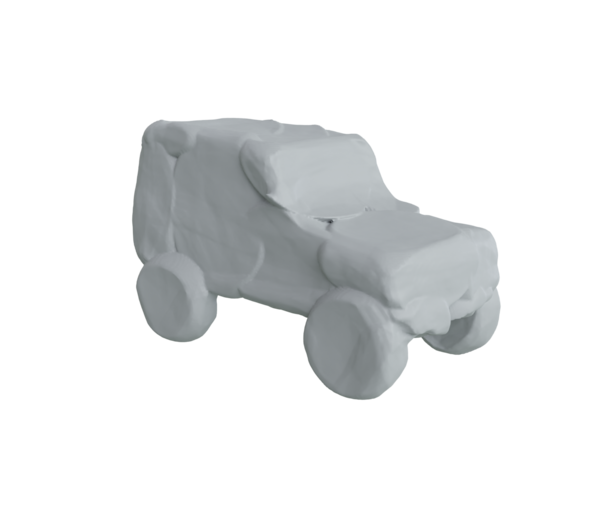}
    \end{subfigure}%
    \vskip\baselineskip%
    \vspace{-2.2em}
    \begin{subfigure}[b]{0.15\linewidth}
		\centering
		\includegraphics[width=\linewidth]{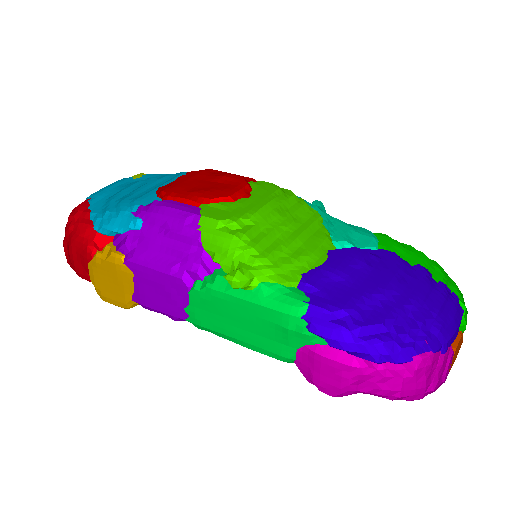}
    \end{subfigure}%
    \hfill%
    \begin{subfigure}[b]{0.15\linewidth}
		\centering
		\includegraphics[width=\linewidth]{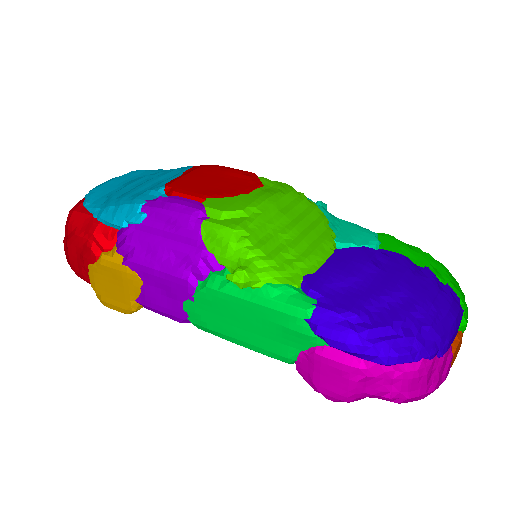}
    \end{subfigure}%
    \hfill%
    \begin{subfigure}[b]{0.15\linewidth}
		\centering
		\includegraphics[width=\linewidth]{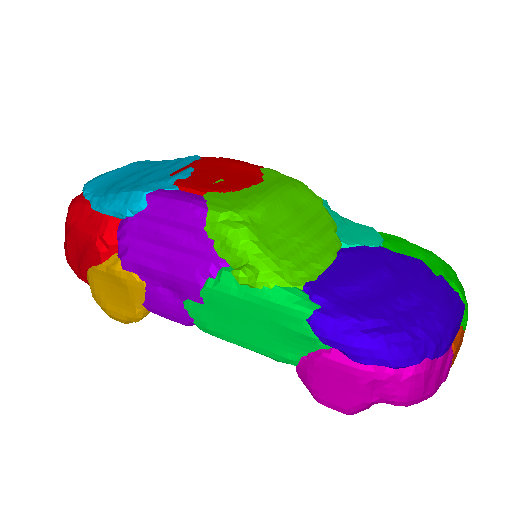}
    \end{subfigure}%
    \hfill%
    \begin{subfigure}[b]{0.15\linewidth}
		\centering
		\includegraphics[width=\linewidth]{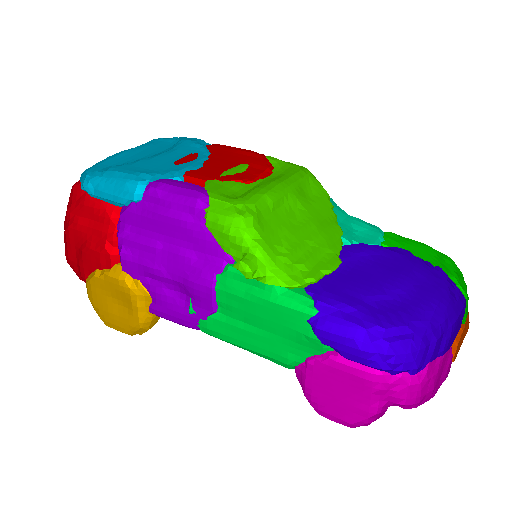}
    \end{subfigure}%
    \hfill%
    \begin{subfigure}[b]{0.15\linewidth}
		\centering
		\includegraphics[width=\linewidth]{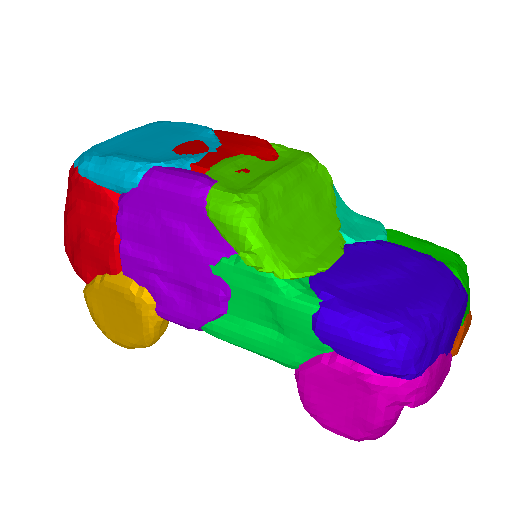}
    \end{subfigure}%
    \hfill%
    \begin{subfigure}[b]{0.15\linewidth}
		\centering
		\includegraphics[width=\linewidth]{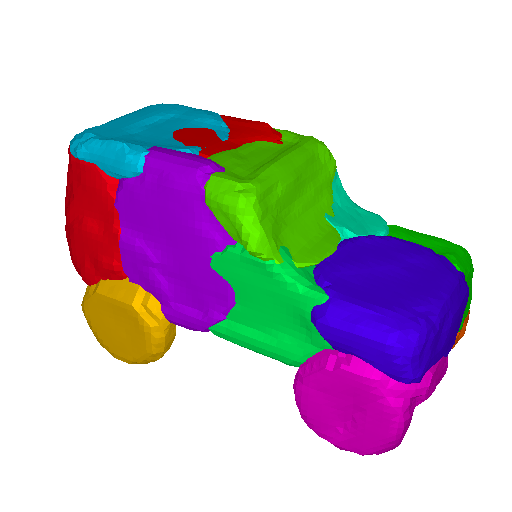}
    \end{subfigure}%
    \vskip\baselineskip%
    \vspace{-1.5em}
    \caption{{\bf Shape Interpolation}. From left to right, we interpolate between the geometry and texture latent codes of the two shapes.}
    \label{fig:shape_interpolations}
    \vspace{-1.2em}
\end{figure}

\boldparagraph{Soft Ray-Part Assignment}%
To investigate the impact of our hard ray-part assignment, we train a variant
of our model without, namely the color of a ray can be determined from multiple
NeRFs. We note that when we apply a transformation on a part of the
object, also the color of other parts changes, as illustrated in \figref{fig:ablation}.
Moreover, note that with this variant of our model it is not possible to change
the color of specific parts of the generated object, as in
\figref{fig:scene_specific_editing}.

\begin{figure}
    \centering
    \scriptsize
    \begin{minipage}[b]{0.15\linewidth}
		\centering
		Shape 1
    \end{minipage}%
    \hfill%
    \begin{minipage}[b]{0.15\linewidth}
		\centering
		Shape 2
    \end{minipage}%
    \hfill%
    \begin{minipage}[b]{0.33\linewidth}
		\centering
		Shape Code Interpolation
    \end{minipage}%
    \hfill%
    \begin{minipage}[b]{0.33\linewidth}
		\centering
		Texture Code Interpolation
    \end{minipage}%
    \hfill%
    \begin{subfigure}[b]{0.15\linewidth}
		\centering
		\includegraphics[width=\linewidth]{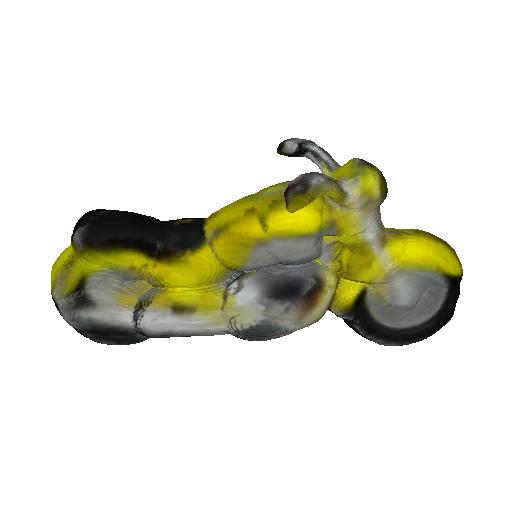}
    \end{subfigure}%
    \hfill%
    \begin{subfigure}[b]{0.15\linewidth}
		\centering
		\includegraphics[width=\linewidth]{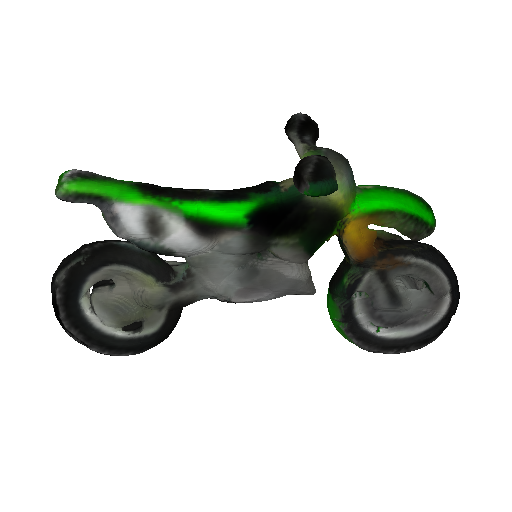}
    \end{subfigure}%
    \hfill%
    \begin{subfigure}[b]{0.15\linewidth}
		\centering
		\includegraphics[width=\linewidth]{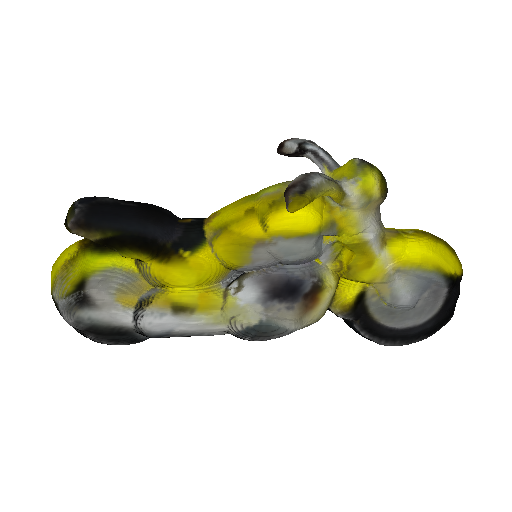}
    \end{subfigure}%
    \hfill%
    \begin{subfigure}[b]{0.15\linewidth}
		\centering
		\includegraphics[width=\linewidth]{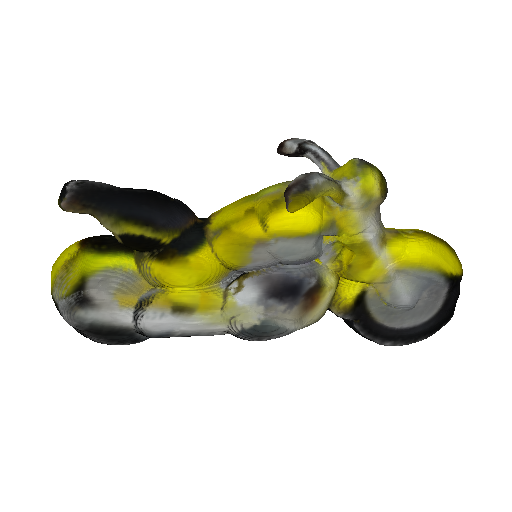}
    \end{subfigure}%
    \hfill%
    \begin{subfigure}[b]{0.15\linewidth}
		\centering
		\includegraphics[width=\linewidth]{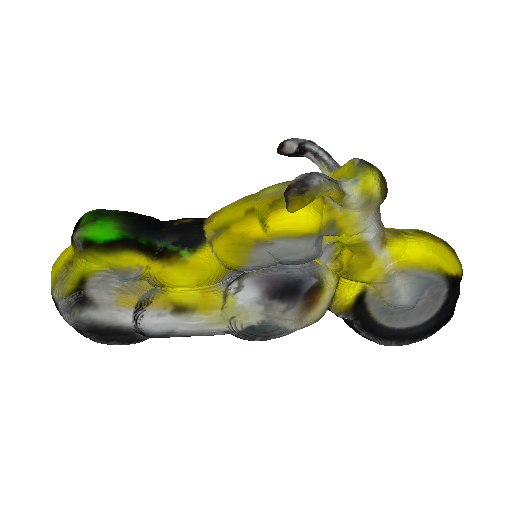}
    \end{subfigure}%
    \hfill%
    \begin{subfigure}[b]{0.15\linewidth}
		\centering
		\includegraphics[width=\linewidth]{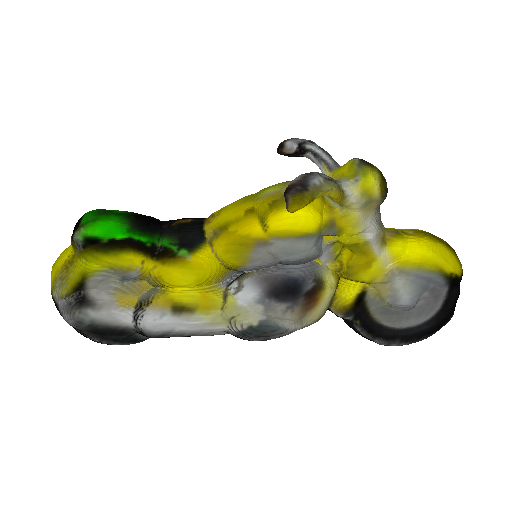}
    \end{subfigure}%
    \vskip\baselineskip%
    \vspace{-2.2em}
    \begin{subfigure}[b]{0.15\linewidth}
		\centering
		\includegraphics[width=\linewidth]{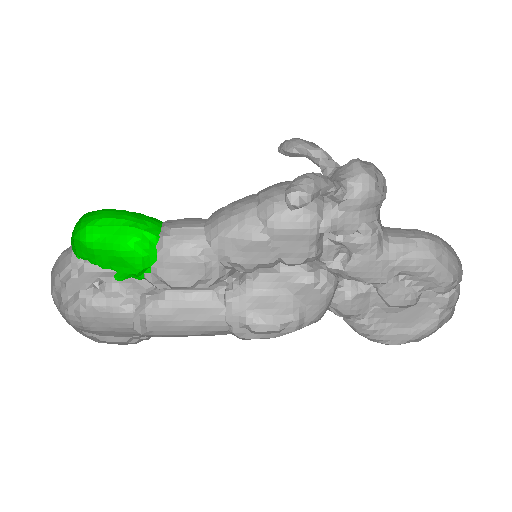}
    \end{subfigure}%
    \hfill%
    \begin{subfigure}[b]{0.15\linewidth}
		\centering
		\includegraphics[width=\linewidth]{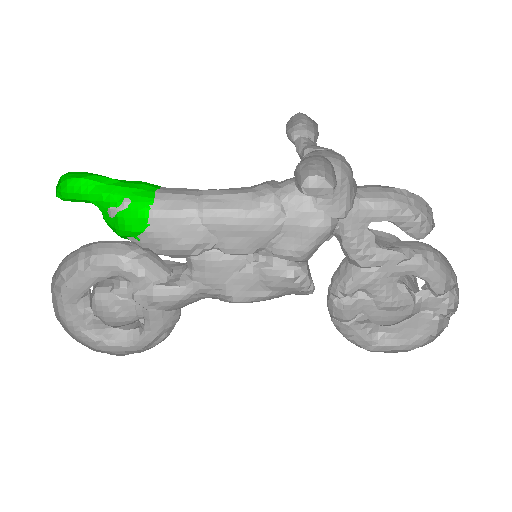}
    \end{subfigure}%
    \hfill%
    \begin{subfigure}[b]{0.15\linewidth}
		\centering
		\includegraphics[width=\linewidth]{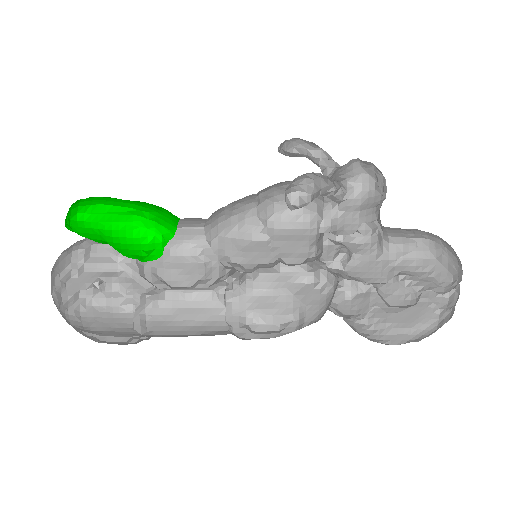}
    \end{subfigure}%
    \hfill%
    \begin{subfigure}[b]{0.15\linewidth}
		\centering
		\includegraphics[width=\linewidth]{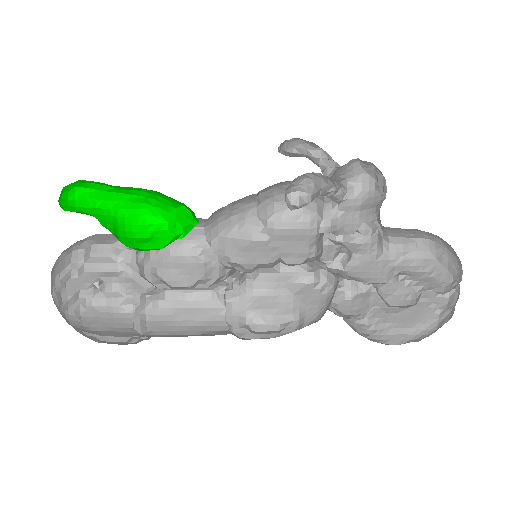}
    \end{subfigure}%
    \hfill%
    \begin{subfigure}[b]{0.15\linewidth}
		\centering
		\includegraphics[width=\linewidth]{part_interpolation_first_shape_parts}
    \end{subfigure}%
    \hfill%
    \begin{subfigure}[b]{0.15\linewidth}
		\centering
		\includegraphics[width=\linewidth]{part_interpolation_first_shape_parts}
    \end{subfigure}%
    \vskip\baselineskip%
    \vspace{-2.2em}
    \begin{subfigure}[b]{0.15\linewidth}
		\centering
		\includegraphics[width=\linewidth]{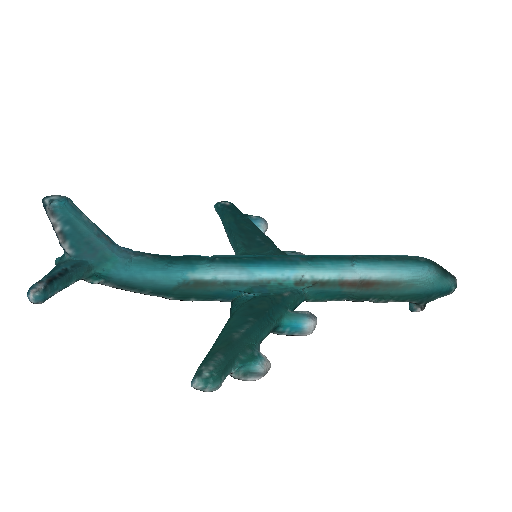}
    \end{subfigure}%
    \hfill%
    \begin{subfigure}[b]{0.15\linewidth}
		\centering
		\includegraphics[width=\linewidth]{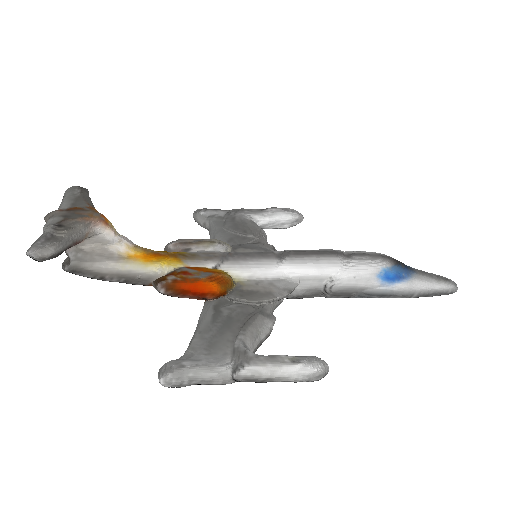}
    \end{subfigure}%
    \hfill%
    \begin{subfigure}[b]{0.15\linewidth}
		\centering
		\includegraphics[width=\linewidth]{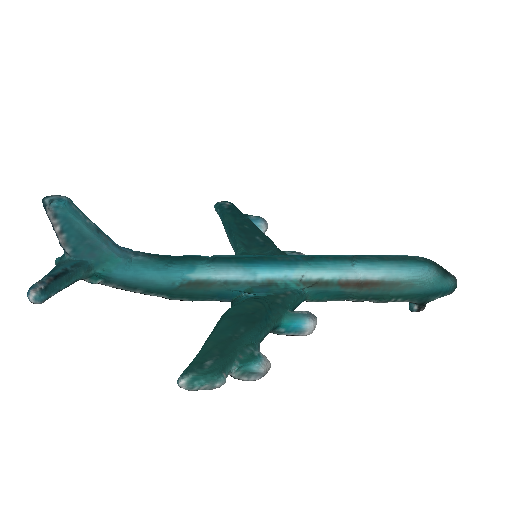}
    \end{subfigure}%
    \hfill%
    \begin{subfigure}[b]{0.15\linewidth}
		\centering
		\includegraphics[width=\linewidth]{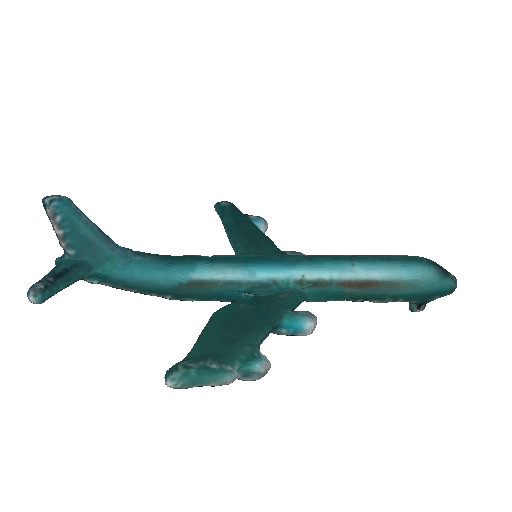}
    \end{subfigure}%
    \hfill%
    \begin{subfigure}[b]{0.15\linewidth}
		\centering
		\includegraphics[width=\linewidth]{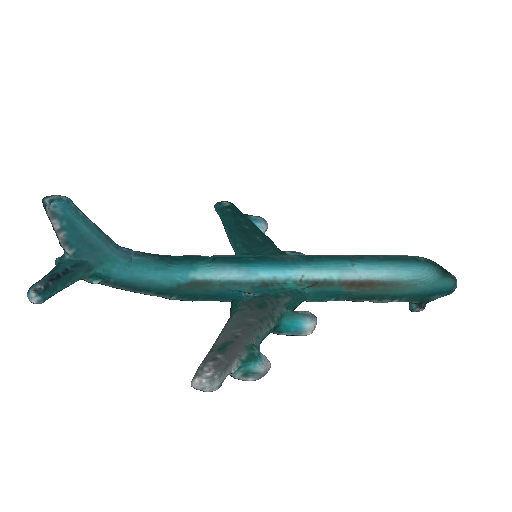}
    \end{subfigure}%
    \hfill%
    \begin{subfigure}[b]{0.15\linewidth}
		\centering
		\includegraphics[width=\linewidth]{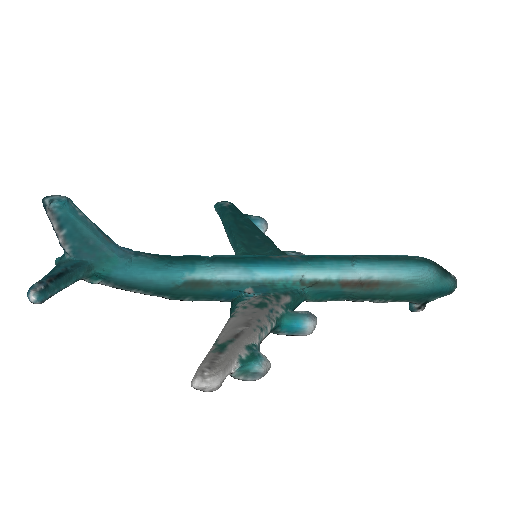}
    \end{subfigure}%
    \vskip\baselineskip%
    \vspace{-2.2em}
    \begin{subfigure}[b]{0.15\linewidth}
		\centering
		\includegraphics[width=\linewidth]{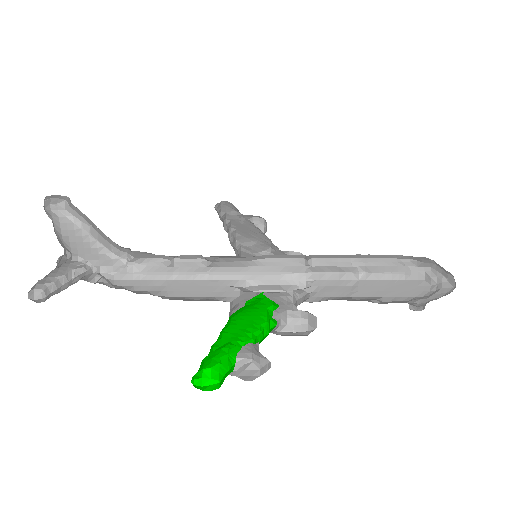}
    \end{subfigure}%
    \hfill%
    \begin{subfigure}[b]{0.15\linewidth}
		\centering
		\includegraphics[width=\linewidth]{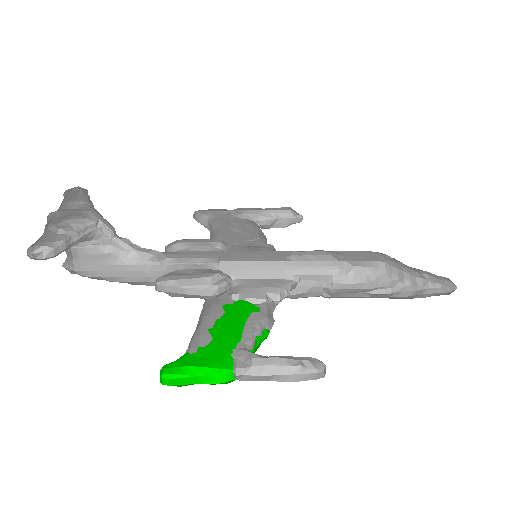}
    \end{subfigure}%
    \hfill%
    \begin{subfigure}[b]{0.15\linewidth}
		\centering
		\includegraphics[width=\linewidth]{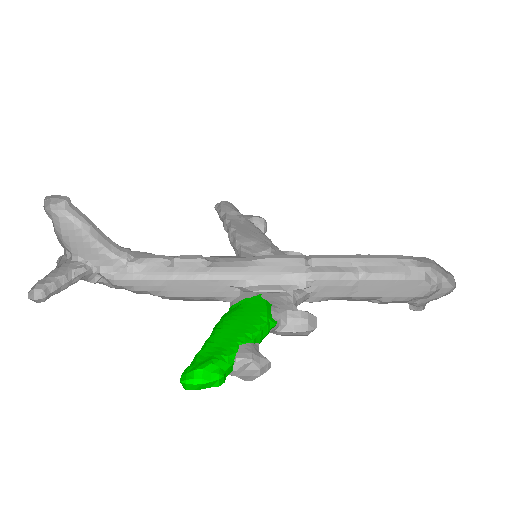}
    \end{subfigure}%
    \hfill%
    \begin{subfigure}[b]{0.15\linewidth}
		\centering
		\includegraphics[width=\linewidth]{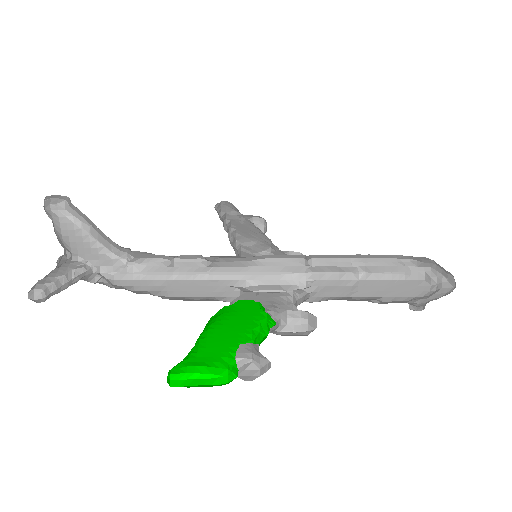}
    \end{subfigure}%
    \hfill%
    \begin{subfigure}[b]{0.15\linewidth}
		\centering
		\includegraphics[width=\linewidth]{part_interpolation_third_shape_parts}
    \end{subfigure}%
    \hfill%
    \begin{subfigure}[b]{0.15\linewidth}
		\centering
		\includegraphics[width=\linewidth]{part_interpolation_third_shape_parts}
    \end{subfigure}%
    \vskip\baselineskip%
    \vspace{-1.7em}
    \caption{{\bf Part-Level Interpolation}. From left to right, we interpolate
    between two parts (colored in green) from two shapes. For the motorbikes, we select the saddle and for
    the airplanes the right wing. In the 3rd and 4th columns,
    we interpolate between the shape codes of the two parts, whereas in the 5th and 6th
    we interpolate between the texture codes of the two parts.}
    \vspace{-1.5em}
    \label{fig:part_interpolations}
\end{figure}

\subsection{Shape Generation}
In \tabref{tab:shape_generation_quantitative}, we compare the quality of our
generations with NeRF-based generative models and observe that it outperforms
existing approaches in terms of MMD and COV
on Motorbikes and Chairs, while being better than \cite{Schwarz2020NEURIPS, Chan2021CVPR}
also for Cars. Furthermore, it performs on par with the
concurrent work of \cite{Gao2022NEURIPS} that relies on a
highly optimized differentiable graphics renderer \cite{Laine2020SIGGRAPH} and
requires training for approximately 16 GPU days (8 A100 for 2
days). Instead our model employs a simpler volumetric renderer and
training takes approximately 5 GPU days (1 RTX 3090).
Compared to
\cite{Chan2021CVPR, Schwarz2020NEURIPS}, our generations are sharper
with more crisp colors (see \figref{fig:shapenet_qualitative_comparison}).
However, compared to \cite{Gao2022NEURIPS,
Chan2022CVPR} that employ tri-plane representations and hence can generate
high-resolution textures, our textures are less detailed.

We also compare PartNeRF with state-of-the-art part-based
approaches that require 3D supervision. While our model is trained
from posed images and object masks, it consistently generates plausible 3D
geometries (see \figref{fig:shapenet_qualitative_comparison_parts}). This
is also valitated quantitatively in \tabref{tab:shape_generation_parts_quantitative},
where we observe that our model outperforms
both \cite{Hao2020CVPR, Hertz2022SIGGRAPH} on Airplanes while being better in
terms of MMD for Tables.

\boldparagraph{Shape Interpolation}%
In \figref{fig:shape_interpolations}, from left to right, we show interpolations
between the shape and texture codes of two cars. We observe that our
model smoothly interpolates between two shapes, while preserving the
shape structure and the the part-based structure.

\boldparagraph{Part-level Interpolation}%
\figref{fig:part_interpolations} shows part-level interpolation, where we pick
two shapes and select a part from both.
For the motorbikes, we choose the saddle part
and linearly interpolate its shape and texture codes. We repeat for the
airplanes, where we select the right wing. When we interpolate the shape codes,
the geometry changes, while the texture
remains the same. In contrast, when we interpolate the texture codes, the geometry
remains unchanged
and only the part texture changes smoothly. Across all interpolations
our model consistently generates realistic shapes.

\subsection{Shape and Texture Editing}

\begin{figure}
    \centering
    \scriptsize
    \begin{minipage}[b]{0.15\linewidth}
		\centering
		Shape 1
    \end{minipage}%
    \hfill%
    \begin{minipage}[b]{0.15\linewidth}
		\centering
		Shape 2
    \end{minipage}%
    \hfill%
    \begin{minipage}[b]{0.22\linewidth}
		\centering
        Geometry Mixing
    \end{minipage}%
    \hfill%
    \begin{minipage}[b]{0.2\linewidth}
		\centering
		Texture Mixing
    \end{minipage}%
    \hfill%
    \begin{minipage}[b]{0.2\linewidth}
		\centering
        Combined
    \end{minipage}%
    \hfill%
    \begin{subfigure}[b]{0.15\linewidth}
		\centering
		\includegraphics[width=\linewidth]{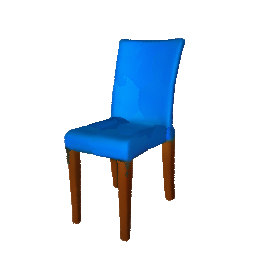}
    \end{subfigure}%
    \hfill%
    \begin{subfigure}[b]{0.15\linewidth}
		\centering
		\includegraphics[width=\linewidth]{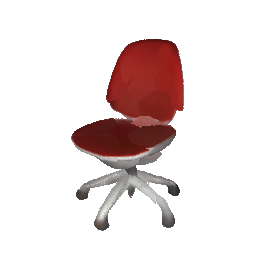}
    \end{subfigure}%
    \hfill%
    \begin{subfigure}[b]{0.15\linewidth}
		\centering
		\includegraphics[width=\linewidth]{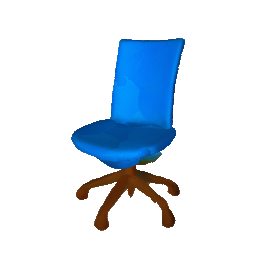}
    \end{subfigure}%
    \hfill%
    \begin{subfigure}[b]{0.15\linewidth}
		\centering
		\includegraphics[width=\linewidth]{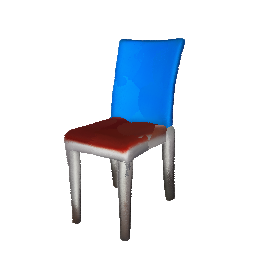}
    \end{subfigure}%
    \hfill%
    \begin{subfigure}[b]{0.15\linewidth}
		\centering
		\includegraphics[width=\linewidth]{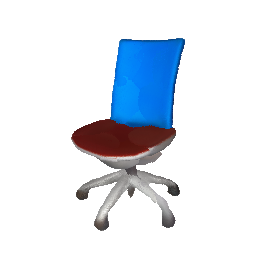}
    \end{subfigure}%
    \vskip\baselineskip%
    \vspace{-1.3em}
    \begin{subfigure}[b]{0.15\linewidth}
		\centering
		\includegraphics[width=\linewidth]{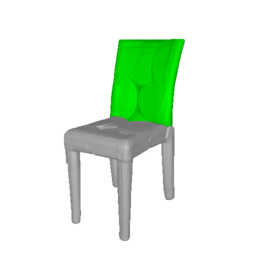}
    \end{subfigure}%
    \hfill%
    \begin{subfigure}[b]{0.15\linewidth}
		\centering
		\includegraphics[width=\linewidth]{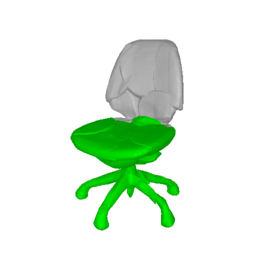}
    \end{subfigure}%
    \hfill%
    \begin{subfigure}[b]{0.15\linewidth}
		\centering
		\includegraphics[width=\linewidth]{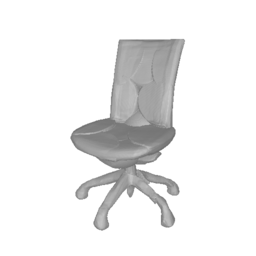}
    \end{subfigure}%
    \hfill%
    \begin{subfigure}[b]{0.15\linewidth}
		\centering
		\includegraphics[width=\linewidth]{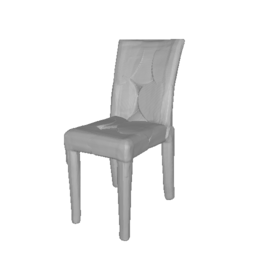}
    \end{subfigure}%
    \hfill%
    \begin{subfigure}[b]{0.15\linewidth}
		\centering
		\includegraphics[width=\linewidth]{shape_mixing_first_shape_mixing_parts}
    \end{subfigure}%
    \vskip\baselineskip%
    \vspace{-1.3em}
    \begin{subfigure}[b]{0.15\linewidth}
		\centering
		\includegraphics[width=\linewidth]{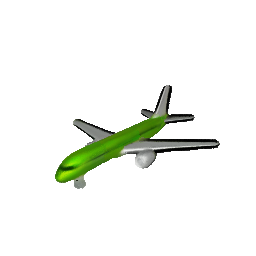}
    \end{subfigure}%
    \hfill%
    \begin{subfigure}[b]{0.15\linewidth}
		\centering
		\includegraphics[width=\linewidth]{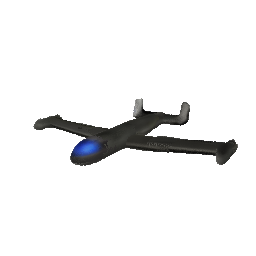}
    \end{subfigure}%
    \hfill%
    \begin{subfigure}[b]{0.15\linewidth}
		\centering
		\includegraphics[width=\linewidth]{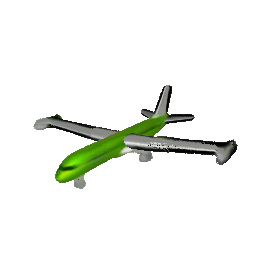}
    \end{subfigure}%
    \hfill%
    \begin{subfigure}[b]{0.15\linewidth}
		\centering
		\includegraphics[width=\linewidth]{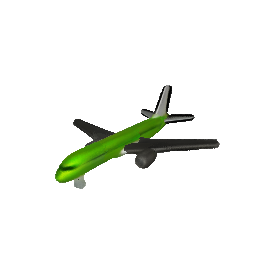}
    \end{subfigure}%
    \hfill%
    \begin{subfigure}[b]{0.15\linewidth}
		\centering
		\includegraphics[width=\linewidth]{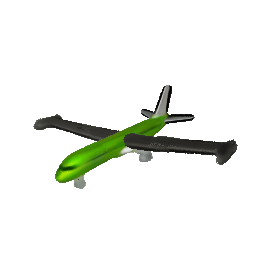}
    \end{subfigure}%
    \vskip\baselineskip%
    \vspace{-1.9em}
    \begin{subfigure}[b]{0.15\linewidth}
		\centering
		\includegraphics[width=\linewidth]{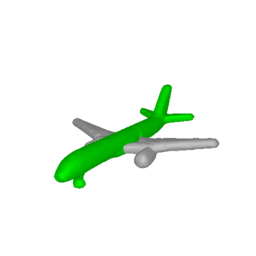}
    \end{subfigure}%
    \hfill%
    \begin{subfigure}[b]{0.15\linewidth}
		\centering
		\includegraphics[width=\linewidth]{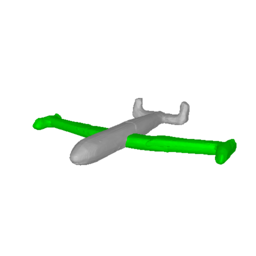}
    \end{subfigure}%
    \hfill%
    \begin{subfigure}[b]{0.15\linewidth}
		\centering
		\includegraphics[width=\linewidth]{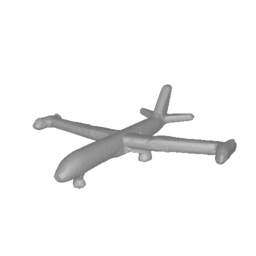}
    \end{subfigure}%
    \hfill%
    \begin{subfigure}[b]{0.15\linewidth}
		\centering
		\includegraphics[width=\linewidth]{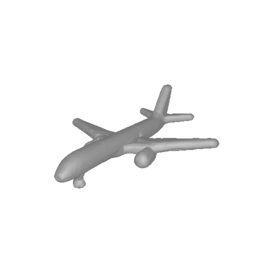}
    \end{subfigure}%
    \hfill%
    \begin{subfigure}[b]{0.15\linewidth}
		\centering
		\includegraphics[width=\linewidth]{shape_mixing_third_shape_mixing_parts}
    \end{subfigure}%
    \vskip\baselineskip%
    \vspace{-1.8em}
    \caption{{\bf Shape Mixing}. We mix parts from two shapes and show
    \emph{geometry} (3rd column), \emph{texture} (4th column), and
    combined geometry and texture mixing (5th column). The selected parts for mixing are colored
    in green.}
    \label{fig:shape_mixing}
    \vspace{-1.2em}
\end{figure}

\boldparagraph{Shape Mixing}%
Starting from two shapes, the task is to select and combine parts in a
meaningful way. As shown in \figref{fig:shape_mixing}, we consider two types of
mixing operations:
\emph{geometry} and \emph{texture} mixing. In geometry mixing, we
combine parts from two objects and generate a new one, whose texture is
determined by the texture codes from one of the two, while shape codes
are taken from different parts of the two objects (third column).
In texture mixing, we mix the shapes only in
terms of texture, while the shape is determined by the shape codes of
one object (fourth column).  We also combine both mixing modes.
PartNeRF consistently
generates plausible shapes across all editing operations.

\boldparagraph{Shape Editing}%
Our method also allows several shape editing operations on the part level, such
as applying rigid and non-rigid transformations and inserting or removing
parts. Affine transformations are directly applied on the rotation and
translation vector that define the per-part coordinate frame. To remove a part,
it suffices to ignore its associated NeRF in the rendering process.
Likewise, adding a part amounts to incorporating its NeRF during
rendering. Examples of these operations are summarized in
\figref{fig:teaser}.

\section{Conclusion}

In this paper, we introduce PartNeRF, the first part-aware generative model that
parametrizes parts as NeRFs. As our work considers the decomposition of objects
into parts, it enables intuitive part-level control and several editing
operations not previously possible. Furthermore, it is trained without explicit 3D
supervision, using only posed images and object masks.
Our experiments showcase the ability of our model to generate plausible 3D
shapes with texture. Moreover, we demonstrate several editing operations both
on the texture and the shape of the generated object. In future work, we plan
to investigate incorporating more complex representations such as triplanes
\cite{Chan2022CVPR}, or using differentiable rendering techniques
\cite{Gao2022NEURIPS} in order to better represent the object's texture.
Another exciting direction for future research is
extending our model to moving objects.

\section*{Acknowledgments}

This research was supported by an ARL grant W911NF-21-2-0104 and an Adobe
research gift.
Konstantinos Tertikas and Ioannis Emiris have received funding from the European 
Union’s Horizon 2020 research and innovation programme under the Marie Skłodowska-Curie 
grant agreement No. 860843.
Despoina Paschalidou is supported by the Swiss National Science Foundation
under grant number P500PT\_206946. Leonidas Guibas is supported by a Vannevar
Bush Faculty Fellowship.

{\small
	\bibliographystyle{ieee_fullname}
	\bibliography{bibliography_long,bibliography,bibliography_custom}
}

\newpage
\clearpage
\appendix
\onecolumn

\begin{minipage}{\linewidth}
   \begin{center}
      \setlength\parindent{0pt}
      {\Large \bf Supplementary Material for\\ PartNeRF: Generating Part-Aware Editable 3D Shapes\\
      without 3D Supervision \par}
            \vspace*{24pt}
      {
      \large
      \begin{tabular}[t]{c}
        Konstantinos Tertikas$^{1,3}$ \quad Despoina Paschalidou$^{2}$ \quad
        Boxiao Pan$^{2}$ \quad Jeong Joon Park$^{2}$ \\ Mikaela Angelina Uy$^{2}$ \quad Ioannis Emiris$^{3,1}$\quad
        Yannis Avrithis$^{4}$ \quad Leonidas Guibas$^{2}$\\[0.4em]
        $^1$National and Kapodistrian University of Athens\quad
        $^2$Stanford University\\
        $^3$Athena RC, Greece \quad
        $^4$Institute of Advanced Research in Artificial Intelligence (IARAI)\\
        \vspace*{1pt}
      \end{tabular}
      \par
      }
   \end{center}
\end{minipage}

\begin{abstract}

In this \textbf{supplementary document}, we first present a detailed overview
of our network architecture, the training and generation procedure, in
\secref{sec:implementation_details}.  We then provide ablations on how
different components of our pipeline impact the performance of our model, in
\secref{sec:ablations}. Subsequently, in \secref{sec:generations} we provide
additional results both on the shape generation task as well as on the shape
editing task. Next, we demonstrate various applications of our model such
as shape and image inversion, in \secref{sec:shape_inversion} and
\secref{sec:image_inversion}, respectively. Finally, in
\secref{sec:parts_discussion}, we analyse the limitations, future research
directions and in \secref{sec:negative_societal_impact} we discuss potential
negative impact of our method on society. 
\end{abstract}

\section{Implementation Details}
\label{sec:implementation_details}

In this section, we provide a detailed description of the several components of
our network architecture (\secref{subsec:network}). Next, we describe our
training procedure (\secref{subsec:training}) and the  generation protocol (\secref{subsec:generation}).
Then, we provide details regarding the various editing operations demonstrated in the main
submission (\secref{subsec:editing}). Finally, we detail our metrics computation
(\secref{subsec:metrics}) and discuss our baselines (\secref{subsec:baselines}).

\subsection{Network Architecture}
\label{subsec:network}

Here, we describe the architecture of each individual component of our model,
as illustrated in Fig. 2 in the main submission. As already discussed in Sec
3.3 of our main submission, we implement our generative model using an
auto-decoder \cite{Bojanowski2018ICML, Park2019CVPR}. In particular, the input
to our model is a set of embedding vectors $\cZ = \{\bz_i\}_{i=1}^n$, uniquely
identifying each one from the $n$ training samples. Each $\bz_i=\{\bz^s, \bz^t\}$,
comprises two learnable embeddings that capture the shape and appearance
properties of each sample. Note that both the shape $\bz^s$ and texture $\bz^t$
codes are initialized by sampling from the normal distribution $\bz^s, \bz^t
\sim N(\mathbf{0}, I)$.
To avoid notation clutter, we omit the object index
$i$ in the rest of this section.

Our network consists of three main components: (i) the \emph{decomposition
network} that maps the instance-specific latent codes $\{\bz^s, \bz^t\}$ to
$M$ per-part shape $\{\bz_m^s\}_{i=1}^M$ and texture
$\{\bz_m^t\}_{i=1}^M$ embeddings, (ii) the \emph{structure network} that
takes the per-part shape code $\bz_m^s$ and predicts an affine transformation
$\{\bR_m, \bt_m\}$ and scale $\bs_m$ that controls the pose and the area of
influence of this part, and (iii) the \emph{neural rendering module} that
renders a 2D image using $M$ NeRFs. 
As supervision, we use the observed RGB color
$C(r) \in \mathbb{R}^3$ and the object mask $I(r) \in \{0, 1\}$ for each ray $r
\in \cR$. We also associate $r$ with a binary label $\ell_r = I(r)$,
as shown in \figref{fig:inside_outside_rays}.
Namely, we label a ray $r$ as
\emph{inside}, if $\ell_r=1$ and \emph{outside} if $\ell_r=0$.

\boldparagraph{Decomposition Network}%
The decomposition network maps the instance-specific shape and texture $\{\bz^s,
\bz^t\}$ latent codes into a set of $M$ latent codes that control the shape and
appearance of each part. We implement the decomposition network using two
multi-head attention transformers without positional encoding \cite{Vaswani2017NIPS},
$\tau^s_\theta$ and $\tau^t_\theta$, that predict $M$ shape $\{\bz_m^s\}_{m=1}^M$ and
texture $\{\bz_m^t\}_{m=1}^M$ codes, as follows:
\begin{equation}
    \begin{aligned}
        \tau^s_{\theta}: \mathbb{R}^{L_d} &\rightarrow \mathbb{R}^{L_d \times M} &
        \{\bz_m^s\}_{m=1}^M &= \tau^s_\theta(f_\theta(\bz^s)) \\
        \tau^t_{\theta}: \mathbb{R}^{L_d} &\rightarrow \mathbb{R}^{L_d \times M} & 
        \{\bz_m^t\}_{m=1}^M &= \tau^t_\theta(f_\theta(\bz^t))
        \label{eq:decomposition_network}
    \end{aligned}
\end{equation}
where $L_d$ is the dimensionality of the instance-specific embeddings and is
set to $128$. The input to transformer $\tau^s_{\theta}$ is the set of $M$
per-part shape codes $\{\hat{\bz}^s\}_{m=1}^M$, where $\hat{\bz}_m^s \in
\mathbb{R}^{128}$. Likewise, the input to transformer $\tau^t_{\theta}$ is the
set of $M$ per-part texture codes $\{\hat{\bz}^t\}_{m=1}^M$, where $\hat{\bz}_m^t \in
\mathbb{R}^{128}$. Note that the $M$ shape $\{\hat{\bz}^s\}_{m=1}^M$ and texture
$\{\hat{\bz}^t\}_{m=1}^M$ codes are produced from $\bz_s$ and $\bz_t$ respectively using $M$
linear projections $f_\theta(\cdot)$. The output of each transformer is the set of
per-part shape $\{\bz_m^s\}_{i=1}^M$ and texture $\{\bz_m^t\}_{i=1}^M$, where $\bz_m^s \in \mathbb{R}^{128}$ and
$\bz_m^t \in \mathbb{R}^{128}$.
Both transformers consist of $2$ layers
with $4$ attention heads. The queries, keys and values have 128 dimensions and
the intermediate representations for the MLPs have 1024 dimensions.
To implement the transformer architecture we use the transformer library
provided
by Wightman~\cite{rw2019timm}\footnote{
\href{https://github.com/rwightman/pytorch-image-models}{https://github.com/rwightman/pytorch-image-models}
}.

\begin{figure}[!h]
    \vspace{1.2em}
    \centering
    \begin{subfigure}[t]{0.48\columnwidth}
        \centering
        \includegraphics[width=\textwidth]{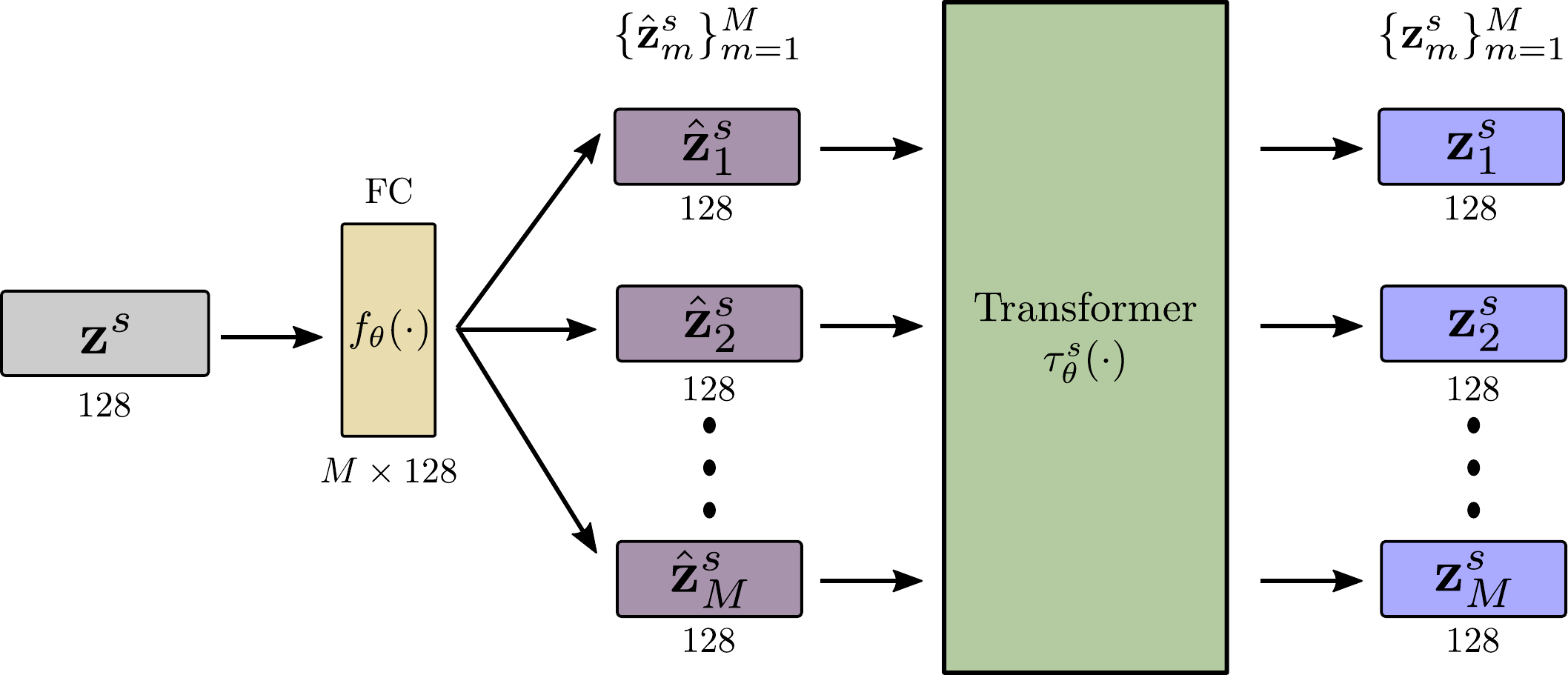}
        \vspace{-1.2em}
        \caption{$\tau^s_{\theta}(\cdot)$ predicts the $M$ per-part shape codes $\{\bz_m^s\}_{i=1}^M$.}
        \label{fig:mlp_translation}
    \end{subfigure}%
    \hfill
    \begin{subfigure}[t]{0.48\columnwidth}
        \centering
        \includegraphics[width=\textwidth]{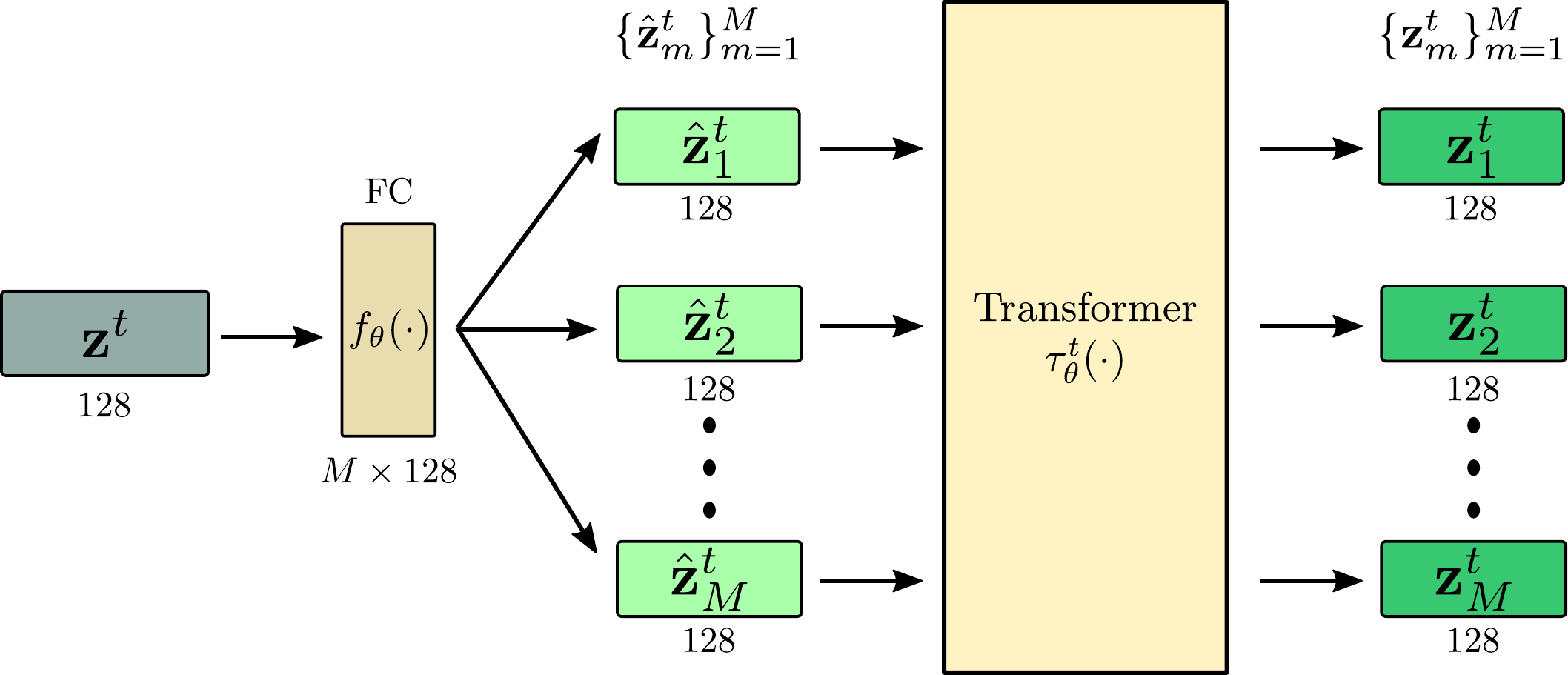}
        \vspace{-1.2em}
        \caption{$\tau^t_{\theta}(\cdot)$ predicts the $M$ per-part shape codes $\{\bz_m^t\}_{i=1}^M$.}
        \label{fig:mlp_size}
    \end{subfigure}
    \caption{{\bf{Decomposition Network.}} The decomposition network consists
    of two two multi-head attention transformers, $\tau^s_{\theta}$ and
    $\tau^t_{\theta}$, without positional encoding, that map the object-specific
    shape and texture embeddings $\bz^s, \bz^t$ to $M$ learnable codes
    $\{\bz_m^s\}_{m=1}^M$ and $\{\bz_m^t\}_{m=1}^M$ that control the shape and
    texture of each part respectively.}
    \label{fig:decomposition_network}
    \vspace{-0.8em}
\end{figure}

\begin{figure}[!h]
    \centering
    \includegraphics[width=0.3\textwidth]{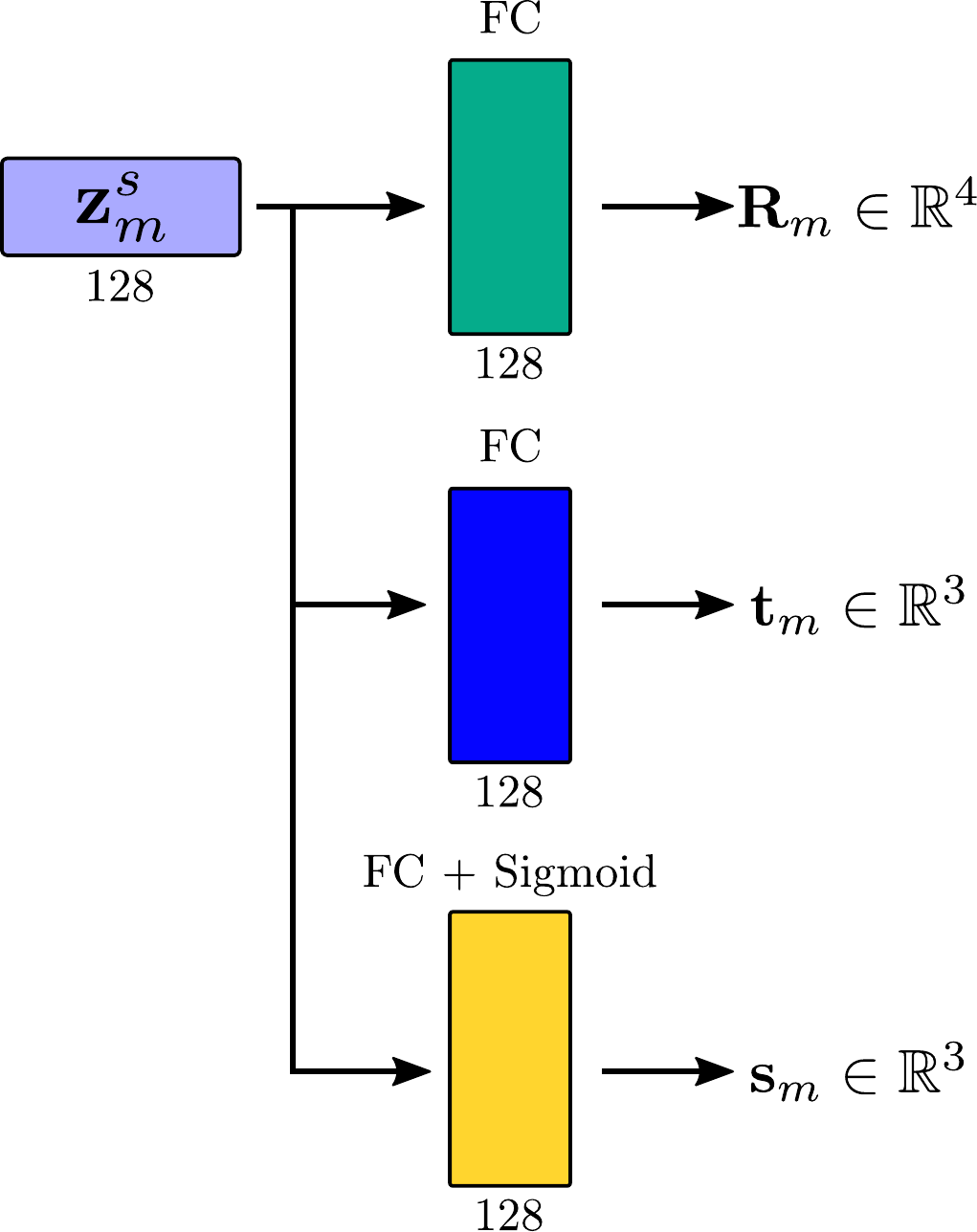}
    \caption{{\bf{Structure Network.}} The structure network $s_\theta$ maps the
    shape latent code $\bz_m^s \in \mathbb{R}^{L_d}$ to a translation vector
    $\bt_m$, a rotation matrix $\bR_m$ and a scale vector $\bs_m$, using an MLP
    with shared weights across parts.}
    \label{fig:structure_network}
\end{figure}

\boldparagraph{Structure Network}%
The structure network learns a function $s_\theta: \mathbb{R}^{L_d} \rightarrow
\mathbb{R}^{4} \times \mathbb{R}^{3} \times \mathbb{R}^{3}$ that maps the
per-part shape code $\bz_m^s$ to an affine transformation,
defined through a rotation matrix $\bR_m \in SO(3)$ and a translation vector
$\bt_m \in \mathbb{R}^3$, and a scale vector $\bs_m \in \mathbb{R}^3$. We follow
\cite{Paschalidou2021CVPR} and parametrize the rotation matrices using
quaternions \cite{Hamilton1848}, whereas the translation and the scale vectors are
represented using 3 scalars along the axes. Specifically the per-part $\bR_m, \bt_m$ and 
$\bs_m$ are predicted from an MLP with shared weights across parts. A
pictorial representation for the structure network is provided in
\figref{fig:structure_network}.
Starting from the shape latent code for part $m$, we feed it to a linear layer
with $128$ hidden dimensions that predicts $4$ values that define the rotation
matrix $\bR_m$ as a quaternion. Similarly, we feed $\bz_m^s$ to another another
linear layer with $128$ hidden dimensions that predicts the $3$ values that
define the translation vector $\bt_m$ of the $m$-th part. Finally, to predict
the scale vector $\bs_m$, we use a linear layer with $128$ hidden dimensions,
followed by a sigmoid activation function. As discussed in Sec. 3.2 of our main
submission, we associate every ray with a specific part and transform it to its
coordinate system using the predicted rotation and translation matrices. To
transform a 3D point $\bx$ to the local coordinate system of the $m$-th part,
we simply apply $\cT_m(\bx) = \bR_m (\bx + \bt_m)$, whereas to transform the
ray direction $\bd$ it simply suffices to multiply it with the rotation matrix
$\bR_m$. The scale vector $\bs_m$ represents the spatial extent of each part.
To ensure that each part captures continuous regions of the object, we multiply
its occupancy function with the occupancy function of an axis-aligned 3D
ellipsoid centered at the origin of the coordinate system, with axis lengths
given by the scale vector $\bs_m$. Specifically the parametric function
of the $m$-th ellipsoid is $f^m_{\theta}(\bx)=f(\cT_m(\bx), \bs_m)$, where
\begin{equation}
    f(\bx,\bs) = \norm{\diag(\bs)^{-1} \bx}^2,
    \label{eq:ellipsoid}
\end{equation}
is less than $1$ when a 3D point $\bx$ is inside the ellipsoid, greater than $1$ when $\bx$ is
outside and equal to one when $\bx$ is on the surface. We
convert the parametric function of each ellipsoid to an occupancy
function $g : \mathbb{R}^3 \rightarrow [0, 1]$ with a sigmoid function $\sigma(\cdot)$,
as follows:
\begin{equation}
    g(\bx, \bs) = \sigma \left(\beta \left(1 - f(\bx, \bs)\right) \right),
    \label{eq:ellipsoid_occupancy}
\end{equation}
where $\beta=100$ controls the sharpness of the transition of the occupancy
function. As already discussed in Sec. 3.2 of our main submission, note that
before estimating the occupancy function of the $m$-the ellipsoid
$g_\theta^m(\bx) = g(T_m(\bx), \bs_m)$, we first transform $\bx$ to the
coordinate frame of the $m$-th part. This ensures that any transformation of
the part, also transforms its ellipsoid.

\boldparagraph{Neural Rendering}%
The last component of our pipeline performs volumetric rendering using $M$
NeRFs using the assignment of rays to parts from Eq. 9 in our main submission
and the per-part rendering equation
\begin{equation}
	\hat{C}_m(r) = \sum_{i=1}^N
		h_\theta^m(\bx_i^r) \prod_{j<i} \left(1 - h_\theta^m (\bx_i^r) \right)
		c_\theta^m(\bx_i^r, \bd^r),
    \label{eq:rendering_ours_supp}
\end{equation}
as follows:
\begin{equation}
	\hat{C}(r) = \sum_{m=1}^M \ind_{r \in \cR_m} \hat{C}_m(r),
    \label{eq:rendering_equation_image_supp}
\end{equation}
where $\hat{C}(r)$ is the predicted color value for ray $r$ and
$h_\theta^m(\bx) = o_\theta^m(\bx) g_\theta^m(\bx)$ is the joint occupancy
function of the $m$-th part. Namely, we use the $m$-th NeRF to render ray $r$ if it is assigned to the
$m$-th part. Whereas, if a ray is not associated with any part its color is
simply black, namely $\hat{C}(r) = 0$.

We follow \cite{Oechsle2021ICCV} and employ two
separate networks: an \emph{occupancy network} $o_{\theta}(\cdot)$ and a
\emph{color network} $c_{\theta}(\cdot)$ to predict the color and occupancy
values. The occupancy network $o_{\theta}$ maps a 3D point $\bx$
and the shape code $\bz_m^s$ to an occupancy value $o$. For the $m$-th part, the
predicted occupancy value of a 3D point $\bx$ is defined as:
\begin{equation}
    \begin{aligned}
        o_{\theta}: \mathbb{R}^3 \times \mathbb{R}^{128} & \rightarrow [0, 1] & \quad
        o = \sigma(\tau o^m_{\theta}(\cT_m(\bx), \bz_m^s)),
    \end{aligned}
    \label{eq:nerf_occupancy_network_supp}
\end{equation}
where $\sigma(\cdot)$ is the sigmoid function and $\tau$ is a hyperparameter
that controls the sharpness of the transition of the occupancy function and is
set to $100$. Likewise, the color network $c_{\theta}$ maps the 3D point $\bx$ and the
viewing direction $\bd$ after applying positional encoding on its
components, as well as the shape code $\bz_m^s$ and the texture code $\bz_m^t$ into a color
value $\bc \in \mathbb{R}^3$. For the $m$-th part the predicted color value for
a 3D point $\bx$ along a ray with direction $d$ can be defined as:
\begin{equation}
    \begin{aligned}
        c_{\theta}: \mathbb{R}^{L_x} \times \mathbb{R}^{L_d} \times \mathbb{R}^{128} \times \mathbb{R}^{128}
        &\rightarrow \mathbb{R}^{+} & \quad
        \bc = c^m_{\theta}(\gamma(\cT_m(\bx)), \gamma(\bR_m\bd), \bz_m^s, \bz_m^t).
    \end{aligned}
    \label{eq:nerf_color_network_supp}
\end{equation}
Here, $\gamma(\cdot)$ is the positional encoding of
\cite{Mildenhall2020ECCV} that is applied element-wise
as follows
\begin{equation}
    \gamma(p) = [p; \sin(2^0\pi p), \cos(2^0\pi p), \dots, \sin(2^{L-1}\pi p), \cos(2^{L-1}\pi p)],
    \label{eq:positional_encoding}
\end{equation}
where $p \in \real$ can be any dimension of the input point $\bx$ and the corresponding ray
direction $\bd$ and $[\cdot ; \cdot]$ denotes concatenation. In our
experiments, we set $L=10$, hence $L_x = L_d = 3(2L+1) = 63$.

\begin{figure}[!h]
    \centering
    \includegraphics[width=\textwidth]{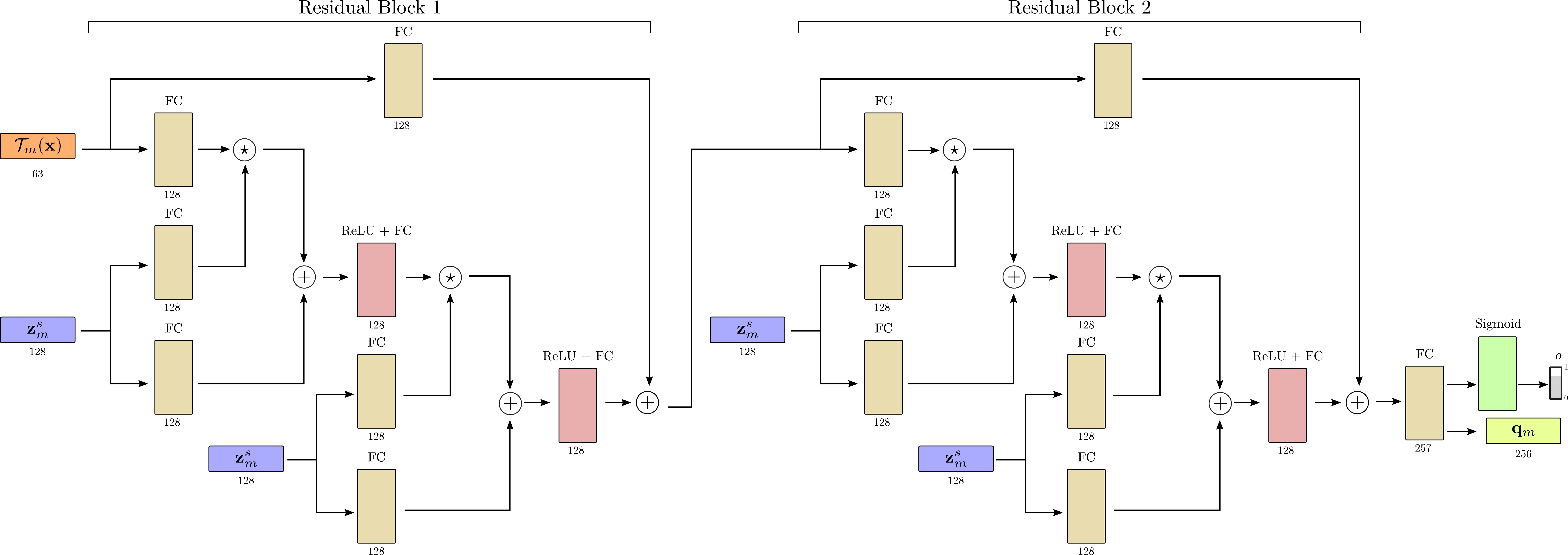}
    \caption{{\bf{Occupancy Network.}} The occupancy network
    $o_{\theta}(\cdot)$ maps a 3D point $\bx$ transformed in the local
    coordinate system of the $m$-th part and its shape code $\bz_m^s$ to an
    occupancy value $o \in [0, 1]$ and an intermediate feature vector $\bq_m
    \in \mathbb{R}^{256}$ which is passed to the color network
    $c_{\theta}(\cdot)$.}
    \label{fig:occupancy_network}
\end{figure}

\boldparagraph{Occupancy Network}%
The occupancy network is implemented using $2$ residual blocks as shown in
\figref{fig:occupancy_network} and is similar to \cite{Mescheder2019CVPR}. The
input to our occupancy network is the per-part shape latent code $\bz_m^s
\in \mathbb{R}^{128}$ and a batch of 3D points sampled along rays that are
transformed to the local coordinate system of the $m$-th part. Note that we do
not apply positional encoding on the input as we empirically observed that
projecting the 3D points into a higher dimensional space using
\eqref{eq:positional_encoding} does not improve our model's performance. The
input points are passed through a fully-connected layer that produce a
$128$-dimensional feature vector for each point. This feature vector, together
with the per-part shape code $\bz_m^s$, is then passed to $2$ residual blocks.
Each residual block first applies Conditional Scaling and Translation (CST) to
the input features followed by a ReLU. The output is
then fed into a fully-connected layer, a second CST, a ReLU activation and
another fully-connected layer. The output of this operation is then added to
the input of the residual block. The Conditional Scaling and Translation layer
(CST) is the same as Conditional Batch-Normalization (CBN)
\cite{Vries2017NIPS}, using in \cite{Mescheder2019CVPR}, but we replace the 
normalization layer with the identity function. The output of the two residual blocks
is fed to a fully-connected layer that produces a $257$-dimensional output.
This is then split into a $1$-dimensional feature vector that we use to compute
the occupancy value, simply by passing it to a sigmoid function, as defined
in \eqref{eq:nerf_occupancy_network_supp}, and a $256$-dimensional feature vector, which
we pass to the color network that produces the corresponding color.

\begin{figure}[!h]
    \centering
    \includegraphics[width=\textwidth]{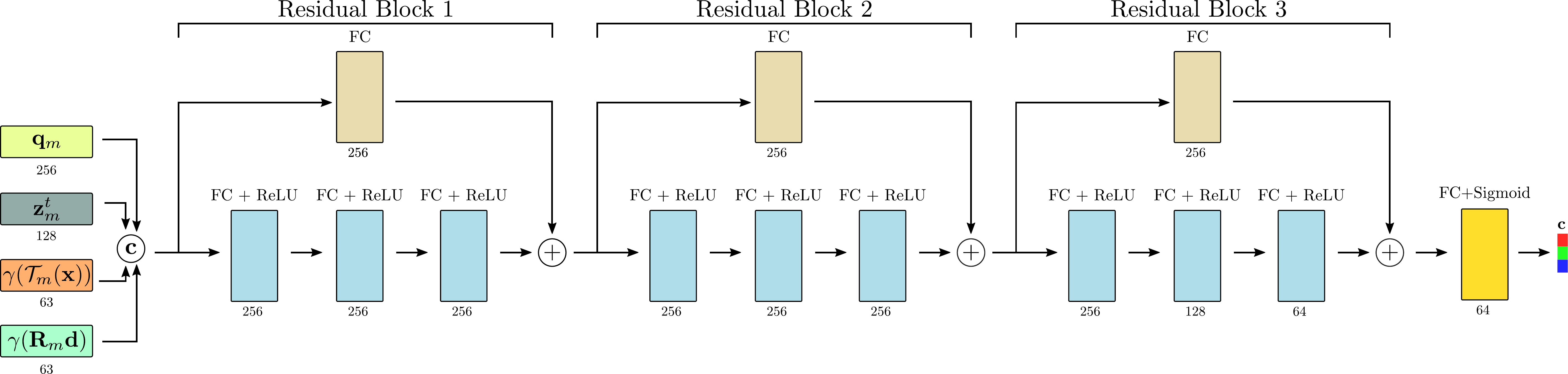}
    \caption{{\bf{Color Network.}} The color network
    $c_{\theta}(\cdot)$ maps a 3D point $\bx$ and a viewing direction $\bd$
    after applying element-wise positional encoding on their components, the 
    texture code $\bz_m^t$ and the feature vector $\bq_m$ into a color value $\bc \in
    \mathbb{R}^3$. Note that $\bq_m$ is predicted from the occupancy network $o_\theta(\cdot)$
    from $\bz_m^s$.}
    \label{fig:color_network}
\end{figure}

\begin{figure}
    \centering
    \includegraphics[width=0.8\columnwidth]{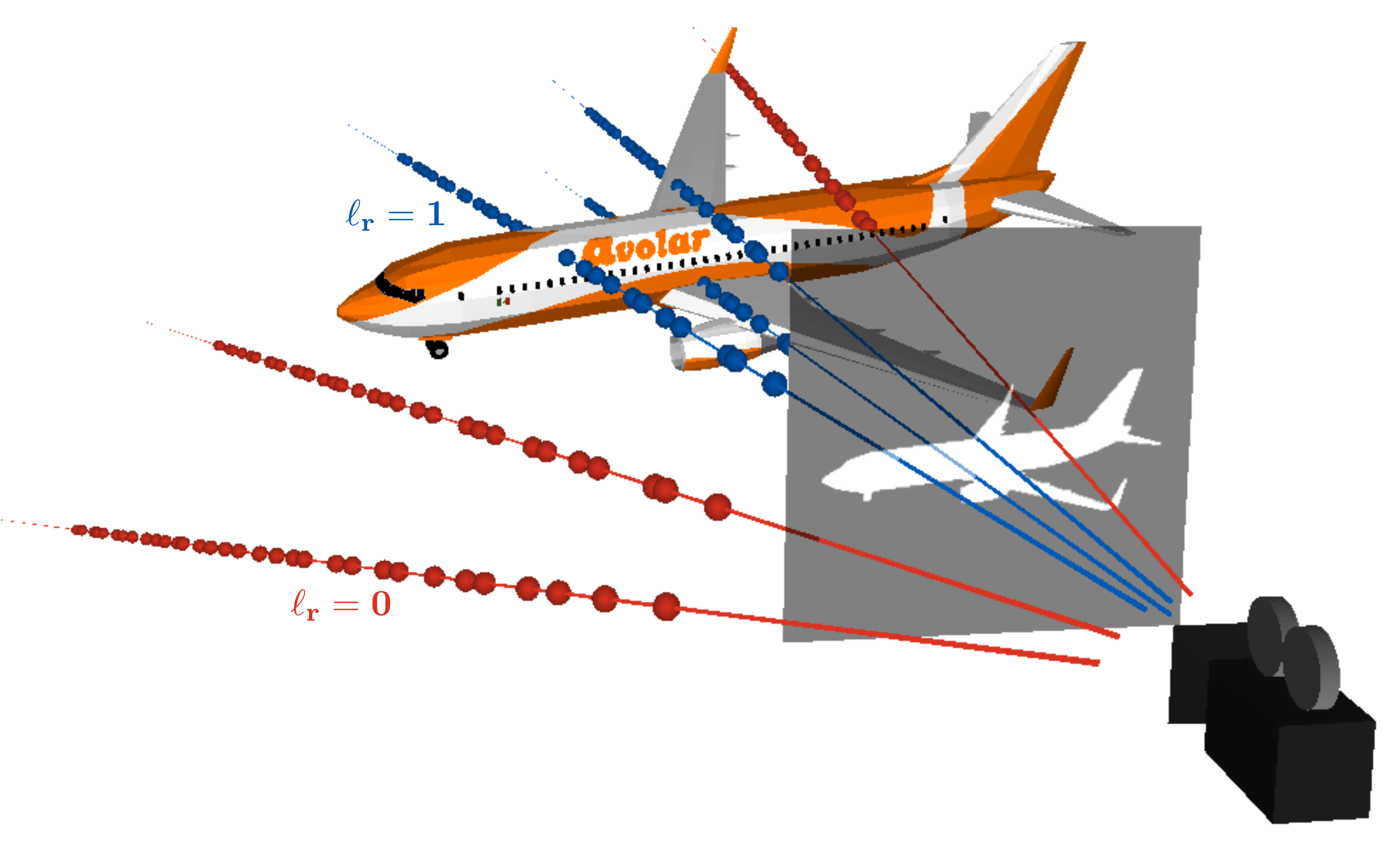}
    \caption{
        {\bf{Inside-Outside Rays.}} \darkblue{Inside} rays correspond to pixels
        within the object mask (\ie white pixel), whereas \darkred{outside} rays
        correspond pixels outside the object mask (\ie black pixel).
    }
    \label{fig:inside_outside_rays}
    \vspace{-1.2em}
\end{figure}

\boldparagraph{Color Network}%
The color network is implemented using $3$ residual blocks, as shown in
\figref{fig:color_network}. Each residual block \cite{He2016CVPR} is
implemented as a 3-layer MLP with ReLU non-linearities. The hidden dimensions
of the 3 linear layers of each block are: $510, 256, 256$ for the first,
$256,256,256$ for the second and $256, 128, 64$ for the third. The inputs to the
color network are: the per-part texture code $\bz_m^t \in \mathbb{R}^{128}$,
the 3D point $\gamma(\cT_m(\bx)) \in \mathbb{R}^{63}$ and
the ray direction $\gamma(\bR_m\bd) \in \mathbb{R}^{63}$
transformed in the coordinate system of the $m$-th part and projected in
a higher dimensional space using \eqref{eq:positional_encoding}, and the
intermediate feature representation $\bq_m \in \mathbb{R}^{256}$ that is
predicted from the occupancy network $o_\theta(\cdot)$. We concatenate all
inputs and produce a $510$-dimensional feature vector. Each residual block
first applies a linear projection to the current feature vector, followed by a
ReLU activation. The output is then fed into a fully-connected layer, a ReLU
activation and another fully-connected layer followed by a ReLU non linearity.
The output of this operation is then added to the input of the residual
block after feeding it to a fully connected layer. To predict the color, we use
one linear layer with hidden size $64$ and output size $3$, followed by a
sigmoid activation.

\subsection{Training Protocol}
\label{subsec:training}

In all our experiments, we use the Adam optimizer \cite{Kingma2015ICLR} with
batch size of $32$ and a learning rate that begins at $\eta=5\times
10^{-4}$ and decays to $\eta=5\times 10^{-6}$ over the course of optimization.
We employ a cosine annealing learning rate schedule with warm up that is set to
$500$ steps. For the other hyperparameters of Adam, we use the
PyTorch~\cite{Paszke2016ARXIV} defaults ($\beta_1=0.9$, $\beta_2=0.999$ and
$\epsilon=10^7$) and we have no weight decay. Furthermore, our input
images and object masks are rendered at $256^2$ resolution and we approximate
each image with $512$ rays. Specifically, we adopt a similar strategy as
\cite{Yu2021CVPR} and  sample an equal number of rays inside and outside the
object mask, as shown in \figref{fig:inside_outside_rays}, during training. We
observe that this sampling strategy results in faster convergence and prevents
several local minima. In addition, during training, we sample $64$ random points
along each ray, whereas during inference, we sample $128$ points. Note that
unlike \cite{Mildenhall2020ECCV}, we do not consider a coarse and fine volume
of point coordinates along rays. While, we anticipate that this would improve
the quality of our renderings, we were not able to adopt this sampling strategy
due to our limited computational resources, as it would significantly
increase the computational requirements of our model. Finally, we train our model with
$16$ parts, for each object category, for approximately $125$k iterations. All
our experiments were conducted on a single NVIDIA RTX 3090, with $24$GB and
training per category takes approximately 1 week.

We weigh the loss terms of Eq. 14 in our main submission with $1.0, 1.0, 1.0,
0.01, 0.01, 1.0, 0.0001, 0.0001$. Note that we use smaller weights for
$\cL_{overlap}(\cR)$ and $\cL_{cov}(\cR)$, as they act as regularizers on the
recovered parts and we want our model to focus primarily on generating part
arrangements that can render plausible objects from any novel camera. For
$\cL_{occ}(\cR)$, we weight the two terms of Eq. 18 in our main submission,
by $0.1, 0.01$. We empirically observe that having both terms when computing
the occupancy loss significantly helps performance. Finally, we weigh the
control loss $\cL_{control}$ with $1$ as it takes relatively smaller values and
weighting it with a smaller number would make its contribution insignificant.
The impact of each term is discussed in detail in \secref{subsec:losses}.
In addition, the $k$ term of $\cL_{cov}(\cR)$ is set to $16$, since we have
$256$ rays and $16$ parts. This encourages that each part will ``cover''
approximately the same number of rays, hence parts would uniformly cover the generated shape.
Having uniformly distributed parts across the generated shape is essential for
our application, as we want to enable uniform control over the generated shape.
Finally, we set parameter $\lambda$ of $\cL_{overlap}(\cR)$ to $3$ as we
empirically observe that it leads to good performance. Note
that setting $\lambda=1$ would enforce that each ray can intersect, \ie
be predicted as internal, only with one part, thus yielding completely
disjoint parts. However, this is quite limiting when trying to generate complex
geometries. Therefore, we instead set $\lambda=3$ in order to enable more
flexibility when capturing complex geometries.  We anticipate that setting
$\lambda=2$ would lead to similar performance, however in our experiments we
use $\lambda=3$ as it seemed to consistently yield plausible generations.

\subsection{Generation Protocol}
\label{subsec:generation}

In this section, we describe the sampling process used for the generation of
new shapes. In particular, we follow \cite{Hertz2022SIGGRAPH} and draw
a set of shape and texture latent codes $\{\bz^s, \bz^t \}$ from a multivariate normal distribution. Namely,
\begin{equation}
    \begin{aligned}
	\bz^s \sim \cN(\bmu_{train}^s,\bSigma_{train}^s) \\
	\bz^t \sim \cN(\bmu_{train}^t,\bSigma_{train}^t).
        \label{eq:generation}
    \end{aligned}
\end{equation}
Here, the mean $(\bmu_{train}^s, \bmu_{train}^t)$ and covariance
$(\bSigma_{train}^s, \bSigma_{train}^t)$ values are calculated from the shape
and texture embeddings learnt during training. At inference/generation, we
randomly sample $\bz_s$ and $\bz_t$ and for a given camera we cast rays and
sample points along these rays. For each point we predict the
corresponding color and opacity value. Note that during generation, our model
does not need to condition on the object mask to be generated. To generate a
novel shape from a novel camera, our model associates the casted rays with the
learned parts as described in Eq. 9 of our main submission and performs
rendering. Note that while for training we sample $64$ points along each ray, for
generation we increase this number to $128$ in order to improve the generation
quality.

\subsection{Editing}
\label{subsec:editing}

In this section, we provide additional information regarding the editing
capabilities of our model and analyze how the editing operations are performed.
As we are interested in editing specific parts of the object, we want
to be able to modify the pose, size and appearance of each part independently.
This can be enforced by making sure that each NeRF/part receives geometric inputs in
its own local coordinate system, which is enforced by augmenting each part with:
\begin{enumerate}    \item an \emph{affine transformation} $T_m(\bx) = \bR_m (\bx + \bt_m)$ that
    maps a 3D point $\bx$ to the local coordinate system of the part, where
    $\bt_m \in \real^3$ is the translation vector and $\bR_m \in SO(3)$ is the
    rotation matrix;
	\item a \emph{scale} vector $\bs_m \in \real^3$, representing spatial extent;
	\item latent codes: \emph{shape} $\bz_m^s \in \mathbb{R}^{128}$ and \emph{texture} $\bz_m^t \in \mathbb{R}^{128}$.
\end{enumerate}
To select and edit a specific part of a 3D shape, it suffices to select its corresponding latent
vectors $\bz_m^s$ and $\bz_m^t$ and manipulate them based on the type of the
editing operation. Below, we provide more details regarding the editing
operations presented in the main paper.

\boldparagraph{Rigid Transformations}%
To apply a rigid transformation on a specific part of a shape, it suffices to
alter the per-part translation vector $\bt_m$ and rotation matrix $\bR_m$. For
example, to translate a part we add an appropriate  displacement vector to the
$\bt_m$. Similarly, to rotate a part, we can multiply $\bR_m$ with the desired
rotation matrix.

\boldparagraph{Non-Rigid Transformations}%
Similarly, a non-rigid transformation that changes the size of a specific part can be implemented
by multiplying the part's rotation matrix with an appropriate scale matrix.

\boldparagraph{Geometry Mixing}%
To perform geometry mixing, we select part shape latent codes $\bz_m^s$ from an
arbitrary number of initial shapes, forming a set of part shape latent codes
that describe the new mixed shape. To generate a shape with a desired texture, we allow
selecting the part texture latent codes either from a single shape or from an arbitrary
number of shapes. For example, to generate the results of the shape mixing experiment
presented in Section 4.3 of our main submission and Fig. 10, we select texture
codes only from Shape 1 for the chair and the airplane in the third column, whereas we select
texture codes from both shapes to generate the mixed shapes for the chair and
the airplane in the last column.
As soon as the selection process is done, we feed the new part latent codes to
the Neural Rendering module and render the new mixed shape. Note that the
resulting number of parts in the generated shape can be arbitrary, as our
model's occupancy and color networks can produce meaningful results for any
given number of parts.

\boldparagraph{Texture Mixing}%
For texture mixing, we follow the same process as in the geometry mixing
procedure, and select part shape and texture latent codes either from a single shape, or
from multiple shapes and feed them to the Neural Rendering module to generate
the new shape. To generate the results for the experiment presented in Section 4.3 and Fig. 10
of our main submission, we select the shape codes from Shape 1 and for the legs
of the chair, we select the texture codes of the legs of Shape 2.

\boldparagraph{Part Addition}%
Part addition can be seen as the equivalent operation of our shape mixing
strategy, where we select all the part latent codes from an entire shape, and
append extra part geometry and texture codes from other shapes. Similarly to the
previously discussed editing operations, the per-part shape and texture codes
of the new shape after the addition are fed to the Neural Rendering module that
renders the new shape.

\boldparagraph{Part Removal}%
To remove parts from a shape, it suffices to remove the corresponding
part shape and texture embeddings, and feed only the remaining part embeddings to
the Neural Rendering module.

\boldparagraph{Color Change}%
Our hard ray-part assignment enables changing the colors of selected parts, by
directly replacing the predicted colors of the associated part rays with a
color of our preference.

\boldparagraph{Part-Level Interpolation}
For the part-level interpolation operation, we select corresponding parts from
two different shape instances, and linearly interpolate the part shape and
texture embeddings, producing new part shape and texture latent codes. The interpolated
vectors are then passed to the Neural Rendering module to generate a new shape.

\subsection{Metrics}
\label{subsec:metrics}

As mentioned in the main submission, we evaluate our model and our baselines
using two metrics that measure the plausibility and the diversity of set of
generated shapes $G$ in comparison to a set of reference shapes $R$. In
particular, we report the Coverage score (COV) and the Minimum Matching
Distance (MMD) \cite{Achlioptas2018ICML} using Chamfer Distance (CD).  MMD
measures the \emph{quality of the generated shapes} by computing how likely it
is that a generated shape looks like a shape from the reference set of shapes.
On the other hand, COV measures how many shape variations are covered by the
generated shapes, by computing the percentage of reference shapes that are closest
to at least one generated shape.

To ensure fair comparison with our baselines, when we compare with NeRF-based
generative models (\ie Table $1$ in our main submission), we follow
\cite{Gao2022NEURIPS}, using the test set as the set of reference shapes and
synthesizing five times as many generated shapes.
In contrary, when comparing to part-based generative models (\ie Table
$2$ in our main submission), we follow \cite{Hertz2022SIGGRAPH} and randomly
generate $1000$ shapes which are compared to $500$ shapes from the training set
and $500$ shapes from the test set.

To estimate the similarity between
two shapes from the two sets, we use the Chamfer Distance (CD), which is simply
the distance between a set of points sampled on the surface of the reference
and the generated mesh. Namely, given a set of $N$ sampled points on the
surface of the reference $\cX=\{\bx_i\}_{i=1}^N$ and the generated shape
$\cY=\{\by_i\}_{i=1}^N$ the Chamfer Distance (CD) becomes
\begin{equation}
    {\text{CD}}(\cX, \cY) = \frac{1}{N} \sum_{\bx \in \cX}
    \min_{\by \in \cY}\norm{\bx - \by}_2^2 +
    \frac{1}{M} \sum_{\by \in \cY}
    \min_{\bx \in \cX}\norm{\by - \bx}_2^2.
    \label{eq:chamfer}
\end{equation}
For both our NeRF-based and part-based baselines, we set $N=2048$.

The Minimum Matching Distance (MMD) is the average distance
between each shape from the generated set $G$ to its closest shape in the reference
set $R$ and can be defined as:
\begin{equation}
    {\text{MMD}}(G, R) = \frac{1}{\lvert R \rvert} \sum_{\cX \in R} \min_{\cY \in G} {\text{CD}}(\cX, \cY).
    \label{eq:mmd}
\end{equation}
Intuitively, MMD measures how likely it is that a generated shape is similar to
a reference shape in terms of Chamfer Distance and is a metric of the plausibility
of the generated shapes. Namely, a high MMD score indicates that the shapes in
the generated set $G$ faithfully represent the shapes in the reference set $R$.

The Coverage score (COV) measures the percentage of shapes in the reference set that are
closest to each shape from the generated set. In particular, for each shape in
the generated set $G$, we assign its closest shape from the reference set $R$.
In our measurement, we only consider shapes from $R$ that are closest to at
least one shape in $G$. Formally, COV is defined as
\begin{equation}
    {\text{COV}}(G, R) = \frac{\lvert \{\argmin_{\cX \in R}{\text{CD}}(\cX, \cY) \mid \cY \in G \}\rvert}{\lvert G \rvert}
    \label{eq:coverage}
\end{equation}
Intuitively, COV measures the diversity of the generated shapes in comparison
to the reference set. In other words, a high Coverage indicates that most
of the shapes in the reference set $R$ are roughly represented by the set of
generated shapes $G$. To ensure a fair comparison, with our baselines, we use
the evaluation code of \cite{Gao2022NEURIPS}.

\subsection{Baselines}
\label{subsec:baselines}

In this section, we provide additional details regarding our baselines. To the
best of our knowledge, our work is the first part-aware generative model that
does not require explicit 3D supervision, thus there are no other works that are
directly comparable to our model. Therefore, in our evaluation, we consider
three types of baselines: NeRF-based generative models
\cite{Schwarz2020NEURIPS, Chan2021CVPR, Chan2022CVPR} that are part agnostic,
part-based generative models \cite{Hao2020CVPR, Hertz2022SIGGRAPH} that require
explicit 3D supervision in the form of watertight meshes and can only generate
shapes without textures and finally the concurrent GET3D~\cite{Gao2022NEURIPS} 
that relies on differentiable rendering and again does not consider any parts.
A concise comparison between our model and these baselines is provided in
\tabref{tab:comparison_prior_works}. Furthermore, we also compare our model
with our baselines \wrt the editing capabilities of each model in
\tabref{tab:comparison_prior_works_editing}.

\begin{table}
    \begin{center}
    \begin{tabular}{l|c|c|c|c|c}
        \toprule
        Method & Rendering & Representation & Textures & Supervision & Parts \\
        \toprule
        GET3D~\cite{Gao2022NEURIPS} & Differentiable & Mesh & \cmark & 2D & \xmark\\
        \midrule
        GRAF~\cite{Schwarz2020NEURIPS} & Volumetric & Neural Field & \cmark & 2D & \xmark\\
        Pi-GAN~\cite{Chan2021CVPR} & Volumetric & Neural Field & \cmark & 2D & \xmark\\
        EG3D~\cite{Chan2022CVPR} & Volumetric & Neural Field & \cmark & 2D & \xmark\\
        \midrule
        DualSDF~\cite{Hao2020CVPR} & - & Implicit SDF & \xmark & 3D & \cmark\\
        SPAGHETTI~\cite{Hertz2022SIGGRAPH} & - & Implicit Occupancy & \xmark & 3D & \cmark\\
        \midrule
        Ours & Volumetric & Neural Field & \cmark & 2D & \cmark\\
        \bottomrule
    \end{tabular}
    \end{center}
    \vspace{-0.2em}
    \caption{{\bf Comparison to Prior Work.} We introduce the first part-based generative model that does
    not require explicit 3D supervision. We compare our model with NeRF-based
    generative models \cite{Schwarz2020NEURIPS, Chan2021CVPR, Chan2022CVPR}, part-based generative models 
    \cite{Hao2020CVPR, Hertz2022SIGGRAPH} and the concurrent
    \cite{Gao2022NEURIPS} that unlike our model relies on differentiable
    rendering.
    }
    \label{tab:comparison_prior_works}
    \vspace{-0.2em}
\end{table}

\boldparagraph{GET3D}%
In concurrent work,
GET3D~\cite{Gao2022NEURIPS}\footnote{\href{https://nv-tlabs.github.io/GET3D/}{https://nv-tlabs.github.io/GET3D/}},
proposed a novel generative model for textured meshes that relies on a 
highly optimized differentiable graphics renderer \cite{Laine2020SIGGRAPH}. In particular, their model
comprises a geometry generator that employs DMTet~\cite{Gao2020NeurIPS} to
recover surfaces of arbitrary topologies and a texture generator that generates
textures as a texture field \cite{Oechsle2021ICCV}. During training, they rely
on two discriminator losses on the rendered image and the 2D silhouette. As the
underlying representation of GET3D is a textured mesh, their generated textures
and geometries are more detailed compared to ours. However, as they do not
consider parts, they cannot perform several editing operations that our model
is capable of. As already mentioned in our main submission, training GET3D requires
approximately 16 GPU days (8 NVIDIA A100 for 2 days), whereas training our
model takes approximately one third of the time (5 GPU days). Due to their rigorous
evaluation and to ensure a fair comparison, we report their results in Table 1
and Figure 6 in our main submission.

\boldparagraph{GRAF}%
GRAF~\cite{Schwarz2020NEURIPS}\footnote{\href{https://github.com/autonomousvision/graf}{https://github.com/autonomousvision/graf}}
was among the first 3D-aware generative models that combined a 2D
GAN~\cite{Goodfellow2014NIPS} with volumetric rendering as in
NeRF~\cite{Mildenhall2020ECCV}. In particular, they introduce a conditional
NeRF parametrized as an MLP that maps the positional encoding of a 3D point and
a viewing direction to a color and a volume density value, conditioned on a
shape and an appearance latent codes. Unlike \cite{Mildenhall2020ECCV}, GRAF is
trained from a collection of unposed images and employs a
PatchGAN~\cite{Isola2017CVPR} discriminator to improve the quality of the
rendered images from a novel view. While GRAF's code is publicly available, to
generate the qualitative results for Fig 6 and the quantitative comparison in
Table 1 from our main submission, we do not train their model. Instead, we
directly report the results from \cite{Gao2022NEURIPS}. Unlike our model, GRAF
employs a discriminator to improve the quality of the rendered images from
novel view points. Furthermore, as \cite{Schwarz2020NEURIPS} does not consider
parts, it only enables few editing operations on the entire image, \ie changing
the color of the depicted object.

\boldparagraph{Pi-GAN}%
Similar to GRAF~\cite{Schwarz2020NEURIPS},
Pi-GAN~\cite{Chan2021CVPR}\footnote{\href{https://marcoamonteiro.github.io/pi-GAN-website/}{https://marcoamonteiro.github.io/pi-GAN-website/}}
addresses the 3D-aware image synthesis task and leverages a neural
representation with periodic activation functions~\cite{Sitzmann2020NEURIPS}. Unlike \cite{Schwarz2020NEURIPS}
that conditions on a shape and appearance latent code, Pi-GAN conditions the MLP
on a single input noise vector, as in StyleGAN~\cite{Karras2020CVPR}. Also,
Pi-GAN uses a discriminator to further improve the quality of the volumetric
rendering. As they do not consider parts, \cite{Chan2021CVPR} have limited editing capabilities.
Again, to ensure a fair comparison, we do not train their model from scratch
but instead report the results from \cite{Gao2022NEURIPS}.

\boldparagraph{EG3D}%
More recently, EG3D~\cite{Chan2022CVPR}\footnote{\href{https://nvlabs.github.io/eg3d/}{https://nvlabs.github.io/eg3d/}}
proposed a StyleGAN-based generator \cite{Karras2020CVPR} that relies on a tri-plane representation instead of
positional encoding. Their model is capable of generating images at high
resolutions using a CNN-based upsampling module. Despite its impressive results, EG3D does not consider parts, hence
it cannot perform local edits on the generated shapes. Similar to \cite{Schwarz2020NEURIPS, Chan2021CVPR},
for the qualitative and quantitative evaluation in Sec. 4.2 of our main submission, we report the results
from \cite{Gao2022NEURIPS}. In addition, note that as EG3D~\cite{Chan2022CVPR}
requires training of approximately $8$ days on $8$ NVIDIA Tesla V100 GPUs, we are not able to run their code due to our
limited GPU resources.

\boldparagraph{DualSDF}%
DualSDF~\cite{Hao2020CVPR}\footnote{\href{https://www.cs.cornell.edu/~hadarelor/dualsdf/}{https://www.cs.cornell.edu/~hadarelor/dualsdf/}}
was among the first to introduce a model capable of editing neural implicit shapes. In particular, they
proposed a representation with two levels of granularity, where a user can manipulate a 3D shape through
the coarse primitive-based representation and these edits are reflected to a
high-resolution implicit shape, via a shared latent code. Unlike our model,
DualSDF assumes explicit 3D supervision in the form of a watertight mesh and cannot generate shapes
with texture. To ensure a fair comparison, we report the results from
\cite{Hertz2022SIGGRAPH} in Table 2 of our main submission. Although our model
considers 2D supervision in the form of posed images and object masks, our
model outperforms DualSDF in terms of MMD and COV both for airplanes and tables
(see Table 2 in our main submission).

\boldparagraph{SPAGHETTI}%
SPAGHETTI~\cite{Hertz2022SIGGRAPH}\footnote{\href{https://github.com/amirhertz/spaghetti}{https://github.com/amirhertz/spaghetti}}
is the state-of-the-art part-based generative model that similar to
\cite{Hao2020CVPR} permits manipulating neural implicit shapes. Similar to our
model, they implement their generative model as an auto-decoder
\cite{Bojanowski2018ICML, Park2019CVPR} that takes an instant-specific latent
code as input and decomposes it to $M$ per-part latent codes, representing
different parts of the object. Unlike our model,
SPAGHETTI~\cite{Hertz2022SIGGRAPH}, employs a transformer decoder that merges
the parts and produces the final implicit shape as an occupancy field. Unlike
our model, \cite{Hertz2022SIGGRAPH} assumes explicit 3D supervision in the form
of a watertight mesh and cannot generate meshes with texture. As already
mentioned previously, to ensure a fair comparison, the numbers in Table 2 of
our main submission are from \cite{Hertz2022SIGGRAPH}. For the qualitative
comparison in Fig. 7 of our main submission, we use the pre-trained models
provided to us by the authors.

\begin{table}
    \begin{center}
    \begin{tabular}{l|c|c|c|c|c|c}
        \toprule
        Method & Image Inversion & Shape Inversion & Generation & Shape Mixing & Shape Editing & Texture Editing \\
        \toprule
        GET3D~\cite{Gao2022NEURIPS} & \cmark & \cmark & \cmark & \xmark & \xmark & \xmark\\
        \midrule
        GRAF~\cite{Schwarz2020NEURIPS} & \cmark & \cmark & \cmark & \xmark & \xmark & \xmark\\
        Pi-GAN~\cite{Chan2021CVPR} & \cmark & \cmark & \cmark & \xmark & \xmark & \xmark\\
        EG3D~\cite{Chan2022CVPR} & \cmark & \cmark & \cmark & \xmark & \xmark & \xmark\\
        \midrule
        DualSDF~\cite{Hao2020CVPR} & \xmark & \cmark & \cmark & \xmark & \cmark & \xmark\\
        SPAGHETTI~\cite{Hertz2022SIGGRAPH} & \xmark & \cmark & \cmark & \cmark & \cmark & \xmark\\
        \midrule
        Ours & \cmark & \cmark & \cmark & \cmark & \cmark & \cmark \\
        \bottomrule
    \end{tabular}
    \end{center}
    \vspace{-0.2em}
    \caption{{\bf Comparison to Prior Work \wrt Editing Operations.} We compare
    the editing capabilities of our model and our baselines. Note that
    \emph{Shape Editing} and \emph{Texture Editing} refer to {\bf{local}} changes
    on the texture and the shape of the generated object.
    }
    \label{tab:comparison_prior_works_editing}
    \vspace{-0.2em}
\end{table}

\subsection{Mesh Extraction}

To extract meshes from the predicted occupancy field, we
employ the Multiresolution IsoSurface Extraction (MISE) technique introduced in
\cite{Mescheder2019CVPR}. In particular, we start from a voxel grid of $32^3$
initial resolution for which we predict occupancy values. Next, we follow the
process proposed in \cite{Mescheder2019CVPR} and extract the approximate
isosurface with Marching Cubes \cite{Lorensen1987SIGGRAPH}, which is then
refined using $3$ optimization steps. For the mesh extraction, we use the code
provided by Mescheder \etal \cite{Mescheder2019CVPR}. Note that for the
individual parts, we only extract approximate surfaces with Marching Cubes
\cite{Lorensen1987SIGGRAPH}.

\clearpage
\section{Ablation Study}
\label{sec:ablations}

In this section, we ablate various components of our model and demonstrate
their impact on its generation capabilities. We conduct all our ablations on a subset of
ShapeNet~\cite{Chang2015ARXIV} Airplanes, which comprises approximately $10\%$
of the original train set. Namely, we use $178$ training samples and $100$
views for training all our model variants. In \secref{subsec:losses}, we
discuss the effect of our loss terms on the overall performance of our model
and in \secref{subsec:n_parts}, we demonstrate the impact of the number
of parts on the generation capabilities of our model. Finally, we provide
additional information regarding the impact of our hard ray-part assignment,
presented in our main submission (see \secref{subsec:hard_ray_part_assignment})

\subsection{Loss Functions}
\label{subsec:losses}

Our optimization objective $\cL$ is the sum of six terms combined with two
regularizers on the shape and texture embeddings $\bz^s, \bz^t$, namely
\begin{equation}
	\cL = 
		\cL_{rgb}(\cR) + \cL_{mask}(\cR) +
		\cL_{occ}(\cR) + \cL_{cov}(\cR) +
		\cL_{overlap}(\cR) + \cL_{control} +
		\norm{\bz^s}_2 + \norm{\bz^t}_2.
        \label{eq:optimization_objective_supp}
\end{equation}
As supervision, we use the observed RGB color
$C(r) \in \mathbb{R}^3$ and the object mask $I(r) \in \{0, 1\}$ for each ray $r
\in \cR$. In addition, we associate $r$ with a binary label $\ell_r = I(r)$, as shown
in \figref{fig:inside_outside_rays}.  Namely, we label a ray $r$ as
\emph{inside}, if $\ell_r=1$ and \emph{outside} if $\ell_r=0$.
In this section, we discuss how the loss terms affect the performance of our
model \wrt the plausibility of the generated shapes. In particular, we train $4$
variants of our model and for each one we omit one of the loss terms. We
provide both quantitative (see \tabref{tab:ablation}) and qualitative
comparison (see \figref{fig:ablation_supp}) for these scenarios.
For a fair comparison, all models are trained for the same number of epochs,
which is set to $50$. Furthermore, in this experiment, we set the number of
parts to $10$. To compute the metrics in \tabref{tab:ablation}, we consider a
subset of the test set, which amounts to $46$ shapes, namely roughly $10\%$ of the original
test set.
\begin{table}[h!]
    \centering
    \begin{tabular}{l||cccc|c}
        \toprule
        & w/o $\cL_{mask}$ & w/o  $\cL_{occ}$ & w/o $\cL_{cov}$ & w/o $\cL_{overlap}$ & Ours \\
        \midrule
        MMD-CD ($\downarrow$) & 1.787 & 1.823 & 1.737 & 1.757 & 1.732 \\
        \bottomrule
    \end{tabular}
    \caption{{\bf Ablation Study on Loss Terms.} We investigate the impact of
    each loss term by training our model without each one of them. We report
    the MMD-CD $(\downarrow)$ for all variants.}
    \label{tab:ablation}
\end{table}

{
\renewcommand{\thesubfigure}{\arabic{subfigure}}

\begin{figure}
    \begin{subfigure}[b]{\columnwidth}
        \begin{subfigure}[t]{0.20\columnwidth}
            \centering
            \caption{w/o $\cL_{mask}$}\vspace{-0.1em}
            \includegraphics[width=\textwidth]{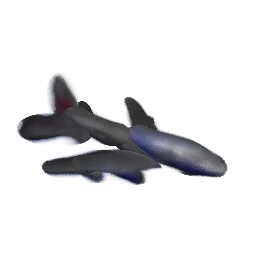}\vspace{-2.5em}
            \includegraphics[width=\textwidth]{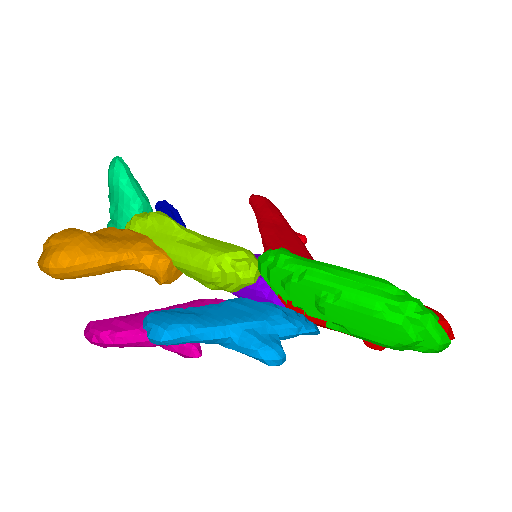}
        \end{subfigure}%
        \begin{subfigure}[t]{0.20\columnwidth}
            \centering
            \caption{w/o  $\cL_{occ}$}\vspace{-0.1em}
            \includegraphics[width=\textwidth]{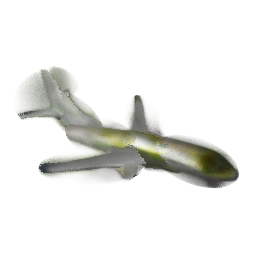}\vspace{-2.5em}
            \includegraphics[width=\textwidth]{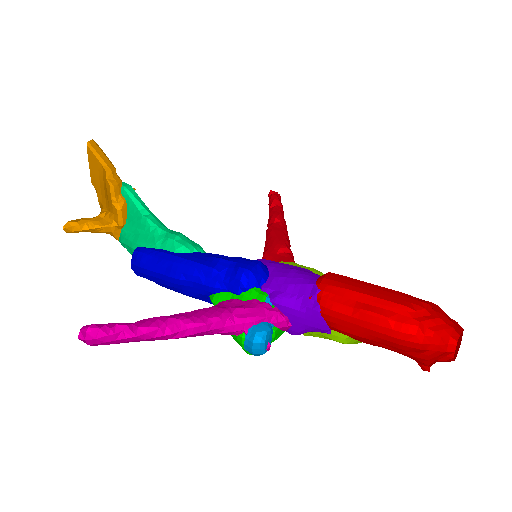}
        \end{subfigure}%
        \begin{subfigure}[t]{0.20\columnwidth}
            \centering
            \caption{w/o $\cL_{cov}$}\vspace{-0.1em}
            \includegraphics[width=\textwidth]{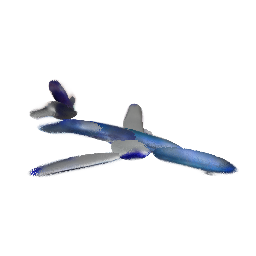}\vspace{-2.5em}
            \includegraphics[width=\textwidth]{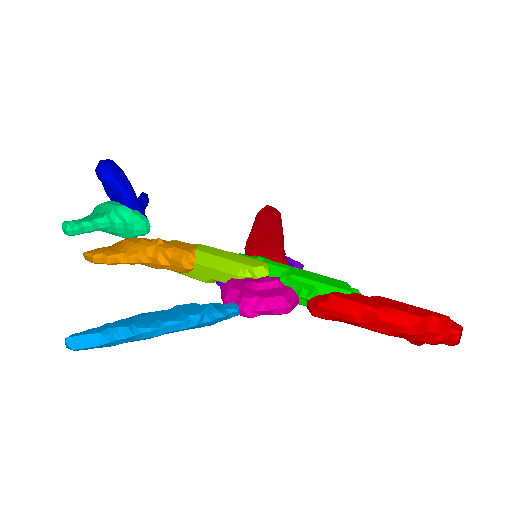}
        \end{subfigure}%
        \begin{subfigure}[t]{0.20\columnwidth}
            \centering
            \caption{w/o $\cL_{overlap}$}\vspace{-0.1em}
            \includegraphics[width=\textwidth]{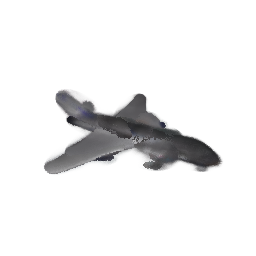}\vspace{-2.5em}
            \includegraphics[width=\textwidth]{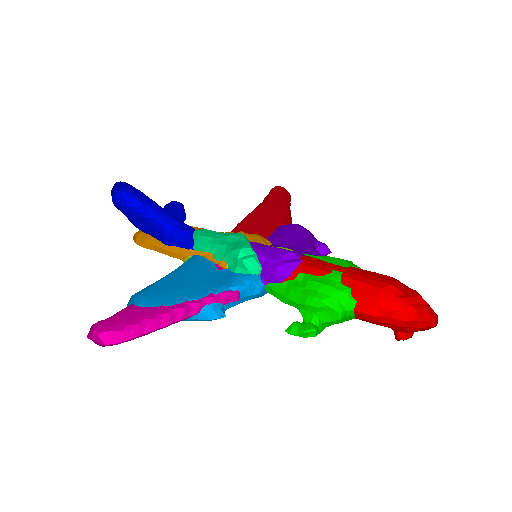}
        \end{subfigure}%
        \begin{subfigure}[t]{0.20\columnwidth}
            \centering
            \caption{Ours}\vspace{-0.1em}
            \includegraphics[width=\textwidth]{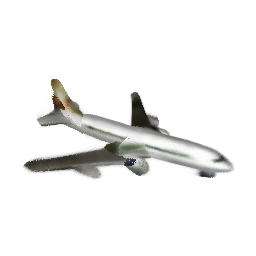}\vspace{-2.5em}
            \includegraphics[width=\textwidth]{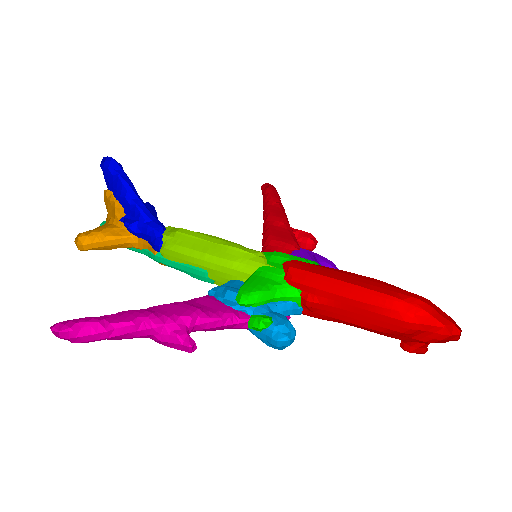}
        \end{subfigure}
    \end{subfigure}\\[0em]
    \begin{subfigure}[b]{\columnwidth}
        \begin{subfigure}[t]{0.20\columnwidth}
            \centering
            \caption{w/o $\cL_{mask}$}\vspace{-0.1em}
            \includegraphics[width=\textwidth]{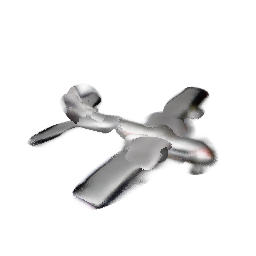}\vspace{-2.0em}
            \includegraphics[width=\textwidth]{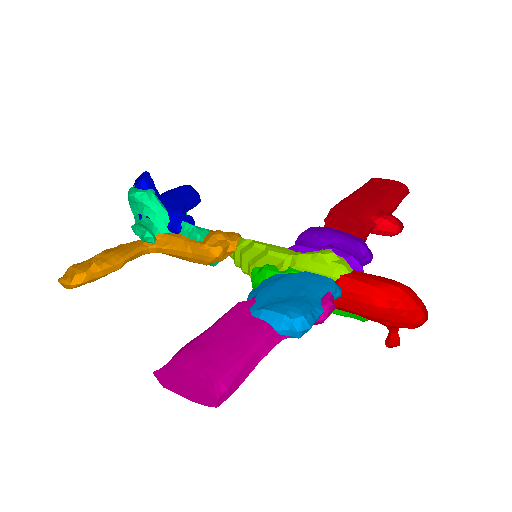}
        \end{subfigure}%
        \begin{subfigure}[t]{0.20\columnwidth}
            \centering
            \caption{w/o  $\cL_{occ}$}\vspace{-0.1em}
            \includegraphics[width=\textwidth]{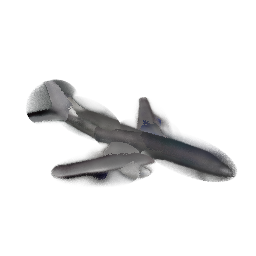}\vspace{-2.0em}
            \includegraphics[width=\textwidth]{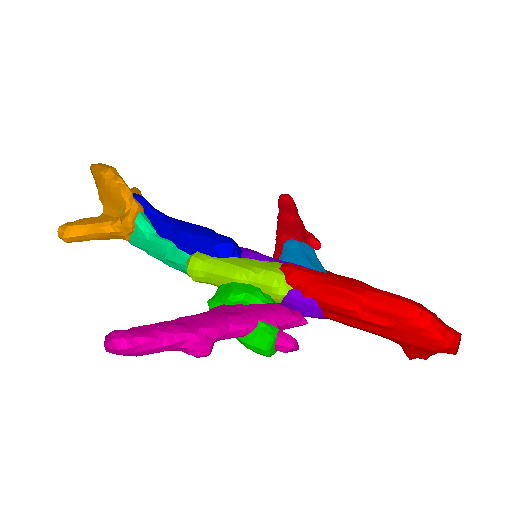}
        \end{subfigure}%
        \begin{subfigure}[t]{0.20\columnwidth}
            \centering
            \caption{w/o $\cL_{cov}$}\vspace{-0.1em}
            \includegraphics[width=\textwidth]{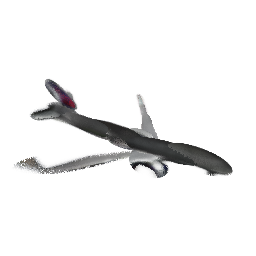}\vspace{-2.0em}
            \includegraphics[width=\textwidth]{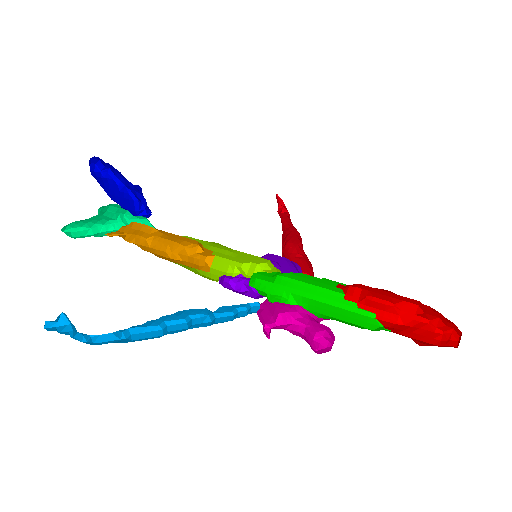}
        \end{subfigure}%
        \begin{subfigure}[t]{0.20\columnwidth}
            \centering
            \caption{w/o $\cL_{overlap}$}\vspace{-0.1em}
            \includegraphics[width=\textwidth]{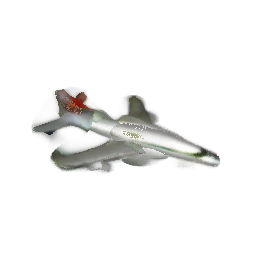}\vspace{-2.0em}
            \includegraphics[width=\textwidth]{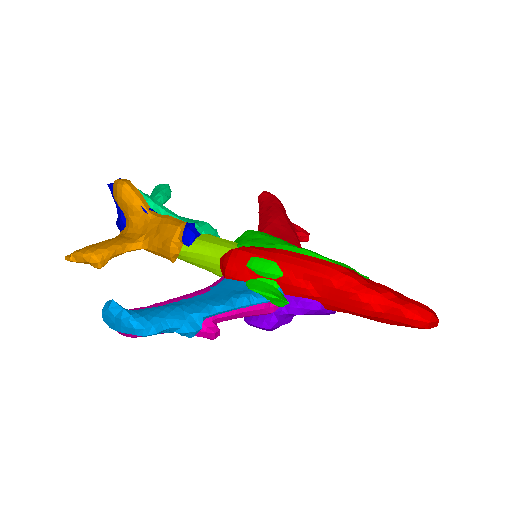}
        \end{subfigure}%
        \begin{subfigure}[t]{0.20\columnwidth}
            \centering
            \caption{Ours}\vspace{-0.1em}
            \includegraphics[width=\textwidth]{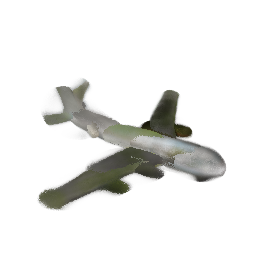}\vspace{-2.0em}
            \includegraphics[width=\textwidth]{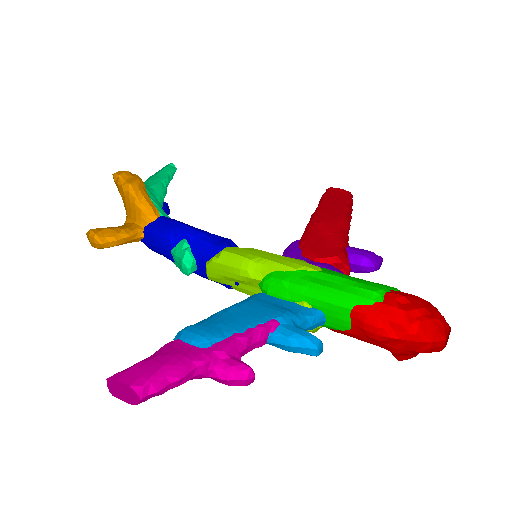}
        \end{subfigure}
    \end{subfigure}\\[0em]
    \begin{subfigure}[b]{\columnwidth}
        \begin{subfigure}[t]{0.20\columnwidth}
            \centering
            \caption{w/o $\cL_{mask}$}\vspace{-0.1em}
            \includegraphics[width=\textwidth]{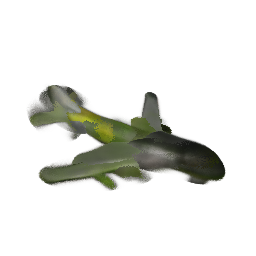}\vspace{-2.0em}
            \includegraphics[width=\textwidth]{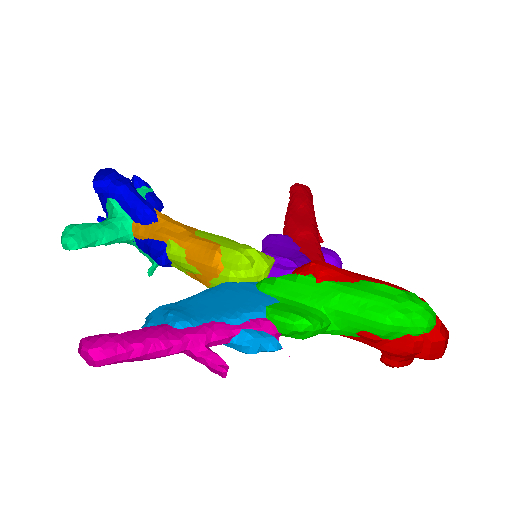}
        \end{subfigure}%
        \begin{subfigure}[t]{0.20\columnwidth}
            \centering
            \caption{w/o  $\cL_{occ}$}\vspace{-0.1em}
            \includegraphics[width=\textwidth]{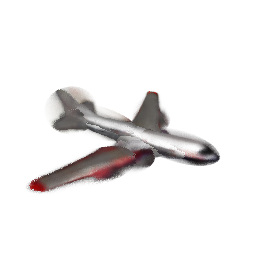}\vspace{-2.0em}
            \includegraphics[width=\textwidth]{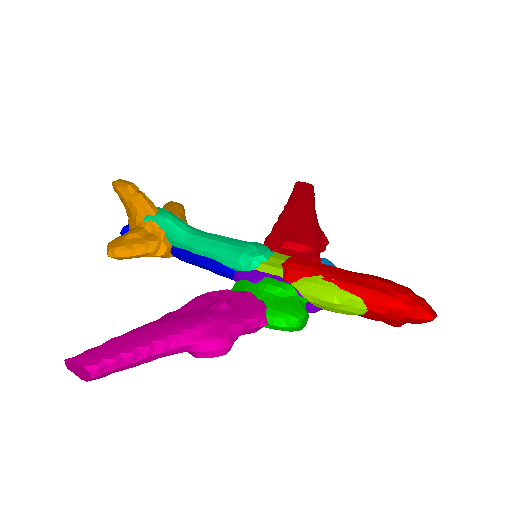}
        \end{subfigure}%
        \begin{subfigure}[t]{0.20\columnwidth}
            \centering
            \caption{w/o $\cL_{cov}$}\vspace{-0.1em}
            \includegraphics[width=\textwidth]{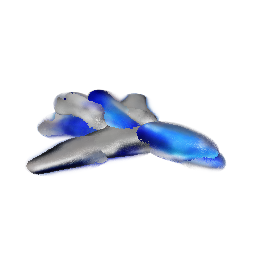}\vspace{-2.0em}
            \includegraphics[width=\textwidth]{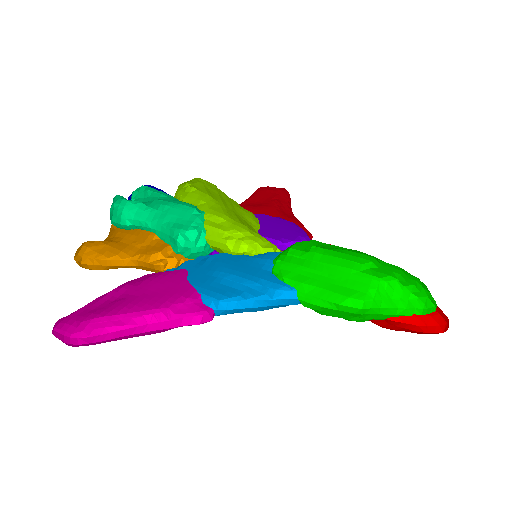}
        \end{subfigure}%
        \begin{subfigure}[t]{0.20\columnwidth}
            \centering
            \caption{w/o $\cL_{overlap}$}\vspace{-0.1em}
            \includegraphics[width=\textwidth]{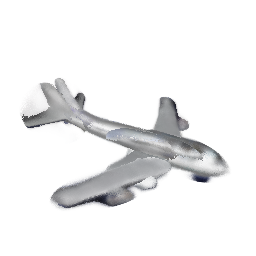}\vspace{-2.0em}
            \includegraphics[width=\textwidth]{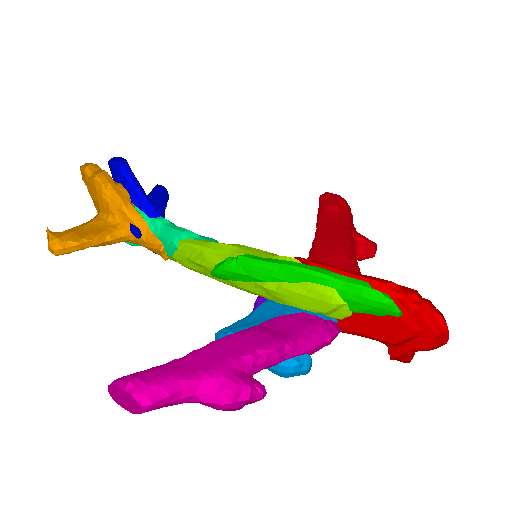}
        \end{subfigure}%
        \begin{subfigure}[t]{0.20\columnwidth}
            \centering
            \caption{Ours}\vspace{-0.1em}
            \includegraphics[width=\textwidth]{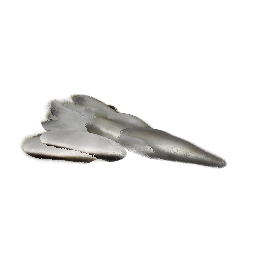}\vspace{-2.0em}
            \includegraphics[width=\textwidth]{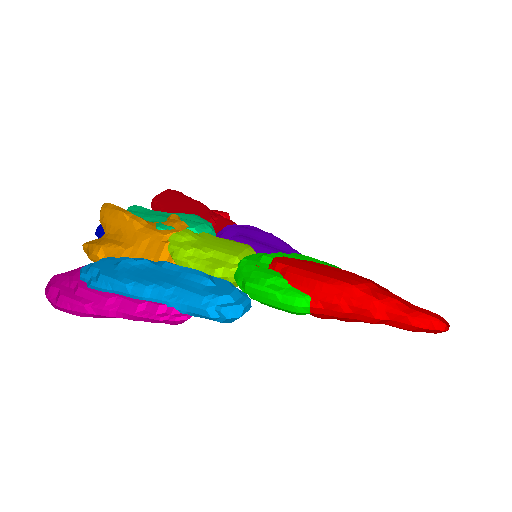}
        \end{subfigure}
    \end{subfigure}\\[0em]
    \caption{{\bf Ablation Study on Loss Terms.} Qualitative evaluation of the
    impact of the $4$ loss terms on the performance of our model. We train our
    model, without each loss term and generate novel shapes.}
    \label{fig:ablation_supp}
\end{figure}
}

\boldparagraph{w/o Mask Loss}%
The mask loss is the squared error between the observed and the rendered
pixel value of the object mask for a set of rays. We observe that removing the
mask loss $\cL_{mask}(\cR)$ results in more noisy/blurry renderings, in particular
around the surface boundaries (see first column in \figref{fig:ablation_supp}).
We experimentally observed that $\cL_{mask}(\cR)$ helps the network
find the object boundaries faster, which naturally improves the overall
rendering quality. On the part-level, we also observe that the generated parts
are less smooth (\eg see the tail parts of the two last airplanes in the first
column of \figref{fig:ablation_supp}).

\boldparagraph{w/o Occupancy Loss}%
The occupancy loss makes sure that the generated shape does not occupy empty
space. This is enforced by making sure that the generated parts only
contain/cover \emph{inside rays}, namely rays with $\ell_r=1$. Therefore,
removing $\cL_{occ}(\cR)$ from our optimization objective results in
renderings of worse quality, particularly around the object boundaries (see
second column in \figref{fig:ablation_supp}). Note that in our evaluation, even
if we remove this loss, we still use $\cL_{mask}(\cR)$, which forces, to some
extent, the network to capture the surface boundary.

\boldparagraph{w/o Coverage Loss}%
The coverage loss prevents degenerate parts, that are either too thin or
too large. Therefore, when we remove $\cL_{cov}(\cR)$ from our optimization, we
note that the generated parts become either very small (first and second
example in third column of \figref{fig:ablation_supp}) or too large (last example in the same column).

\boldparagraph{w/o Overlapping Loss}%
The overlapping loss encourages the generated parts to not capture the same
regions of the object. We identify overlapping parts by a ray being
internal to more than $\lambda$ parts and we set $\lambda=3$. Note that this done using the predicted 
label for ray $r$ \wrt to part $m$,
$\displaystyle \hat{\ell}_r^m = \max_{\bx_i^r \in \cX_r}
h_\theta^m(\bx_i^r)$.
Removing $\cL_{overlap}(\cR)$ results in generated parts that capture the same regions of the object,
as shown in the fourth column of \figref{fig:ablation_supp}.

\subsection{Number of Parts}
\label{subsec:n_parts}

In this section, we analyze the impact of the number of parts on the
quality of the generated shapes. Specifically, we train our model with: $5$,
$10$, $16$ and $20$ parts. Unlike the results presented in our main submission, for
this experiment, we train all model variants on a subset of the Airplanes
category, consisting of $178$ training samples, for $50$ epochs. To evaluate the
plausibility of the generated shapes, we compute the Minimum Matching Distance
(MMD-CD) score between shapes generated with a different number of parts and a
set of reference shapes.  As reference shapes, we take $10\%$ of the entire
test set, which amounts to approximately $46$ shapes. Results are summarized
in \tabref{tab:n_parts_ablation}.
\begin{table}[!h]
    \centering
    \begin{tabular}{l||cccc}
        \toprule
        & $5$ Parts & $10$ Parts & $16$ Parts  & $20$ Parts \\
        \midrule
        MMD-CD ($\downarrow$) & 1.87 & 1.73 & 1.53  & 1.52 \\
        \bottomrule
    \end{tabular}
    \caption{{\bf Ablation Study on the Number of Parts.} This table shows a
    quantitative comparison of our approach with different numbers of parts
    \wrt MMD-CD ($\downarrow$).}
    \label{tab:n_parts_ablation}
\end{table}
Note that we do not compare our model variants \wrt Coverage (COV), which is a metric
indicative of the diversity of the generated shapes, as we only consider a very
small reference set of shapes and the results could have been misleading.
\begin{figure}[!h]
    \centering
    \begin{subfigure}[b]{0.15\linewidth}
		\centering
		\includegraphics[width=\linewidth]{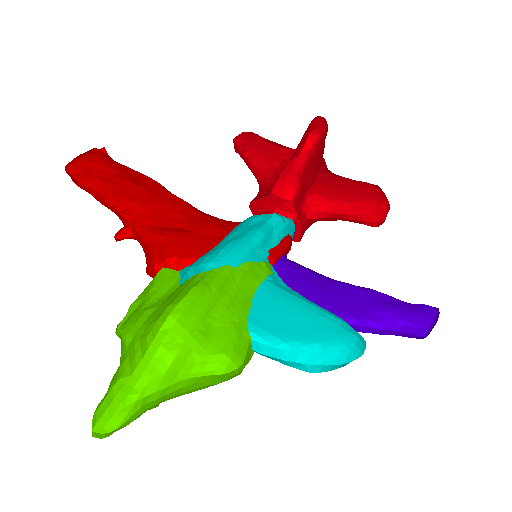}
    \end{subfigure}%
    \hfill%
    \begin{subfigure}[b]{0.15\linewidth}
		\centering
		\includegraphics[width=\linewidth]{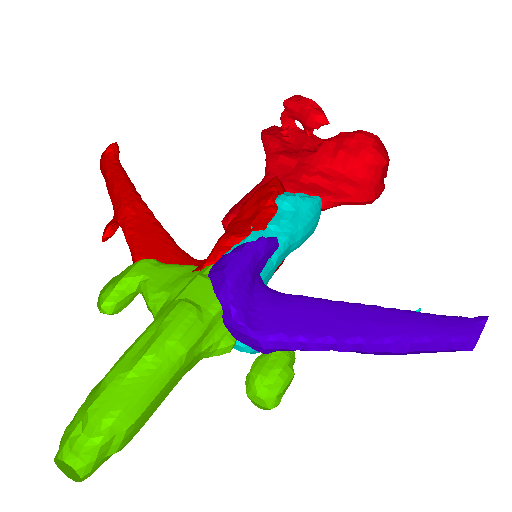}
    \end{subfigure}%
    \hfill%
    \begin{subfigure}[b]{0.15\linewidth}
		\centering
		\includegraphics[width=\linewidth]{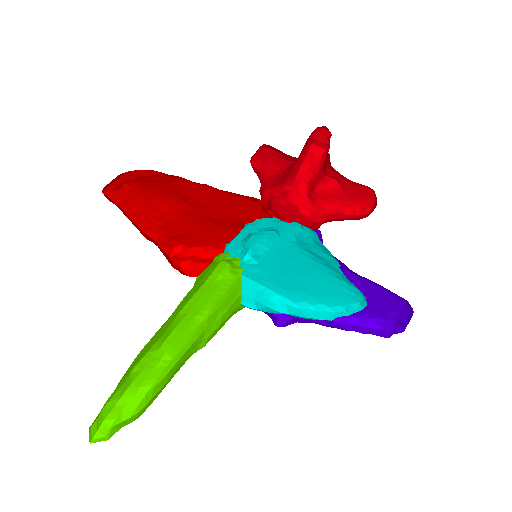}
    \end{subfigure}%
    \hfill%
    \begin{subfigure}[b]{0.15\linewidth}
		\centering
		\includegraphics[width=\linewidth]{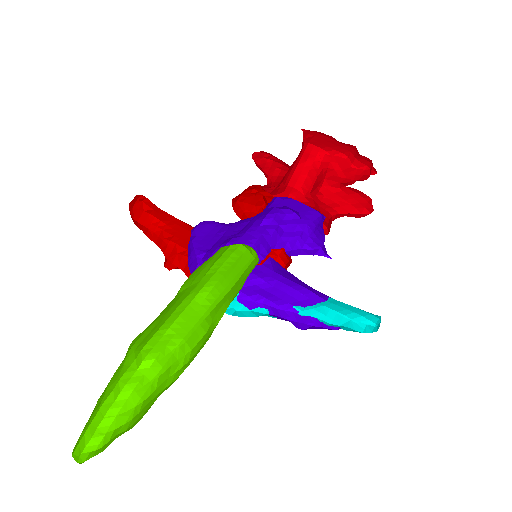}
    \end{subfigure}%
    \hfill%
    \begin{subfigure}[b]{0.15\linewidth}
		\centering
		\includegraphics[width=\linewidth]{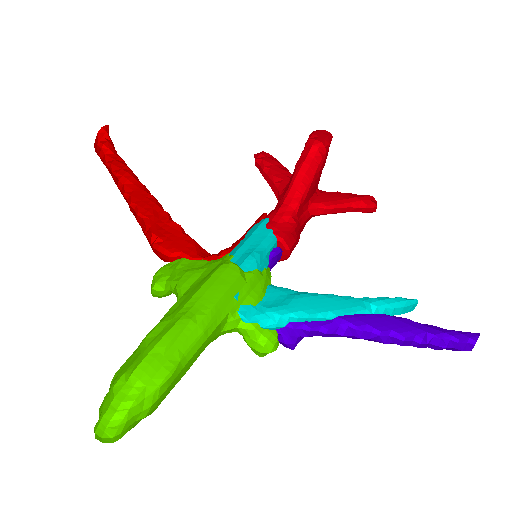}
    \end{subfigure}%
    \hfill%
    \begin{subfigure}[b]{0.15\linewidth}
		\centering
		\includegraphics[width=\linewidth]{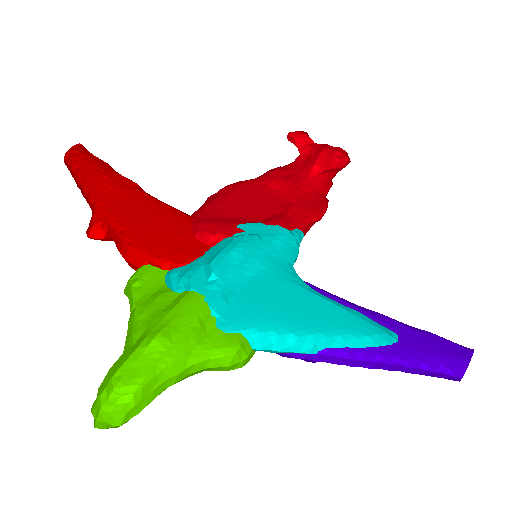}
    \end{subfigure}%
    \vskip\baselineskip%
    \vspace{-1.5em}
    \begin{minipage}[b]{\linewidth}
		\centering 5 Parts
    \end{minipage}%
    \vskip\baselineskip%
    \vspace{-1.2em}
    \begin{subfigure}[b]{0.15\linewidth}
		\centering
		\includegraphics[width=\linewidth]{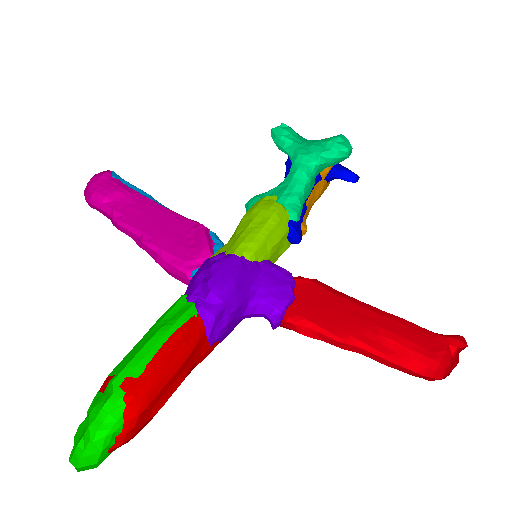}
    \end{subfigure}%
    \hfill%
    \begin{subfigure}[b]{0.15\linewidth}
		\centering
		\includegraphics[width=\linewidth]{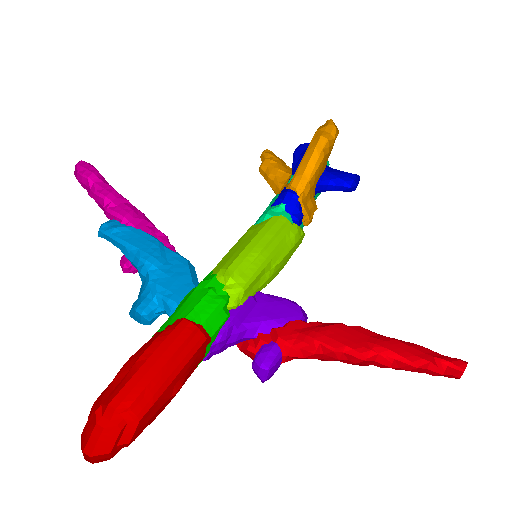}
    \end{subfigure}%
    \hfill%
    \begin{subfigure}[b]{0.15\linewidth}
		\centering
		\includegraphics[width=\linewidth]{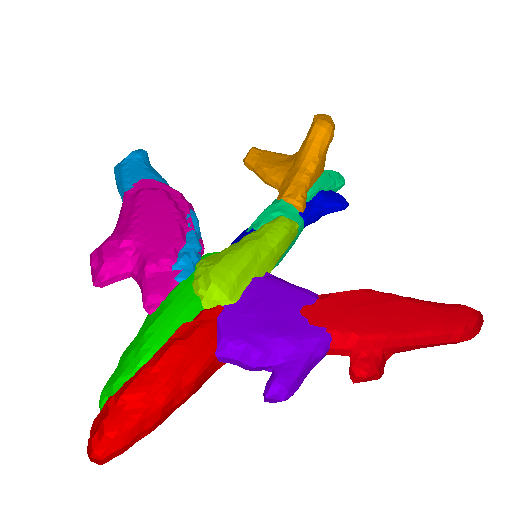}
    \end{subfigure}%
    \hfill%
    \begin{subfigure}[b]{0.15\linewidth}
		\centering
		\includegraphics[width=\linewidth]{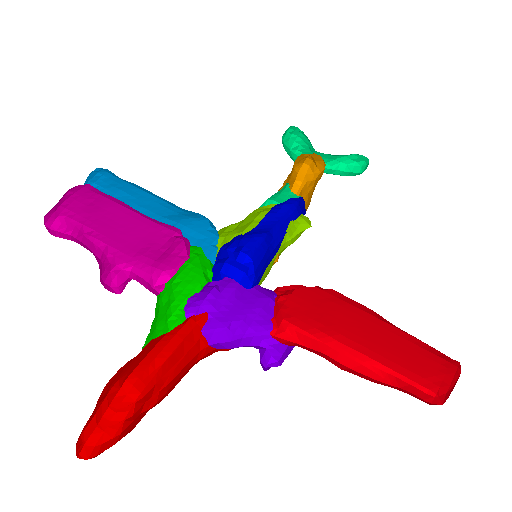}
    \end{subfigure}%
    \hfill%
    \begin{subfigure}[b]{0.15\linewidth}
		\centering
		\includegraphics[width=\linewidth]{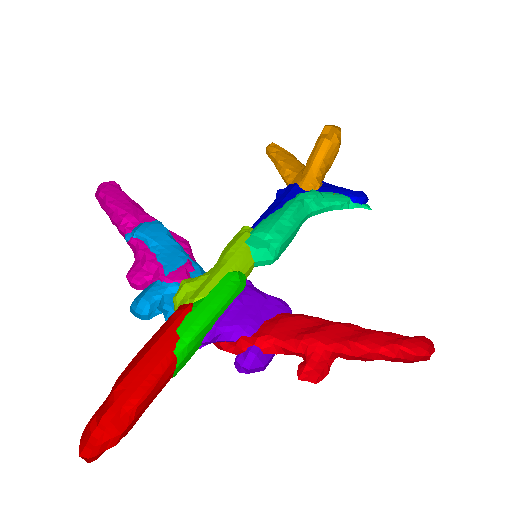}
    \end{subfigure}%
    \hfill%
    \begin{subfigure}[b]{0.15\linewidth}
		\centering
		\includegraphics[width=\linewidth]{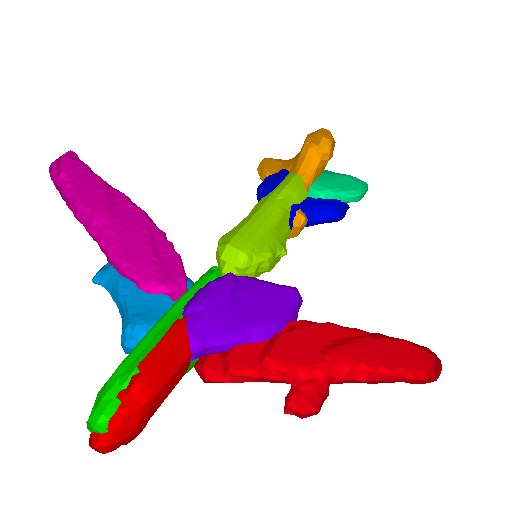}
    \end{subfigure}%
    \vskip\baselineskip%
    \vspace{-1.2em}
    \begin{minipage}[b]{\linewidth}
		\centering 10 Parts
    \end{minipage}%
    \vskip\baselineskip%
    \vspace{-1.2em}
    \begin{subfigure}[b]{0.15\linewidth}
		\centering
		\includegraphics[width=\linewidth]{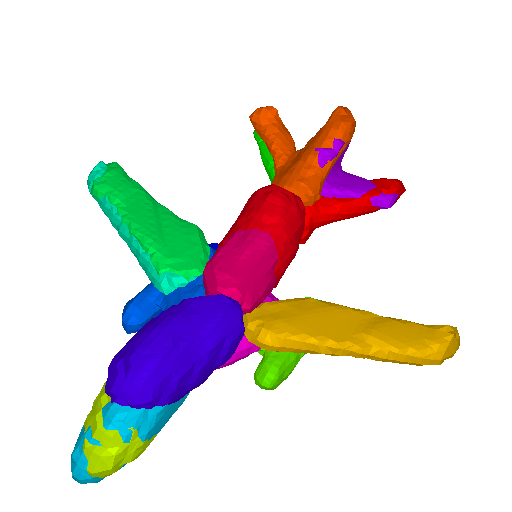}
    \end{subfigure}%
    \hfill%
    \begin{subfigure}[b]{0.15\linewidth}
		\centering
		\includegraphics[width=\linewidth]{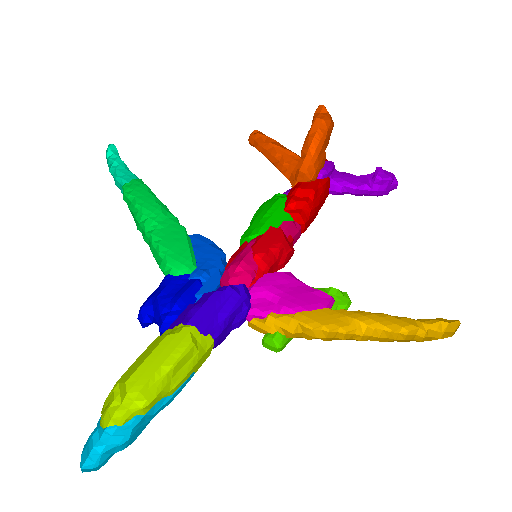}
    \end{subfigure}%
    \hfill%
    \begin{subfigure}[b]{0.15\linewidth}
		\centering
		\includegraphics[width=\linewidth]{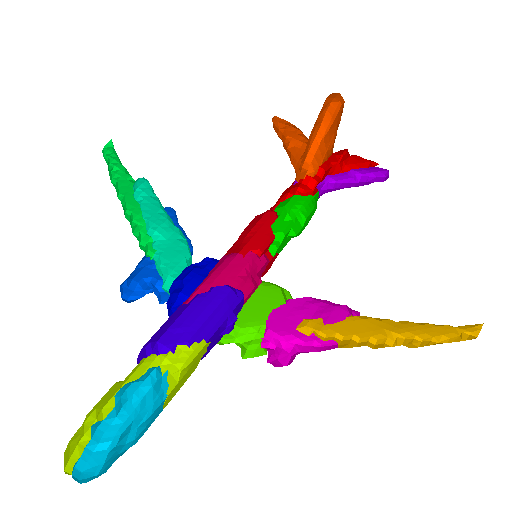}
    \end{subfigure}%
    \hfill%
    \begin{subfigure}[b]{0.15\linewidth}
		\centering
		\includegraphics[width=\linewidth]{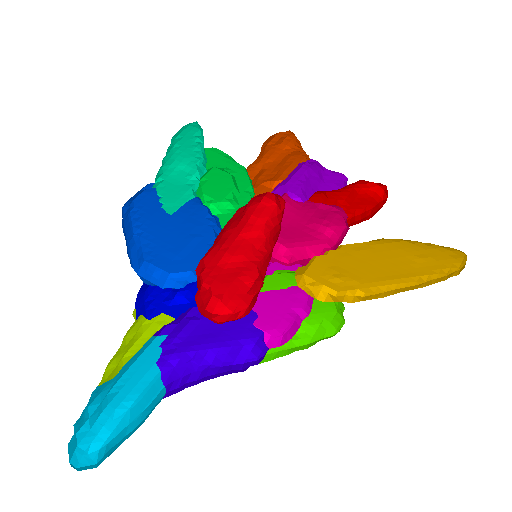}
    \end{subfigure}%
    \hfill%
    \begin{subfigure}[b]{0.15\linewidth}
		\centering
		\includegraphics[width=\linewidth]{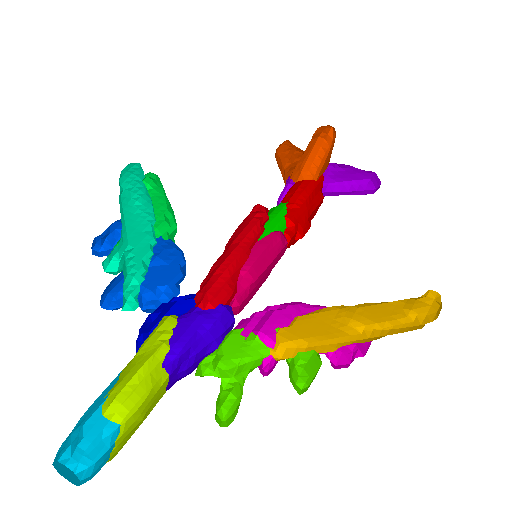}
    \end{subfigure}%
    \hfill%
    \begin{subfigure}[b]{0.15\linewidth}
		\centering
		\includegraphics[width=\linewidth]{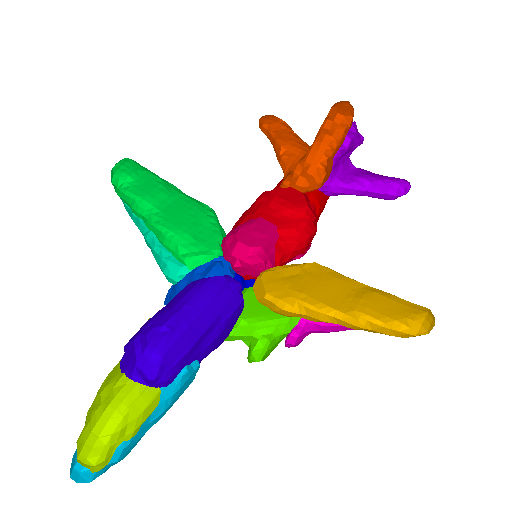}
    \end{subfigure}%
    \vskip\baselineskip%
    \vspace{-1.2em}
    \begin{minipage}[b]{\linewidth}
		\centering 16 Parts
    \end{minipage}%
    \vskip\baselineskip%
    \vspace{-1.2em}
    \begin{subfigure}[b]{0.15\linewidth}
		\centering
		\includegraphics[width=\linewidth]{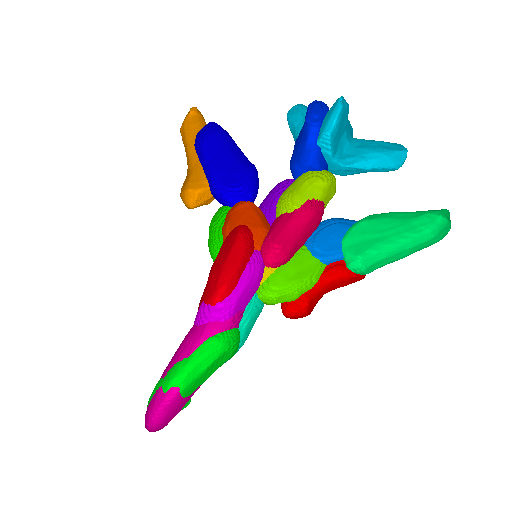}
    \end{subfigure}%
    \hfill%
    \begin{subfigure}[b]{0.15\linewidth}
		\centering
		\includegraphics[width=\linewidth]{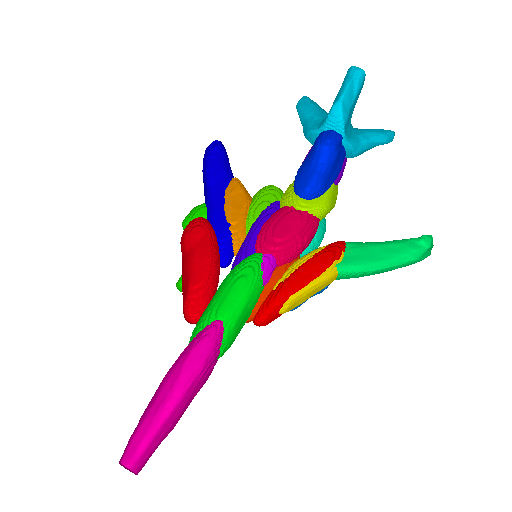}
    \end{subfigure}%
    \hfill%
    \begin{subfigure}[b]{0.15\linewidth}
		\centering
		\includegraphics[width=\linewidth]{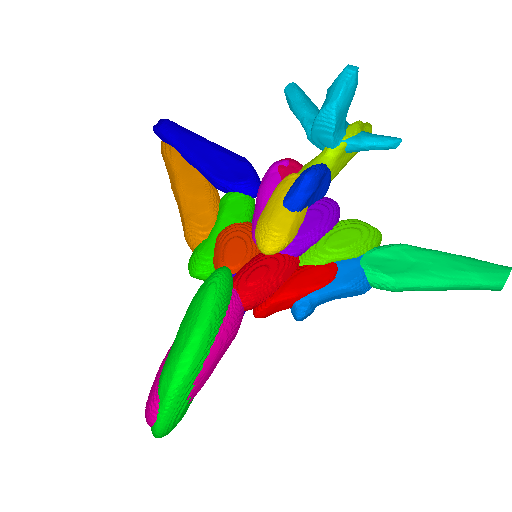}
    \end{subfigure}%
    \hfill%
    \begin{subfigure}[b]{0.15\linewidth}
		\centering
		\includegraphics[width=\linewidth]{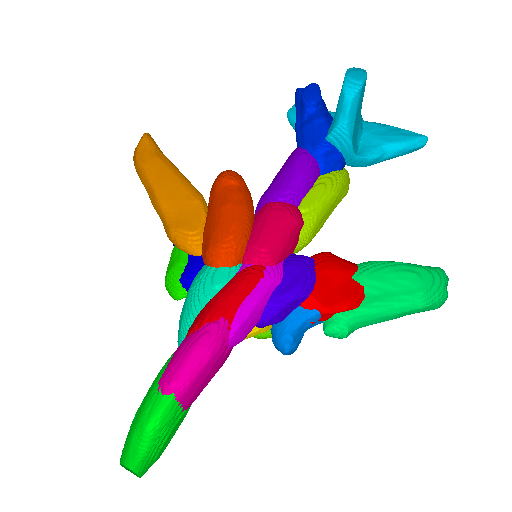}
    \end{subfigure}%
    \hfill%
    \begin{subfigure}[b]{0.15\linewidth}
		\centering
		\includegraphics[width=\linewidth]{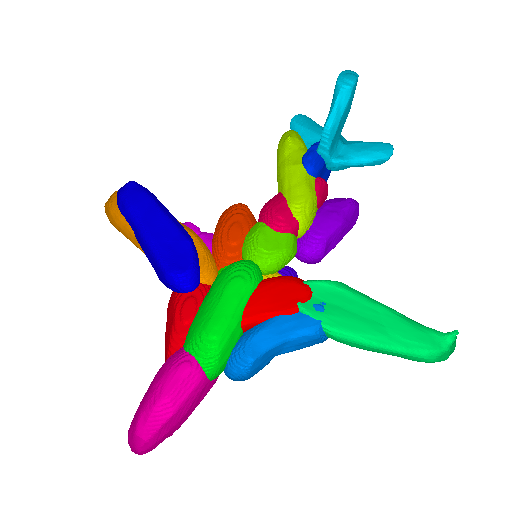}
    \end{subfigure}%
    \hfill%
    \begin{subfigure}[b]{0.15\linewidth}
		\centering
		\includegraphics[width=\linewidth]{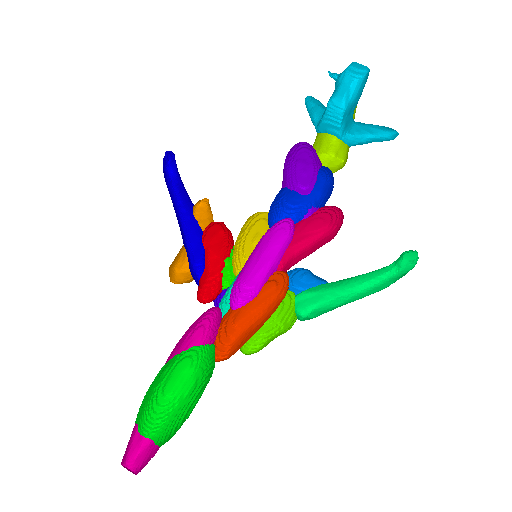}
    \end{subfigure}%
    \vskip\baselineskip%
    \vspace{-1.2em}
    \begin{minipage}[b]{\linewidth}
		\centering 20 Parts
    \end{minipage}%
    \caption{{\bf Ablation Study on the Number of Parts}. Qualitative
    evaluation of the impact of the number of parts. The first row shows
    randomly generated airplanes with $5$ parts, the second row with $10$, the
    third with $16$, and the last row with $20$ parts. Note that we train a
    different model for the four different numbers of parts.}
    \label{fig:n_parts_ablation}
\end{figure}

From the quantitative comparison in \tabref{tab:n_parts_ablation}, we observe
that the quality of the generated meshes increases for a larger number of
parts, namely using more parts yields higher MMD-CD scores. From the
qualitative comparison of \figref{fig:n_parts_ablation}, we note that using
fewer parts results in parts that capture multiple regions of the object, \eg the red
part represents the left wing and the tail. Since our main goal is to enable
editability and part-level control, we want to ensure that there is an adequate
number of parts so that the user can choose what they want to edit/change. For
the case of $10$ parts, we note that now parts are capable of recovering
more distinct regions of the object.

\subsection{Hard Ray-Part Assignment}
\label{subsec:hard_ray_part_assignment}

In Sec. 4.1 of our main submission, we demonstrate that enabling a soft
assignment between parts and rays, namely the color of one ray being
determined from multiple parts, prevents several editing operations (see Fig. 5
in our main submission). This is to be expected as for the case of the soft
ray-part assignment, the colors and opacities of all rays are determined jointly
by all NeRFs. As a result, changing one NeRF naturally affects the others,
hence the colors and opacities of parts of the object that we did not intend
to change.

\clearpage
\section{Additional Experimental Results}
\label{sec:generations}

\begin{figure}[!h]
    \centering
    \begin{subfigure}[b]{0.20\linewidth}
		\centering
		\includegraphics[width=\linewidth]{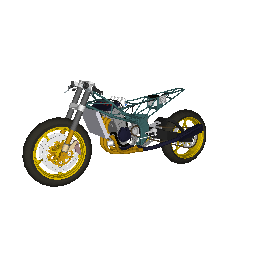}
    \end{subfigure}%
    \hfill%
    \begin{subfigure}[b]{0.20\linewidth}
		\centering
		\includegraphics[width=\linewidth]{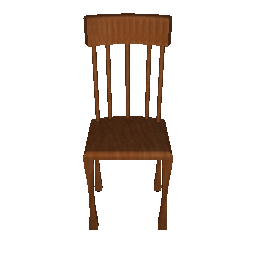}
    \end{subfigure}%
    \hfill%
    \begin{subfigure}[b]{0.20\linewidth}
		\centering
		\includegraphics[width=\linewidth]{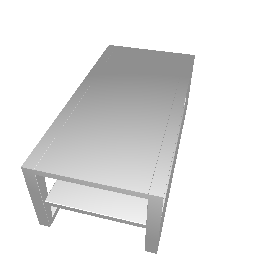}
    \end{subfigure}%
    \hfill%
    \begin{subfigure}[b]{0.20\linewidth}
		\centering
		\includegraphics[width=\linewidth]{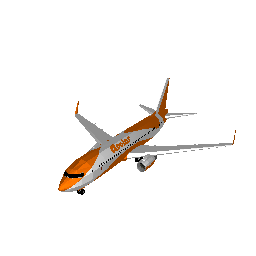}
    \end{subfigure}%
    \hfill%
    \begin{subfigure}[b]{0.20\linewidth}
		\centering
		\includegraphics[width=\linewidth]{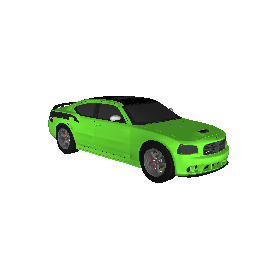}
    \end{subfigure}%
    \vskip\baselineskip%
    \vspace{-1.2em}
    \begin{subfigure}[b]{0.20\linewidth}
		\centering
		\includegraphics[width=\linewidth]{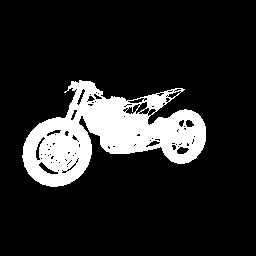}
    \end{subfigure}%
    \hfill%
    \begin{subfigure}[b]{0.20\linewidth}
		\centering
		\includegraphics[width=\linewidth]{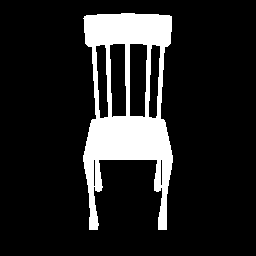}
    \end{subfigure}%
    \hfill%
    \begin{subfigure}[b]{0.20\linewidth}
		\centering
		\includegraphics[width=\linewidth]{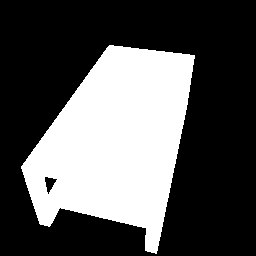}
    \end{subfigure}%
    \hfill%
    \begin{subfigure}[b]{0.20\linewidth}
		\centering
		\includegraphics[width=\linewidth]{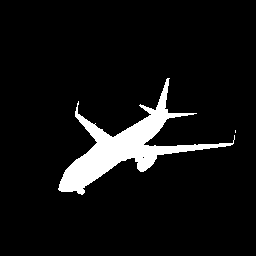}
    \end{subfigure}%
    \hfill%
    \begin{subfigure}[b]{0.20\linewidth}
		\centering
		\includegraphics[width=\linewidth]{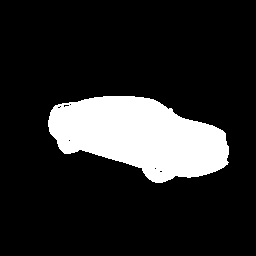}
    \end{subfigure}%
    \vskip\baselineskip%
    \vspace{-1.2em}
    \caption{{\bf Rendered RGB Images and Object Masks}. We provide examples of the rendered RGB images
    and object masks that we use for training.}
    \label{fig:rendering_examples}
\end{figure}

In this section, we provide additional information regarding our experiments on
ShapeNet~\cite{Chang2015ARXIV}. In particular, we consider five categories:
\emph{Motorbike, Chair, Table, Airplane, Car}, which contain 337, 6678, 8509,
4045 and 7497 shapes respectively. For the \emph{Car, Table, Airplane and
Chair}, we use $24$ random views, while for Motorbike we use $100$, as it
contains fewer training samples. To ensure fair comparison with our baselines, we
use the train-test splits of \cite{Gao2022NEURIPS} for the \emph{Motorbike,
Car, Chair} object categories and the train-test splits of
\cite{Hertz2022SIGGRAPH} for the \emph{Airplane, Table} category. To
render our training data, we randomly sample camera poses from the upper
hemisphere of each shape and render images at $256^2$ resolution as in
\cite{Gao2022NEURIPS}. To render both the RGB images and the corresponding
object masks we use
\emph{simple-3dviz}\cite{Katharopoulos2020simple3dviz}\footnote{\href{https://simple-3dviz.com/}{https://simple-3dviz.com/}}.
Examples of the rendered RGB images and object masks per category are provided in \figref{fig:rendering_examples}.

\subsection{Shape Generation}

\begin{figure}[!h]
    \centering
    \begin{subfigure}[b]{0.15\linewidth}
		\centering
		\includegraphics[width=\linewidth]{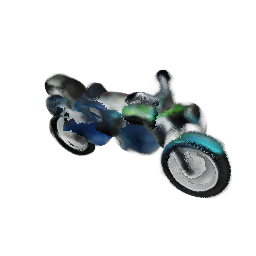}
    \end{subfigure}%
    \hfill%
    \begin{subfigure}[b]{0.15\linewidth}
		\centering
		\includegraphics[width=\linewidth]{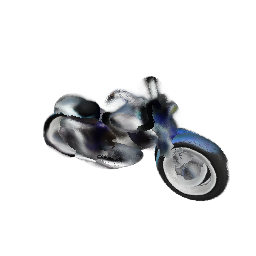}
    \end{subfigure}%
    \hfill%
    \begin{subfigure}[b]{0.15\linewidth}
		\centering
		\includegraphics[width=\linewidth]{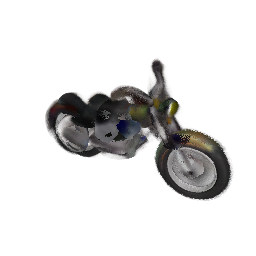}
    \end{subfigure}%
    \hfill%
    \begin{subfigure}[b]{0.15\linewidth}
		\centering
		\includegraphics[width=\linewidth]{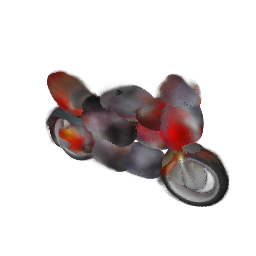}
    \end{subfigure}%
    \hfill%
    \begin{subfigure}[b]{0.15\linewidth}
		\centering
		\includegraphics[width=\linewidth]{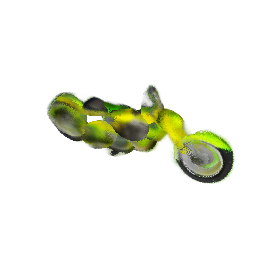}
    \end{subfigure}%
    \hfill%
    \begin{subfigure}[b]{0.15\linewidth}
		\centering
		\includegraphics[width=\linewidth]{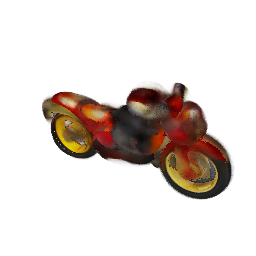}
    \end{subfigure}%
    \vskip\baselineskip%
    \vspace{-2.2em}
    \begin{subfigure}[b]{0.15\linewidth}
		\centering
		\includegraphics[width=\linewidth]{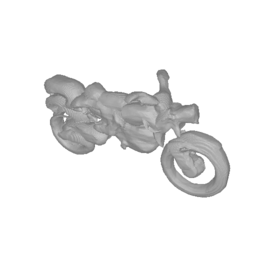}
    \end{subfigure}%
    \hfill%
    \begin{subfigure}[b]{0.15\linewidth}
		\centering
		\includegraphics[width=\linewidth]{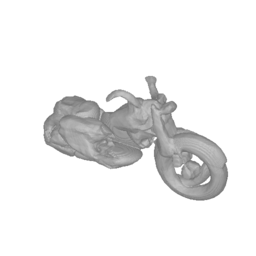}
    \end{subfigure}%
    \hfill%
    \begin{subfigure}[b]{0.15\linewidth}
		\centering
		\includegraphics[width=\linewidth]{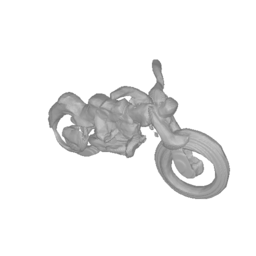}
    \end{subfigure}%
    \hfill%
    \begin{subfigure}[b]{0.15\linewidth}
		\centering
		\includegraphics[width=\linewidth]{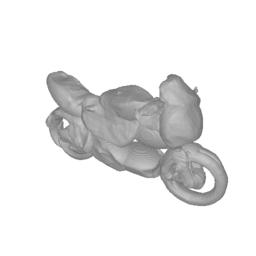}
    \end{subfigure}%
    \hfill%
    \begin{subfigure}[b]{0.15\linewidth}
		\centering
		\includegraphics[width=\linewidth]{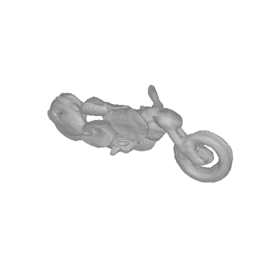}
    \end{subfigure}%
    \hfill%
    \begin{subfigure}[b]{0.15\linewidth}
		\centering
		\includegraphics[width=\linewidth]{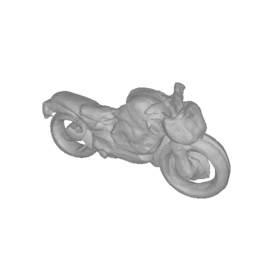}
    \end{subfigure}%
    \vskip\baselineskip%
    \vspace{-2.2em}
    \begin{subfigure}[b]{0.15\linewidth}
		\centering
		\includegraphics[width=\linewidth]{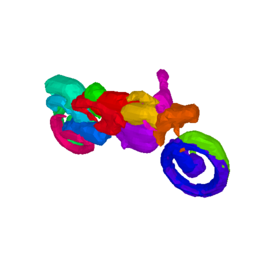}
    \end{subfigure}%
    \hfill%
    \begin{subfigure}[b]{0.15\linewidth}
		\centering
		\includegraphics[width=\linewidth]{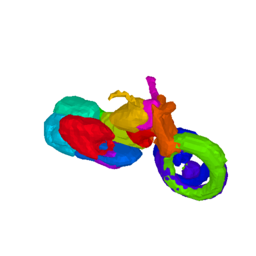}
    \end{subfigure}%
    \hfill%
    \begin{subfigure}[b]{0.15\linewidth}
		\centering
		\includegraphics[width=\linewidth]{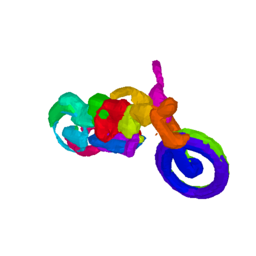}
    \end{subfigure}%
    \hfill%
    \begin{subfigure}[b]{0.15\linewidth}
		\centering
		\includegraphics[width=\linewidth]{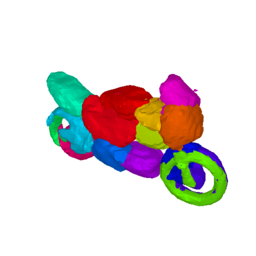}
    \end{subfigure}%
    \hfill%
    \begin{subfigure}[b]{0.15\linewidth}
		\centering
		\includegraphics[width=\linewidth]{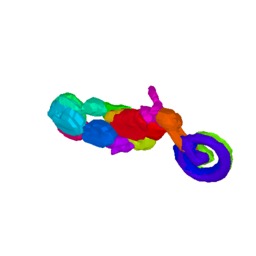}
    \end{subfigure}%
    \hfill%
    \begin{subfigure}[b]{0.15\linewidth}
		\centering
		\includegraphics[width=\linewidth]{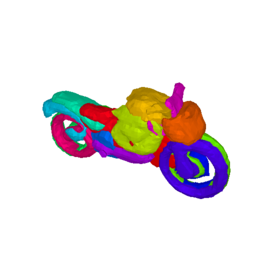}
    \end{subfigure}%
    \vskip\baselineskip%
    \vspace{-2.2em}
    \begin{subfigure}[b]{0.15\linewidth}
		\centering
		\includegraphics[width=\linewidth]{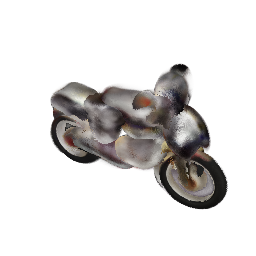}
    \end{subfigure}%
    \hfill%
    \begin{subfigure}[b]{0.15\linewidth}
		\centering
		\includegraphics[width=\linewidth]{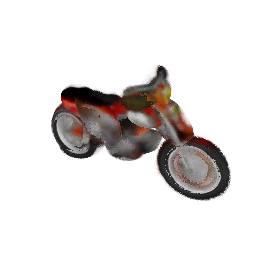}
    \end{subfigure}%
    \hfill%
    \begin{subfigure}[b]{0.15\linewidth}
		\centering
		\includegraphics[width=\linewidth]{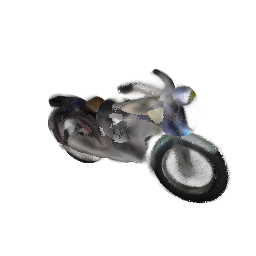}
    \end{subfigure}%
    \hfill%
    \begin{subfigure}[b]{0.15\linewidth}
		\centering
		\includegraphics[width=\linewidth]{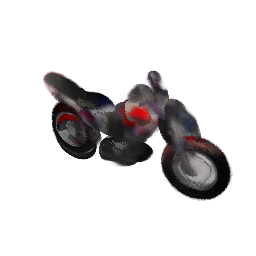}
    \end{subfigure}%
    \hfill%
    \begin{subfigure}[b]{0.15\linewidth}
		\centering
		\includegraphics[width=\linewidth]{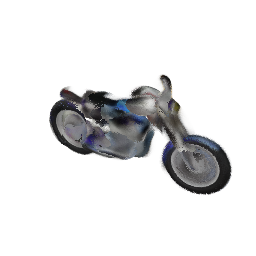}
    \end{subfigure}%
    \hfill%
    \begin{subfigure}[b]{0.15\linewidth}
		\centering
		\includegraphics[width=\linewidth]{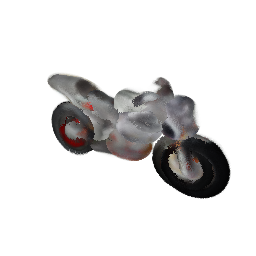}
    \end{subfigure}%
    \vskip\baselineskip%
    \vspace{-2.2em}
    \begin{subfigure}[b]{0.15\linewidth}
		\centering
		\includegraphics[width=\linewidth]{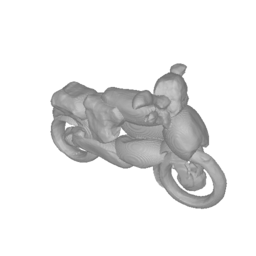}
    \end{subfigure}%
    \hfill%
    \begin{subfigure}[b]{0.15\linewidth}
		\centering
		\includegraphics[width=\linewidth]{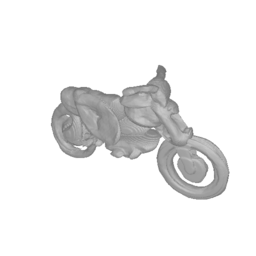}
    \end{subfigure}%
    \hfill%
    \begin{subfigure}[b]{0.15\linewidth}
		\centering
		\includegraphics[width=\linewidth]{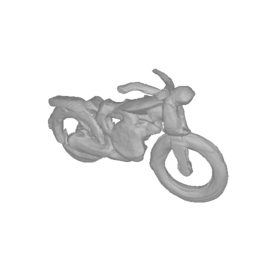}
    \end{subfigure}%
    \hfill%
    \begin{subfigure}[b]{0.15\linewidth}
		\centering
		\includegraphics[width=\linewidth]{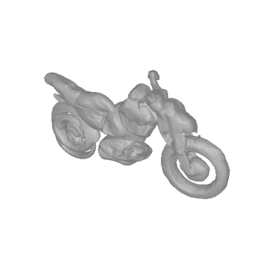}
    \end{subfigure}%
    \hfill%
    \begin{subfigure}[b]{0.15\linewidth}
		\centering
		\includegraphics[width=\linewidth]{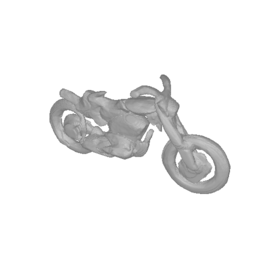}
    \end{subfigure}%
    \hfill%
    \begin{subfigure}[b]{0.15\linewidth}
		\centering
		\includegraphics[width=\linewidth]{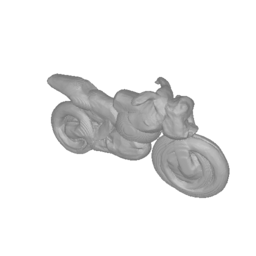}
    \end{subfigure}%
    \vskip\baselineskip%
    \vspace{-2.2em}
    \begin{subfigure}[b]{0.15\linewidth}
		\centering
		\includegraphics[width=\linewidth]{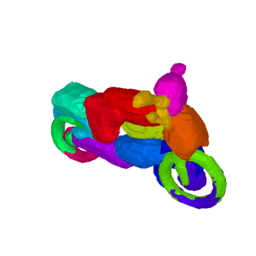}
    \end{subfigure}%
    \hfill%
    \begin{subfigure}[b]{0.15\linewidth}
		\centering
		\includegraphics[width=\linewidth]{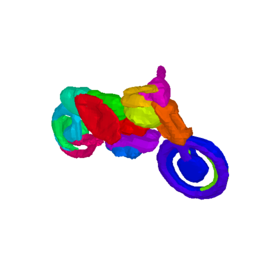}
    \end{subfigure}%
    \hfill%
    \begin{subfigure}[b]{0.15\linewidth}
		\centering
		\includegraphics[width=\linewidth]{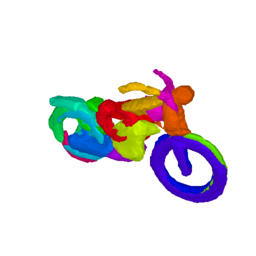}
    \end{subfigure}%
    \hfill%
    \begin{subfigure}[b]{0.15\linewidth}
		\centering
		\includegraphics[width=\linewidth]{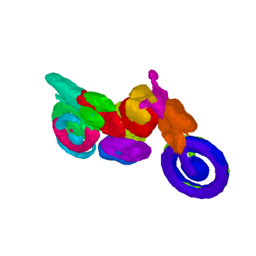}
    \end{subfigure}%
    \hfill%
    \begin{subfigure}[b]{0.15\linewidth}
		\centering
		\includegraphics[width=\linewidth]{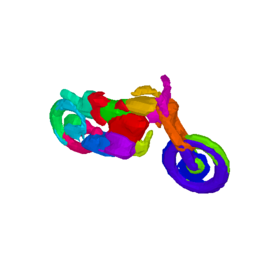}
    \end{subfigure}%
    \hfill%
    \begin{subfigure}[b]{0.15\linewidth}
		\centering
		\includegraphics[width=\linewidth]{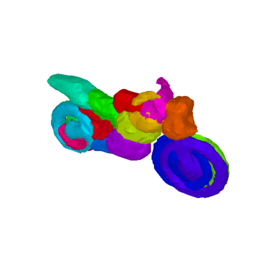}
    \end{subfigure}%
    \vskip\baselineskip%
    \vspace{-2.2em}
    \begin{subfigure}[b]{0.15\linewidth}
		\centering
		\includegraphics[width=\linewidth]{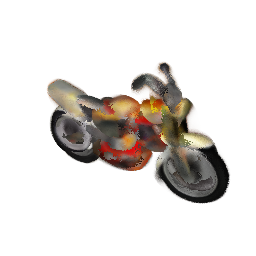}
    \end{subfigure}%
    \hfill%
    \begin{subfigure}[b]{0.15\linewidth}
		\centering
		\includegraphics[width=\linewidth]{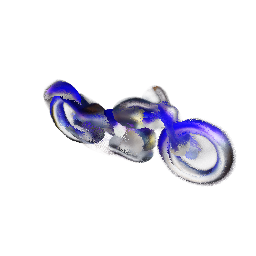}
    \end{subfigure}%
    \hfill%
    \begin{subfigure}[b]{0.15\linewidth}
		\centering
		\includegraphics[width=\linewidth]{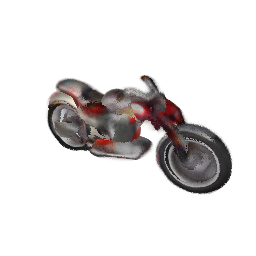}
    \end{subfigure}%
    \hfill%
    \begin{subfigure}[b]{0.15\linewidth}
		\centering
		\includegraphics[width=\linewidth]{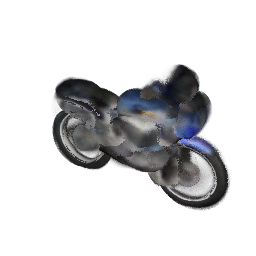}
    \end{subfigure}%
    \hfill%
    \begin{subfigure}[b]{0.15\linewidth}
		\centering
		\includegraphics[width=\linewidth]{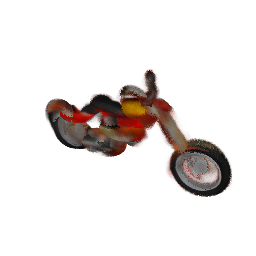}
    \end{subfigure}%
    \hfill%
    \begin{subfigure}[b]{0.15\linewidth}
		\centering
		\includegraphics[width=\linewidth]{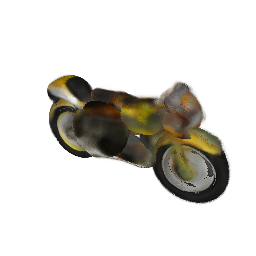}
    \end{subfigure}%
    \vskip\baselineskip%
    \vspace{-2.2em}
    \begin{subfigure}[b]{0.15\linewidth}
		\centering
		\includegraphics[width=\linewidth]{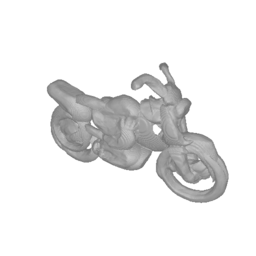}
    \end{subfigure}%
    \hfill%
    \begin{subfigure}[b]{0.15\linewidth}
		\centering
		\includegraphics[width=\linewidth]{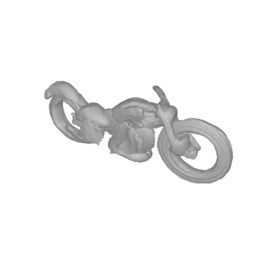}
    \end{subfigure}%
    \hfill%
    \begin{subfigure}[b]{0.15\linewidth}
		\centering
		\includegraphics[width=\linewidth]{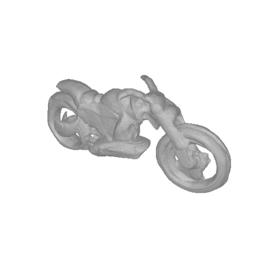}
    \end{subfigure}%
    \hfill%
    \begin{subfigure}[b]{0.15\linewidth}
		\centering
		\includegraphics[width=\linewidth]{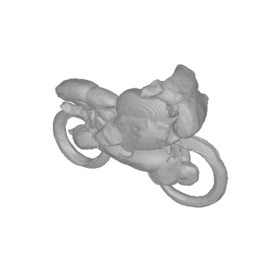}
    \end{subfigure}%
    \hfill%
    \begin{subfigure}[b]{0.15\linewidth}
		\centering
		\includegraphics[width=\linewidth]{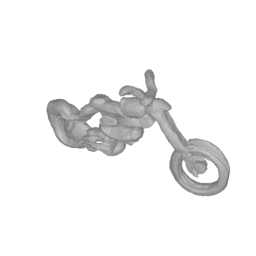}
    \end{subfigure}%
    \hfill%
    \begin{subfigure}[b]{0.15\linewidth}
		\centering
		\includegraphics[width=\linewidth]{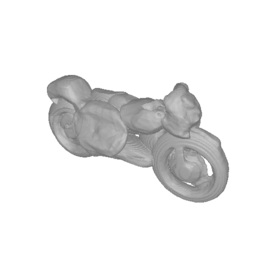}
    \end{subfigure}%
    \vskip\baselineskip%
    \vspace{-2.2em}
    \begin{subfigure}[b]{0.15\linewidth}
		\centering
		\includegraphics[width=\linewidth]{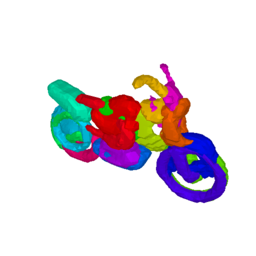}
    \end{subfigure}%
    \hfill%
    \begin{subfigure}[b]{0.15\linewidth}
		\centering
		\includegraphics[width=\linewidth]{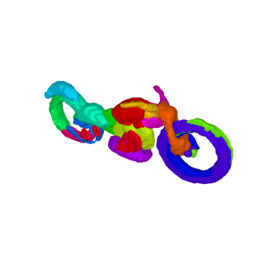}
    \end{subfigure}%
    \hfill%
    \begin{subfigure}[b]{0.15\linewidth}
		\centering
		\includegraphics[width=\linewidth]{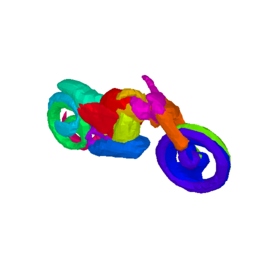}
    \end{subfigure}%
    \hfill%
    \begin{subfigure}[b]{0.15\linewidth}
		\centering
		\includegraphics[width=\linewidth]{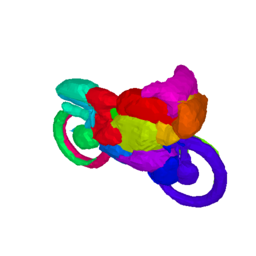}
    \end{subfigure}%
    \hfill%
    \begin{subfigure}[b]{0.15\linewidth}
		\centering
		\includegraphics[width=\linewidth]{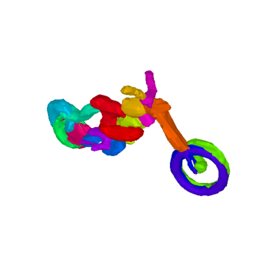}
    \end{subfigure}%
    \hfill%
    \begin{subfigure}[b]{0.15\linewidth}
		\centering
		\includegraphics[width=\linewidth]{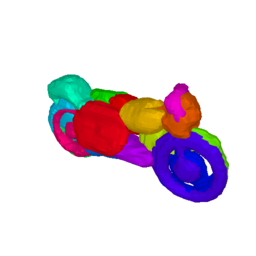}
    \end{subfigure}%
    \caption{{\bf Generated Motorbikes}. Generated motorbikes accompanied by
    their corresponding 3D geometries and part-based geometries.}
    \label{fig:generated_motorbikes}
    \vspace{-1.2em}
\end{figure}

\begin{figure}[!h]
    \centering
    \begin{subfigure}[b]{0.15\linewidth}
		\centering
		\includegraphics[width=\linewidth]{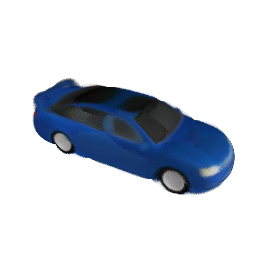}
    \end{subfigure}%
    \hfill%
    \begin{subfigure}[b]{0.15\linewidth}
		\centering
		\includegraphics[width=\linewidth]{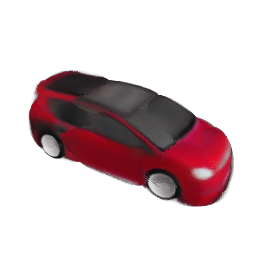}
    \end{subfigure}%
    \hfill%
    \begin{subfigure}[b]{0.15\linewidth}
		\centering
		\includegraphics[width=\linewidth]{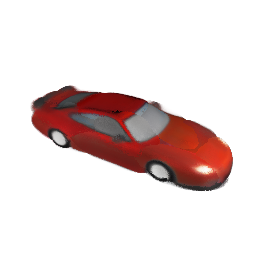}
    \end{subfigure}%
    \hfill%
    \begin{subfigure}[b]{0.15\linewidth}
		\centering
		\includegraphics[width=\linewidth]{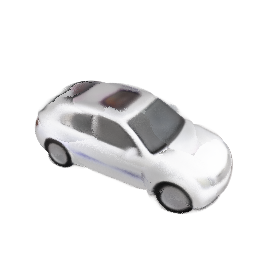}
    \end{subfigure}%
    \hfill%
    \begin{subfigure}[b]{0.15\linewidth}
		\centering
		\includegraphics[width=\linewidth]{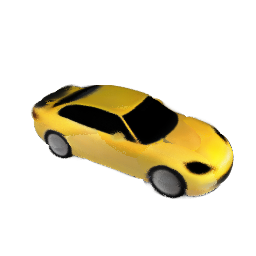}
    \end{subfigure}%
    \hfill%
    \begin{subfigure}[b]{0.15\linewidth}
		\centering
		\includegraphics[width=\linewidth]{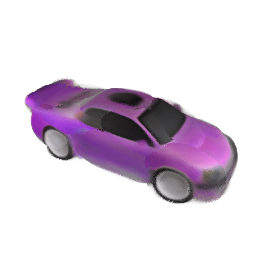}
    \end{subfigure}%
    \vskip\baselineskip%
    \vspace{-2.2em}
    \begin{subfigure}[b]{0.15\linewidth}
		\centering
		\includegraphics[width=\linewidth]{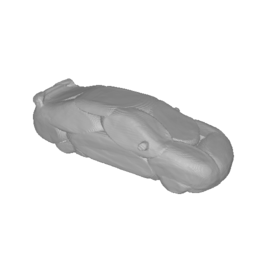}
    \end{subfigure}%
    \hfill%
    \begin{subfigure}[b]{0.15\linewidth}
		\centering
		\includegraphics[width=\linewidth]{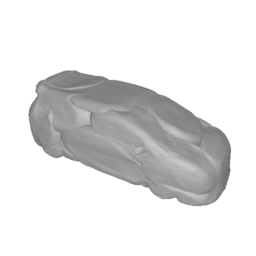}
    \end{subfigure}%
    \hfill%
    \begin{subfigure}[b]{0.15\linewidth}
		\centering
		\includegraphics[width=\linewidth]{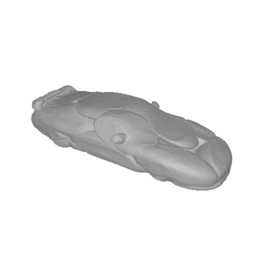}
    \end{subfigure}%
    \hfill%
    \begin{subfigure}[b]{0.15\linewidth}
		\centering
		\includegraphics[width=\linewidth]{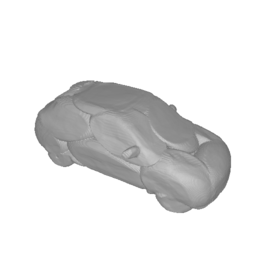}
    \end{subfigure}%
    \hfill%
    \begin{subfigure}[b]{0.15\linewidth}
		\centering
		\includegraphics[width=\linewidth]{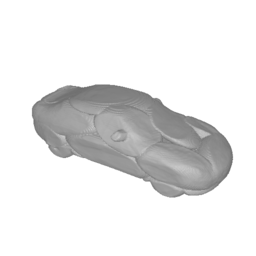}
    \end{subfigure}%
    \hfill%
    \begin{subfigure}[b]{0.15\linewidth}
		\centering
		\includegraphics[width=\linewidth]{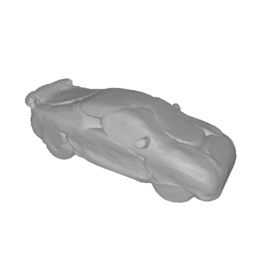}
    \end{subfigure}%
    \vskip\baselineskip%
    \vspace{-2.2em}
    \begin{subfigure}[b]{0.15\linewidth}
		\centering
		\includegraphics[width=\linewidth]{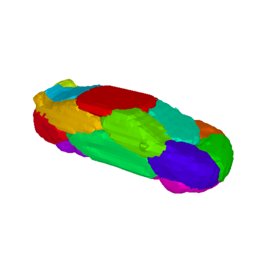}
    \end{subfigure}%
    \hfill%
    \begin{subfigure}[b]{0.15\linewidth}
		\centering
		\includegraphics[width=\linewidth]{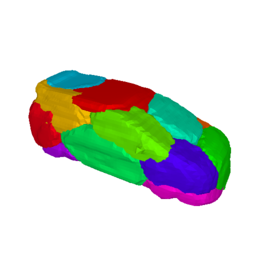}
    \end{subfigure}%
    \hfill%
    \begin{subfigure}[b]{0.15\linewidth}
		\centering
		\includegraphics[width=\linewidth]{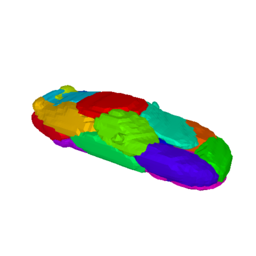}
    \end{subfigure}%
    \hfill%
    \begin{subfigure}[b]{0.15\linewidth}
		\centering
		\includegraphics[width=\linewidth]{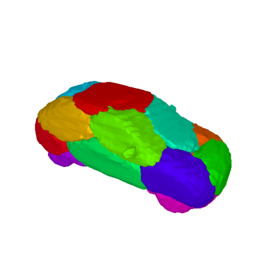}
    \end{subfigure}%
    \hfill%
    \begin{subfigure}[b]{0.15\linewidth}
		\centering
		\includegraphics[width=\linewidth]{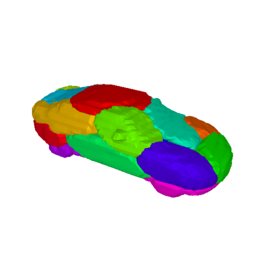}
    \end{subfigure}%
    \hfill%
    \begin{subfigure}[b]{0.15\linewidth}
		\centering
		\includegraphics[width=\linewidth]{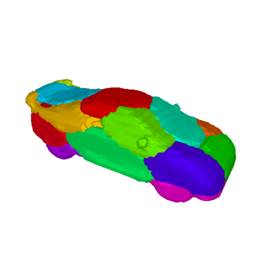}
    \end{subfigure}%
    \vskip\baselineskip%
    \vspace{-2.2em}
    \begin{subfigure}[b]{0.15\linewidth}
		\centering
		\includegraphics[width=\linewidth]{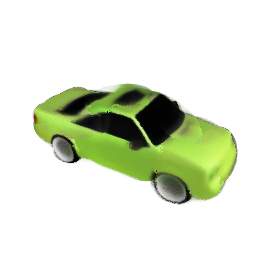}
    \end{subfigure}%
    \hfill%
    \begin{subfigure}[b]{0.15\linewidth}
		\centering
		\includegraphics[width=\linewidth]{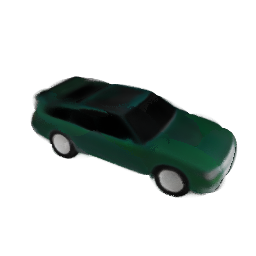}
    \end{subfigure}%
    \hfill%
    \begin{subfigure}[b]{0.15\linewidth}
		\centering
		\includegraphics[width=\linewidth]{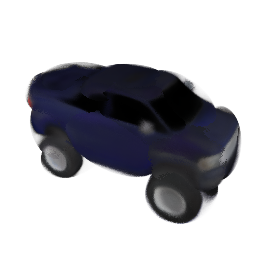}
    \end{subfigure}%
    \hfill%
    \begin{subfigure}[b]{0.15\linewidth}
		\centering
		\includegraphics[width=\linewidth]{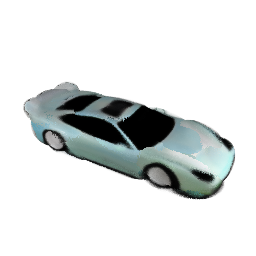}
    \end{subfigure}%
    \hfill%
    \begin{subfigure}[b]{0.15\linewidth}
		\centering
		\includegraphics[width=\linewidth]{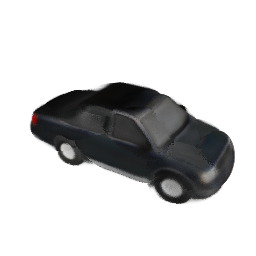}
    \end{subfigure}%
    \hfill%
    \begin{subfigure}[b]{0.15\linewidth}
		\centering
		\includegraphics[width=\linewidth]{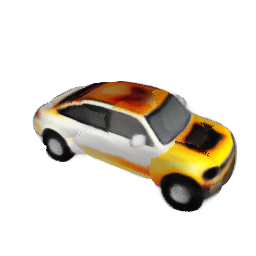}
    \end{subfigure}%
    \vskip\baselineskip%
    \vspace{-2.2em}
    \begin{subfigure}[b]{0.15\linewidth}
		\centering
		\includegraphics[width=\linewidth]{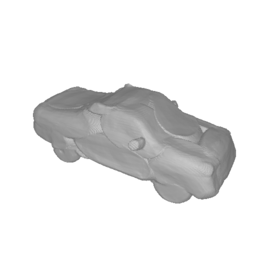}
    \end{subfigure}%
    \hfill%
    \begin{subfigure}[b]{0.15\linewidth}
		\centering
		\includegraphics[width=\linewidth]{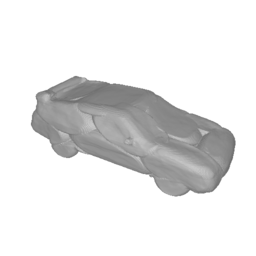}
    \end{subfigure}%
    \hfill%
    \begin{subfigure}[b]{0.15\linewidth}
		\centering
		\includegraphics[width=\linewidth]{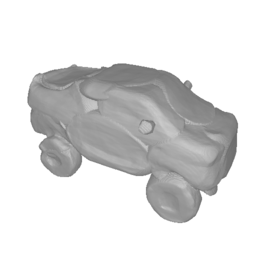}
    \end{subfigure}%
    \hfill%
    \begin{subfigure}[b]{0.15\linewidth}
		\centering
		\includegraphics[width=\linewidth]{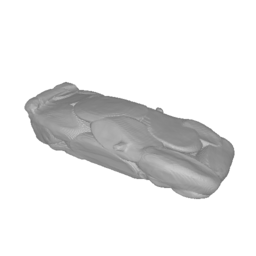}
    \end{subfigure}%
    \hfill%
    \begin{subfigure}[b]{0.15\linewidth}
		\centering
		\includegraphics[width=\linewidth]{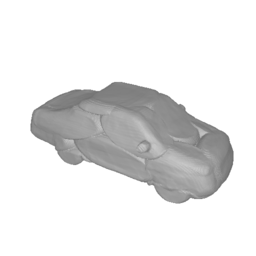}
    \end{subfigure}%
    \hfill%
    \begin{subfigure}[b]{0.15\linewidth}
		\centering
		\includegraphics[width=\linewidth]{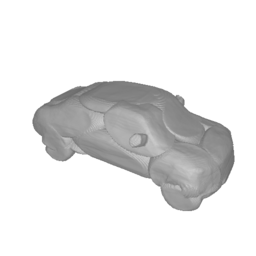}
    \end{subfigure}%
    \vskip\baselineskip%
    \vspace{-2.2em}
    \begin{subfigure}[b]{0.15\linewidth}
		\centering
		\includegraphics[width=\linewidth]{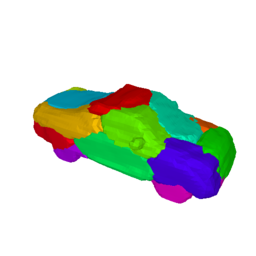}
    \end{subfigure}%
    \hfill%
    \begin{subfigure}[b]{0.15\linewidth}
		\centering
		\includegraphics[width=\linewidth]{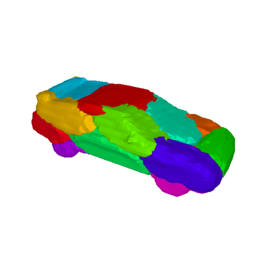}
    \end{subfigure}%
    \hfill%
    \begin{subfigure}[b]{0.15\linewidth}
		\centering
		\includegraphics[width=\linewidth]{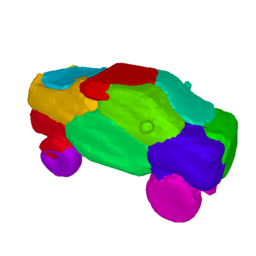}
    \end{subfigure}%
    \hfill%
    \begin{subfigure}[b]{0.15\linewidth}
		\centering
		\includegraphics[width=\linewidth]{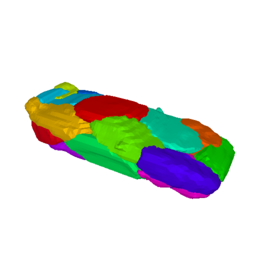}
    \end{subfigure}%
    \hfill%
    \begin{subfigure}[b]{0.15\linewidth}
		\centering
		\includegraphics[width=\linewidth]{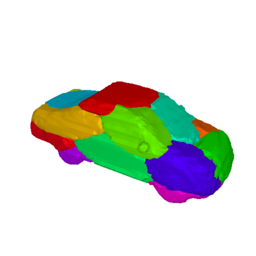}
    \end{subfigure}%
    \hfill%
    \begin{subfigure}[b]{0.15\linewidth}
		\centering
		\includegraphics[width=\linewidth]{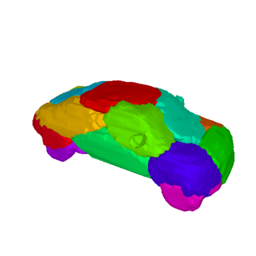}
    \end{subfigure}%
    \vskip\baselineskip%
    \vspace{-2.2em}
    \begin{subfigure}[b]{0.15\linewidth}
		\centering
		\includegraphics[width=\linewidth]{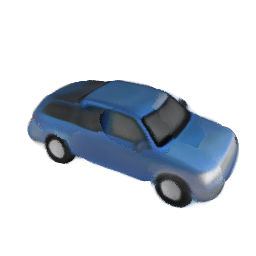}
    \end{subfigure}%
    \hfill%
    \begin{subfigure}[b]{0.15\linewidth}
		\centering
		\includegraphics[width=\linewidth]{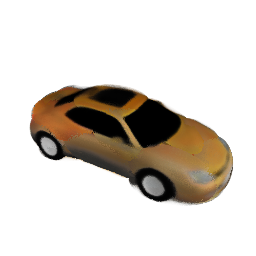}
    \end{subfigure}%
    \hfill%
    \begin{subfigure}[b]{0.15\linewidth}
		\centering
		\includegraphics[width=\linewidth]{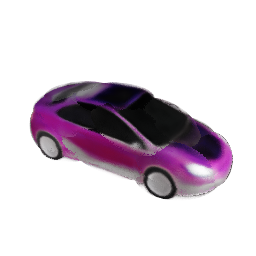}
    \end{subfigure}%
    \hfill%
    \begin{subfigure}[b]{0.15\linewidth}
		\centering
		\includegraphics[width=\linewidth]{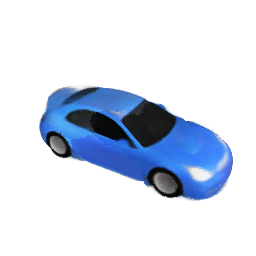}
    \end{subfigure}%
    \hfill%
    \begin{subfigure}[b]{0.15\linewidth}
		\centering
		\includegraphics[width=\linewidth]{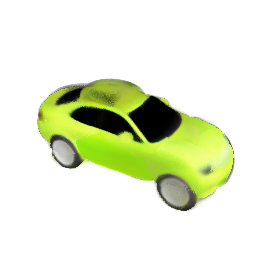}
    \end{subfigure}%
    \hfill%
    \begin{subfigure}[b]{0.15\linewidth}
		\centering
		\includegraphics[width=\linewidth]{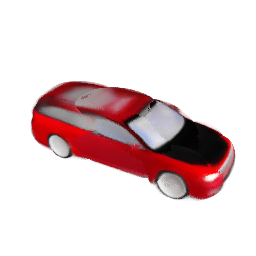}
    \end{subfigure}%
    \vskip\baselineskip%
    \vspace{-2.2em}
    \begin{subfigure}[b]{0.15\linewidth}
		\centering
		\includegraphics[width=\linewidth]{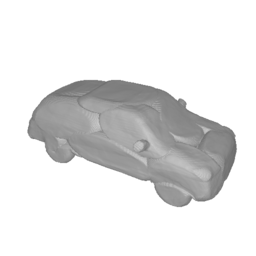}
    \end{subfigure}%
    \hfill%
    \begin{subfigure}[b]{0.15\linewidth}
		\centering
		\includegraphics[width=\linewidth]{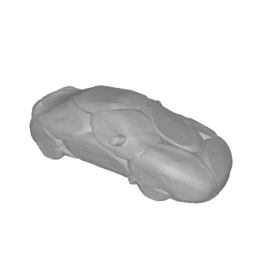}
    \end{subfigure}%
    \hfill%
    \begin{subfigure}[b]{0.15\linewidth}
		\centering
		\includegraphics[width=\linewidth]{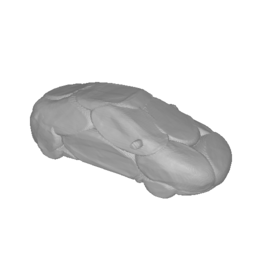}
    \end{subfigure}%
    \hfill%
    \begin{subfigure}[b]{0.15\linewidth}
		\centering
		\includegraphics[width=\linewidth]{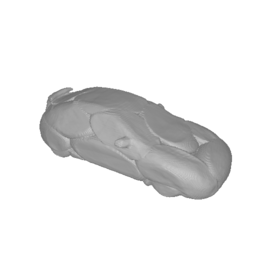}
    \end{subfigure}%
    \hfill%
    \begin{subfigure}[b]{0.15\linewidth}
		\centering
		\includegraphics[width=\linewidth]{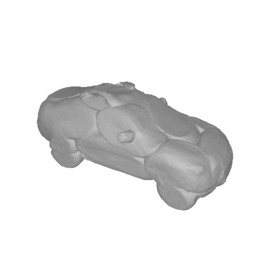}
    \end{subfigure}%
    \hfill%
    \begin{subfigure}[b]{0.15\linewidth}
		\centering
		\includegraphics[width=\linewidth]{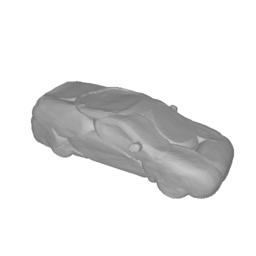}
    \end{subfigure}%
    \vskip\baselineskip%
    \vspace{-2.2em}
    \begin{subfigure}[b]{0.15\linewidth}
		\centering
		\includegraphics[width=\linewidth]{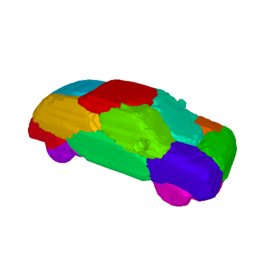}
    \end{subfigure}%
    \hfill%
    \begin{subfigure}[b]{0.15\linewidth}
		\centering
		\includegraphics[width=\linewidth]{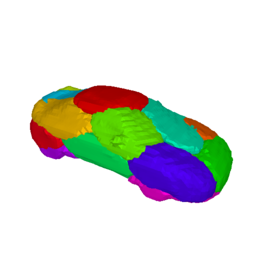}
    \end{subfigure}%
    \hfill%
    \begin{subfigure}[b]{0.15\linewidth}
		\centering
		\includegraphics[width=\linewidth]{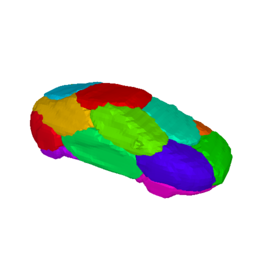}
    \end{subfigure}%
    \hfill%
    \begin{subfigure}[b]{0.15\linewidth}
		\centering
		\includegraphics[width=\linewidth]{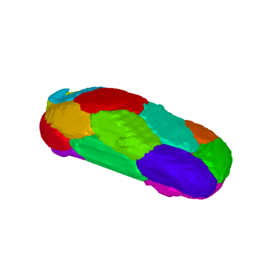}
    \end{subfigure}%
    \hfill%
    \begin{subfigure}[b]{0.15\linewidth}
		\centering
		\includegraphics[width=\linewidth]{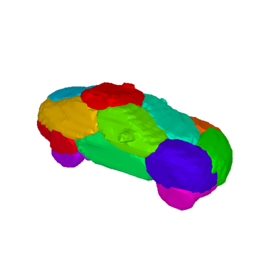}
    \end{subfigure}%
    \hfill%
    \begin{subfigure}[b]{0.15\linewidth}
		\centering
		\includegraphics[width=\linewidth]{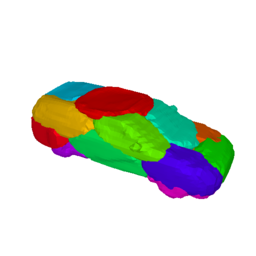}
    \end{subfigure}%
    \caption{{\bf Generated Cars}. Generated cars accompanied by
    their corresponding 3D geometries and part-based geometries.}
    \label{fig:generated_cars}
    \vspace{-1.2em}
\end{figure}

\begin{figure}[!h]
    \centering
    \begin{subfigure}[b]{0.15\linewidth}
		\centering
		\includegraphics[width=\linewidth]{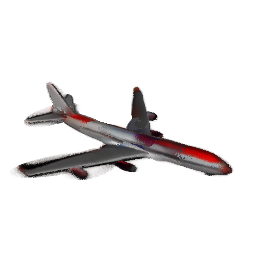}
    \end{subfigure}%
    \hfill%
    \begin{subfigure}[b]{0.15\linewidth}
		\centering
		\includegraphics[width=\linewidth]{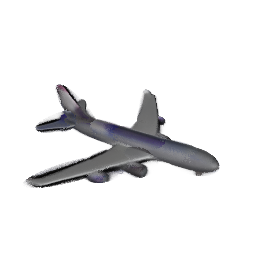}
    \end{subfigure}%
    \hfill%
    \begin{subfigure}[b]{0.15\linewidth}
		\centering
		\includegraphics[width=\linewidth]{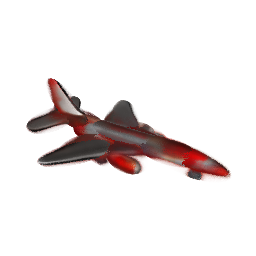}
    \end{subfigure}%
    \hfill%
    \begin{subfigure}[b]{0.15\linewidth}
		\centering
		\includegraphics[width=\linewidth]{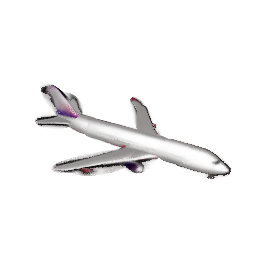}
    \end{subfigure}%
    \hfill%
    \begin{subfigure}[b]{0.15\linewidth}
		\centering
		\includegraphics[width=\linewidth]{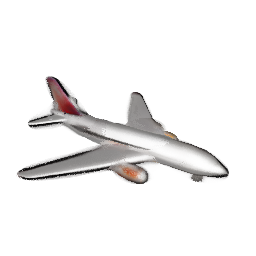}
    \end{subfigure}%
    \hfill%
    \begin{subfigure}[b]{0.15\linewidth}
		\centering
		\includegraphics[width=\linewidth]{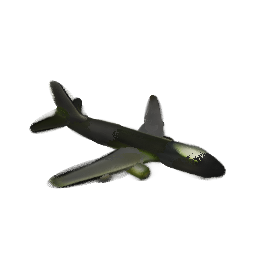}
    \end{subfigure}%
    \vskip\baselineskip%
    \vspace{-2.2em}
    \begin{subfigure}[b]{0.15\linewidth}
		\centering
		\includegraphics[width=\linewidth]{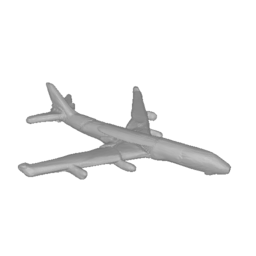}
    \end{subfigure}%
    \hfill%
    \begin{subfigure}[b]{0.15\linewidth}
		\centering
		\includegraphics[width=\linewidth]{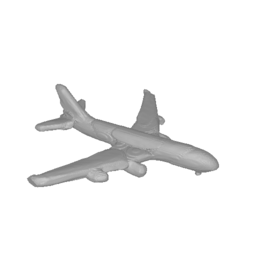}
    \end{subfigure}%
    \hfill%
    \begin{subfigure}[b]{0.15\linewidth}
		\centering
		\includegraphics[width=\linewidth]{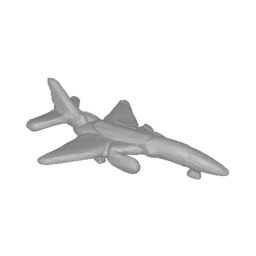}
    \end{subfigure}%
    \hfill%
    \begin{subfigure}[b]{0.15\linewidth}
		\centering
		\includegraphics[width=\linewidth]{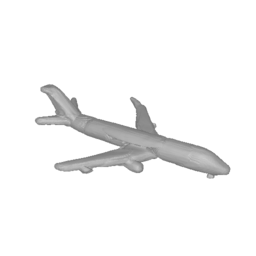}
    \end{subfigure}%
    \hfill%
    \begin{subfigure}[b]{0.15\linewidth}
		\centering
		\includegraphics[width=\linewidth]{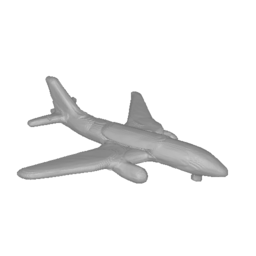}
    \end{subfigure}%
    \hfill%
    \begin{subfigure}[b]{0.15\linewidth}
		\centering
		\includegraphics[width=\linewidth]{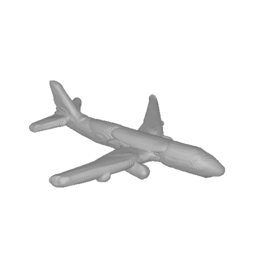}
    \end{subfigure}%
    \vskip\baselineskip%
    \vspace{-2.2em}
    \begin{subfigure}[b]{0.15\linewidth}
		\centering
		\includegraphics[width=\linewidth]{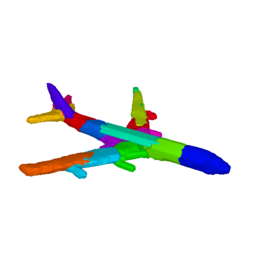}
    \end{subfigure}%
    \hfill%
    \begin{subfigure}[b]{0.15\linewidth}
		\centering
		\includegraphics[width=\linewidth]{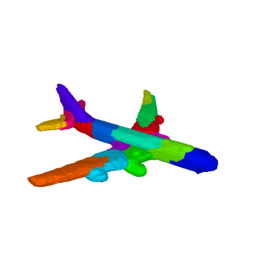}
    \end{subfigure}%
    \hfill%
    \begin{subfigure}[b]{0.15\linewidth}
		\centering
		\includegraphics[width=\linewidth]{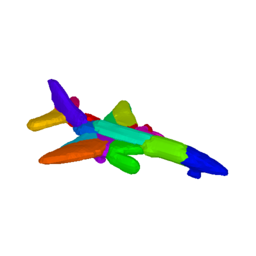}
    \end{subfigure}%
    \hfill%
    \begin{subfigure}[b]{0.15\linewidth}
		\centering
		\includegraphics[width=\linewidth]{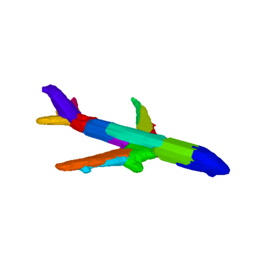}
    \end{subfigure}%
    \hfill%
    \begin{subfigure}[b]{0.15\linewidth}
		\centering
		\includegraphics[width=\linewidth]{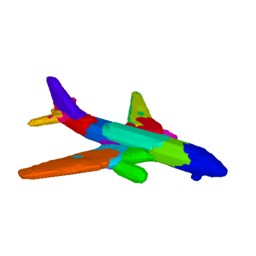}
    \end{subfigure}%
    \hfill%
    \begin{subfigure}[b]{0.15\linewidth}
		\centering
		\includegraphics[width=\linewidth]{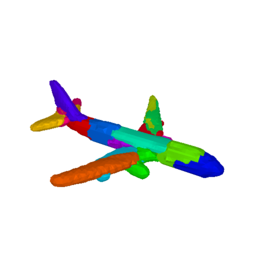}
    \end{subfigure}%
    \vskip\baselineskip%
    \vspace{-2.2em}
    \begin{subfigure}[b]{0.15\linewidth}
		\centering
		\includegraphics[width=\linewidth]{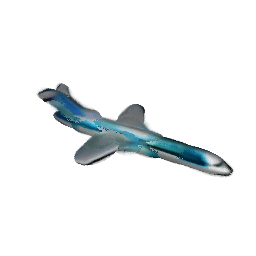}
    \end{subfigure}%
    \hfill%
    \begin{subfigure}[b]{0.15\linewidth}
		\centering
		\includegraphics[width=\linewidth]{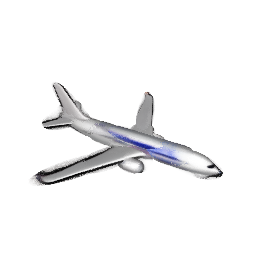}
    \end{subfigure}%
    \hfill%
    \begin{subfigure}[b]{0.15\linewidth}
		\centering
		\includegraphics[width=\linewidth]{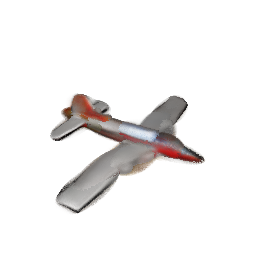}
    \end{subfigure}%
    \hfill%
    \begin{subfigure}[b]{0.15\linewidth}
		\centering
		\includegraphics[width=\linewidth]{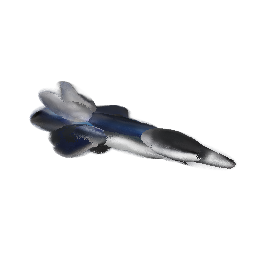}
    \end{subfigure}%
    \hfill%
    \begin{subfigure}[b]{0.15\linewidth}
		\centering
		\includegraphics[width=\linewidth]{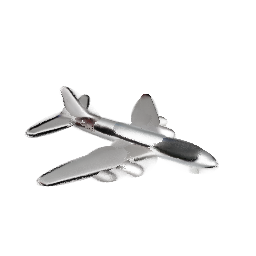}
    \end{subfigure}%
    \hfill%
    \begin{subfigure}[b]{0.15\linewidth}
		\centering
		\includegraphics[width=\linewidth]{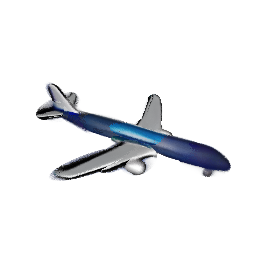}
    \end{subfigure}%
    \vskip\baselineskip%
    \vspace{-2.2em}
    \begin{subfigure}[b]{0.15\linewidth}
		\centering
		\includegraphics[width=\linewidth]{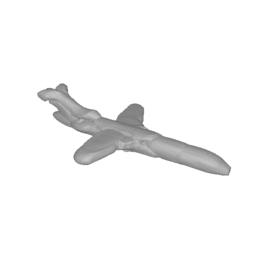}
    \end{subfigure}%
    \hfill%
    \begin{subfigure}[b]{0.15\linewidth}
		\centering
		\includegraphics[width=\linewidth]{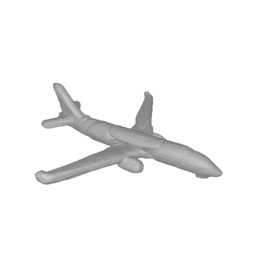}
    \end{subfigure}%
    \hfill%
    \begin{subfigure}[b]{0.15\linewidth}
		\centering
		\includegraphics[width=\linewidth]{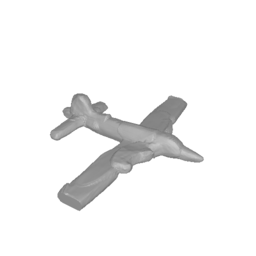}
    \end{subfigure}%
    \hfill%
    \begin{subfigure}[b]{0.15\linewidth}
		\centering
		\includegraphics[width=\linewidth]{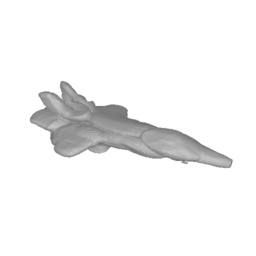}
    \end{subfigure}%
    \hfill%
    \begin{subfigure}[b]{0.15\linewidth}
		\centering
		\includegraphics[width=\linewidth]{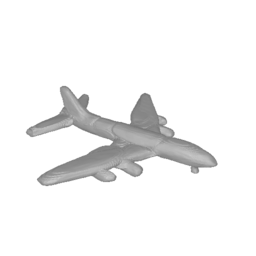}
    \end{subfigure}%
    \hfill%
    \begin{subfigure}[b]{0.15\linewidth}
		\centering
		\includegraphics[width=\linewidth]{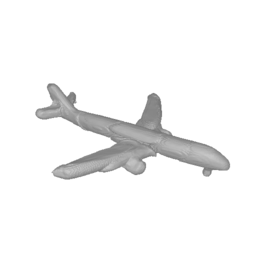}
    \end{subfigure}%
    \vskip\baselineskip%
    \vspace{-2.2em}
    \begin{subfigure}[b]{0.15\linewidth}
		\centering
		\includegraphics[width=\linewidth]{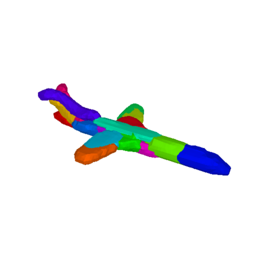}
    \end{subfigure}%
    \hfill%
    \begin{subfigure}[b]{0.15\linewidth}
		\centering
		\includegraphics[width=\linewidth]{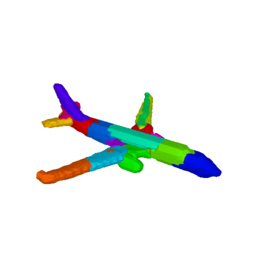}
    \end{subfigure}%
    \hfill%
    \begin{subfigure}[b]{0.15\linewidth}
		\centering
		\includegraphics[width=\linewidth]{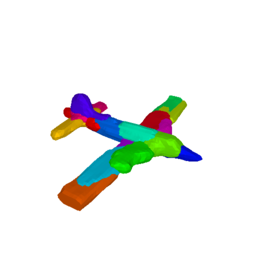}
    \end{subfigure}%
    \hfill%
    \begin{subfigure}[b]{0.15\linewidth}
		\centering
		\includegraphics[width=\linewidth]{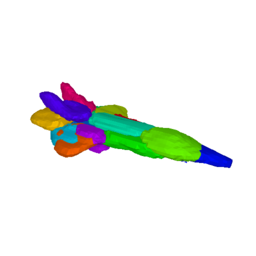}
    \end{subfigure}%
    \hfill%
    \begin{subfigure}[b]{0.15\linewidth}
		\centering
		\includegraphics[width=\linewidth]{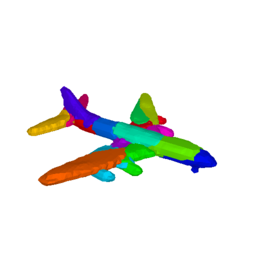}
    \end{subfigure}%
    \hfill%
    \begin{subfigure}[b]{0.15\linewidth}
		\centering
		\includegraphics[width=\linewidth]{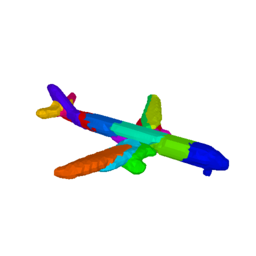}
    \end{subfigure}%
    \vskip\baselineskip%
    \vspace{-2.2em}
    \begin{subfigure}[b]{0.15\linewidth}
		\centering
		\includegraphics[width=\linewidth]{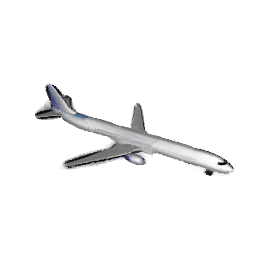}
    \end{subfigure}%
    \hfill%
    \begin{subfigure}[b]{0.15\linewidth}
		\centering
		\includegraphics[width=\linewidth]{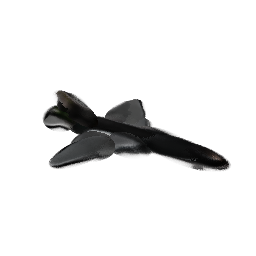}
    \end{subfigure}%
    \hfill%
    \begin{subfigure}[b]{0.15\linewidth}
		\centering
		\includegraphics[width=\linewidth]{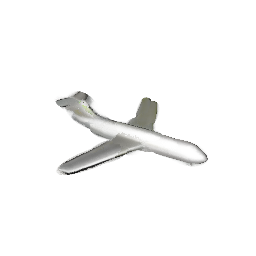}
    \end{subfigure}%
    \hfill%
    \begin{subfigure}[b]{0.15\linewidth}
		\centering
		\includegraphics[width=\linewidth]{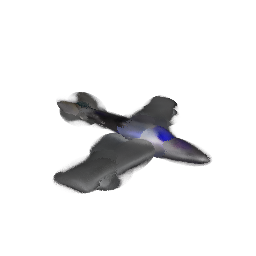}
    \end{subfigure}%
    \hfill%
    \begin{subfigure}[b]{0.15\linewidth}
		\centering
		\includegraphics[width=\linewidth]{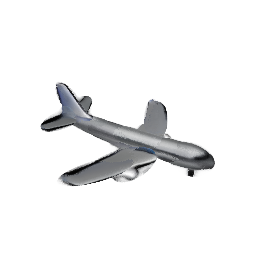}
    \end{subfigure}%
    \hfill%
    \begin{subfigure}[b]{0.15\linewidth}
		\centering
		\includegraphics[width=\linewidth]{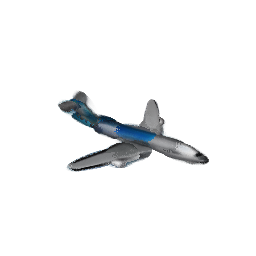}
    \end{subfigure}%
    \vskip\baselineskip%
    \vspace{-2.2em}
    \begin{subfigure}[b]{0.15\linewidth}
		\centering
		\includegraphics[width=\linewidth]{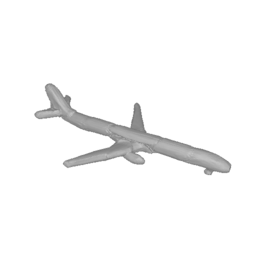}
    \end{subfigure}%
    \hfill%
    \begin{subfigure}[b]{0.15\linewidth}
		\centering
		\includegraphics[width=\linewidth]{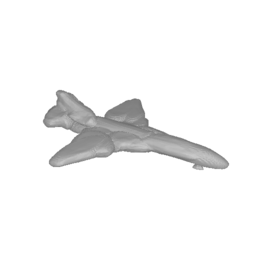}
    \end{subfigure}%
    \hfill%
    \begin{subfigure}[b]{0.15\linewidth}
		\centering
		\includegraphics[width=\linewidth]{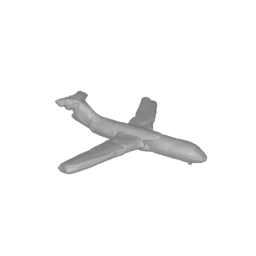}
    \end{subfigure}%
    \hfill%
    \begin{subfigure}[b]{0.15\linewidth}
		\centering
		\includegraphics[width=\linewidth]{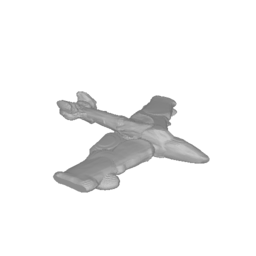}
    \end{subfigure}%
    \hfill%
    \begin{subfigure}[b]{0.15\linewidth}
		\centering
		\includegraphics[width=\linewidth]{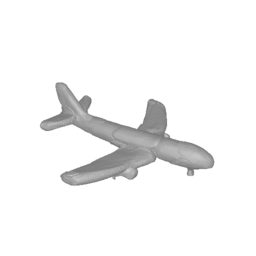}
    \end{subfigure}%
    \hfill%
    \begin{subfigure}[b]{0.15\linewidth}
		\centering
		\includegraphics[width=\linewidth]{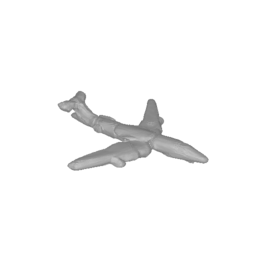}
    \end{subfigure}%
    \vskip\baselineskip%
    \vspace{-2.2em}
    \begin{subfigure}[b]{0.15\linewidth}
		\centering
		\includegraphics[width=\linewidth]{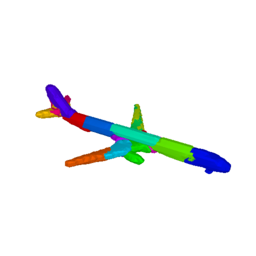}
    \end{subfigure}%
    \hfill%
    \begin{subfigure}[b]{0.15\linewidth}
		\centering
		\includegraphics[width=\linewidth]{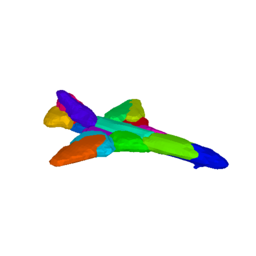}
    \end{subfigure}%
    \hfill%
    \begin{subfigure}[b]{0.15\linewidth}
		\centering
		\includegraphics[width=\linewidth]{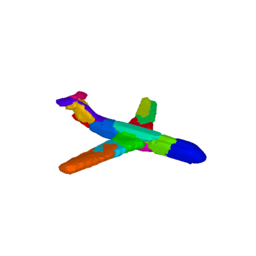}
    \end{subfigure}%
    \hfill%
    \begin{subfigure}[b]{0.15\linewidth}
		\centering
		\includegraphics[width=\linewidth]{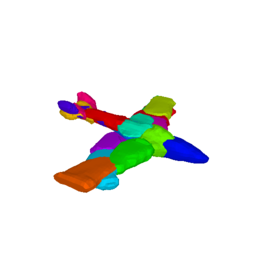}
    \end{subfigure}%
    \hfill%
    \begin{subfigure}[b]{0.15\linewidth}
		\centering
		\includegraphics[width=\linewidth]{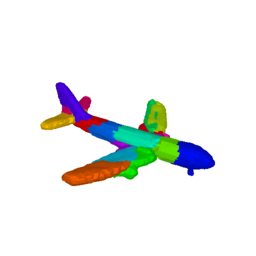}
    \end{subfigure}%
    \hfill%
    \begin{subfigure}[b]{0.15\linewidth}
		\centering
		\includegraphics[width=\linewidth]{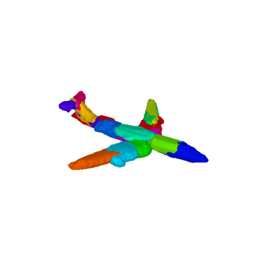}
    \end{subfigure}%
    \caption{{\bf Generated Airplanes}. Generated airplanes accompanied by
    their corresponding 3D geometries and part-based geometries.}
    \label{fig:generated_airplanes}
    \vspace{-1.2em}
\end{figure}

\begin{figure}[!h]
    \centering
    \begin{subfigure}[b]{0.14\linewidth}
		\centering
		\includegraphics[width=\linewidth]{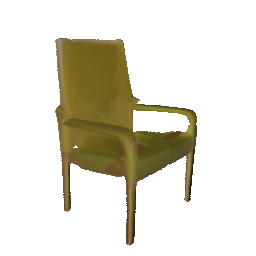}
    \end{subfigure}%
    \hfill%
    \begin{subfigure}[b]{0.14\linewidth}
		\centering
		\includegraphics[width=\linewidth]{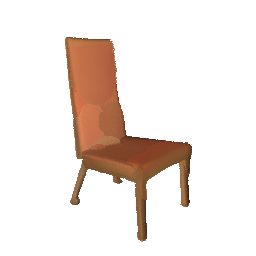}
    \end{subfigure}%
    \hfill%
    \begin{subfigure}[b]{0.14\linewidth}
		\centering
		\includegraphics[width=\linewidth]{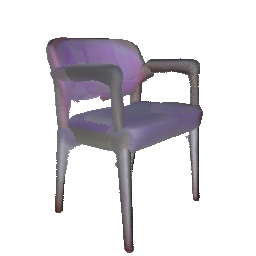}
    \end{subfigure}%
    \hfill%
    \begin{subfigure}[b]{0.14\linewidth}
		\centering
		\includegraphics[width=\linewidth]{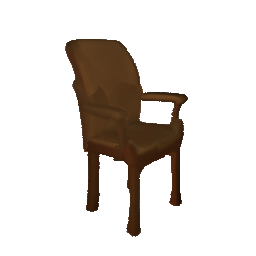}
    \end{subfigure}%
    \hfill%
    \begin{subfigure}[b]{0.14\linewidth}
		\centering
		\includegraphics[width=\linewidth]{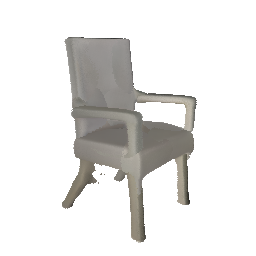}
    \end{subfigure}%
    \hfill%
    \begin{subfigure}[b]{0.14\linewidth}
		\centering
		\includegraphics[width=\linewidth]{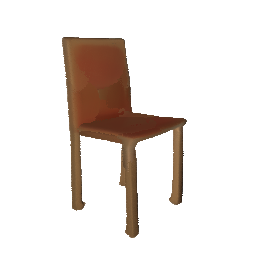}
    \end{subfigure}%
    \vskip\baselineskip%
    \vspace{-1.2em}
    \begin{subfigure}[b]{0.14\linewidth}
		\centering
		\includegraphics[width=\linewidth]{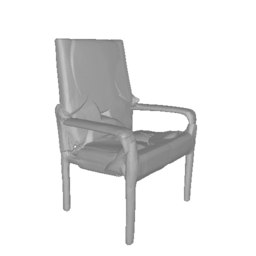}
    \end{subfigure}%
    \hfill%
    \begin{subfigure}[b]{0.14\linewidth}
		\centering
		\includegraphics[width=\linewidth]{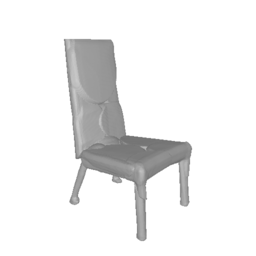}
    \end{subfigure}%
    \hfill%
    \begin{subfigure}[b]{0.14\linewidth}
		\centering
		\includegraphics[width=\linewidth]{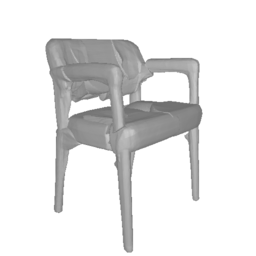}
    \end{subfigure}%
    \hfill%
    \begin{subfigure}[b]{0.14\linewidth}
		\centering
		\includegraphics[width=\linewidth]{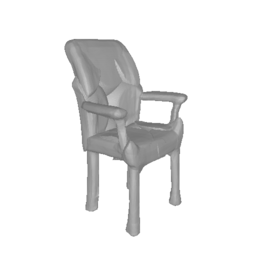}
    \end{subfigure}%
    \hfill%
    \begin{subfigure}[b]{0.14\linewidth}
		\centering
		\includegraphics[width=\linewidth]{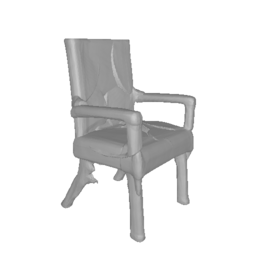}
    \end{subfigure}%
    \hfill%
    \begin{subfigure}[b]{0.14\linewidth}
		\centering
		\includegraphics[width=\linewidth]{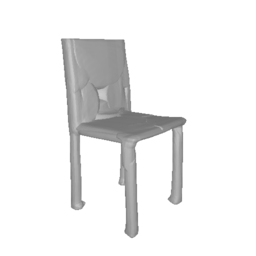}
    \end{subfigure}%
    \vskip\baselineskip%
    \vspace{-1.2em}
    \begin{subfigure}[b]{0.14\linewidth}
		\centering
		\includegraphics[width=\linewidth]{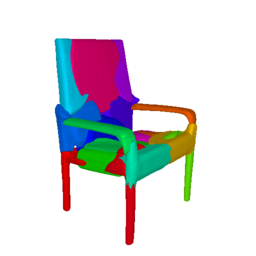}
    \end{subfigure}%
    \hfill%
    \begin{subfigure}[b]{0.14\linewidth}
		\centering
		\includegraphics[width=\linewidth]{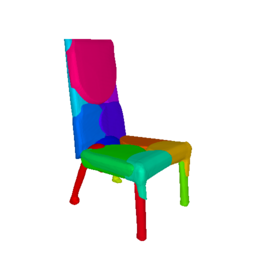}
    \end{subfigure}%
    \hfill%
    \begin{subfigure}[b]{0.14\linewidth}
		\centering
		\includegraphics[width=\linewidth]{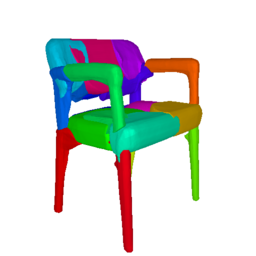}
    \end{subfigure}%
    \hfill%
    \begin{subfigure}[b]{0.14\linewidth}
		\centering
		\includegraphics[width=\linewidth]{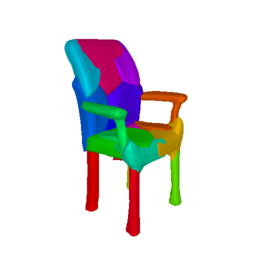}
    \end{subfigure}%
    \hfill%
    \begin{subfigure}[b]{0.14\linewidth}
		\centering
		\includegraphics[width=\linewidth]{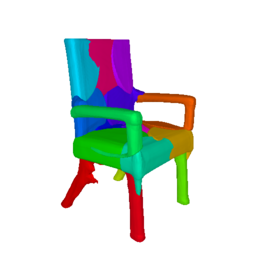}
    \end{subfigure}%
    \hfill%
    \begin{subfigure}[b]{0.14\linewidth}
		\centering
		\includegraphics[width=\linewidth]{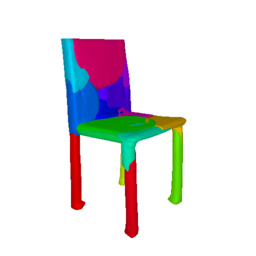}
    \end{subfigure}%
    \vskip\baselineskip%
    \vspace{-1.2em}
    \begin{subfigure}[b]{0.14\linewidth}
		\centering
		\includegraphics[width=\linewidth]{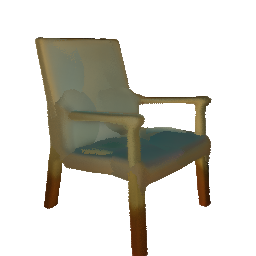}
    \end{subfigure}%
    \hfill%
    \begin{subfigure}[b]{0.14\linewidth}
		\centering
		\includegraphics[width=\linewidth]{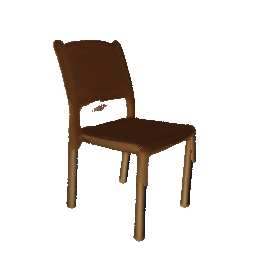}
    \end{subfigure}%
    \hfill%
    \begin{subfigure}[b]{0.14\linewidth}
		\centering
		\includegraphics[width=\linewidth]{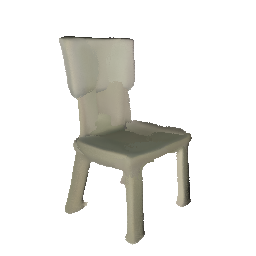}
    \end{subfigure}%
    \hfill%
    \begin{subfigure}[b]{0.14\linewidth}
		\centering
		\includegraphics[width=\linewidth]{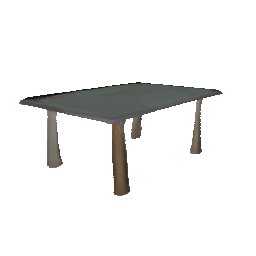}
    \end{subfigure}%
    \hfill%
    \begin{subfigure}[b]{0.14\linewidth}
		\centering
		\includegraphics[width=\linewidth]{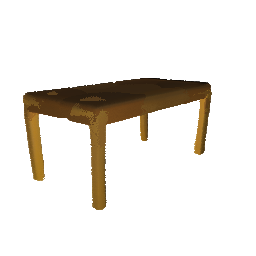}
    \end{subfigure}%
    \hfill%
    \begin{subfigure}[b]{0.14\linewidth}
		\centering
		\includegraphics[width=\linewidth]{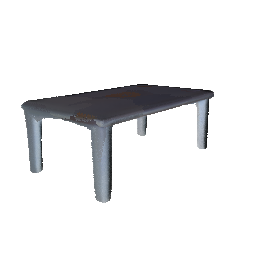}
    \end{subfigure}%
    \vskip\baselineskip%
    \vspace{-1.2em}
    \begin{subfigure}[b]{0.14\linewidth}
		\centering
		\includegraphics[width=\linewidth]{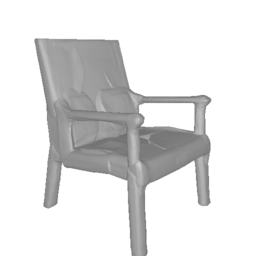}
    \end{subfigure}%
    \hfill%
    \begin{subfigure}[b]{0.14\linewidth}
		\centering
		\includegraphics[width=\linewidth]{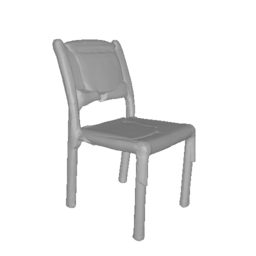}
    \end{subfigure}%
    \hfill%
    \begin{subfigure}[b]{0.14\linewidth}
		\centering
		\includegraphics[width=\linewidth]{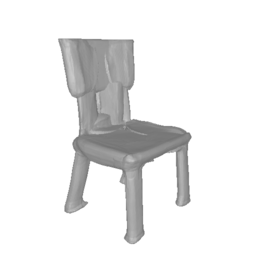}
    \end{subfigure}%
    \hfill%
    \begin{subfigure}[b]{0.14\linewidth}
		\centering
		\includegraphics[width=\linewidth]{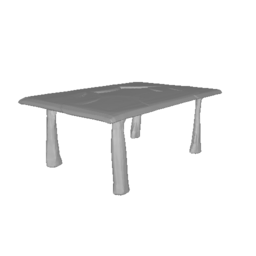}
    \end{subfigure}%
    \hfill%
    \begin{subfigure}[b]{0.14\linewidth}
		\centering
		\includegraphics[width=\linewidth]{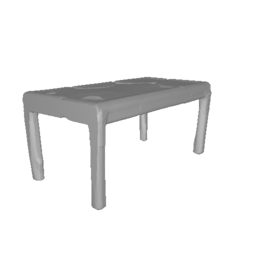}
    \end{subfigure}%
    \hfill%
    \begin{subfigure}[b]{0.14\linewidth}
		\centering
		\includegraphics[width=\linewidth]{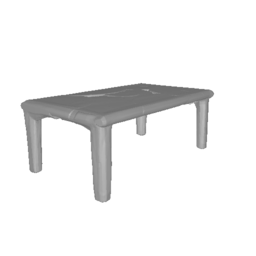}
    \end{subfigure}%
    \vskip\baselineskip%
    \vspace{-1.2em}
    \begin{subfigure}[b]{0.14\linewidth}
		\centering
		\includegraphics[width=\linewidth]{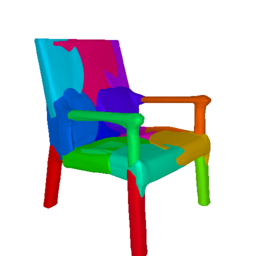}
    \end{subfigure}%
    \hfill%
    \begin{subfigure}[b]{0.14\linewidth}
		\centering
		\includegraphics[width=\linewidth]{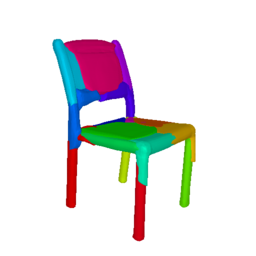}
    \end{subfigure}%
    \hfill%
    \begin{subfigure}[b]{0.14\linewidth}
		\centering
		\includegraphics[width=\linewidth]{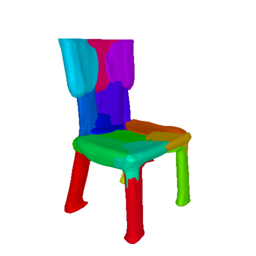}
    \end{subfigure}%
    \hfill%
    \begin{subfigure}[b]{0.14\linewidth}
		\centering
		\includegraphics[width=\linewidth]{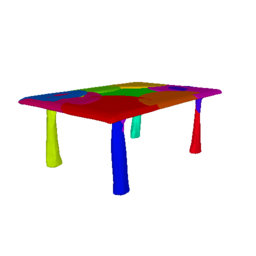}
    \end{subfigure}%
    \hfill%
    \begin{subfigure}[b]{0.14\linewidth}
		\centering
		\includegraphics[width=\linewidth]{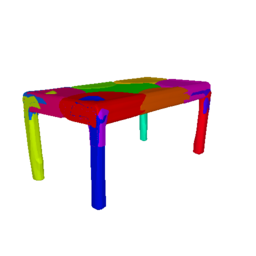}
    \end{subfigure}%
    \hfill%
    \begin{subfigure}[b]{0.14\linewidth}
		\centering
		\includegraphics[width=\linewidth]{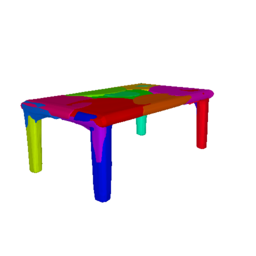}
    \end{subfigure}%
    \vskip\baselineskip%
    \vspace{-1.2em}
    \begin{subfigure}[b]{0.14\linewidth}
		\centering
		\includegraphics[width=\linewidth]{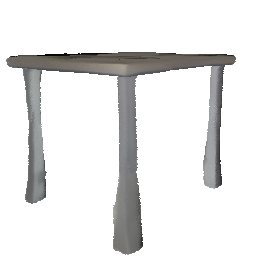}
    \end{subfigure}%
    \hfill%
    \begin{subfigure}[b]{0.14\linewidth}
		\centering
		\includegraphics[width=\linewidth]{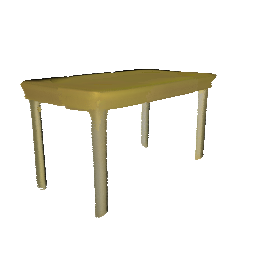}
    \end{subfigure}%
    \hfill%
    \begin{subfigure}[b]{0.14\linewidth}
		\centering
		\includegraphics[width=\linewidth]{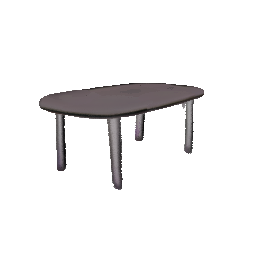}
    \end{subfigure}%
    \hfill
    \begin{subfigure}[b]{0.14\linewidth}
		\centering
		\includegraphics[width=\linewidth]{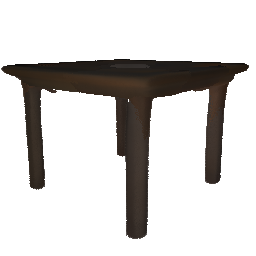}
    \end{subfigure}%
    \hfill%
    \begin{subfigure}[b]{0.14\linewidth}
		\centering
		\includegraphics[width=\linewidth]{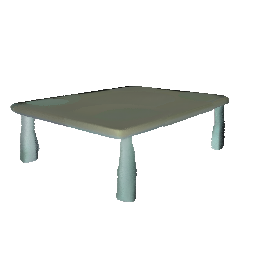}
    \end{subfigure}%
    \hfill%
    \begin{subfigure}[b]{0.14\linewidth}
		\centering
		\includegraphics[width=\linewidth]{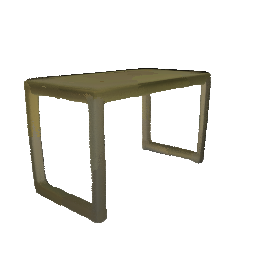}
    \end{subfigure}%
    \vskip\baselineskip%
    \vspace{-1.2em}
    \begin{subfigure}[b]{0.14\linewidth}
		\centering
		\includegraphics[width=\linewidth]{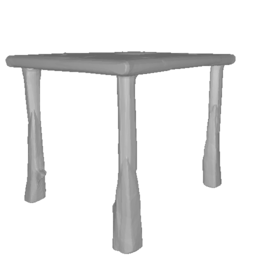}
    \end{subfigure}%
    \hfill%
    \begin{subfigure}[b]{0.14\linewidth}
		\centering
		\includegraphics[width=\linewidth]{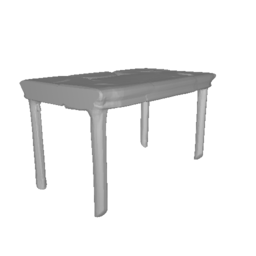}
    \end{subfigure}%
    \hfill%
    \begin{subfigure}[b]{0.14\linewidth}
		\centering
		\includegraphics[width=\linewidth]{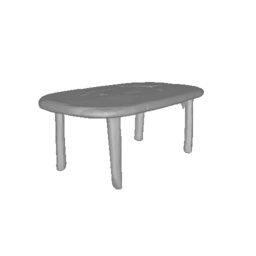}
    \end{subfigure}%
    \hfill
    \begin{subfigure}[b]{0.14\linewidth}
		\centering
		\includegraphics[width=\linewidth]{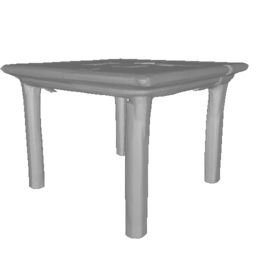}
    \end{subfigure}%
    \hfill%
    \begin{subfigure}[b]{0.14\linewidth}
		\centering
		\includegraphics[width=\linewidth]{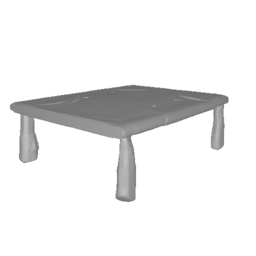}
    \end{subfigure}%
    \hfill%
    \begin{subfigure}[b]{0.14\linewidth}
		\centering
		\includegraphics[width=\linewidth]{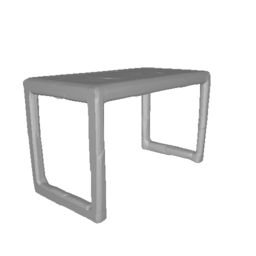}
    \end{subfigure}%
    \vskip\baselineskip%
    \vspace{-1.2em}
    \begin{subfigure}[b]{0.14\linewidth}
		\centering
		\includegraphics[width=\linewidth]{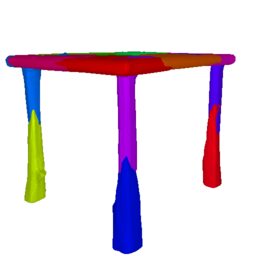}
    \end{subfigure}%
    \hfill%
    \begin{subfigure}[b]{0.14\linewidth}
		\centering
		\includegraphics[width=\linewidth]{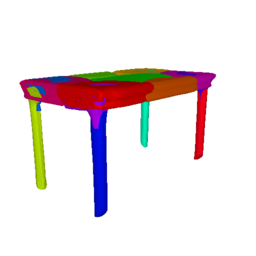}
    \end{subfigure}%
    \hfill%
    \begin{subfigure}[b]{0.14\linewidth}
		\centering
		\includegraphics[width=\linewidth]{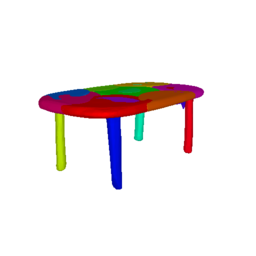}
    \end{subfigure}%
    \hfill
    \begin{subfigure}[b]{0.14\linewidth}
		\centering
		\includegraphics[width=\linewidth]{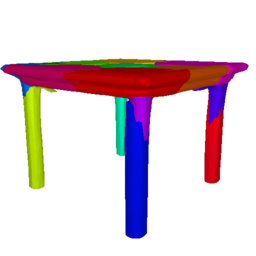}
    \end{subfigure}%
    \hfill%
    \begin{subfigure}[b]{0.14\linewidth}
		\centering
		\includegraphics[width=\linewidth]{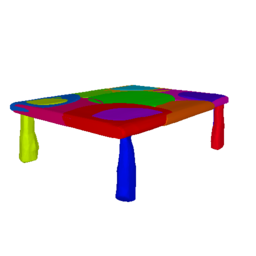}
    \end{subfigure}%
    \hfill%
    \begin{subfigure}[b]{0.14\linewidth}
		\centering
		\includegraphics[width=\linewidth]{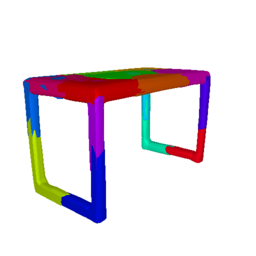}
    \end{subfigure}%
    \caption{{\bf Generated Chairs \& Tables}. Generated chairs and tables accompanied by
    their corresponding 3D geometries and part-based geometries.}
    \label{fig:generated_chairs_tables}
    \vspace{-1.2em}
\end{figure}

In this section, we provide additional qualitative results for our shape
generation experiment on the five ShapeNet~\cite{Chang2015ARXIV} categories:
\emph{Motorbike, Chair, Table, Airplane, Car}. In particular, we visualize
several random samples per category, rendered from a novel view point, not seen
during training. In addition, we also show the corresponding 3D meshes and
parts. Specifically, in \figref{fig:generated_motorbikes}, we illustrate
randomly generated motorbikes. We notice that all motorbikes are consistently
plausible and the underlying parts meaningfully capture several motorbike's
geometry. Likewise, for the case of generated cars (see
\figref{fig:generated_cars}) and airplanes (see
\figref{fig:generated_airplanes}), we observe that our model is capable of generating diverse 
geometries and textures that are consistently plausible. For the case of
chairs and tables (see \figref{fig:generated_chairs_tables}), we notice that the geometries are consistently
plausible, as also evidenced from Table 1 in our main submission, however the
generated textures are less crisp in comparison to the other categories. We
hypothesize that this discrepancy stems from the fact that tables and chairs
have thinner parts than the other objects. We believe that incorporating a
coarse and a fine volume of 3D coordinates would potentially improve this.

\clearpage
\newpage

\subsubsection{Comparison with Part-Based Generative Models}

\begin{figure}[!h]
    \centering
    \begin{subfigure}[b]{0.14\linewidth}
		\centering
		\includegraphics[width=\linewidth]{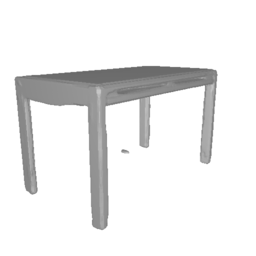}
    \end{subfigure}%
    \hfill%
    \begin{subfigure}[b]{0.14\linewidth}
		\centering
		\includegraphics[width=\linewidth]{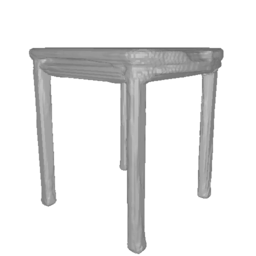}
    \end{subfigure}%
    \hfill%
    \begin{subfigure}[b]{0.14\linewidth}
		\centering
		\includegraphics[width=\linewidth]{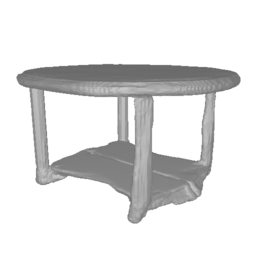}
    \end{subfigure}%
    \hfill%
    \begin{subfigure}[b]{0.14\linewidth}
		\centering
		\includegraphics[width=\linewidth]{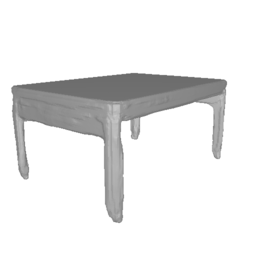}
    \end{subfigure}%
    \hfill%
    \begin{subfigure}[b]{0.14\linewidth}
		\centering
		\includegraphics[width=\linewidth]{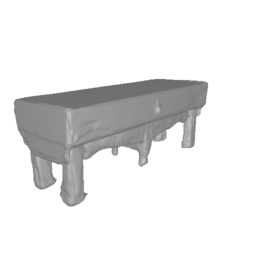}
    \end{subfigure}%
    \hfill%
    \begin{subfigure}[b]{0.14\linewidth}
		\centering
		\includegraphics[width=\linewidth]{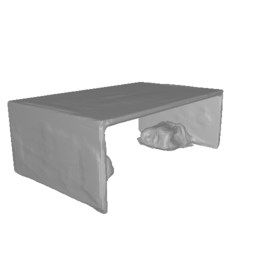}
    \end{subfigure}%
    \vskip\baselineskip%
    \vspace{-2.2em}
    \begin{minipage}[b]{\linewidth}
		\centering
		\scriptsize{SPAGHETTI \cite{Hertz2022SIGGRAPH}}
    \end{minipage}%
    \vskip\baselineskip%
    \vspace{-1.3em}
    \begin{subfigure}[b]{0.14\linewidth}
		\centering
		\includegraphics[width=\linewidth]{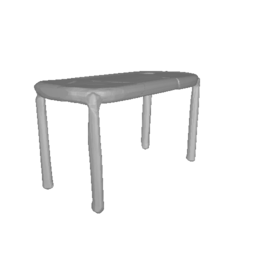}
    \end{subfigure}%
    \hfill%
    \begin{subfigure}[b]{0.14\linewidth}
		\centering
		\includegraphics[width=\linewidth]{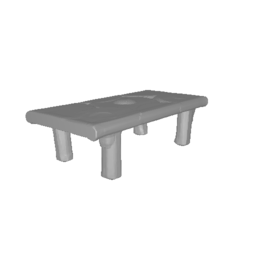}
    \end{subfigure}%
    \hfill%
    \begin{subfigure}[b]{0.14\linewidth}
		\centering
		\includegraphics[width=\linewidth]{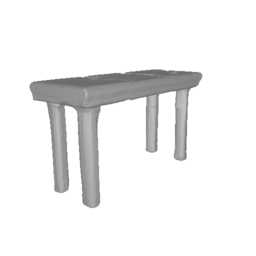}
    \end{subfigure}%
    \hfill%
    \begin{subfigure}[b]{0.14\linewidth}
		\centering
		\includegraphics[width=\linewidth]{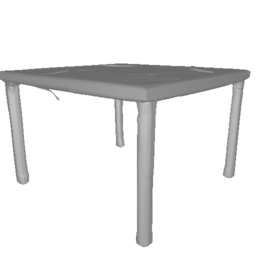}
    \end{subfigure}%
    \hfill%
    \begin{subfigure}[b]{0.14\linewidth}
		\centering
		\includegraphics[width=\linewidth]{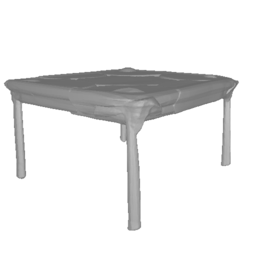}
    \end{subfigure}%
    \hfill%
    \begin{subfigure}[b]{0.14\linewidth}
		\centering
		\includegraphics[width=\linewidth]{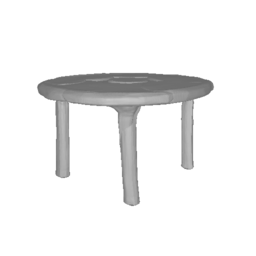}
    \end{subfigure}%
    \vskip\baselineskip%
    \vspace{-2.5em}
    \begin{minipage}[b]{\linewidth}
		\centering
		\scriptsize{Ours}
    \end{minipage}%
    \vskip\baselineskip%
    \vspace{-1.3em}
    \begin{subfigure}[b]{0.14\linewidth}
		\centering
		\includegraphics[width=\linewidth]{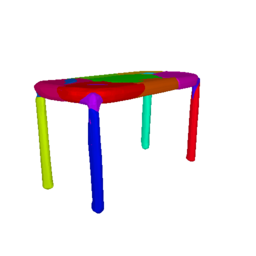}
    \end{subfigure}%
    \hfill%
    \begin{subfigure}[b]{0.14\linewidth}
		\centering
		\includegraphics[width=\linewidth]{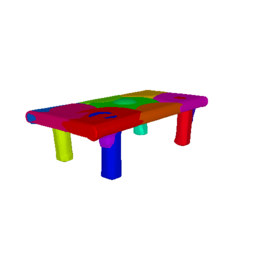}
    \end{subfigure}%
    \hfill%
    \begin{subfigure}[b]{0.14\linewidth}
		\centering
		\includegraphics[width=\linewidth]{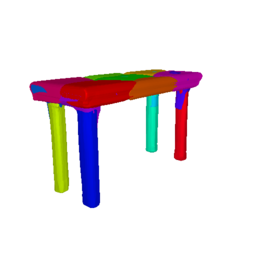}
    \end{subfigure}%
    \hfill%
    \begin{subfigure}[b]{0.14\linewidth}
		\centering
		\includegraphics[width=\linewidth]{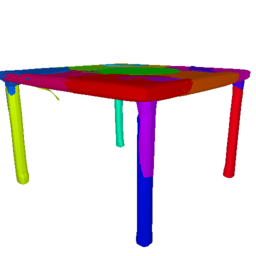}
    \end{subfigure}%
    \hfill%
    \begin{subfigure}[b]{0.14\linewidth}
		\centering
		\includegraphics[width=\linewidth]{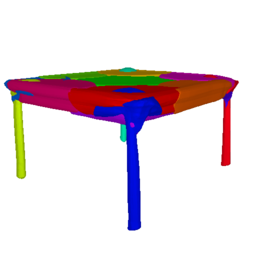}
    \end{subfigure}%
    \hfill%
    \begin{subfigure}[b]{0.14\linewidth}
		\centering
		\includegraphics[width=\linewidth]{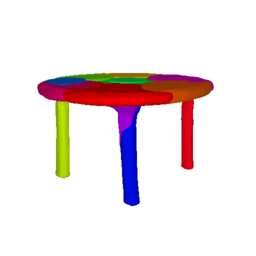}
    \end{subfigure}%
    \vspace{-2.2em}
    \begin{minipage}[b]{\linewidth}
		\centering
		\scriptsize{Ours with Parts}
    \end{minipage}%
    \vskip\baselineskip%
    \vspace{-1.1em}
    \begin{subfigure}[b]{0.14\linewidth}
		\centering
		\includegraphics[width=\linewidth]{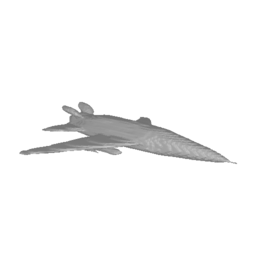}
    \end{subfigure}%
    \hfill%
    \begin{subfigure}[b]{0.14\linewidth}
		\centering
		\includegraphics[width=\linewidth]{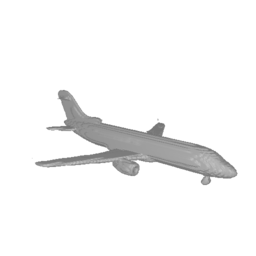}
    \end{subfigure}%
    \hfill%
    \begin{subfigure}[b]{0.14\linewidth}
		\centering
		\includegraphics[width=\linewidth]{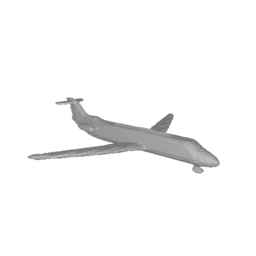}
    \end{subfigure}%
    \hfill%
    \begin{subfigure}[b]{0.14\linewidth}
		\centering
		\includegraphics[width=\linewidth]{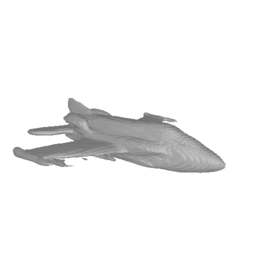}
    \end{subfigure}%
    \hfill%
    \begin{subfigure}[b]{0.14\linewidth}
		\centering
		\includegraphics[width=\linewidth]{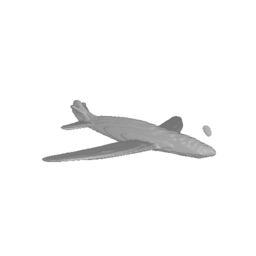}
    \end{subfigure}%
    \hfill%
    \begin{subfigure}[b]{0.14\linewidth}
		\centering
		\includegraphics[width=\linewidth]{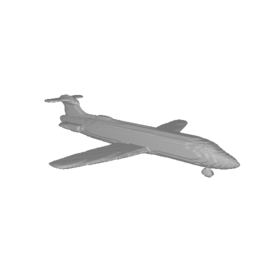}
    \end{subfigure}%
    \vskip\baselineskip%
    \vspace{-2.2em}
    \begin{minipage}[b]{\linewidth}
		\centering
		\scriptsize{SPAGHETTI \cite{Hertz2022SIGGRAPH}}
    \end{minipage}%
    \vskip\baselineskip%
    \vspace{-1.3em}
    \begin{subfigure}[b]{0.14\linewidth}
		\centering
		\includegraphics[width=\linewidth]{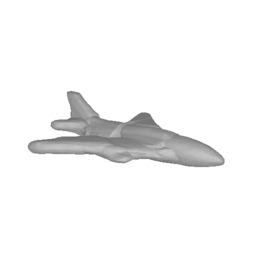}
    \end{subfigure}%
    \hfill%
    \begin{subfigure}[b]{0.14\linewidth}
		\centering
		\includegraphics[width=\linewidth]{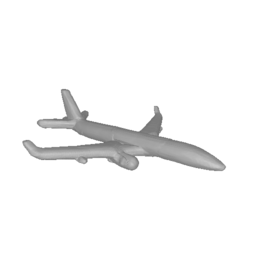}
    \end{subfigure}%
    \hfill%
    \begin{subfigure}[b]{0.14\linewidth}
		\centering
		\includegraphics[width=\linewidth]{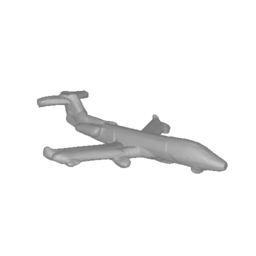}
    \end{subfigure}%
    \hfill%
    \begin{subfigure}[b]{0.14\linewidth}
		\centering
		\includegraphics[width=\linewidth]{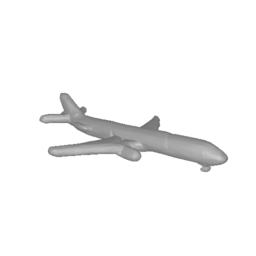}
    \end{subfigure}%
    \hfill%
    \begin{subfigure}[b]{0.14\linewidth}
		\centering
		\includegraphics[width=\linewidth]{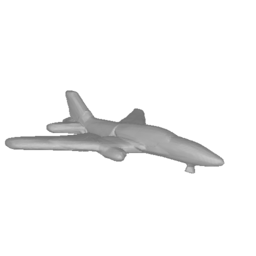}
    \end{subfigure}%
    \hfill%
    \begin{subfigure}[b]{0.14\linewidth}
		\centering
		\includegraphics[width=\linewidth]{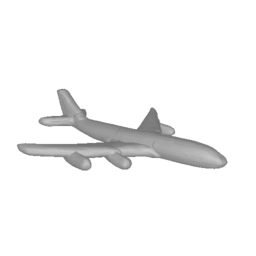}
    \end{subfigure}%
    \vspace{-2.5em}
    \begin{minipage}[b]{\linewidth}
		\centering
		\scriptsize{Ours}
    \end{minipage}%
    \vskip\baselineskip%
    \vspace{-1.3em}
    \begin{subfigure}[b]{0.14\linewidth}
		\centering
		\includegraphics[width=\linewidth]{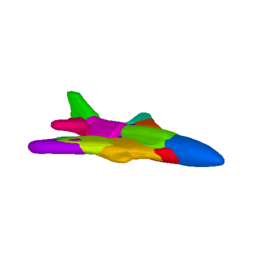}
    \end{subfigure}%
    \hfill%
    \begin{subfigure}[b]{0.14\linewidth}
		\centering
		\includegraphics[width=\linewidth]{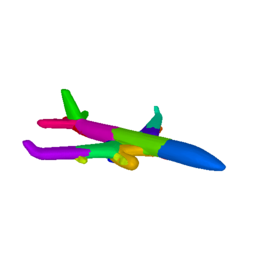}
    \end{subfigure}%
    \hfill%
    \begin{subfigure}[b]{0.14\linewidth}
		\centering
		\includegraphics[width=\linewidth]{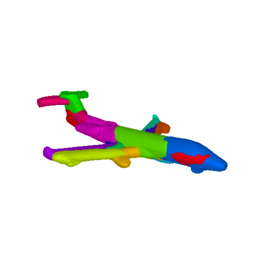}
    \end{subfigure}%
    \hfill%
    \begin{subfigure}[b]{0.14\linewidth}
		\centering
		\includegraphics[width=\linewidth]{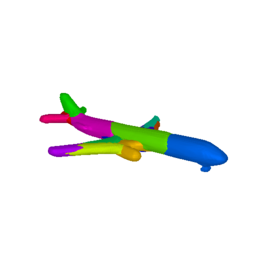}
    \end{subfigure}%
    \hfill%
    \begin{subfigure}[b]{0.14\linewidth}
		\centering
		\includegraphics[width=\linewidth]{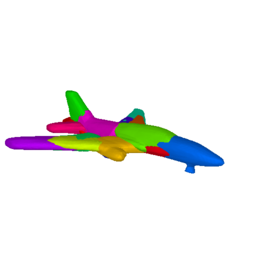}
    \end{subfigure}%
    \hfill%
    \begin{subfigure}[b]{0.14\linewidth}
		\centering
		\includegraphics[width=\linewidth]{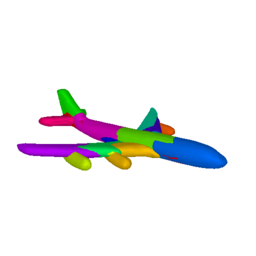}
    \end{subfigure}%
    \vspace{-2.2em}
    \begin{minipage}[b]{\linewidth}
		\centering
		\scriptsize{Ours with Parts}
    \end{minipage}%
    \vskip\baselineskip%
    \vspace{-1.1em}
    \begin{subfigure}[b]{0.14\linewidth}
		\centering
		\includegraphics[width=\linewidth]{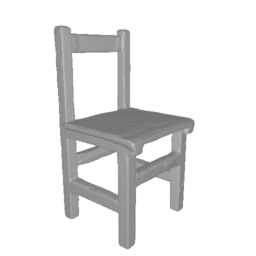}
    \end{subfigure}%
    \hfill%
    \begin{subfigure}[b]{0.14\linewidth}
		\centering
		\includegraphics[width=\linewidth]{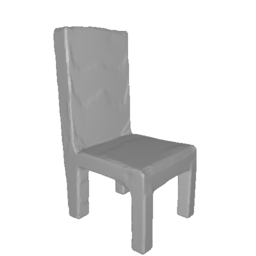}
    \end{subfigure}%
    \hfill%
    \begin{subfigure}[b]{0.14\linewidth}
		\centering
		\includegraphics[width=\linewidth]{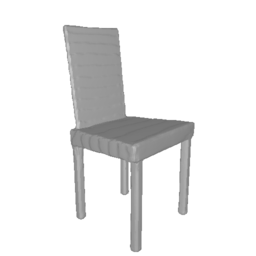}
    \end{subfigure}%
    \hfill%
    \begin{subfigure}[b]{0.14\linewidth}
		\centering
		\includegraphics[width=\linewidth]{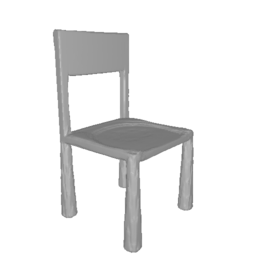}
    \end{subfigure}%
    \hfill%
    \begin{subfigure}[b]{0.14\linewidth}
		\centering
		\includegraphics[width=\linewidth]{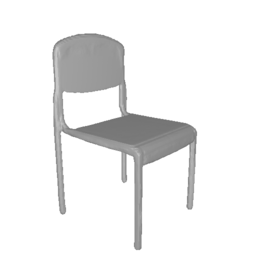}
    \end{subfigure}%
    \hfill%
    \begin{subfigure}[b]{0.14\linewidth}
		\centering
		\includegraphics[width=\linewidth]{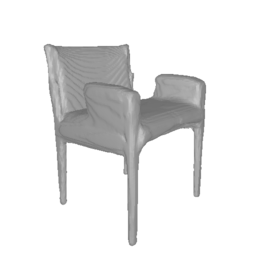}
    \end{subfigure}%
    \vskip\baselineskip%
    \vspace{-2.2em}
    \begin{minipage}[b]{\linewidth}
		\centering
		\scriptsize{SPAGHETTI \cite{Hertz2022SIGGRAPH}}
    \end{minipage}%
    \vskip\baselineskip%
    \vspace{-1.3em}
    \begin{subfigure}[b]{0.14\linewidth}
		\centering
		\includegraphics[width=\linewidth]{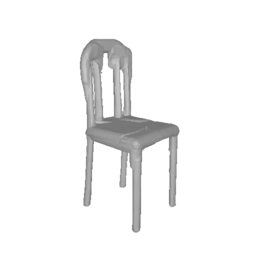}
    \end{subfigure}%
    \hfill%
    \begin{subfigure}[b]{0.14\linewidth}
		\centering
		\includegraphics[width=\linewidth]{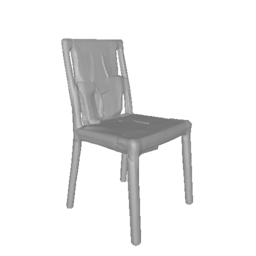}
    \end{subfigure}%
    \hfill%
    \begin{subfigure}[b]{0.14\linewidth}
		\centering
		\includegraphics[width=\linewidth]{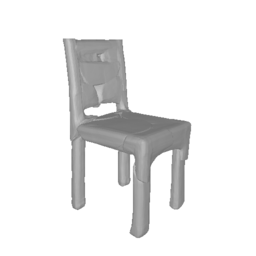}
    \end{subfigure}%
    \hfill%
    \begin{subfigure}[b]{0.14\linewidth}
		\centering
		\includegraphics[width=\linewidth]{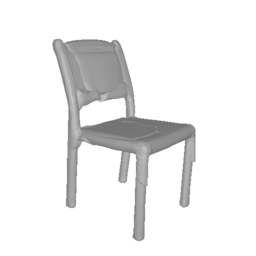}
    \end{subfigure}%
    \hfill%
    \begin{subfigure}[b]{0.14\linewidth}
		\centering
		\includegraphics[width=\linewidth]{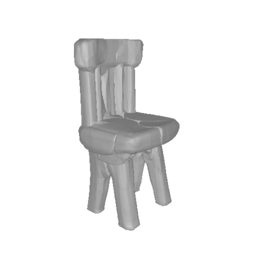}
    \end{subfigure}%
    \hfill%
    \begin{subfigure}[b]{0.14\linewidth}
		\centering
		\includegraphics[width=\linewidth]{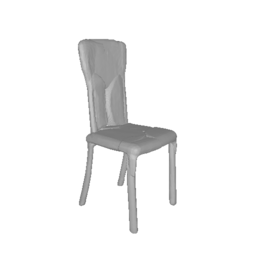}
    \end{subfigure}%
    \vskip\baselineskip%
    \vspace{-2.5em}
    \begin{minipage}[b]{\linewidth}
		\centering
		\scriptsize{Ours}
    \end{minipage}%
    \vskip\baselineskip%
    \vspace{-1.3em}
    \begin{subfigure}[b]{0.14\linewidth}
		\centering
		\includegraphics[width=\linewidth]{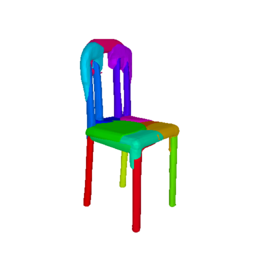}
    \end{subfigure}%
    \hfill%
    \begin{subfigure}[b]{0.14\linewidth}
		\centering
		\includegraphics[width=\linewidth]{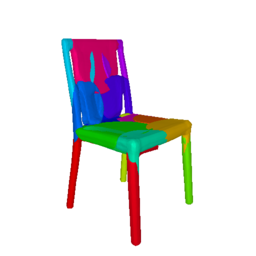}
    \end{subfigure}%
    \hfill%
    \begin{subfigure}[b]{0.14\linewidth}
		\centering
		\includegraphics[width=\linewidth]{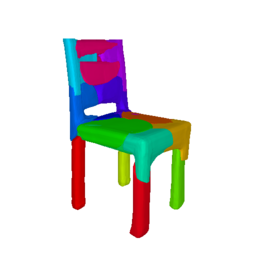}
    \end{subfigure}%
    \hfill%
    \begin{subfigure}[b]{0.14\linewidth}
		\centering
		\includegraphics[width=\linewidth]{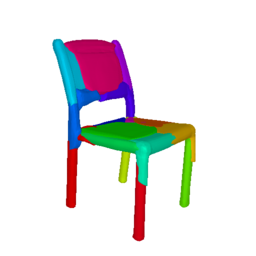}
    \end{subfigure}%
    \hfill%
    \begin{subfigure}[b]{0.14\linewidth}
		\centering
		\includegraphics[width=\linewidth]{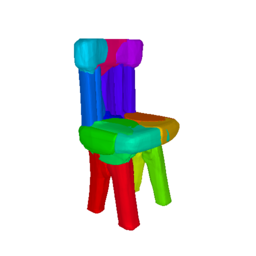}
    \end{subfigure}%
    \hfill%
    \begin{subfigure}[b]{0.14\linewidth}
		\centering
		\includegraphics[width=\linewidth]{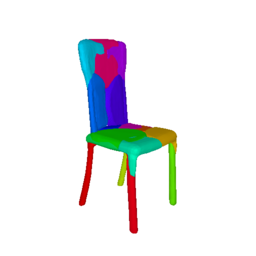}
    \end{subfigure}%
    \vskip\baselineskip%
    \vspace{-2.5em}
    \begin{minipage}[b]{\linewidth}
		\centering
		\scriptsize{Ours}
    \end{minipage}%
    \caption{{\bf Shape Generation Comparisons}. We compare our model with
    \cite{Hertz2022SIGGRAPH} and show six randomly generated samples per
    category. The first row per category corresponds to samples from
    \cite{Hertz2022SIGGRAPH}, the second row to our results, and the third row
    to our part geometry results.}
    \label{fig:spaghetti_comparisons_supp}
    \vspace{-1.2em}
\end{figure}

In this section, we provide more qualitative comparisons of our generated
shapes with existing part-based generative methods~\cite{Hertz2022SIGGRAPH},
that require explicit 3D supervision in the form of occupancy labels. In
particular, we visualize random samples for three
ShapeNet~\cite{Chang2015ARXIV} categories: \emph{Table, Airplane, Chair}.
Results are summarized in \figref{fig:spaghetti_comparisons_supp}. Our method can
generate geometries of similar quality to~\cite{Hertz2022SIGGRAPH}, without the
need of explicit 3D supervision.

\subsection{Shape Interpolation}
\begin{figure}[!h]
    \centering
    \begin{subfigure}[b]{0.15\linewidth}
		\centering
		\includegraphics[width=\linewidth]{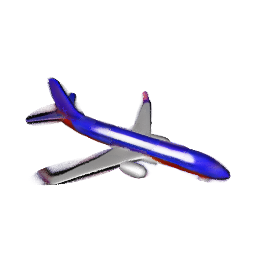}
    \end{subfigure}%
    \hfill%
    \begin{subfigure}[b]{0.15\linewidth}
		\centering
		\includegraphics[width=\linewidth]{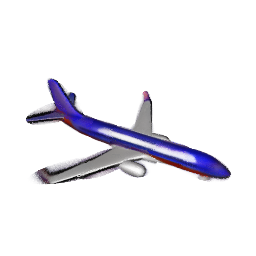}
    \end{subfigure}%
    \hfill%
    \begin{subfigure}[b]{0.15\linewidth}
		\centering
		\includegraphics[width=\linewidth]{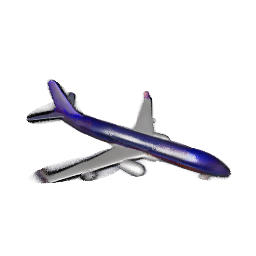}
    \end{subfigure}%
    \hfill%
    \begin{subfigure}[b]{0.15\linewidth}
		\centering
		\includegraphics[width=\linewidth]{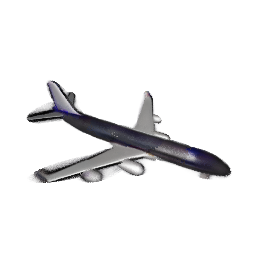}
    \end{subfigure}%
    \hfill%
    \begin{subfigure}[b]{0.15\linewidth}
		\centering
		\includegraphics[width=\linewidth]{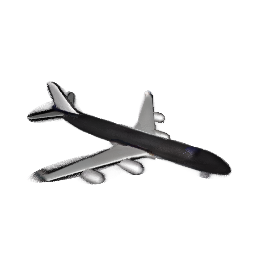}
    \end{subfigure}%
    \hfill%
    \begin{subfigure}[b]{0.15\linewidth}
		\centering
		\includegraphics[width=\linewidth]{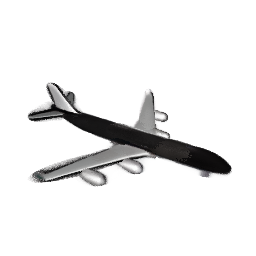}
    \end{subfigure}%
    \vskip\baselineskip%
    \vspace{-2.2em}
    \begin{subfigure}[b]{0.15\linewidth}
		\centering
		\includegraphics[width=\linewidth]{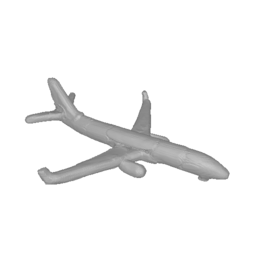}
    \end{subfigure}%
    \hfill%
    \begin{subfigure}[b]{0.15\linewidth}
		\centering
		\includegraphics[width=\linewidth]{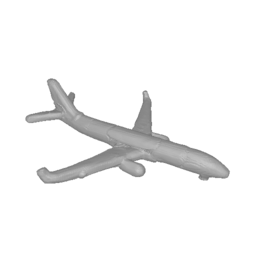}
    \end{subfigure}%
    \hfill%
    \begin{subfigure}[b]{0.15\linewidth}
		\centering
		\includegraphics[width=\linewidth]{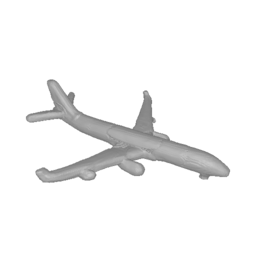}
    \end{subfigure}%
    \hfill%
    \begin{subfigure}[b]{0.15\linewidth}
		\centering
		\includegraphics[width=\linewidth]{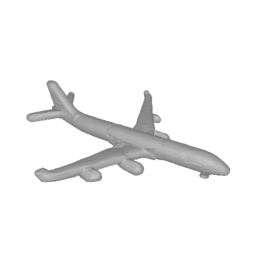}
    \end{subfigure}%
    \hfill%
    \begin{subfigure}[b]{0.15\linewidth}
		\centering
		\includegraphics[width=\linewidth]{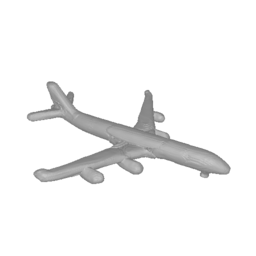}
    \end{subfigure}%
    \hfill%
    \begin{subfigure}[b]{0.15\linewidth}
		\centering
		\includegraphics[width=\linewidth]{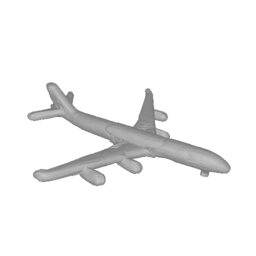}
    \end{subfigure}%
    \vskip\baselineskip%
    \vspace{-2.2em}
    \begin{subfigure}[b]{0.15\linewidth}
		\centering
		\includegraphics[width=\linewidth]{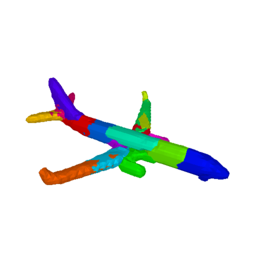}
    \end{subfigure}%
    \hfill%
    \begin{subfigure}[b]{0.15\linewidth}
		\centering
		\includegraphics[width=\linewidth]{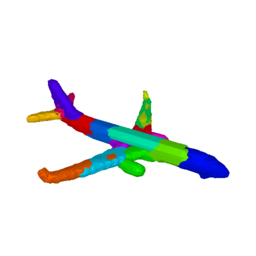}
    \end{subfigure}%
    \hfill%
    \begin{subfigure}[b]{0.15\linewidth}
		\centering
		\includegraphics[width=\linewidth]{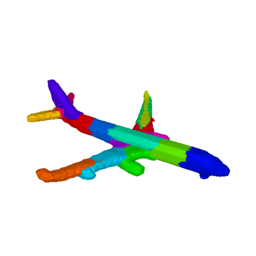}
    \end{subfigure}%
    \hfill%
    \begin{subfigure}[b]{0.15\linewidth}
		\centering
		\includegraphics[width=\linewidth]{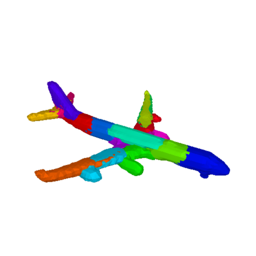}
    \end{subfigure}%
    \hfill%
    \begin{subfigure}[b]{0.15\linewidth}
		\centering
		\includegraphics[width=\linewidth]{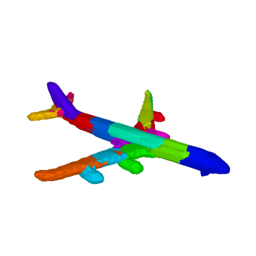}
    \end{subfigure}%
    \hfill%
    \begin{subfigure}[b]{0.15\linewidth}
		\centering
		\includegraphics[width=\linewidth]{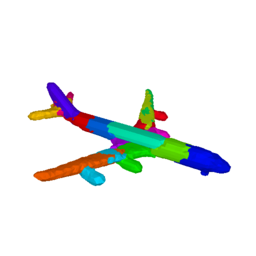}
    \end{subfigure}%
    \vskip\baselineskip%
    \vspace{-1.5em}
    \begin{subfigure}[b]{0.15\linewidth}
		\centering
		\includegraphics[width=\linewidth]{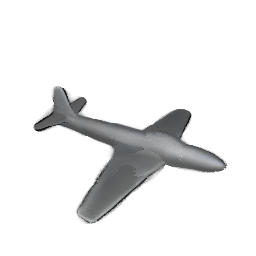}
    \end{subfigure}%
    \hfill%
    \begin{subfigure}[b]{0.15\linewidth}
		\centering
		\includegraphics[width=\linewidth]{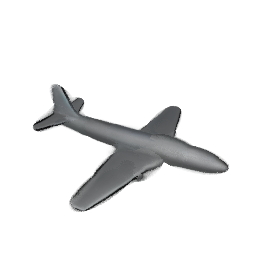}
    \end{subfigure}%
    \hfill%
    \begin{subfigure}[b]{0.15\linewidth}
		\centering
		\includegraphics[width=\linewidth]{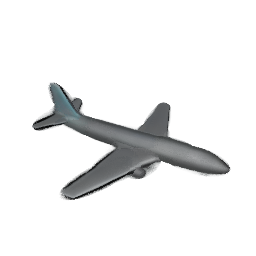}
    \end{subfigure}%
    \hfill%
    \begin{subfigure}[b]{0.15\linewidth}
		\centering
		\includegraphics[width=\linewidth]{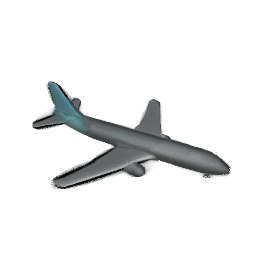}
    \end{subfigure}%
    \hfill%
    \begin{subfigure}[b]{0.15\linewidth}
		\centering
		\includegraphics[width=\linewidth]{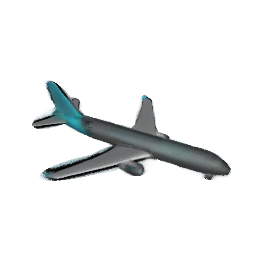}
    \end{subfigure}%
    \hfill%
    \begin{subfigure}[b]{0.15\linewidth}
		\centering
		\includegraphics[width=\linewidth]{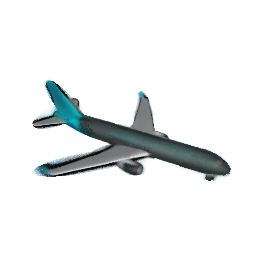}
    \end{subfigure}%
    \vskip\baselineskip%
    \vspace{-2.2em}
    \begin{subfigure}[b]{0.15\linewidth}
		\centering
		\includegraphics[width=\linewidth]{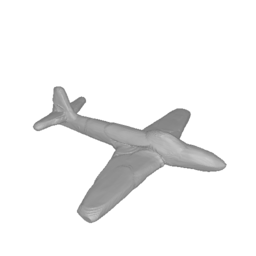}
    \end{subfigure}%
    \hfill%
    \begin{subfigure}[b]{0.15\linewidth}
		\centering
		\includegraphics[width=\linewidth]{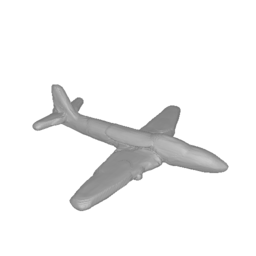}
    \end{subfigure}%
    \hfill%
    \begin{subfigure}[b]{0.15\linewidth}
		\centering
		\includegraphics[width=\linewidth]{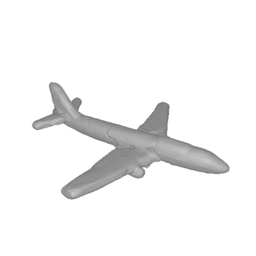}
    \end{subfigure}%
    \hfill%
    \begin{subfigure}[b]{0.15\linewidth}
		\centering
		\includegraphics[width=\linewidth]{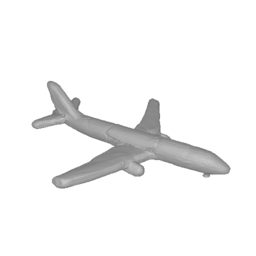}
    \end{subfigure}%
    \hfill%
    \begin{subfigure}[b]{0.15\linewidth}
		\centering
		\includegraphics[width=\linewidth]{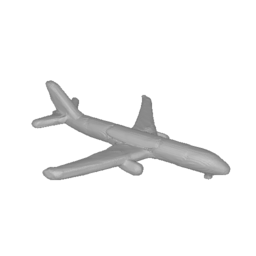}
    \end{subfigure}%
    \hfill%
    \begin{subfigure}[b]{0.15\linewidth}
		\centering
		\includegraphics[width=\linewidth]{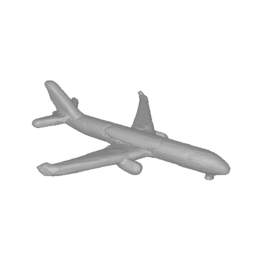}
    \end{subfigure}%
    \vskip\baselineskip%
    \vspace{-2.2em}
    \begin{subfigure}[b]{0.15\linewidth}
		\centering
		\includegraphics[width=\linewidth]{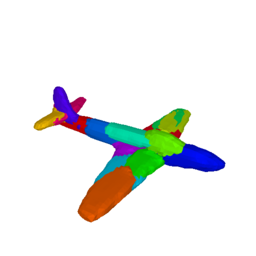}
    \end{subfigure}%
    \hfill%
    \begin{subfigure}[b]{0.15\linewidth}
		\centering
		\includegraphics[width=\linewidth]{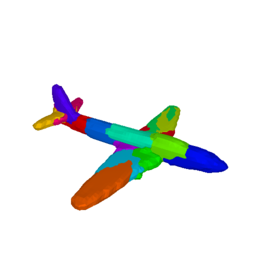}
    \end{subfigure}%
    \hfill%
    \begin{subfigure}[b]{0.15\linewidth}
		\centering
		\includegraphics[width=\linewidth]{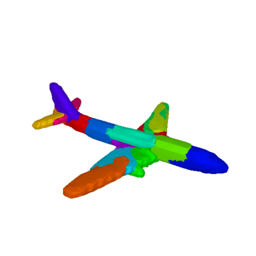}
    \end{subfigure}%
    \hfill%
    \begin{subfigure}[b]{0.15\linewidth}
		\centering
		\includegraphics[width=\linewidth]{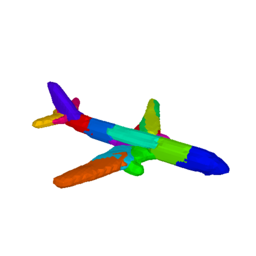}
    \end{subfigure}%
    \hfill%
    \begin{subfigure}[b]{0.15\linewidth}
		\centering
		\includegraphics[width=\linewidth]{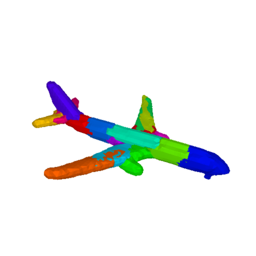}
    \end{subfigure}%
    \hfill%
    \begin{subfigure}[b]{0.15\linewidth}
		\centering
		\includegraphics[width=\linewidth]{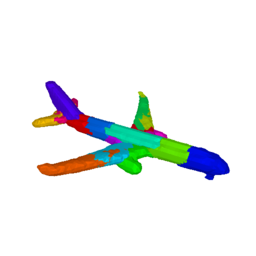}
    \end{subfigure}%
    \vskip\baselineskip%
    \vspace{-1.5em}
    \begin{subfigure}[b]{0.15\linewidth}
		\centering
		\includegraphics[width=\linewidth]{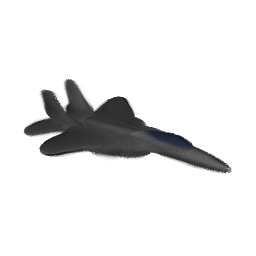}
    \end{subfigure}%
    \hfill%
    \begin{subfigure}[b]{0.15\linewidth}
		\centering
		\includegraphics[width=\linewidth]{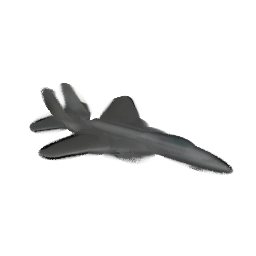}
    \end{subfigure}%
    \hfill%
    \begin{subfigure}[b]{0.15\linewidth}
		\centering
		\includegraphics[width=\linewidth]{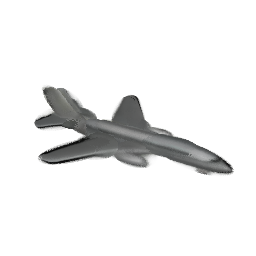}
    \end{subfigure}%
    \hfill%
    \begin{subfigure}[b]{0.15\linewidth}
		\centering
		\includegraphics[width=\linewidth]{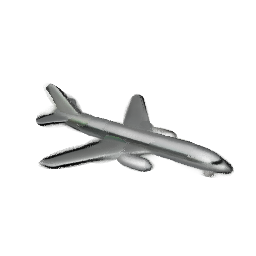}
    \end{subfigure}%
    \hfill%
    \begin{subfigure}[b]{0.15\linewidth}
		\centering
		\includegraphics[width=\linewidth]{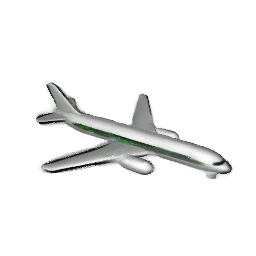}
    \end{subfigure}%
    \hfill%
    \begin{subfigure}[b]{0.15\linewidth}
		\centering
		\includegraphics[width=\linewidth]{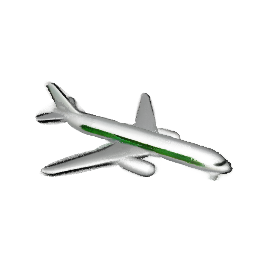}
    \end{subfigure}%
    \vskip\baselineskip%
    \vspace{-2.2em}
    \begin{subfigure}[b]{0.15\linewidth}
		\centering
		\includegraphics[width=\linewidth]{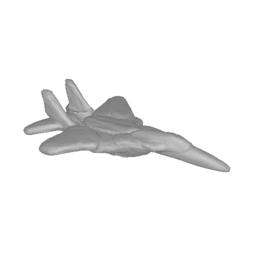}
    \end{subfigure}%
    \hfill%
    \begin{subfigure}[b]{0.15\linewidth}
		\centering
		\includegraphics[width=\linewidth]{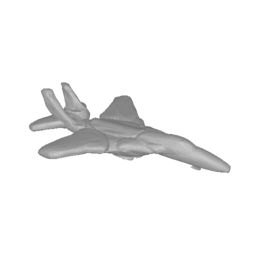}
    \end{subfigure}%
    \hfill%
    \begin{subfigure}[b]{0.15\linewidth}
		\centering
		\includegraphics[width=\linewidth]{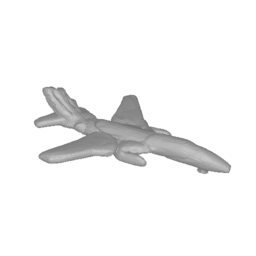}
    \end{subfigure}%
    \hfill%
    \begin{subfigure}[b]{0.15\linewidth}
		\centering
		\includegraphics[width=\linewidth]{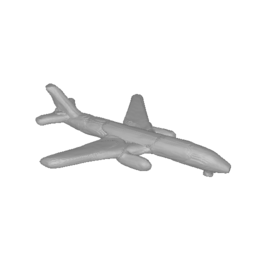}
    \end{subfigure}%
    \hfill%
    \begin{subfigure}[b]{0.15\linewidth}
		\centering
		\includegraphics[width=\linewidth]{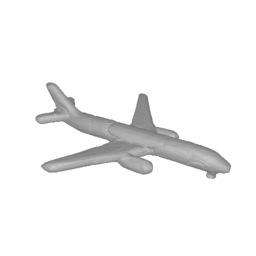}
    \end{subfigure}%
    \hfill%
    \begin{subfigure}[b]{0.15\linewidth}
		\centering
		\includegraphics[width=\linewidth]{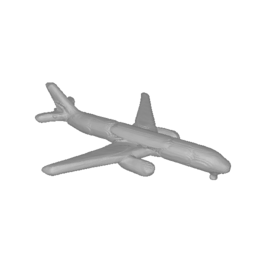}
    \end{subfigure}%
    \vskip\baselineskip%
    \vspace{-2.2em}
    \begin{subfigure}[b]{0.15\linewidth}
		\centering
		\includegraphics[width=\linewidth]{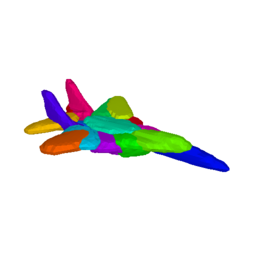}
    \end{subfigure}%
    \hfill%
    \begin{subfigure}[b]{0.15\linewidth}
		\centering
		\includegraphics[width=\linewidth]{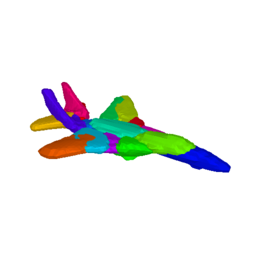}
    \end{subfigure}%
    \hfill%
    \begin{subfigure}[b]{0.15\linewidth}
		\centering
		\includegraphics[width=\linewidth]{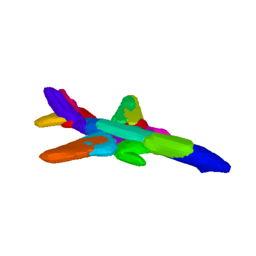}
    \end{subfigure}%
    \hfill%
    \begin{subfigure}[b]{0.15\linewidth}
		\centering
		\includegraphics[width=\linewidth]{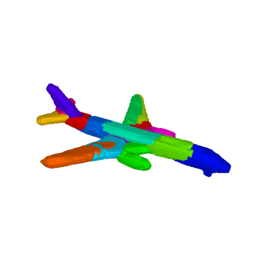}
    \end{subfigure}%
    \hfill%
    \begin{subfigure}[b]{0.15\linewidth}
		\centering
		\includegraphics[width=\linewidth]{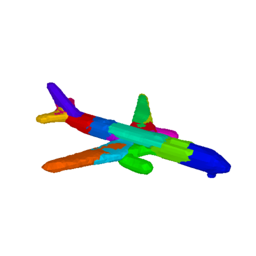}
    \end{subfigure}%
    \hfill%
    \begin{subfigure}[b]{0.15\linewidth}
		\centering
		\includegraphics[width=\linewidth]{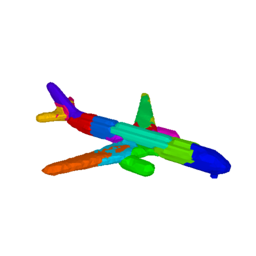}
    \end{subfigure}%
    \caption{{\bf Airplanes Shape Interpolation}.
    From left to right, we interpolate
    between the geometry and texture latent codes of the two airplanes.}
    \label{fig:interpolated_airplanes}
    \vspace{-1.2em}
\end{figure}

\begin{figure}[!h]
    \centering
    \begin{subfigure}[b]{0.15\linewidth}
		\centering
		\includegraphics[width=\linewidth]{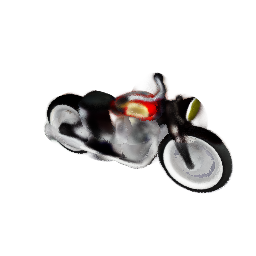}
    \end{subfigure}%
    \hfill%
    \begin{subfigure}[b]{0.15\linewidth}
		\centering
		\includegraphics[width=\linewidth]{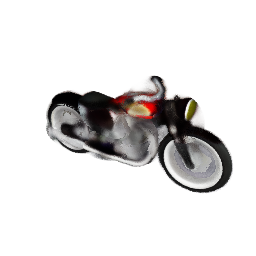}
    \end{subfigure}%
    \hfill%
    \begin{subfigure}[b]{0.15\linewidth}
		\centering
		\includegraphics[width=\linewidth]{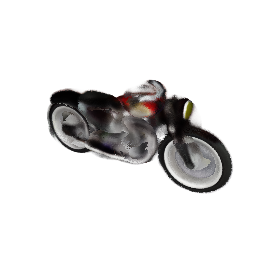}
    \end{subfigure}%
    \hfill%
    \begin{subfigure}[b]{0.15\linewidth}
		\centering
		\includegraphics[width=\linewidth]{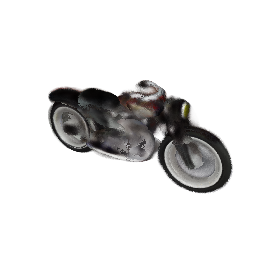}
    \end{subfigure}%
    \hfill%
    \begin{subfigure}[b]{0.15\linewidth}
		\centering
		\includegraphics[width=\linewidth]{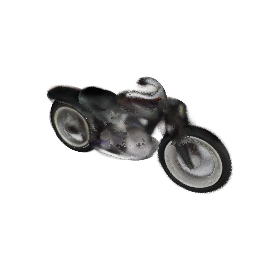}
    \end{subfigure}%
    \hfill%
    \begin{subfigure}[b]{0.15\linewidth}
		\centering
		\includegraphics[width=\linewidth]{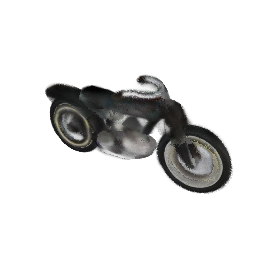}
    \end{subfigure}%
    \vskip\baselineskip%
    \vspace{-2.2em}
    \begin{subfigure}[b]{0.15\linewidth}
		\centering
		\includegraphics[width=\linewidth]{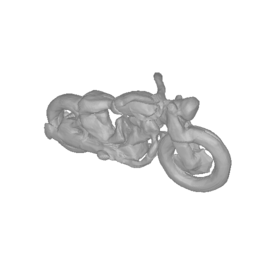}
    \end{subfigure}%
    \hfill%
    \begin{subfigure}[b]{0.15\linewidth}
		\centering
		\includegraphics[width=\linewidth]{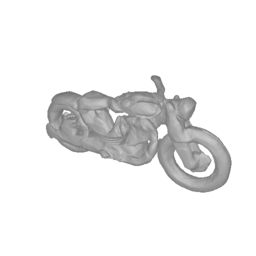}
    \end{subfigure}%
    \hfill%
    \begin{subfigure}[b]{0.15\linewidth}
		\centering
		\includegraphics[width=\linewidth]{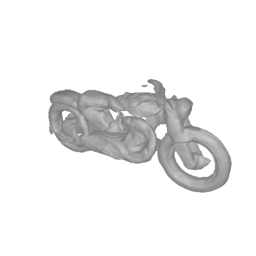}
    \end{subfigure}%
    \hfill%
    \begin{subfigure}[b]{0.15\linewidth}
		\centering
		\includegraphics[width=\linewidth]{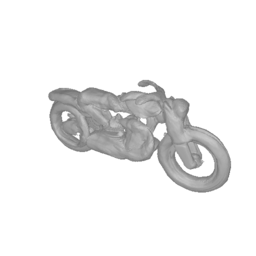}
    \end{subfigure}%
    \hfill%
    \begin{subfigure}[b]{0.15\linewidth}
		\centering
		\includegraphics[width=\linewidth]{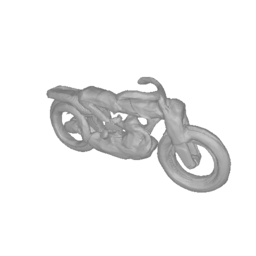}
    \end{subfigure}%
    \hfill%
    \begin{subfigure}[b]{0.15\linewidth}
		\centering
		\includegraphics[width=\linewidth]{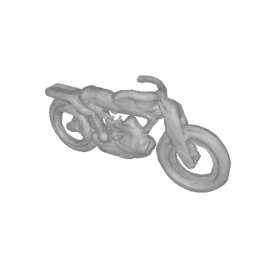}
    \end{subfigure}%
    \vskip\baselineskip%
    \vspace{-2.2em}
    \begin{subfigure}[b]{0.15\linewidth}
		\centering
		\includegraphics[width=\linewidth]{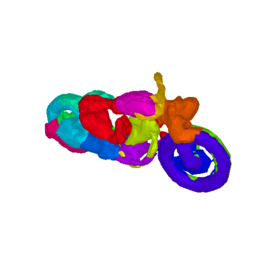}
    \end{subfigure}%
    \hfill%
    \begin{subfigure}[b]{0.15\linewidth}
		\centering
		\includegraphics[width=\linewidth]{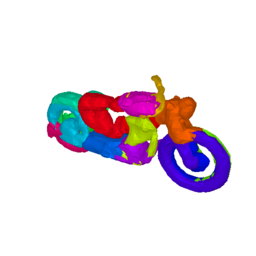}
    \end{subfigure}%
    \hfill%
    \begin{subfigure}[b]{0.15\linewidth}
		\centering
		\includegraphics[width=\linewidth]{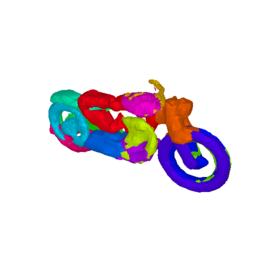}
    \end{subfigure}%
    \hfill%
    \begin{subfigure}[b]{0.15\linewidth}
		\centering
		\includegraphics[width=\linewidth]{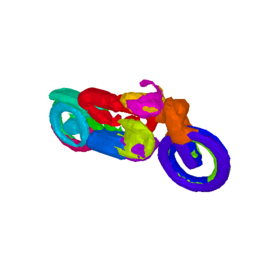}
    \end{subfigure}%
    \hfill%
    \begin{subfigure}[b]{0.15\linewidth}
		\centering
		\includegraphics[width=\linewidth]{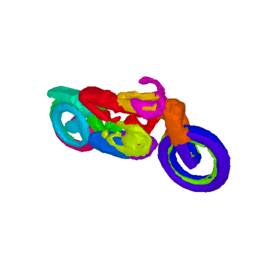}
    \end{subfigure}%
    \hfill%
    \begin{subfigure}[b]{0.15\linewidth}
		\centering
		\includegraphics[width=\linewidth]{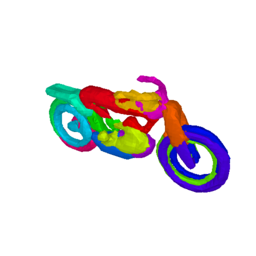}
    \end{subfigure}%
    \vskip\baselineskip%
    \vspace{-1.5em}
    \begin{subfigure}[b]{0.15\linewidth}
		\centering
		\includegraphics[width=\linewidth]{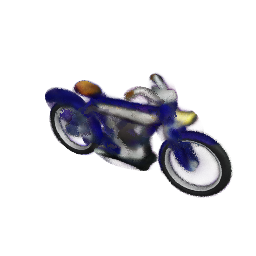}
    \end{subfigure}%
    \hfill%
    \begin{subfigure}[b]{0.15\linewidth}
		\centering
		\includegraphics[width=\linewidth]{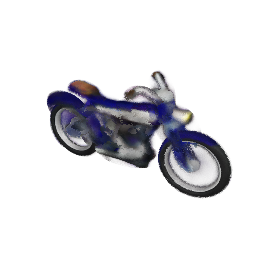}
    \end{subfigure}%
    \hfill%
    \begin{subfigure}[b]{0.15\linewidth}
		\centering
		\includegraphics[width=\linewidth]{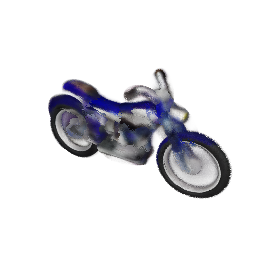}
    \end{subfigure}%
    \hfill%
    \begin{subfigure}[b]{0.15\linewidth}
		\centering
		\includegraphics[width=\linewidth]{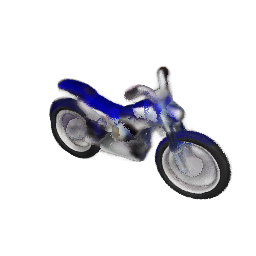}
    \end{subfigure}%
    \hfill%
    \begin{subfigure}[b]{0.15\linewidth}
		\centering
		\includegraphics[width=\linewidth]{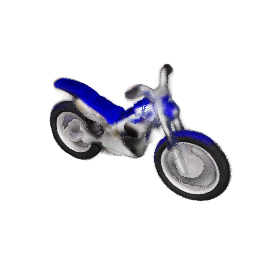}
    \end{subfigure}%
    \hfill%
    \begin{subfigure}[b]{0.15\linewidth}
		\centering
		\includegraphics[width=\linewidth]{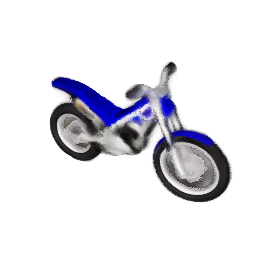}
    \end{subfigure}%
    \vskip\baselineskip%
    \vspace{-2.2em}
    \begin{subfigure}[b]{0.15\linewidth}
		\centering
		\includegraphics[width=\linewidth]{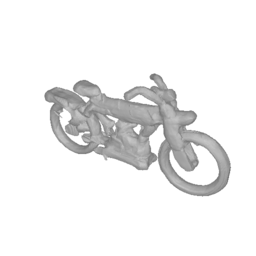}
    \end{subfigure}%
    \hfill%
    \begin{subfigure}[b]{0.15\linewidth}
		\centering
		\includegraphics[width=\linewidth]{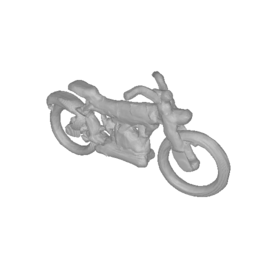}
    \end{subfigure}%
    \hfill%
    \begin{subfigure}[b]{0.15\linewidth}
		\centering
		\includegraphics[width=\linewidth]{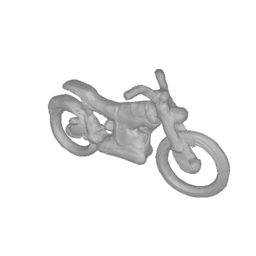}
    \end{subfigure}%
    \hfill%
    \begin{subfigure}[b]{0.15\linewidth}
		\centering
		\includegraphics[width=\linewidth]{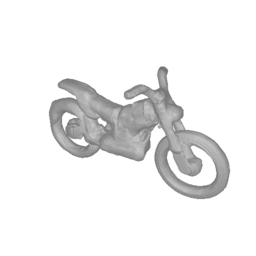}
    \end{subfigure}%
    \hfill%
    \begin{subfigure}[b]{0.15\linewidth}
		\centering
		\includegraphics[width=\linewidth]{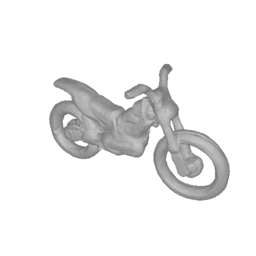}
    \end{subfigure}%
    \hfill%
    \begin{subfigure}[b]{0.15\linewidth}
		\centering
		\includegraphics[width=\linewidth]{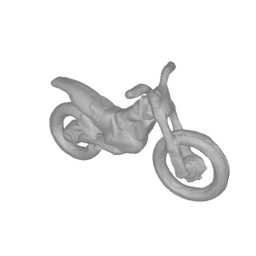}
    \end{subfigure}%
    \vskip\baselineskip%
    \vspace{-2.2em}
    \begin{subfigure}[b]{0.15\linewidth}
		\centering
		\includegraphics[width=\linewidth]{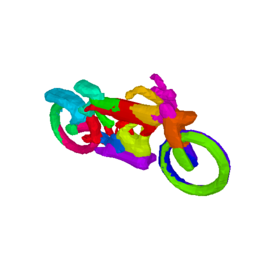}
    \end{subfigure}%
    \hfill%
    \begin{subfigure}[b]{0.15\linewidth}
		\centering
		\includegraphics[width=\linewidth]{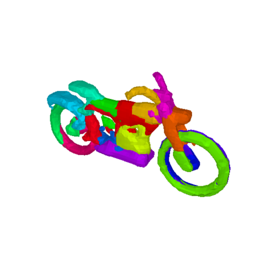}
    \end{subfigure}%
    \hfill%
    \begin{subfigure}[b]{0.15\linewidth}
		\centering
		\includegraphics[width=\linewidth]{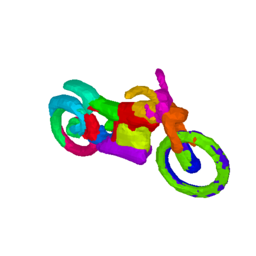}
    \end{subfigure}%
    \hfill%
    \begin{subfigure}[b]{0.15\linewidth}
		\centering
		\includegraphics[width=\linewidth]{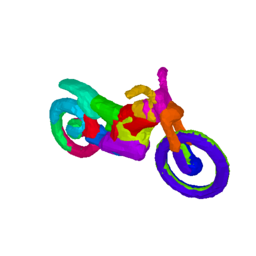}
    \end{subfigure}%
    \hfill%
    \begin{subfigure}[b]{0.15\linewidth}
		\centering
		\includegraphics[width=\linewidth]{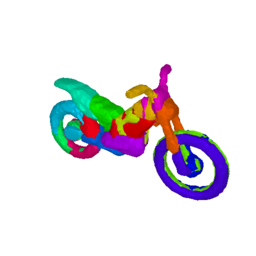}
    \end{subfigure}%
    \hfill%
    \begin{subfigure}[b]{0.15\linewidth}
		\centering
		\includegraphics[width=\linewidth]{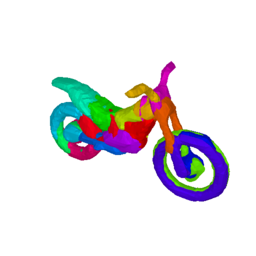}
    \end{subfigure}%
    \vskip\baselineskip%
    \vspace{-1.5em}
    \begin{subfigure}[b]{0.15\linewidth}
		\centering
		\includegraphics[width=\linewidth]{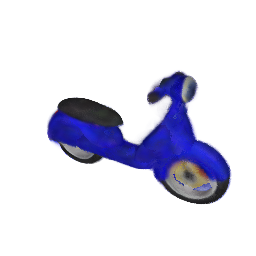}
    \end{subfigure}%
    \hfill%
    \begin{subfigure}[b]{0.15\linewidth}
		\centering
		\includegraphics[width=\linewidth]{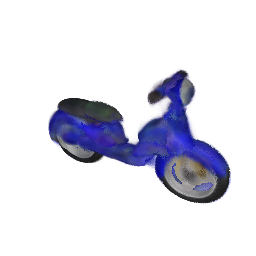}
    \end{subfigure}%
    \hfill%
    \begin{subfigure}[b]{0.15\linewidth}
		\centering
		\includegraphics[width=\linewidth]{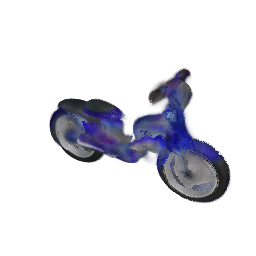}
    \end{subfigure}%
    \hfill%
    \begin{subfigure}[b]{0.15\linewidth}
		\centering
		\includegraphics[width=\linewidth]{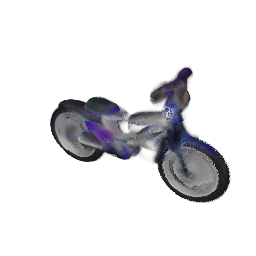}
    \end{subfigure}%
    \hfill%
    \begin{subfigure}[b]{0.15\linewidth}
		\centering
		\includegraphics[width=\linewidth]{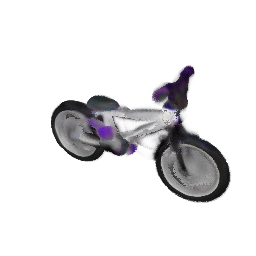}
    \end{subfigure}%
    \hfill%
    \begin{subfigure}[b]{0.15\linewidth}
		\centering
		\includegraphics[width=\linewidth]{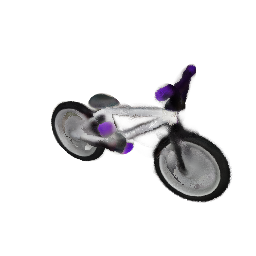}
    \end{subfigure}%
    \vskip\baselineskip%
    \vspace{-2.2em}
    \begin{subfigure}[b]{0.15\linewidth}
		\centering
		\includegraphics[width=\linewidth]{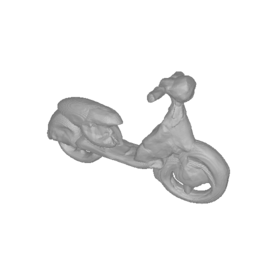}
    \end{subfigure}%
    \hfill%
    \begin{subfigure}[b]{0.15\linewidth}
		\centering
		\includegraphics[width=\linewidth]{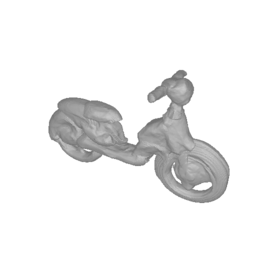}
    \end{subfigure}%
    \hfill%
    \begin{subfigure}[b]{0.15\linewidth}
		\centering
		\includegraphics[width=\linewidth]{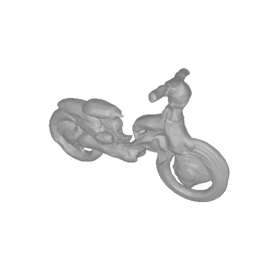}
    \end{subfigure}%
    \hfill%
    \begin{subfigure}[b]{0.15\linewidth}
		\centering
		\includegraphics[width=\linewidth]{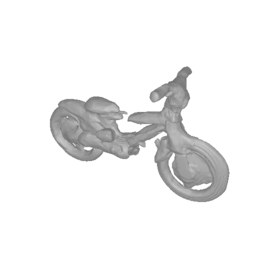}
    \end{subfigure}%
    \hfill%
    \begin{subfigure}[b]{0.15\linewidth}
		\centering
		\includegraphics[width=\linewidth]{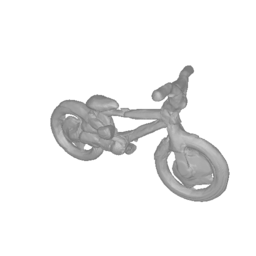}
    \end{subfigure}%
    \hfill%
    \begin{subfigure}[b]{0.15\linewidth}
		\centering
		\includegraphics[width=\linewidth]{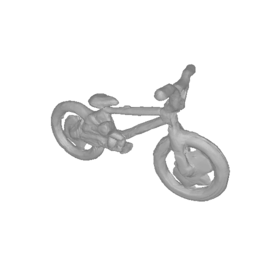}
    \end{subfigure}%
    \vskip\baselineskip%
    \vspace{-2.2em}
    \begin{subfigure}[b]{0.15\linewidth}
		\centering
		\includegraphics[width=\linewidth]{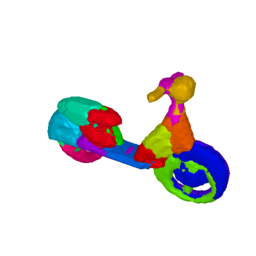}
    \end{subfigure}%
    \hfill%
    \begin{subfigure}[b]{0.15\linewidth}
		\centering
		\includegraphics[width=\linewidth]{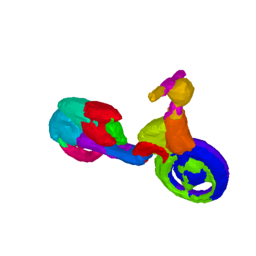}
    \end{subfigure}%
    \hfill%
    \begin{subfigure}[b]{0.15\linewidth}
		\centering
		\includegraphics[width=\linewidth]{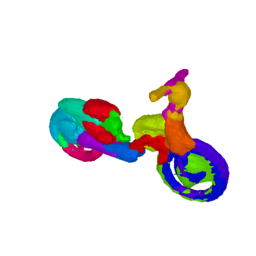}
    \end{subfigure}%
    \hfill%
    \begin{subfigure}[b]{0.15\linewidth}
		\centering
		\includegraphics[width=\linewidth]{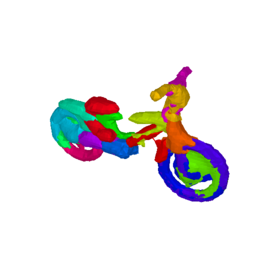}
    \end{subfigure}%
    \hfill%
    \begin{subfigure}[b]{0.15\linewidth}
		\centering
		\includegraphics[width=\linewidth]{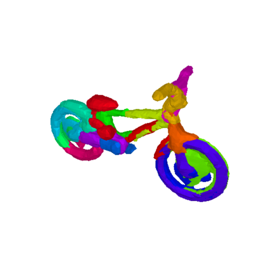}
    \end{subfigure}%
    \hfill%
    \begin{subfigure}[b]{0.15\linewidth}
		\centering
		\includegraphics[width=\linewidth]{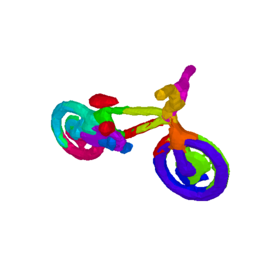}
    \end{subfigure}%
    \caption{{\bf Motorbikes Shape Interpolation}.
    From left to right, we interpolate
    between the geometry and texture latent codes of the two motorbikes.}
    \label{fig:interpolated_motorbikes}
    \vspace{-1.2em}
\end{figure}

\begin{figure}[!h]
    \centering
    \begin{subfigure}[b]{0.15\linewidth}
		\centering
		\includegraphics[width=\linewidth]{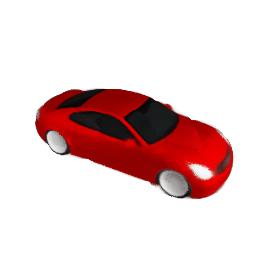}
    \end{subfigure}%
    \hfill%
    \begin{subfigure}[b]{0.15\linewidth}
		\centering
		\includegraphics[width=\linewidth]{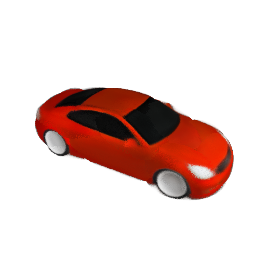}
    \end{subfigure}%
    \hfill%
    \begin{subfigure}[b]{0.15\linewidth}
		\centering
		\includegraphics[width=\linewidth]{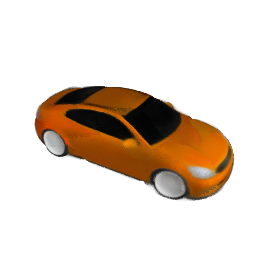}
    \end{subfigure}%
    \hfill%
    \begin{subfigure}[b]{0.15\linewidth}
		\centering
		\includegraphics[width=\linewidth]{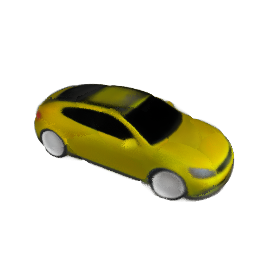}
    \end{subfigure}%
    \hfill%
    \begin{subfigure}[b]{0.15\linewidth}
		\centering
		\includegraphics[width=\linewidth]{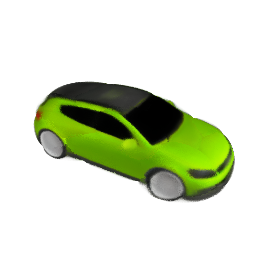}
    \end{subfigure}%
    \hfill%
    \begin{subfigure}[b]{0.15\linewidth}
		\centering
		\includegraphics[width=\linewidth]{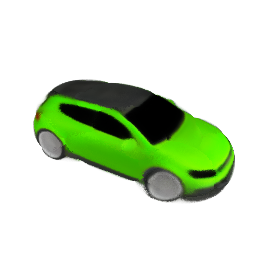}
    \end{subfigure}%
    \vskip\baselineskip%
    \vspace{-2.2em}
    \begin{subfigure}[b]{0.15\linewidth}
		\centering
		\includegraphics[width=\linewidth]{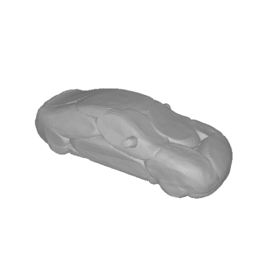}
    \end{subfigure}%
    \hfill%
    \begin{subfigure}[b]{0.15\linewidth}
		\centering
		\includegraphics[width=\linewidth]{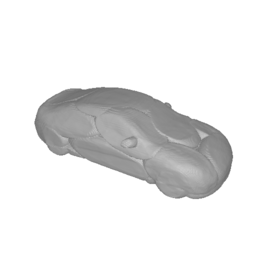}
    \end{subfigure}%
    \hfill%
    \begin{subfigure}[b]{0.15\linewidth}
		\centering
		\includegraphics[width=\linewidth]{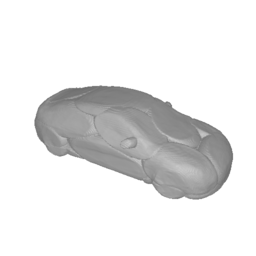}
    \end{subfigure}%
    \hfill%
    \begin{subfigure}[b]{0.15\linewidth}
		\centering
		\includegraphics[width=\linewidth]{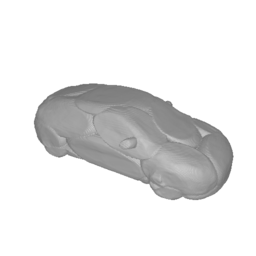}
    \end{subfigure}%
    \hfill%
    \begin{subfigure}[b]{0.15\linewidth}
		\centering
		\includegraphics[width=\linewidth]{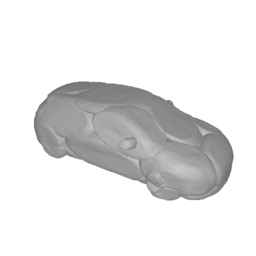}
    \end{subfigure}%
    \hfill%
    \begin{subfigure}[b]{0.15\linewidth}
		\centering
		\includegraphics[width=\linewidth]{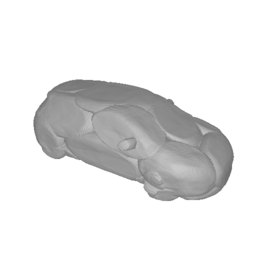}
    \end{subfigure}%
    \vskip\baselineskip%
    \vspace{-2.2em}
    \begin{subfigure}[b]{0.15\linewidth}
		\centering
		\includegraphics[width=\linewidth]{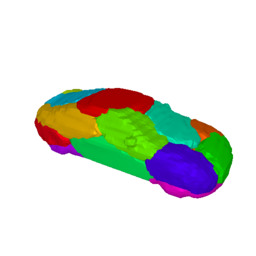}
    \end{subfigure}%
    \hfill%
    \begin{subfigure}[b]{0.15\linewidth}
		\centering
		\includegraphics[width=\linewidth]{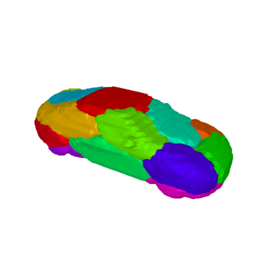}
    \end{subfigure}%
    \hfill%
    \begin{subfigure}[b]{0.15\linewidth}
		\centering
		\includegraphics[width=\linewidth]{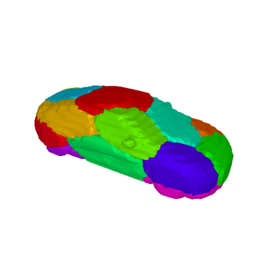}
    \end{subfigure}%
    \hfill%
    \begin{subfigure}[b]{0.15\linewidth}
		\centering
		\includegraphics[width=\linewidth]{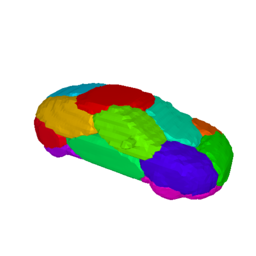}
    \end{subfigure}%
    \hfill%
    \begin{subfigure}[b]{0.15\linewidth}
		\centering
		\includegraphics[width=\linewidth]{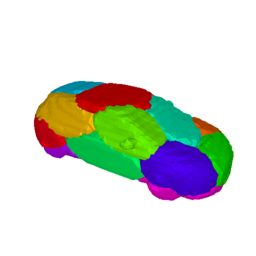}
    \end{subfigure}%
    \hfill%
    \begin{subfigure}[b]{0.15\linewidth}
		\centering
		\includegraphics[width=\linewidth]{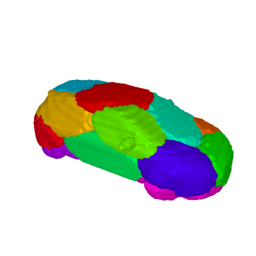}
    \end{subfigure}%
    \vskip\baselineskip%
    \vspace{-1.5em}
    \begin{subfigure}[b]{0.15\linewidth}
		\centering
		\includegraphics[width=\linewidth]{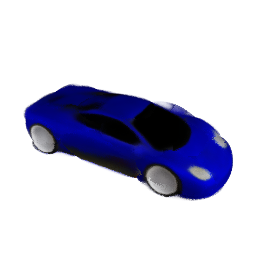}
    \end{subfigure}%
    \hfill%
    \begin{subfigure}[b]{0.15\linewidth}
		\centering
		\includegraphics[width=\linewidth]{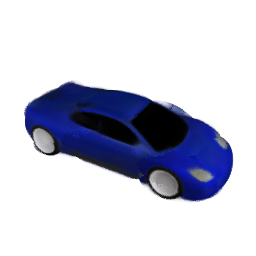}
    \end{subfigure}%
    \hfill%
    \begin{subfigure}[b]{0.15\linewidth}
		\centering
		\includegraphics[width=\linewidth]{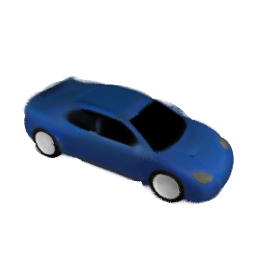}
    \end{subfigure}%
    \hfill%
    \begin{subfigure}[b]{0.15\linewidth}
		\centering
		\includegraphics[width=\linewidth]{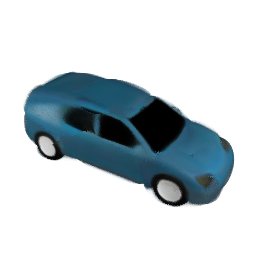}
    \end{subfigure}%
    \hfill%
    \begin{subfigure}[b]{0.15\linewidth}
		\centering
		\includegraphics[width=\linewidth]{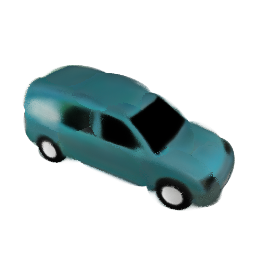}
    \end{subfigure}%
    \hfill%
    \begin{subfigure}[b]{0.15\linewidth}
		\centering
		\includegraphics[width=\linewidth]{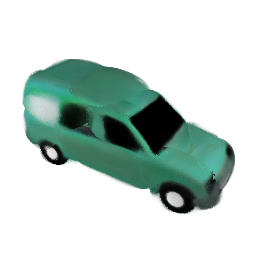}
    \end{subfigure}%
    \vskip\baselineskip%
    \vspace{-2.2em}
    \begin{subfigure}[b]{0.15\linewidth}
		\centering
		\includegraphics[width=\linewidth]{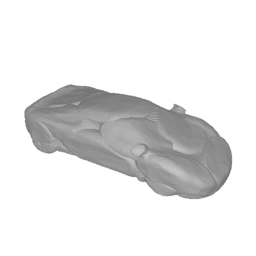}
    \end{subfigure}%
    \hfill%
    \begin{subfigure}[b]{0.15\linewidth}
		\centering
		\includegraphics[width=\linewidth]{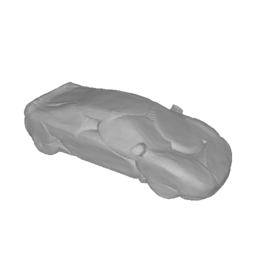}
    \end{subfigure}%
    \hfill%
    \begin{subfigure}[b]{0.15\linewidth}
		\centering
		\includegraphics[width=\linewidth]{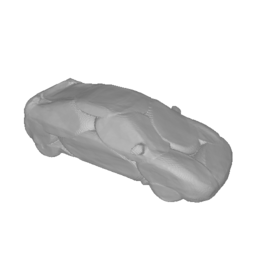}
    \end{subfigure}%
    \hfill%
    \begin{subfigure}[b]{0.15\linewidth}
		\centering
		\includegraphics[width=\linewidth]{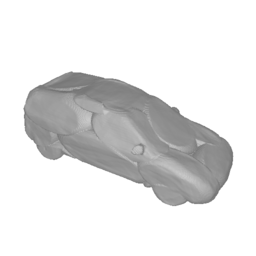}
    \end{subfigure}%
    \hfill%
    \begin{subfigure}[b]{0.15\linewidth}
		\centering
		\includegraphics[width=\linewidth]{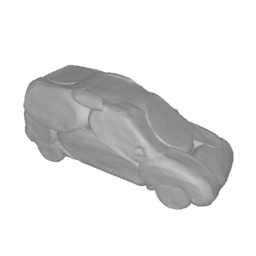}
    \end{subfigure}%
    \hfill%
    \begin{subfigure}[b]{0.15\linewidth}
		\centering
		\includegraphics[width=\linewidth]{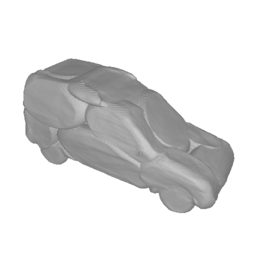}
    \end{subfigure}%
    \vskip\baselineskip%
    \vspace{-2.2em}
    \begin{subfigure}[b]{0.15\linewidth}
		\centering
		\includegraphics[width=\linewidth]{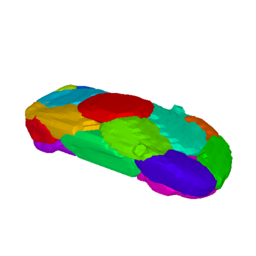}
    \end{subfigure}%
    \hfill%
    \begin{subfigure}[b]{0.15\linewidth}
		\centering
		\includegraphics[width=\linewidth]{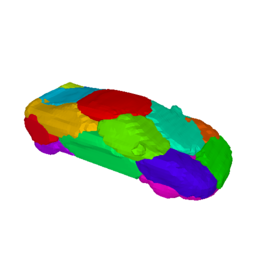}
    \end{subfigure}%
    \hfill%
    \begin{subfigure}[b]{0.15\linewidth}
		\centering
		\includegraphics[width=\linewidth]{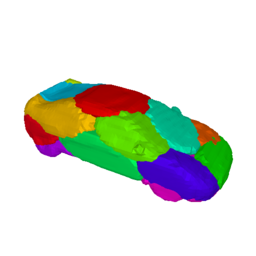}
    \end{subfigure}%
    \hfill%
    \begin{subfigure}[b]{0.15\linewidth}
		\centering
		\includegraphics[width=\linewidth]{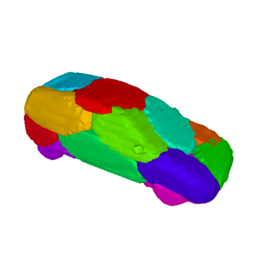}
    \end{subfigure}%
    \hfill%
    \begin{subfigure}[b]{0.15\linewidth}
		\centering
		\includegraphics[width=\linewidth]{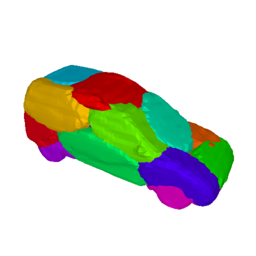}
    \end{subfigure}%
    \hfill%
    \begin{subfigure}[b]{0.15\linewidth}
		\centering
		\includegraphics[width=\linewidth]{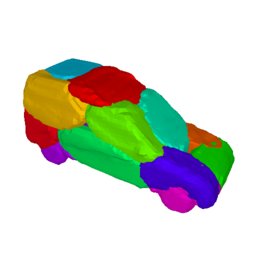}
    \end{subfigure}%
    \vskip\baselineskip%
    \vspace{-1.5em}
    \begin{subfigure}[b]{0.15\linewidth}
		\centering
		\includegraphics[width=\linewidth]{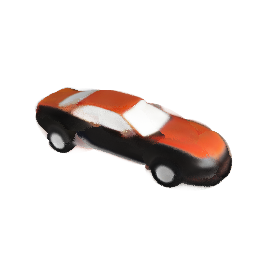}
    \end{subfigure}%
    \hfill%
    \begin{subfigure}[b]{0.15\linewidth}
		\centering
		\includegraphics[width=\linewidth]{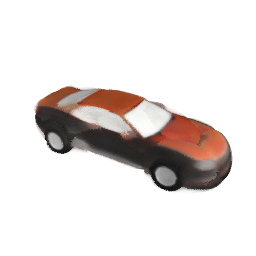}
    \end{subfigure}%
    \hfill%
    \begin{subfigure}[b]{0.15\linewidth}
		\centering
		\includegraphics[width=\linewidth]{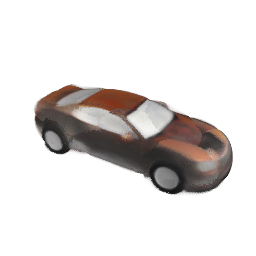}
    \end{subfigure}%
    \hfill%
    \begin{subfigure}[b]{0.15\linewidth}
		\centering
		\includegraphics[width=\linewidth]{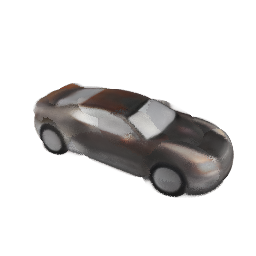}
    \end{subfigure}%
    \hfill%
    \begin{subfigure}[b]{0.15\linewidth}
		\centering
		\includegraphics[width=\linewidth]{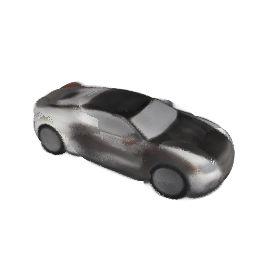}
    \end{subfigure}%
    \hfill%
    \begin{subfigure}[b]{0.15\linewidth}
		\centering
		\includegraphics[width=\linewidth]{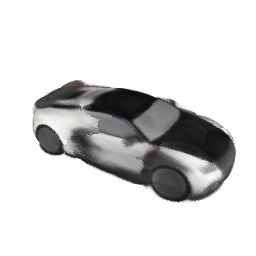}
    \end{subfigure}%
    \vskip\baselineskip%
    \vspace{-2.2em}
    \begin{subfigure}[b]{0.15\linewidth}
		\centering
		\includegraphics[width=\linewidth]{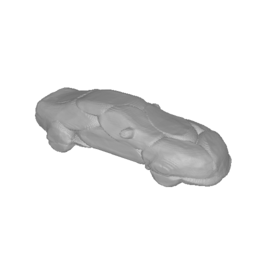}
    \end{subfigure}%
    \hfill%
    \begin{subfigure}[b]{0.15\linewidth}
		\centering
		\includegraphics[width=\linewidth]{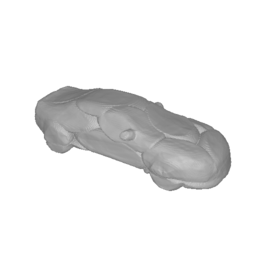}
    \end{subfigure}%
    \hfill%
    \begin{subfigure}[b]{0.15\linewidth}
		\centering
		\includegraphics[width=\linewidth]{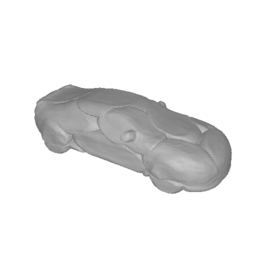}
    \end{subfigure}%
    \hfill%
    \begin{subfigure}[b]{0.15\linewidth}
		\centering
		\includegraphics[width=\linewidth]{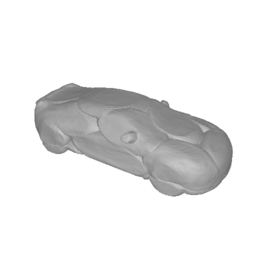}
    \end{subfigure}%
    \hfill%
    \begin{subfigure}[b]{0.15\linewidth}
		\centering
		\includegraphics[width=\linewidth]{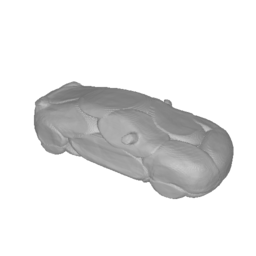}
    \end{subfigure}%
    \hfill%
    \begin{subfigure}[b]{0.15\linewidth}
		\centering
		\includegraphics[width=\linewidth]{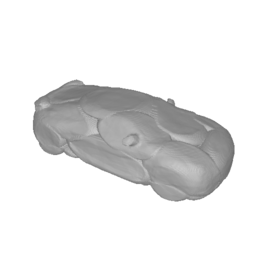}
    \end{subfigure}%
    \vskip\baselineskip%
    \vspace{-2.2em}
    \begin{subfigure}[b]{0.15\linewidth}
		\centering
		\includegraphics[width=\linewidth]{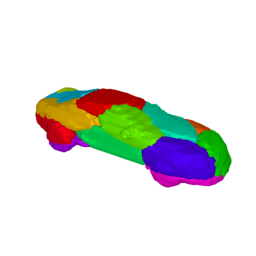}
    \end{subfigure}%
    \hfill%
    \begin{subfigure}[b]{0.15\linewidth}
		\centering
		\includegraphics[width=\linewidth]{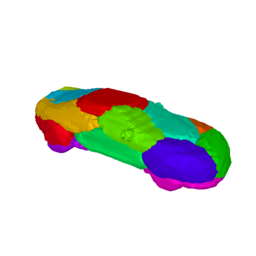}
    \end{subfigure}%
    \hfill%
    \begin{subfigure}[b]{0.15\linewidth}
		\centering
		\includegraphics[width=\linewidth]{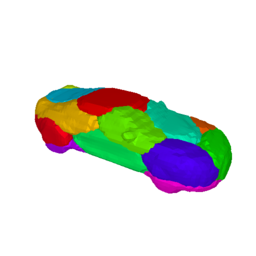}
    \end{subfigure}%
    \hfill%
    \begin{subfigure}[b]{0.15\linewidth}
		\centering
		\includegraphics[width=\linewidth]{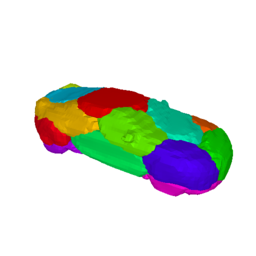}
    \end{subfigure}%
    \hfill%
    \begin{subfigure}[b]{0.15\linewidth}
		\centering
		\includegraphics[width=\linewidth]{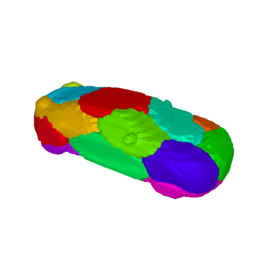}
    \end{subfigure}%
    \hfill%
    \begin{subfigure}[b]{0.15\linewidth}
		\centering
		\includegraphics[width=\linewidth]{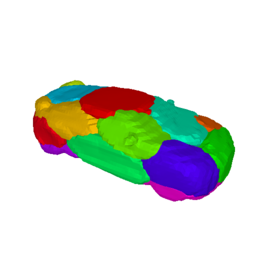}
    \end{subfigure}%
    \caption{{\bf Cars Shape Interpolation}.
    From left to right, we interpolate
    between the geometry and texture latent codes of the two cars.}
    \label{fig:interpolated_cars}
    \vspace{-1.2em}
\end{figure}

In this section, we show interpolations between the shape and texture codes of
two various objects for all object categories. In particular, we interpolate
between the latent codes of two shapes and demonstrate that our model is able to
smoothly interpolate between two shapes, while preserving the shape structure
and the part-based structure for all object categories.
\figref{fig:interpolated_airplanes} shows three interpolations from left to
right between two airplanes. For the case of cars and motorbikes, interpolation
results are summarized in \figref{fig:interpolated_cars} and
\figref{fig:interpolated_motorbikes}. Note that for all interpolations the
transitions from one shape to the other are consistently meaningful.

\clearpage
\section{Shape Inversion}
\label{sec:shape_inversion}

Our generative model is realized by an auto-decoder architecture. In order to be
able to edit a shape that does not belong to the training set, we seek to match
the given shape to a shape embedding $\bz^s$ that can closely reconstruct its
geometry. Following \cite{Hertz2022SIGGRAPH}, we randomly initialize the shape
embedding $\bz^s$, sampling from a multivariate normal distribution using the
mean $\bmu_{train}^s$ and covariance $\bSigma_{train}^s$ of shape embeddings
seen during training:
\begin{equation}
	\bz^s \sim \cN(\bmu_{train}^s,\bSigma_{train}^s).
\end{equation}
Then, we sample points along rays $\cR$ from $N = 5$ different views of the
object, and freeze our entire network, optimizing the latent code $\bz^s$ using
the following loss function:
\begin{equation}
    \cL_{inversion} = \cL_{mask}(\cR) + \cL_{occ}(\cR) + \norm{\bz^s}_2,
\end{equation}
using weights 1.0, 1.0 and 0.0001. We optimize the latent code $\bz^s$
for a fixed number of 700 gradient update steps. Note that this experiment does
not require any color information, as we are solely interested in extracting an
accurate representation of the geometry of the target shape. We therefore
utilize only object masks during optimization. We compare our method with
DualSDF~\cite{Hao2020CVPR} and SPAGHETTI~\cite{Hertz2022SIGGRAPH} on the test
set of the ShapeNet Airplanes category. Our results are summarized in
\tabref{tab:shapenet_shape_inversion}.  To measure the Chamfer-$L_1$ distance,
we follow \cite{Hertz2022SIGGRAPH} and sample $30,000$ points on the surface of the
generated and the target mesh.
\begin{table}[!h]
    \centering
    \begin{tabular}{lccc}
        \toprule
        & DualSDF & SPAGHETTI & Ours \\
        \midrule
        Chamfer-$L_1$ & 0.806 & \bf{0.050} & 0.4536 \\
        \bottomrule
    \end{tabular}
    \caption{{\bf Shape Inversion.} Quantitative evaluation of our method
    against DualSDF~\cite{Hao2020CVPR} and SPAGHETTI~\cite{Hertz2022SIGGRAPH}
    \wrt to Chamfer distance ($\downarrow$) on the test set of the
    \emph{Airplanes} category.}
    \label{tab:shapenet_shape_inversion}
    \vspace{-0.4em}
\end{table}
\begin{figure}[!h]
    \centering
    \scriptsize
    \begin{minipage}[b]{0.14\linewidth}
		\centering Ground Truth
    \end{minipage}
    \begin{subfigure}[b]{0.71\linewidth}
		\centering Input Object Masks
    \end{subfigure}%
    \hfill%
    \begin{subfigure}[b]{0.14\linewidth}
		\centering Ours
    \end{subfigure}
    \vskip\baselineskip%
    \vspace{-1.0em}
    \begin{subfigure}[b]{0.14\linewidth}
		\centering
		\includegraphics[width=\linewidth]{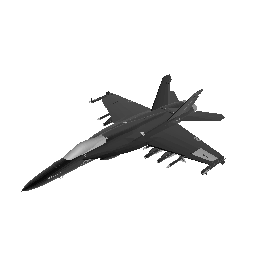}
    \end{subfigure}%
    \hfill%
    \begin{subfigure}[b]{0.14\linewidth}
		\centering
		\includegraphics[width=\linewidth]{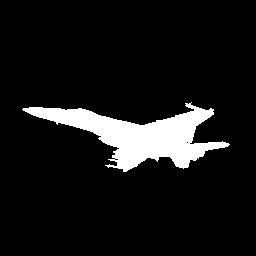}
    \end{subfigure}%
    \hfill%
    \begin{subfigure}[b]{0.14\linewidth}
		\centering
		\includegraphics[width=\linewidth]{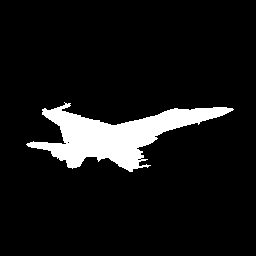}
    \end{subfigure}%
    \hfill%
    \begin{subfigure}[b]{0.14\linewidth}
		\centering
		\includegraphics[width=\linewidth]{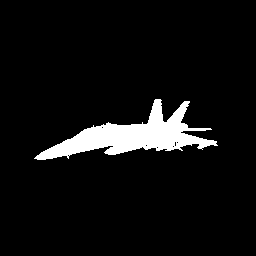}
    \end{subfigure}%
    \hfill%
    \begin{subfigure}[b]{0.14\linewidth}
		\centering
		\includegraphics[width=\linewidth]{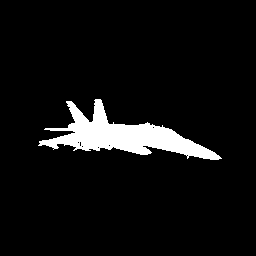}
    \end{subfigure}%
    \hfill%
    \begin{subfigure}[b]{0.14\linewidth}
		\centering
		\includegraphics[width=\linewidth]{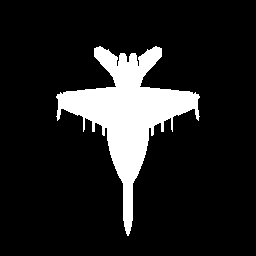}
    \end{subfigure}%
    \hfill%
    \begin{subfigure}[b]{0.14\linewidth}
		\centering
		\includegraphics[width=\linewidth]{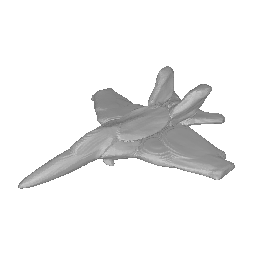}
    \end{subfigure}%
    \vskip\baselineskip%
    \vspace{-1.2em}
    \vskip\baselineskip%
    \vspace{-1.0em}
    \begin{subfigure}[b]{0.14\linewidth}
		\centering
		\includegraphics[width=\linewidth]{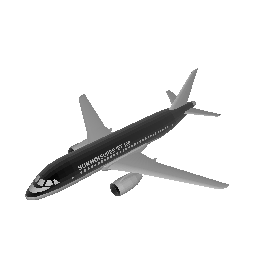}
    \end{subfigure}%
    \hfill%
    \begin{subfigure}[b]{0.14\linewidth}
		\centering
		\includegraphics[width=\linewidth]{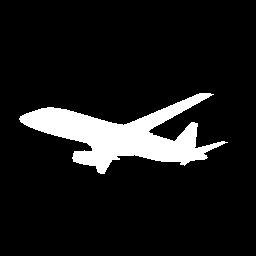}
    \end{subfigure}%
    \hfill%
    \begin{subfigure}[b]{0.14\linewidth}
		\centering
		\includegraphics[width=\linewidth]{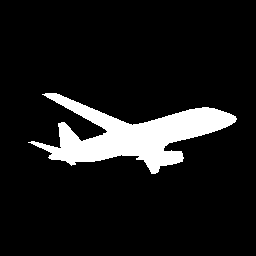}
    \end{subfigure}%
    \hfill%
    \begin{subfigure}[b]{0.14\linewidth}
		\centering
		\includegraphics[width=\linewidth]{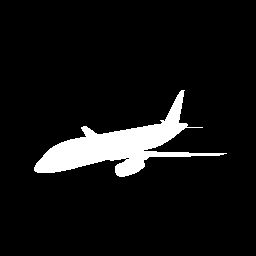}
    \end{subfigure}%
    \hfill%
    \begin{subfigure}[b]{0.14\linewidth}
		\centering
		\includegraphics[width=\linewidth]{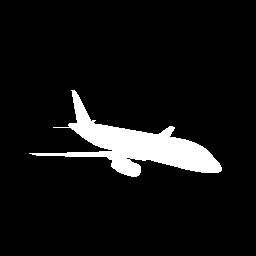}
    \end{subfigure}%
    \hfill%
    \begin{subfigure}[b]{0.14\linewidth}
		\centering
		\includegraphics[width=\linewidth]{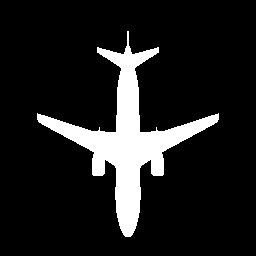}
    \end{subfigure}%
    \hfill%
    \begin{subfigure}[b]{0.14\linewidth}
		\centering
		\includegraphics[width=\linewidth]{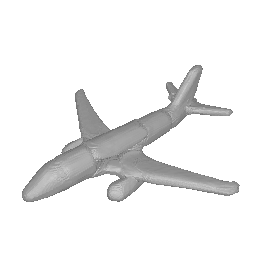}
    \end{subfigure}%
    \vskip\baselineskip%
    \vspace{-1.2em}
    \begin{subfigure}[b]{0.14\linewidth}
		\centering
		\includegraphics[width=\linewidth]{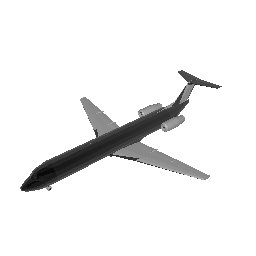}
    \end{subfigure}%
    \hfill%
    \begin{subfigure}[b]{0.14\linewidth}
		\centering
		\includegraphics[width=\linewidth]{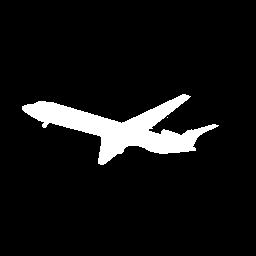}
    \end{subfigure}%
    \hfill%
    \begin{subfigure}[b]{0.14\linewidth}
		\centering
		\includegraphics[width=\linewidth]{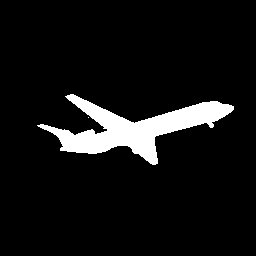}
    \end{subfigure}%
    \hfill%
    \begin{subfigure}[b]{0.14\linewidth}
		\centering
		\includegraphics[width=\linewidth]{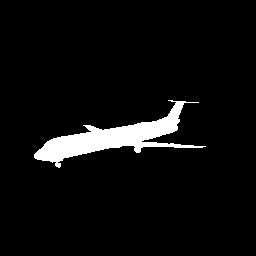}
    \end{subfigure}%
    \hfill%
    \begin{subfigure}[b]{0.14\linewidth}
		\centering
		\includegraphics[width=\linewidth]{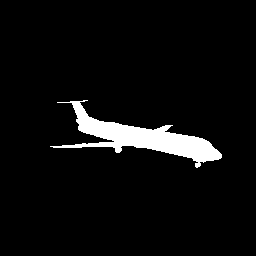}
    \end{subfigure}%
    \hfill%
    \begin{subfigure}[b]{0.14\linewidth}
		\centering
		\includegraphics[width=\linewidth]{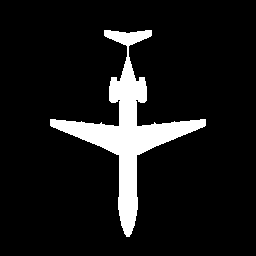}
    \end{subfigure}%
    \hfill%
    \begin{subfigure}[b]{0.14\linewidth}
		\centering
		\includegraphics[width=\linewidth]{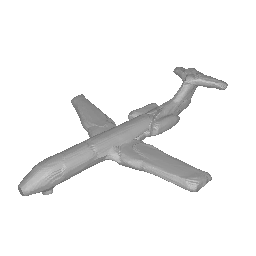}
    \end{subfigure}%
    \vskip\baselineskip%
    \vspace{-1.2em}
    \begin{subfigure}[b]{0.14\linewidth}
		\centering
		\includegraphics[width=\linewidth]{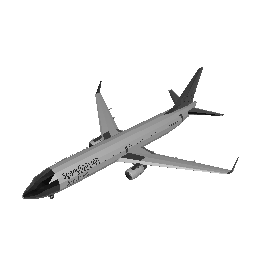}
    \end{subfigure}%
    \hfill%
    \begin{subfigure}[b]{0.14\linewidth}
		\centering
		\includegraphics[width=\linewidth]{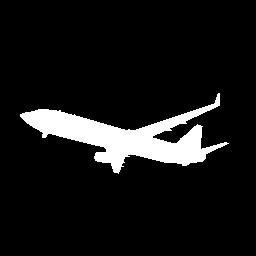}
    \end{subfigure}%
    \hfill%
    \begin{subfigure}[b]{0.14\linewidth}
		\centering
		\includegraphics[width=\linewidth]{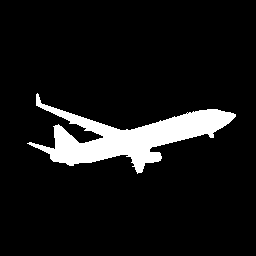}
    \end{subfigure}%
    \hfill%
    \begin{subfigure}[b]{0.14\linewidth}
		\centering
		\includegraphics[width=\linewidth]{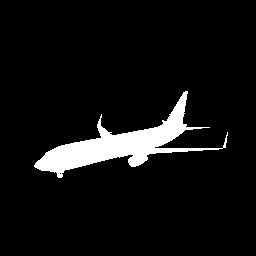}
    \end{subfigure}%
    \hfill%
    \begin{subfigure}[b]{0.14\linewidth}
		\centering
		\includegraphics[width=\linewidth]{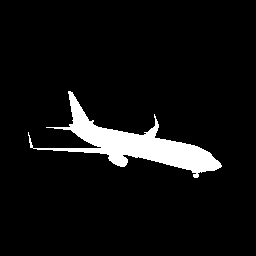}
    \end{subfigure}%
    \hfill%
    \begin{subfigure}[b]{0.14\linewidth}
		\centering
		\includegraphics[width=\linewidth]{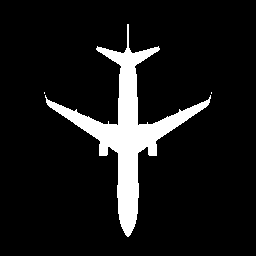}
    \end{subfigure}%
    \hfill%
    \begin{subfigure}[b]{0.14\linewidth}
		\centering
		\includegraphics[width=\linewidth]{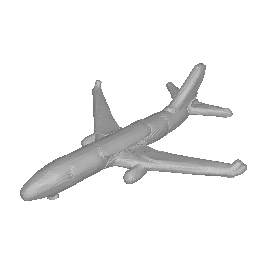}
    \end{subfigure}%
    \vskip\baselineskip%
    \vspace{-1.2em}
    \caption{{\bf Shape Inversion}. We show examples of our results when
    performing shape inversion on the ShapeNet airplanes.}
    \label{fig:shapenet_shape_inversion}
    \vspace{-1.2em}
\end{figure}

Our model yields better reconstructions than DualSDF~\cite{Hao2020CVPR} in terms of Chamfer distance,
even though \cite{Hao2020CVPR} was trained with explicit 3D supervision.
Compared to SPAGHETTI~\cite{Hertz2022SIGGRAPH}, our model yields worse
reconstructions. We hypothesize that conditioning our generations to more than
$5$ views would further improve the performance of our model. In
\figref{fig:shapenet_shape_inversion}, we show examples of our shape inversion experiment.

\section{Image Inversion}
\label{sec:image_inversion}

\begin{figure}[!h]
    \centering
    \scriptsize
    \begin{subfigure}[b]{0.71\linewidth}
		\centering Input Images
    \end{subfigure}%
    \hfill%
    \begin{subfigure}[b]{0.14\linewidth}
		\centering Ours
    \end{subfigure}
    \begin{minipage}[b]{0.14\linewidth}
		\centering Ours - Geometry
    \end{minipage}
    \vskip\baselineskip%
    \vspace{-1.0em}
    \begin{subfigure}[b]{0.14\linewidth}
		\centering
		\includegraphics[width=\linewidth]{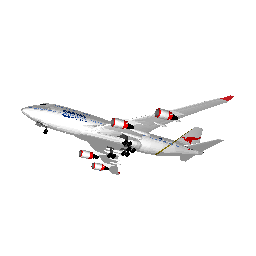}
    \end{subfigure}%
    \hfill%
    \begin{subfigure}[b]{0.14\linewidth}
		\centering
		\includegraphics[width=\linewidth]{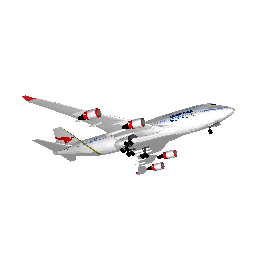}
    \end{subfigure}%
    \hfill%
    \begin{subfigure}[b]{0.14\linewidth}
		\centering
		\includegraphics[width=\linewidth]{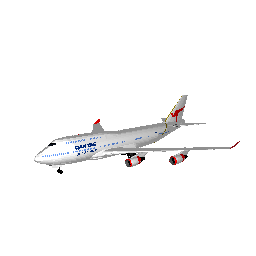}
    \end{subfigure}%
    \hfill%
    \begin{subfigure}[b]{0.14\linewidth}
		\centering
		\includegraphics[width=\linewidth]{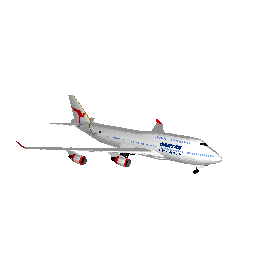}
    \end{subfigure}%
    \hfill%
    \begin{subfigure}[b]{0.14\linewidth}
		\centering
		\includegraphics[width=\linewidth]{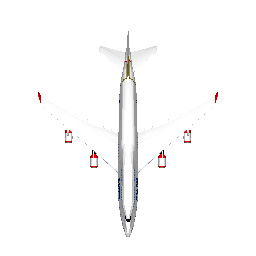}
    \end{subfigure}%
    \hfill%
    \begin{subfigure}[b]{0.14\linewidth}
		\centering
		\includegraphics[width=\linewidth]{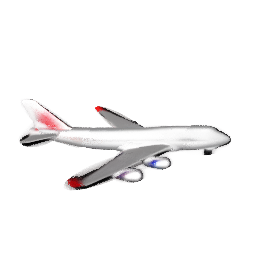}
    \end{subfigure}%
    \hfill%
    \begin{subfigure}[b]{0.14\linewidth}
		\centering
		\includegraphics[width=\linewidth]{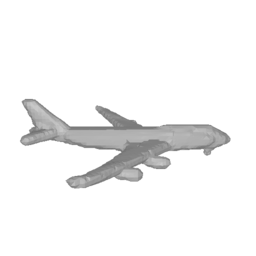}
    \end{subfigure}%
    \vskip\baselineskip%
    \vspace{-1.2em}
    \begin{subfigure}[b]{0.14\linewidth}
		\centering
		\includegraphics[width=\linewidth]{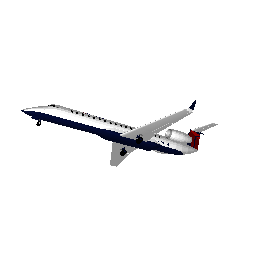}
    \end{subfigure}%
    \hfill%
    \begin{subfigure}[b]{0.14\linewidth}
		\centering
		\includegraphics[width=\linewidth]{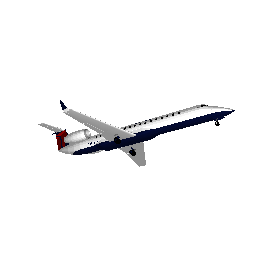}
    \end{subfigure}%
    \hfill%
    \begin{subfigure}[b]{0.14\linewidth}
		\centering
		\includegraphics[width=\linewidth]{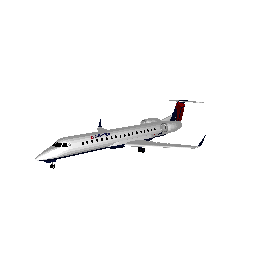}
    \end{subfigure}%
    \hfill%
    \begin{subfigure}[b]{0.14\linewidth}
		\centering
		\includegraphics[width=\linewidth]{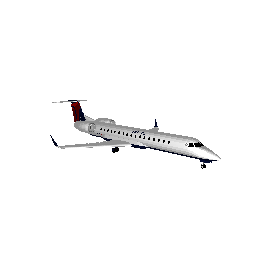}
    \end{subfigure}%
    \hfill%
    \begin{subfigure}[b]{0.14\linewidth}
		\centering
		\includegraphics[width=\linewidth]{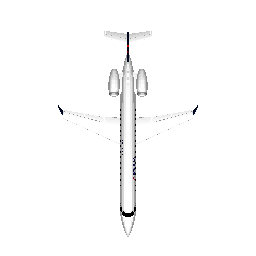}
    \end{subfigure}%
    \hfill%
    \begin{subfigure}[b]{0.14\linewidth}
		\centering
		\includegraphics[width=\linewidth]{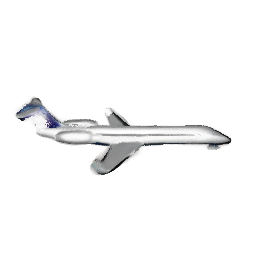}
    \end{subfigure}%
    \hfill%
    \begin{subfigure}[b]{0.14\linewidth}
		\centering
		\includegraphics[width=\linewidth]{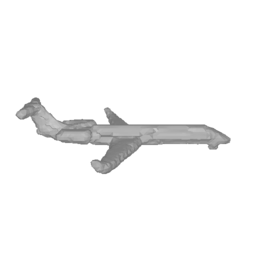}
    \end{subfigure}%
    \vskip\baselineskip%
    \vspace{-1.2em}
    \begin{subfigure}[b]{0.14\linewidth}
		\centering
		\includegraphics[width=\linewidth]{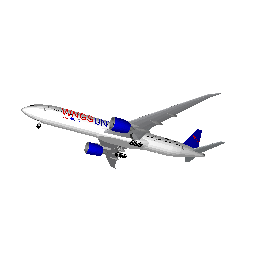}
    \end{subfigure}%
    \hfill%
    \begin{subfigure}[b]{0.14\linewidth}
		\centering
		\includegraphics[width=\linewidth]{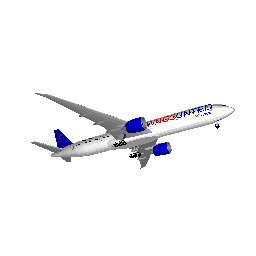}
    \end{subfigure}%
    \hfill%
    \begin{subfigure}[b]{0.14\linewidth}
		\centering
		\includegraphics[width=\linewidth]{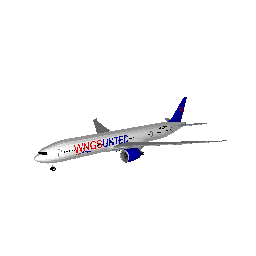}
    \end{subfigure}%
    \hfill%
    \begin{subfigure}[b]{0.14\linewidth}
		\centering
		\includegraphics[width=\linewidth]{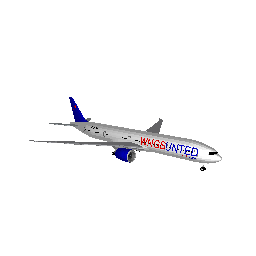}
    \end{subfigure}%
    \hfill%
    \begin{subfigure}[b]{0.14\linewidth}
		\centering
		\includegraphics[width=\linewidth]{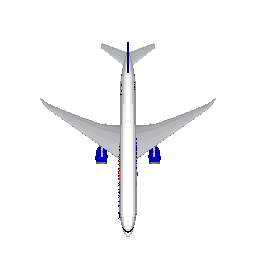}
    \end{subfigure}%
    \hfill%
    \begin{subfigure}[b]{0.14\linewidth}
		\centering
		\includegraphics[width=\linewidth]{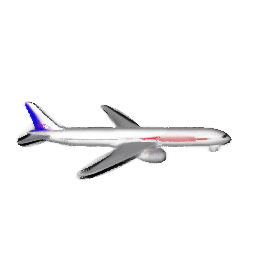}
    \end{subfigure}%
    \hfill%
    \begin{subfigure}[b]{0.14\linewidth}
		\centering
		\includegraphics[width=\linewidth]{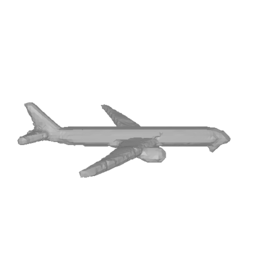}
    \end{subfigure}%
    \vskip\baselineskip%
    \vspace{-1.2em}
    \caption{{\bf Image Inversion}. We show examples of our results when
    performing image inversion on the ShapeNet airplanes.}
    \label{fig:shapenet_image_inversion}
    \vspace{-1.2em}
\end{figure}

The goal of this experiment is to recover both the appearance and geometry of a
3D shape, as opposed to only the shape as for the case of shape inversion
(\secref{sec:shape_inversion}). To this end, we need to find an appropriate match for 
both the shape and texture embeddings $\{\bz^s, \bz^t\}$, given a set of posed
images and masks as targets. In a similar manner to the sampling process of
\secref{subsec:generation}, we randomly initialize the shape and texture
embeddings, as discussed in \secref{subsec:generation}. Then, we freeze the network parameters
and optimize the embeddings using the loss function:
\begin{equation}
    \cL_{image\_inversion} = \cL_{rgb}(\cR) + \cL_{mask}(\cR) + \cL_{occ}(\cR) + \norm{\bz^s}_2 + \norm{\bz^t}_2,
\end{equation}
with weights of 1.0, 1.0, 1.0, 0.0001 and 0.0001. We use $N=5$ posed
images and masks as targets, and optimize the latent codes for a fixed number
of 2000 gradient update steps. We showcase several inverted examples in
\figref{fig:shapenet_image_inversion}.

We observe that our model can faithfully recover the object geometry and
appearance for various objects. Note that Image Inversion enables users to
directly edit shapes, by just providing a few posed images and masks. In
\secref{sec:editing} we showcase several editing functionalities on the
inverted shapes.

\section{Shape Editing}
\label{sec:editing}

\begin{figure}[!h]
    \centering
    \scriptsize
    \begin{minipage}[b]{0.14\linewidth}
		\centering Shape 1
    \end{minipage}
    \begin{minipage}[b]{0.14\linewidth}
		\centering Shape 2
    \end{minipage}
    \begin{minipage}[b]{0.14\linewidth}
		\centering Shape 3
    \end{minipage}
    \begin{minipage}[b]{0.14\linewidth}
		\centering Mixing (Texture 1)
    \end{minipage}
    \begin{minipage}[b]{0.14\linewidth}
		\centering Mixing (Texture 2)
    \end{minipage}
    \begin{minipage}[b]{0.14\linewidth}
		\centering Mixing (Texture 3)
    \end{minipage}
    \begin{subfigure}[b]{0.14\linewidth}
		\centering Mixing (Combined)
    \end{subfigure}
    \vskip\baselineskip%
    \vspace{-1.2em}
    \begin{subfigure}[b]{0.14\linewidth}
		\centering
		\includegraphics[width=\linewidth]{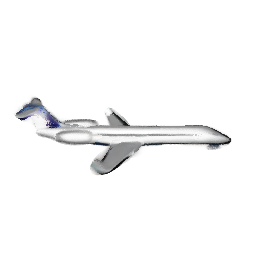}
    \end{subfigure}%
    \hfill%
    \begin{subfigure}[b]{0.14\linewidth}
		\centering
		\includegraphics[width=\linewidth]{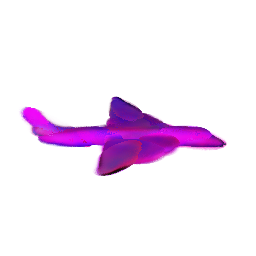}
    \end{subfigure}%
    \hfill%
    \begin{subfigure}[b]{0.14\linewidth}
		\centering
		\includegraphics[width=\linewidth]{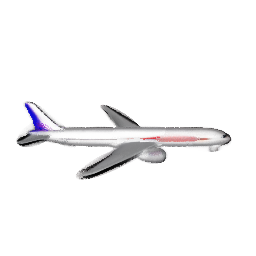}
    \end{subfigure}%
    \hfill%
    \begin{subfigure}[b]{0.14\linewidth}
		\centering
		\includegraphics[width=\linewidth]{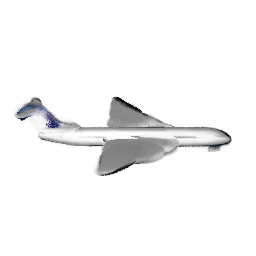}
    \end{subfigure}%
    \hfill%
    \begin{subfigure}[b]{0.14\linewidth}
		\centering
		\includegraphics[width=\linewidth]{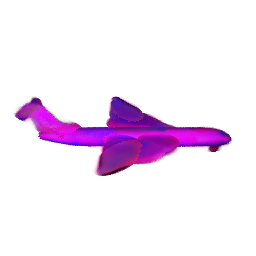}
    \end{subfigure}%
    \hfill%
    \begin{subfigure}[b]{0.14\linewidth}
		\centering
		\includegraphics[width=\linewidth]{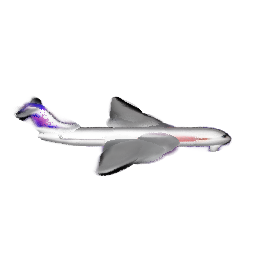}
    \end{subfigure}%
    \hfill%
    \begin{subfigure}[b]{0.14\linewidth}
		\centering
		\includegraphics[width=\linewidth]{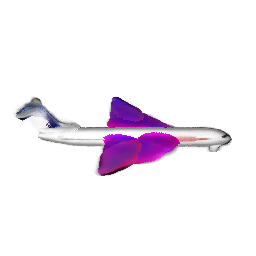}
    \end{subfigure}%
    \vskip\baselineskip%
    \vspace{-2.2em}
    \begin{subfigure}[b]{0.14\linewidth}
		\centering
		\includegraphics[width=\linewidth]{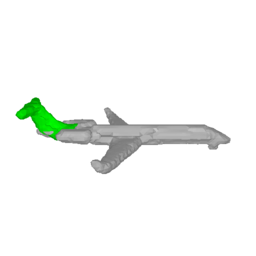}
    \end{subfigure}%
    \hfill%
    \begin{subfigure}[b]{0.14\linewidth}
		\centering
		\includegraphics[width=\linewidth]{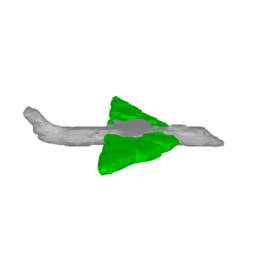}
    \end{subfigure}%
    \hfill%
    \begin{subfigure}[b]{0.14\linewidth}
		\centering
		\includegraphics[width=\linewidth]{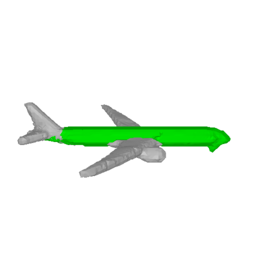}
    \end{subfigure}%
    \hfill%
    \begin{subfigure}[b]{0.14\linewidth}
		\centering
		\includegraphics[width=\linewidth]{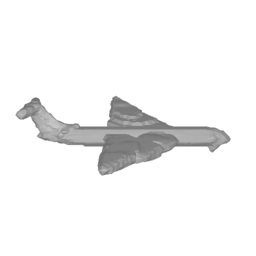}
    \end{subfigure}%
    \hfill%
    \begin{subfigure}[b]{0.14\linewidth}
		\centering
		\includegraphics[width=\linewidth]{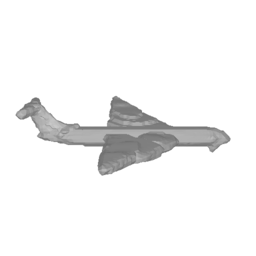}
    \end{subfigure}%
    \hfill%
    \begin{subfigure}[b]{0.14\linewidth}
		\centering
		\includegraphics[width=\linewidth]{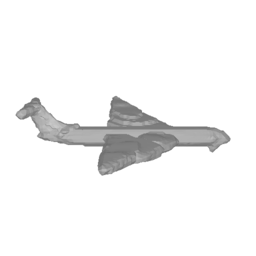}
    \end{subfigure}%
    \hfill%
    \begin{subfigure}[b]{0.14\linewidth}
		\centering
		\includegraphics[width=\linewidth]{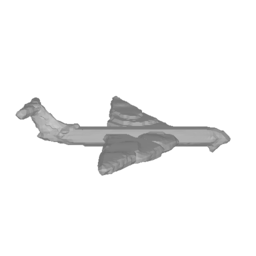}
    \end{subfigure}%
    \vskip\baselineskip%
    \vspace{-2.2em}
    \begin{minipage}[b]{\linewidth}
		\normalsize
	        \centering Geometry Mixing
    \end{minipage}%
    \vskip\baselineskip%
    \vskip\baselineskip%
            \begin{minipage}[b]{0.14\linewidth}
		\centering Texture 1
    \end{minipage}
    \begin{minipage}[b]{0.14\linewidth}
		\centering Texture 2
    \end{minipage}
    \begin{minipage}[b]{0.14\linewidth}
		\centering Texture 3
    \end{minipage}
    \begin{minipage}[b]{0.14\linewidth}
		\centering Mixing (Shape 1)
    \end{minipage}
    \begin{minipage}[b]{0.14\linewidth}
		\centering Mixing (Shape 2)
    \end{minipage}
    \begin{minipage}[b]{0.14\linewidth}
		\centering Mixing (Shape 3)
    \end{minipage}
    \begin{subfigure}[b]{0.14\linewidth}
		\centering Mixing (Combined)
    \end{subfigure}
    \vskip\baselineskip%
    \vspace{-1.2em}
    \begin{subfigure}[b]{0.14\linewidth}
		\centering
		\includegraphics[width=\linewidth]{supplementary_editing_shape1}
    \end{subfigure}%
    \hfill%
    \begin{subfigure}[b]{0.14\linewidth}
		\centering
		\includegraphics[width=\linewidth]{supplementary_editing_shape2}
    \end{subfigure}%
    \hfill%
    \begin{subfigure}[b]{0.14\linewidth}
		\centering
		\includegraphics[width=\linewidth]{supplementary_editing_shape3}
    \end{subfigure}%
    \hfill%
    \begin{subfigure}[b]{0.14\linewidth}
		\centering
		\includegraphics[width=\linewidth]{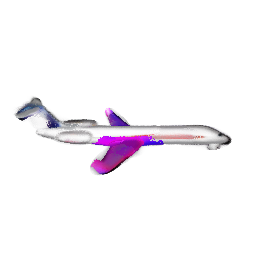}
    \end{subfigure}%
    \hfill%
    \begin{subfigure}[b]{0.14\linewidth}
		\centering
		\includegraphics[width=\linewidth]{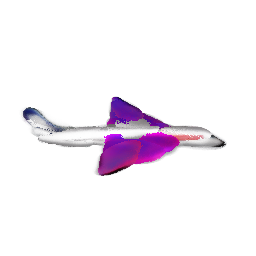}
    \end{subfigure}%
    \hfill%
    \begin{subfigure}[b]{0.14\linewidth}
		\centering
		\includegraphics[width=\linewidth]{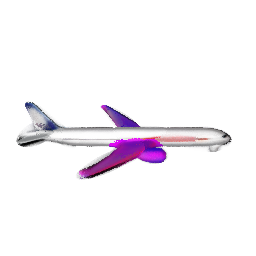}
    \end{subfigure}%
    \hfill%
    \begin{subfigure}[b]{0.14\linewidth}
		\centering
		\includegraphics[width=\linewidth]{supplementary_editing_combined_mixing}
    \end{subfigure}%
    \vskip\baselineskip%
    \vspace{-2.2em}
    \begin{subfigure}[b]{0.14\linewidth}
		\centering
		\includegraphics[width=\linewidth]{supplementary_editing_shape1_parts}
    \end{subfigure}%
    \hfill%
    \begin{subfigure}[b]{0.14\linewidth}
		\centering
		\includegraphics[width=\linewidth]{supplementary_editing_shape2_parts}
    \end{subfigure}%
    \hfill%
    \begin{subfigure}[b]{0.14\linewidth}
		\centering
		\includegraphics[width=\linewidth]{supplementary_editing_shape3_parts}
    \end{subfigure}%
    \hfill%
    \begin{subfigure}[b]{0.14\linewidth}
		\centering
		\includegraphics[width=\linewidth]{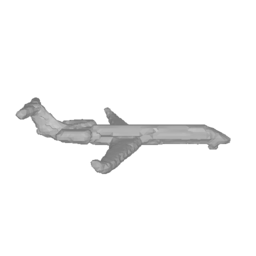}
    \end{subfigure}%
    \hfill%
    \begin{subfigure}[b]{0.14\linewidth}
		\centering
		\includegraphics[width=\linewidth]{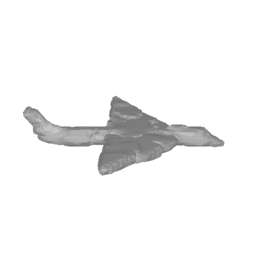}
    \end{subfigure}%
    \hfill%
    \begin{subfigure}[b]{0.14\linewidth}
		\centering
		\includegraphics[width=\linewidth]{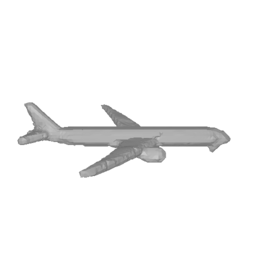}
    \end{subfigure}%
    \hfill%
    \begin{subfigure}[b]{0.14\linewidth}
		\centering
		\includegraphics[width=\linewidth]{supplementary_editing_combined_mixing_parts}
    \end{subfigure}%
    \vskip\baselineskip%
    \vspace{-2.2em}
    \begin{minipage}[b]{\linewidth}
		\normalsize
	        \centering Texture Mixing
    \end{minipage}%
    \vskip\baselineskip%
    \vspace{-1.2em}
    \begin{subfigure}[b]{0.16\linewidth}
		\centering
		\includegraphics[width=\linewidth]{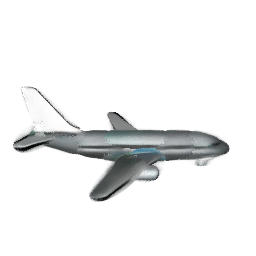}
    \end{subfigure}%
    \hfill%
    \begin{subfigure}[b]{0.16\linewidth}
		\centering
		\includegraphics[width=\linewidth]{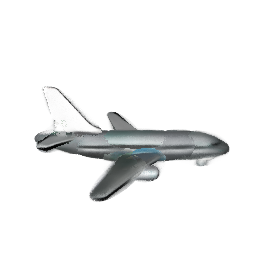}
    \end{subfigure}%
    \hfill%
    \begin{subfigure}[b]{0.16\linewidth}
		\centering
		\includegraphics[width=\linewidth]{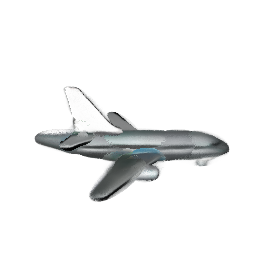}
    \end{subfigure}%
    \hfill%
    \begin{subfigure}[b]{0.16\linewidth}
		\centering
		\includegraphics[width=\linewidth]{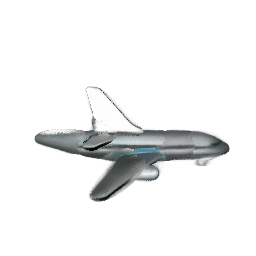}
    \end{subfigure}%
    \hfill%
    \begin{subfigure}[b]{0.16\linewidth}
		\centering
		\includegraphics[width=\linewidth]{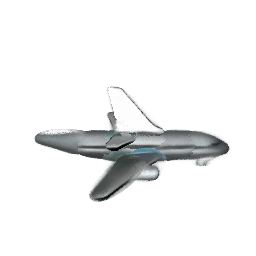}
    \end{subfigure}%
    \hfill%
    \begin{subfigure}[b]{0.16\linewidth}
		\centering
		\includegraphics[width=\linewidth]{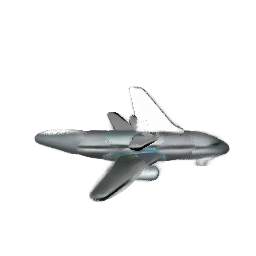}
    \end{subfigure}%
    \vskip\baselineskip%
    \vspace{-3.4em}
    \begin{subfigure}[b]{0.16\linewidth}
		\centering
		\includegraphics[width=\linewidth]{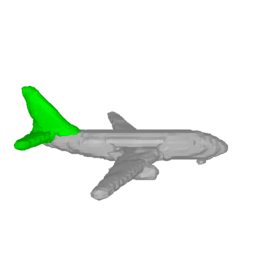}
    \end{subfigure}%
    \hfill%
    \begin{subfigure}[b]{0.16\linewidth}
		\centering
		\includegraphics[width=\linewidth]{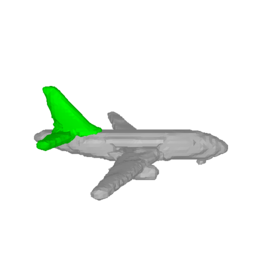}
    \end{subfigure}%
    \hfill%
    \begin{subfigure}[b]{0.16\linewidth}
		\centering
		\includegraphics[width=\linewidth]{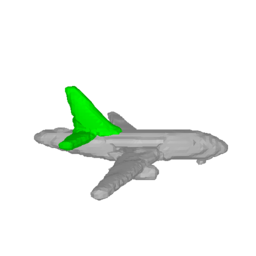}
    \end{subfigure}%
    \hfill%
    \begin{subfigure}[b]{0.16\linewidth}
		\centering
		\includegraphics[width=\linewidth]{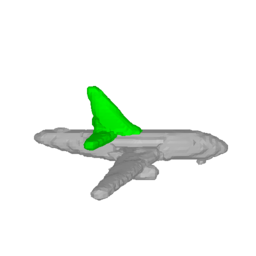}
    \end{subfigure}%
    \hfill%
    \begin{subfigure}[b]{0.16\linewidth}
		\centering
		\includegraphics[width=\linewidth]{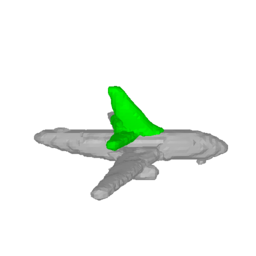}
    \end{subfigure}%
    \hfill%
    \begin{subfigure}[b]{0.16\linewidth}
		\centering
		\includegraphics[width=\linewidth]{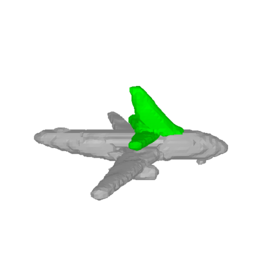}
    \end{subfigure}%
    \vskip\baselineskip%
    \vspace{-2.8em}
    \begin{subfigure}[b]{0.16\linewidth}
		\centering
		\includegraphics[width=\linewidth]{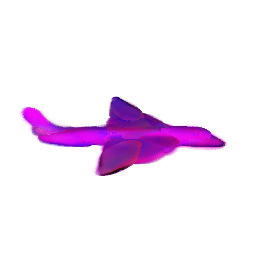}
    \end{subfigure}%
    \hfill%
    \begin{subfigure}[b]{0.16\linewidth}
		\centering
		\includegraphics[width=\linewidth]{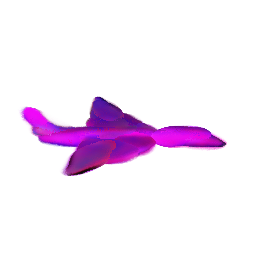}
    \end{subfigure}%
    \hfill%
    \begin{subfigure}[b]{0.16\linewidth}
		\centering
		\includegraphics[width=\linewidth]{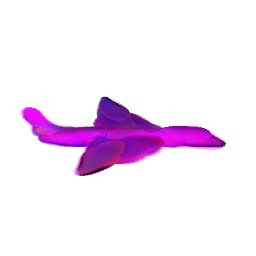}
    \end{subfigure}%
    \hfill%
    \begin{subfigure}[b]{0.16\linewidth}
		\centering
		\includegraphics[width=\linewidth]{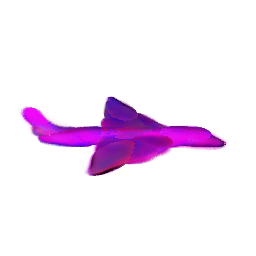}
    \end{subfigure}%
    \hfill%
    \begin{subfigure}[b]{0.16\linewidth}
		\centering
		\includegraphics[width=\linewidth]{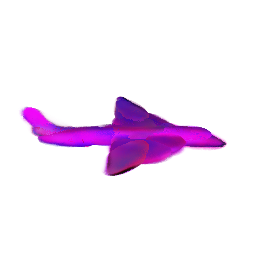}
    \end{subfigure}%
    \hfill%
    \begin{subfigure}[b]{0.16\linewidth}
		\centering
		\includegraphics[width=\linewidth]{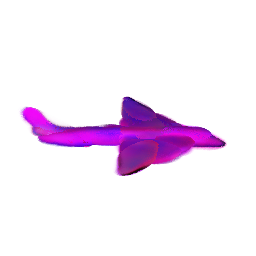}
    \end{subfigure}%
    \vskip\baselineskip%
    \vspace{-3.4em}
    \begin{subfigure}[b]{0.16\linewidth}
		\centering
		\includegraphics[width=\linewidth]{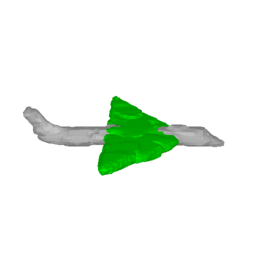}
    \end{subfigure}%
    \hfill%
    \begin{subfigure}[b]{0.16\linewidth}
		\centering
		\includegraphics[width=\linewidth]{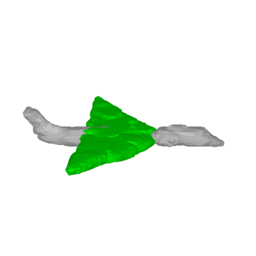}
    \end{subfigure}%
    \hfill%
    \begin{subfigure}[b]{0.16\linewidth}
		\centering
		\includegraphics[width=\linewidth]{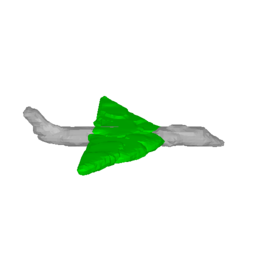}
    \end{subfigure}%
    \hfill%
    \begin{subfigure}[b]{0.16\linewidth}
		\centering
		\includegraphics[width=\linewidth]{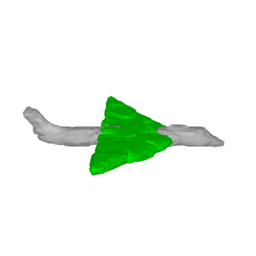}
    \end{subfigure}%
    \hfill%
    \begin{subfigure}[b]{0.16\linewidth}
		\centering
		\includegraphics[width=\linewidth]{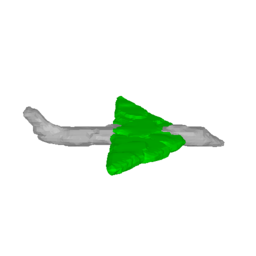}
    \end{subfigure}%
    \hfill%
    \begin{subfigure}[b]{0.16\linewidth}
		\centering
		\includegraphics[width=\linewidth]{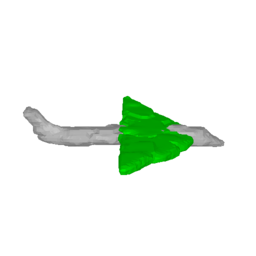}
    \end{subfigure}%
    \vskip\baselineskip%
    \vspace{-2.2em}
    \begin{minipage}[b]{\linewidth}
		\normalsize
	        \centering Shape Editing
    \end{minipage}%
    \vskip\baselineskip%
    \vspace{-1.2em}
    \caption{{\bf Shape Editing}. We show examples of our results when performing geometry mixing, texture mixing and part editing operations on image inverted ShapeNet airplanes.}
    \label{fig:supp_shape_mixing}
    \vspace{-1.2em}
\end{figure}

In this section, we demonstrate the editing capabilities of our model on
inverted shapes from ShapeNet \emph{Airplane}. To perform the image inversion,
we follow the process described in \secref{sec:image_inversion}. In
\figref{fig:supp_shape_mixing}, we show several examples of \emph{geometry},
\emph{texture} mixing and shape editing. In particular, we show two shape
editing examples, (see last four rows in \figref{fig:supp_shape_mixing}), where
we select several parts from two airplanes and we translate them along the body of the
two airplanes. The parts that we select correspond to the tail of
the first airplane and the wings of the second, highlighted in green.
Across both transformations, only the pose of the selected parts
change, while their appearance/texture as well as the shape and appearance of
the other parts does not change. For the case of texture mixing (see third and
fourth rows in \figref{fig:supp_shape_mixing}), we start from three airplanes
and perform several texture mixing operations. For example, in the last column
of the third row of \figref{fig:supp_shape_mixing}, we show a generated
airplane, whose texture and shape for the tail is from the first shape, while
the texture and shape for the wings and the body are from the second and third
shape respectively. Likewise, for the geometry mixing experiment (see two first
rows in \figref{fig:supp_shape_mixing}), we show examples where we mix the
shapes of the three input airplanes, while keeping the texture of the first
(fourth column), the second (fifth column) or the third input airplane (sixth
column).

\clearpage
\section{Discussion and Limitations}
\label{sec:parts_discussion}

In the primitive-based literature, the word part refers to geometric
primitives that are typically simple shapes such as cylinders
\cite{Li2019CVPR}, cuboids \cite{Tulsiani2017CVPRa}, superquadrics
\cite{Paschalidou2019CVPR, Paschalidou2020CVPR}, spheres \cite{Hao2020CVPR} or
more generally convex shapes \cite{Deng2020CVPR}. More recently, the term part
has also been used to describe parts that can capture complex geometries and
are not limited to simple geometric shapes \cite{Genova2019ARXIV,
Paschalidou2021CVPR, Kawana2020ARXIV}. Regardless of the part's expressivity,
primitive-based methods yield semantically consistent reconstructions, where
the same part is consistently used for representing the same part of the
object. Being semantically consistent does not necessarily mean that parts will
also be semantically meaningful, namely refer to humanly identifiable parts.

As we parametrize parts by neural radiance fields, our parts can capture
complex topologies that are not limited to simple geometric shapes. Examples of
our generated parts can be found in
\figref{fig:generated_motorbikes}$-$\figref{fig:interpolated_cars}.  We note
that our parts can be coherent across some objects, e.g. the same part is used
for representing the same region of the object, in particular when the parts
have comparable size and shape. However, for two shapes with different sizes e.g
a big and small car (see \figref{fig:generated_cars}), we note that our parts
are not semantically consistent, namely the part that is used to represent the
tire and the top part of the car for a smaller car is now representing only the
tire of a bigger car. Note that this is not a limitation only of our work but
also of existing part-based methods, as they do not explicitly enforce the
semantic consistency of parts.

\begin{figure}[H]
    \begin{subfigure}[t]{0.49\columnwidth}
        \centering
        \includegraphics[width=\textwidth]{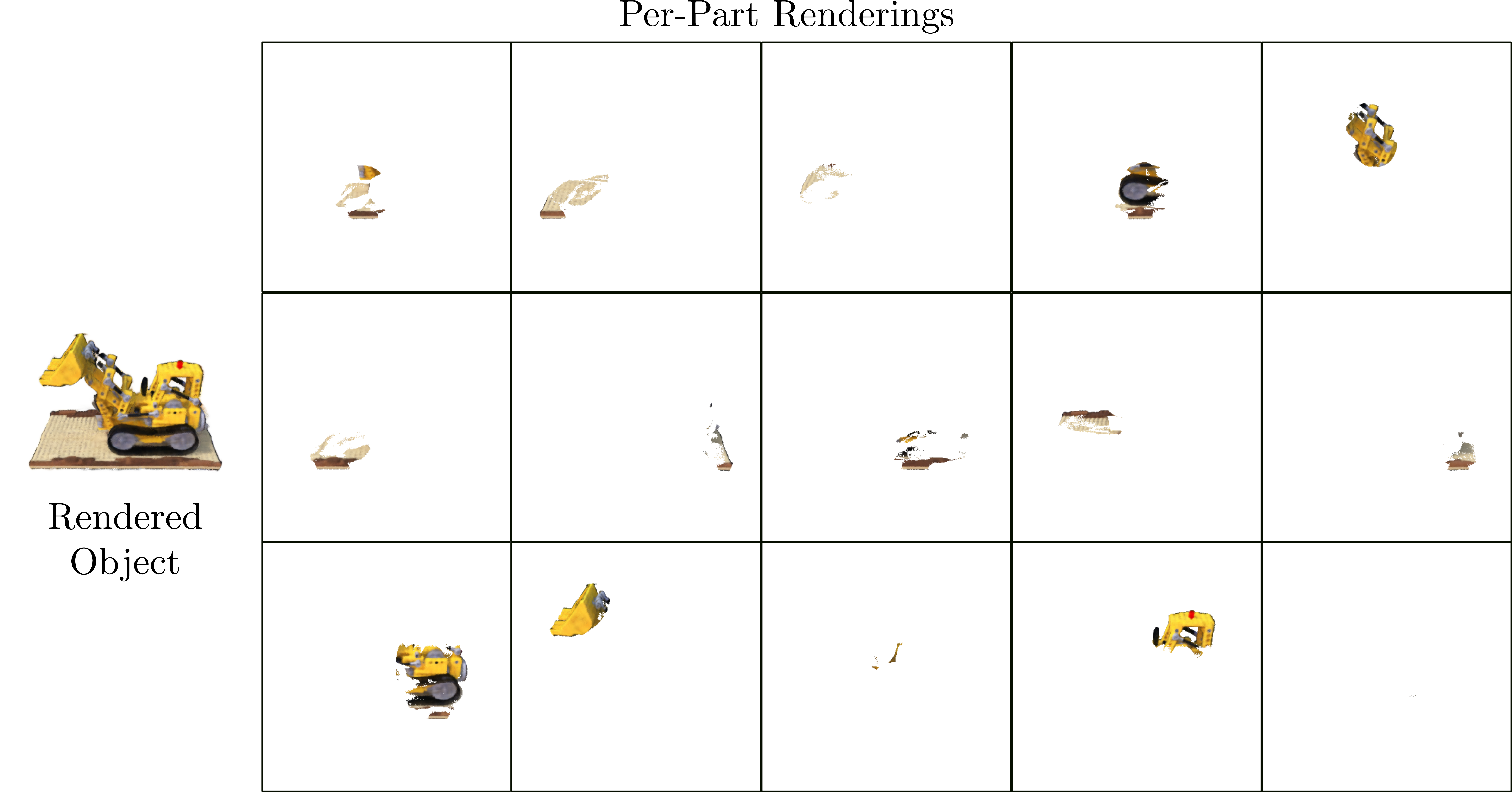}
        \caption{Tractor Scene}
    \end{subfigure}%
    \hfill%
    \begin{subfigure}[t]{0.49\columnwidth}
        \centering
        \includegraphics[width=\textwidth]{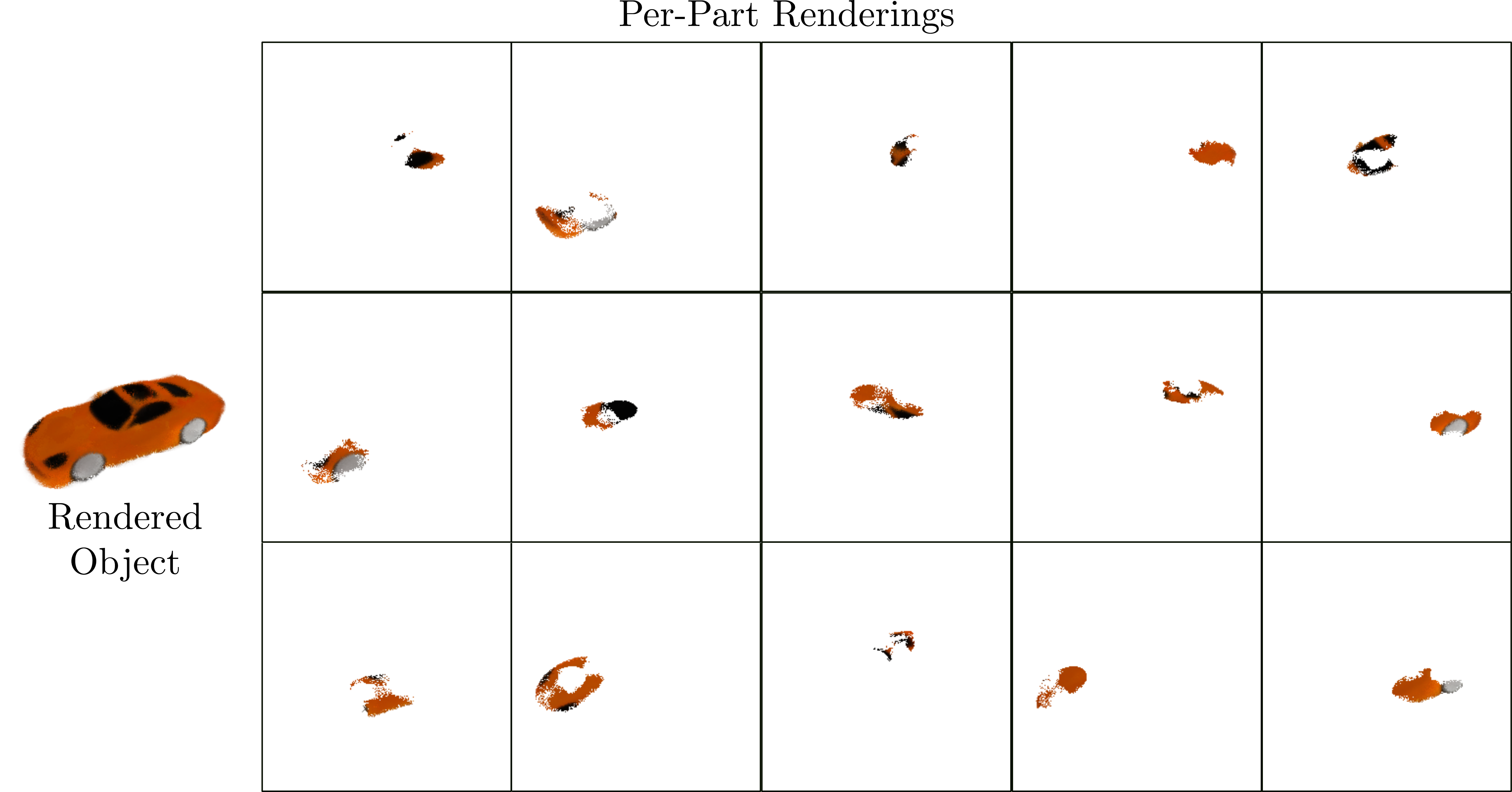}
        \caption{ShapeNet Car}
    \end{subfigure}%
    \caption{{\bf Per-Part Renderings.} We show the per-part renderings produced by
    our $16$ NeRF/parts for a novel view. }
    \label{fig:per_part_renderings}
\end{figure}

In addition, in \figref{fig:per_part_renderings}, we provide two examples of the renderings
produced using our $16$ NeRFs/parts, when rendering the tractor scene from a novel view
and when rendering a ShapeNet car from a novel camera. Although the per-part renderings are consistently
crisp and faithfully capture the appearance details of the object, they do not
always correspond to semantically meaningful parts.
For example, for the case of the tractor scene, our network has associated one
part with the bucket of the tractor, however it uses multiple parts to capture
regions of the floor, whereas ideally we would like one part to capture the
entire floor. Similarly, for the case of the car, while the per-part renderings
have sharp colors, they cover parts of the object that are not humanly
interpretable, such as a single NeRF representing part of the tire and part of the car.
This is again a common issue observed in all existing part-based
methods~\cite{Tulsiani2017CVPRa, Paschalidou2019CVPR, Deng2020CVPR,
Paschalidou2021CVPR}. As these methods do not explicitly enforce the
meaningfulness of their parts, oftentimes parts are associated with regions of
objects that are not meaningful and do not correspond to humanly interpretable
parts. We believe that in future research, it would be valuable to explore
ways to enforce that parts are more interpretable, in order to unlock more
editing operations, that better keep humans in the loop. Our main goal for this
work, was to introduce a generative model that enables local control through
parts.  While increasing the number of parts enables more control on the
generated shape, as we can select multiple small parts that compose a semantic
part such as the wings of the airplane,
we believe that it is useful to also explore other types of more
semantic control, where a user for example could select a part based on
its semantic class e.g. wing of an airplane or saddle of a motorbike, without
having to manually select multiple parts, as is the case for our model.

In contrast to existing NerF-based generative models \cite{Schwarz2020NEURIPS,
Chan2021CVPR, Chan2022CVPR, Gao2020NeurIPS} that rely on adversarial losses to
generate high quality renderings, our model does not employ such losses. We believe that 
incorporating 
triplane-based representations \cite{Chan2022CVPR} or adversarial losses could further improve 
the quality of our generated textures. Moreover, in our current framework,
moving one part of an object to a new location outside the object does not necessarily
generate a contiguous shape. To address this, \cite{Hertz2022SIGGRAPH},
introduced a blending network, implemented as a transformer decoder that merges
the subsequent parts and synthesizes a novel object. Incorporating such a
technique in our model is not possible, as we do not have access to 3D supervision
in the form of a mesh. However, we could utilize \cite{Yin2020THREEDV,
Lin2022NEURIPS} to synthesize connections between parts.

To summarize, while our work makes an important step towards generating
editable 3D shapes, it still has several limitations. In particular, during
training, our supervision comes from posed images and object masks. Nevertheless, 
acquiring detailed masks for objects in the wild is not always possible.
Therefore, we believe it would be useful to explore ways to alleviate the need
for object masks. Likewise, adopting a formulation like
\cite{Schwarz2020NEURIPS} that enables training from unposed images could
further extend the capabilities of our model. In addition, while our model can
perform several editing operations, our current formulation does not support
operations where a part is deformed using techniques such as Neural Cages \cite{Yifan2020CVPR}
or Biharmonic Coordinates \cite{Wang2015SIGGRAPH}. Incorporating deformations in our editing operations
could unlock several editing applications.

\section{Potential Negative Impact on Society}
\label{sec:negative_societal_impact}

Our proposed model enables generating editable 3D meshes with textures. While,
we see this as an important step towards automatic content creation and
enabling a multitude of editing functionalities, it can also lead to negative
consequences, when applied to sensitive data, such as human bodies or faces.
Therefore, we believe it is imperative to always check the license of any 3D
publicly available 3D model. In addition, we see the development of techniques
for identifying real from synthetic data as an essential research direction that
could potential prevent deep fake.

Note that throughout this work, we have only worked with publicly available
datasets and did not use any data that involves privacy or copyright concerns.
For future users that would like to train our model on new data, we recommend to
first remove biases from the training data in order to ensure that our model
can fairly capture the diversities in terms of shapes, sizes and textures.

\end{document}